\PassOptionsToPackage{dvipsnames}{xcolor}
\documentclass[twoside,11pt]{article}

\usepackage[abbrvbib]{jmlr2e}

\usepackage{url}            
\usepackage{booktabs}       
\usepackage{amsfonts}       
\usepackage{nicefrac}       
\usepackage{microtype}      
\usepackage{xcolor}         
\usepackage{mathtools}
\mathtoolsset{showonlyrefs}
\usepackage{bm}
\usepackage{bbm}
\usepackage{tocloft}
\usepackage{enumerate}
\usepackage{footnote}
\usepackage{amssymb}
\usepackage{amsmath}
\usepackage{mathrsfs}
\usepackage{caption}
\usepackage{subcaption}
\usepackage{dsfont}
\usepackage{adjustbox}
\usepackage{array}
\usepackage{colortbl}
\usepackage{graphicx}
\usepackage{multirow, hhline}
\usepackage[ruled,linesnumbered]{algorithm2e}

\usepackage{hyperref}
\hypersetup{ hidelinks }

\allowdisplaybreaks
\usepackage{amsmath}
\makeatletter
\let\save@mathaccent\mathaccent
\newcommand*\if@single[3]{%
  \setbox0\hbox{${\mathaccent"0362{#1}}^H$}%
  \setbox2\hbox{${\mathaccent"0362{\kern0pt#1}}^H$}%
  \ifdim\ht0=\ht2 #3\else #2\fi
  }
\newcommand*\rel@kern[1]{\kern#1\dimexpr\macc@kerna}
\newcommand*\widebar[1]{\@ifnextchar^{{\wide@bar{#1}{0}}}{\wide@bar{#1}{1}}}
\newcommand*\wide@bar[2]{\if@single{#1}{\wide@bar@{#1}{#2}{1}}{\wide@bar@{#1}{#2}{2}}}
\newcommand*\wide@bar@[3]{%
  \begingroup
  \def\mathaccent##1##2{%
    \let\mathaccent\save@mathaccent
    \if#32 \let\macc@nucleus\first@char \fi
    \setbox\z@\hbox{$\macc@style{\macc@nucleus}_{}$}%
    \setbox\tw@\hbox{$\macc@style{\macc@nucleus}{}_{}$}%
    \dimen@\wd\tw@
    \advance\dimen@-\wd\z@
    \divide\dimen@ 3
    \@tempdima\wd\tw@
    \advance\@tempdima-\scriptspace
    \divide\@tempdima 10
    \advance\dimen@-\@tempdima
    \ifdim\dimen@>\z@ \dimen@0pt\fi
    \rel@kern{0.6}\kern-\dimen@
    \if#31
      \overline{\rel@kern{-0.6}\kern\dimen@\macc@nucleus\rel@kern{0.4}\kern\dimen@}%
      \advance\dimen@0.4\dimexpr\macc@kerna
      \let\final@kern#2%
      \ifdim\dimen@<\z@ \let\final@kern1\fi
      \if\final@kern1 \kern-\dimen@\fi
    \else
      \overline{\rel@kern{-0.6}\kern\dimen@#1}%
    \fi
  }%
  \macc@depth\@ne
  \let\math@bgroup\@empty \let\math@egroup\macc@set@skewchar
  \mathsurround\z@ \frozen@everymath{\mathgroup\macc@group\relax}%
  \macc@set@skewchar\relax
  \let\mathaccentV\macc@nested@a
  \if#31
    \macc@nested@a\relax111{#1}%
  \else
    \def\gobble@till@marker##1\endmarker{}%
    \futurelet\first@char\gobble@till@marker#1\endmarker
    \ifcat\noexpand\first@char A\else
      \def\first@char{}%
    \fi
    \macc@nested@a\relax111{\first@char}%
  \fi
  \endgroup
}
\makeatother

\newtheorem{assumption}{Assumption} 

\newcommand{\bP}{\bm{P}}
\newcommand{\tP}{\mathbb{P}}
\newcommand{\tE}{\mathbb{E}}

\newcommand{\bdeltaks}[1]{\bm{\delta}^{(#1)*}}
\newcommand{\bTheta}{\bm{\Theta}}
\newcommand{\btheta}{\bm{\theta}}
\newcommand{\barthetaks}[1]{\bar{\bm{\theta}}^{(#1)*}}

\newcommand{\tildethetaks}[1]{\tilde{\bm{\theta}}^{(#1)*}}

\newcommand{\bthetaks}[1]{\bm{\theta}^{(#1)*}}
\newcommand{\bthetak}[1]{\bm{\theta}^{(#1)}}
\newcommand{\htheta}{\widehat{\bm{\theta}}}
\newcommand{\hthetak}[1]{\widehat{\bm{\theta}}^{(#1)}}
\newcommand{\bbeta}{\bm{\beta}}
\newcommand{\bbetaks}[1]{\bm{\beta}^{(#1)*}}
\newcommand{\bbetak}[1]{\bm{\beta}^{(#1)}}

\newcommand{\hbetak}[1]{\widehat{\bm{\beta}}^{(#1)}}
\newcommand{\wtbbeta}{\widetilde{\bbeta}}
\newcommand{\wtbbetak}[1]{\widetilde{\bbeta}^{(#1)}}
\newcommand{\bA}{\bm{A}}
\newcommand{\bB}{\bm{B}}
\newcommand{\wtbA}{\widetilde{\bm{A}}}
\newcommand{\barA}{\widebar{\bm{A}}}
\newcommand{\hbarA}{\widehat{\widebar{\bm{A}}}}
\newcommand{\bAk}[1]{\bm{A}^{(#1)}}
\newcommand{\bAks}[1]{\bm{A}^{(#1)*}}
\newcommand{\hA}{\widehat{\bm{A}}}
\newcommand{\hAk}[1]{\widehat{\bm{A}}^{(#1)}}
\newcommand{\oA}{\widebar{\bm{A}}}
\newcommand{\hoA}{\widehat{\widebar{\bm{A}}}}
\newcommand{\bx}{\bm{x}}

\newcommand{\bxk}[1]{\bm{x}^{(#1)}}
\newcommand{\yk}[1]{y^{(#1)}}
\newcommand{\bXk}[1]{\bm{X}^{(#1)}}
\newcommand{\bYk}[1]{\bm{Y}^{(#1)}}
\newcommand{\bSigma}{\bm{\Sigma}}
\newcommand{\bSigmak}[1]{\bm{\Sigma}^{(#1)}}

\newcommand{\hSigmak}[1]{\widehat{\bm{\Sigma}}^{(#1)}}
\newcommand{\fk}[1]{f^{(#1)}}

\newcommand{\bepsilonk}[1]{\bm{\epsilon}^{(#1)}}
\newcommand{\epsilonk}[1]{\epsilon^{(#1)}}
\newcommand{\mO}{\mathcal{O}^{p \times r}}
\newcommand{\mOr}{\mathcal{O}^{r \times r}}
\newcommand{\mQ}{\mathbb{Q}}
\newcommand{\mA}{\mathcal{A}}
\newcommand{\mB}{\mathcal{B}}
\newcommand{\bu}{\bm{u}}
\newcommand{\bv}{\bm{v}}
\newcommand{\sigmamin}{\sigma_{\min}}
\newcommand{\lambdamin}{\lambda_{\min}}
\newcommand{\sigmamax}{\sigma_{\max}}
\newcommand{\lambdamax}{\lambda_{\max}}
\newcommand{\bDelta}{\bm{\Delta}}
\newcommand{\mS}{\mathcal{S}}
\newcommand{\wpr}{w.p. at least $1-e^{-C(r+\log T)}$}
\newcommand{\wpp}{w.p. at least $1-e^{-C(p+\log T)}$}
\newcommand{\wppp}{w.p. at least $1-e^{-C'(p+\log T)}$}

\newcommand{\wprp}{w.p. at least $1-e^{-C'(r+\log T)}$}

\newcommand{\KL}{\textup{KL}}
\newcommand{\dint}{\textup{d}}
\newcommand{\bR}{\bm{R}}
\newcommand{\barzeta}{\bar{\zeta}}
\newcommand{\zetak}[1]{\zeta^{(#1)}}

\makeatletter
\newcommand*{\Rom}[1]{\expandafter\@slowromancap\romannumeral #1@}
\makeatother
\newcommand*{\rom}[1]{\romannumeral #1}

\DeclareMathOperator*{\argmin}{arg\,min}

\newcommand{\norm}[1]{|#1|}
\newcommand{\norma}[1]{\left|#1\right|}

\newcommand{\twonorm}[1]{\|#1\|_{2}}
\newcommand{\twonorma}[1]{\left\|#1\right\|_{2}}
\newcommand{\infnorm}[1]{\|#1\|_{\infty}}

\newcommand{\maxnorm}[1]{\|#1\|_{\max}}

\newcommand{\fnorm}[1]{\|#1\|_{\textup{F}}}
\newcommand{\fnorma}[1]{\left|\left|#1\right|\right|_{\textup{F}}}
\newcommand{\distf}{\textup{dist}_{\textup{F}}}
\newcommand{\disttwo}{\textup{dist}_{2}}

\newcommand{\<}{\langle}
\renewcommand{\>}{\rangle}

\title{Learning from Similar Linear Representations: Adaptivity, Minimaxity, and Robustness}

\author{\name Ye Tian \email ye.t@columbia.edu \\
       \addr Department of Statistics\\
       Columbia University\\
       New York, NY 10027, USA
       \AND
       \name Yuqi Gu \email yuqi.gu@columbia.edu \\
       \addr Department of Statistics\\
       Columbia University\\
       New York, NY 10027, USA
       \AND
       \name Yang Feng \email yang.feng@nyu.edu \\
       \addr Department of Biostatistics, School of Global Public Health\\
       New York University\\
       New York, NY 10003, USA}

\usepackage{lastpage}
\firstpageno{1}
\jmlrheading{26}{2025}{1-\pageref{LastPage}}{7/23; Revised
4/25}{6/25}{23-0902}{Ye Tian, Yuqi Gu, and Yang Feng}
\ShortHeadings{Learning from Similar Linear Representations}{Tian, Gu, and Feng}

\begin{document}

\editor{Ji Zhu}

\maketitle

\begin{abstract}
Representation multi-task learning (MTL) has achieved tremendous success in practice. However, the theoretical understanding of these methods is still lacking. Most existing theoretical works focus on cases where all tasks share the same representation, and claim that MTL almost always improves performance. Nevertheless, as the number of tasks grows, assuming all tasks share the same representation is unrealistic. Furthermore, empirical findings often indicate that a shared representation does not necessarily improve single-task learning performance. In this paper, we aim to understand how to learn from tasks with \textit{similar but not exactly the same} linear representations, while dealing with outlier tasks. Assuming a known intrinsic dimension, we propose a penalized empirical risk minimization method and a spectral method that are \textit{adaptive} to the similarity structure and \textit{robust} to outlier tasks. Both algorithms outperform single-task learning when representations across tasks are sufficiently similar and the proportion of outlier tasks is small. Moreover, they always perform at least as well as single-task learning, even when the representations are dissimilar. We provide information-theoretic lower bounds to demonstrate that both methods are nearly \textit{minimax} optimal in a large regime, with the spectral method being optimal in the absence of outlier tasks. Additionally, we introduce a thresholding algorithm to adapt to an unknown intrinsic dimension. We conduct extensive numerical experiments to validate our theoretical findings.
\end{abstract}

\vspace{0.2cm}


\addtocontents{toc}{\protect\setcounter{tocdepth}{0}}
\section{Introduction}\label{sec: introduction}

\subsection{Representation Multi-task Learning}\label{subsec: RMTL}
With the increased computational power, machine learning systems can now process datasets on a large scale. However, for each machine learning task, we may not have access to a large amount of data due to data privacy restrictions and the high cost of data acquisition. This motivated the idea of multi-task learning (MTL), where we jointly learn many tasks that are similar but not identical to enhance model performance  \citep{zhang2018overview, zhang2021survey}. Related concepts include transfer learning (TL), learning-to-learn, and meta-learning, where model structures learned from multiple tasks can be transferred to new incoming tasks to improve their performance \citep{weiss2016survey, hospedales2021meta}. Among numerous multi-task and transfer learning approaches, representation learning has been one of the most popular and successful methods over the past few years, where a data representation is jointly learned from multiple similar data sets and can be shared across them \citep{rostami2022transfer}. A successful example of multi-task and transfer representation learning is learning the weights of a few initial layers of neural networks from ImageNet pre-training, then retraining final layers on new image classification tasks \citep{donahue2014decaf, goyal2019scaling}. Other applications include multilingual knowledge graph completion \citep{chen2020multilingual} and reinforcement learning \citep{gupta2017learning}.

While representation learning has been successful in practice, its theoretical understanding in the context of multi-task and transfer learning remains limited. Most existing theoretical works assume that the same representation is shared across all tasks, which is not always realistic in scenarios with a large number of tasks \citep{rostami2022transfer}. Furthermore, empirical studies have shown that freezing a representation across tasks from different contexts may not improve model performance and can even be harmful. For example, \cite{raghu2019transfusion} found that pre-training on ImageNet offered little help to target medical tasks, and \cite{wang2019can} found that different target tasks might benefit from different pre-training in natural language understanding. These studies suggested that a frozen representation may not always work well. Additionally, there may be outlier tasks that are dissimilar to other tasks \citep{zhang2021survey} or may be contaminated with adversarial attacks on the data \citep{qiao2018outliers, qiao2018learning, konstantinov2020sample}. If left unaddressed, such issues could severely impact the machine learning system's overall performance.

This paper investigates the effective learning of tasks with \textit{similar representations} in the presence of potential outlier tasks or adversarial attacks. Specifically, we consider the following linear model with linear representations. Suppose there are $T$ tasks in total, and we have collected a sample $\{\bxk{t}_i, \yk{t}_i\}_{i=1}^n$ from the $t$-th task, where $\bxk{t}_i \in \mathbb{R}^p$, $\yk{t}_i \in \mathbb{R}$, and $t \in [T] = \{1, 2, \ldots, T\}$. There exists an unknown subset $S \subseteq [T]$, such that for all $t \in S$, 
\begin{equation}\label{eq: linear model intro}
	\yk{t}_i = (\bxk{t}_i)^\top\bbetaks{t} + \epsilonk{t}_i, \quad i = 1:n,
\end{equation}
where the regression coefficient $\bbetaks{t}=\bAks{t}\bthetaks{t}$, the representation $\bAks{t} \in \mO = \{\bA \in \mathbb{R}^{p \times r}: \bA^\top \bA = \bm{I}_r\}$, low-dimensional parameter $\bthetaks{t} \in \mathbb{R}^r$, $r \leq p$, and $\{\epsilonk{t}_i\}_{i=1}^n$ are random noises. Here $r$ represents the intrinsic dimension of the problem, which is usually much smaller than $p$. The data $\{\bxk{t}_i, \yk{t}_i\}_{i=1}^n$ for $t \notin S$ can be arbitrarily distributed in the worst case, and the corresponding tasks in $S^c = [T]\backslash S$ are outlier or contaminated tasks. We call $\epsilon \coloneqq |S^c|/T$ the \textit{contamination proportion} or the \textit{proportion of outlier tasks}. To ensure effective learning from similar representations, we assume that $\{\bAks{t}\}_{t \in S}$ are similar to each other, in the sense that $\min_{\oA \in \mO}\max_{t \in S}\twonorm{\bAks{t}(\bAks{t})^\top - \oA  (\oA)^\top} \leq h$, where $\oA\in \mO$ achieving the minimum can be understood as a ``central representation" and $h$ is the similarity measure. Our goal is to explore the upper and lower error bounds in estimating $\{\bbetaks{t}\}_{t \in S}$ for all possible cases of $S$ under certain conditions. Furthermore, when the tasks in $S^c$ also satisfy the linear model \eqref{eq: linear model intro}, we aim to ensure the effective estimation of $\{\bbetaks{t}\}_{t \in S^c}$ as well.

It is worth pointing out that we allow the scales of $\{\bthetaks{t}\}_{t \in S}$, i.e. $\{\twonorm{\bthetaks{t}}\}_{t \in S}$, to differ across tasks in $S$, and $\twonorm{\bthetaks{t}}$ can also diverge as $n \rightarrow \infty$. Here, $\twonorm{\bthetaks{t}} = \twonorm{\bbetaks{t}}$ can be viewed as the signal strength of the $t$-th task. It turns out that the performance of representation MTL on each task is highly relevant to $\twonorm{\bthetaks{t}}$. In contrast, existing literature generally assumes $\twonorm{\bthetaks{t}} \lesssim 1$ for all $t \in S=[T]$ and ignores the impact of $\twonorm{\bthetaks{t}}$ on the model performance.

\subsection{Related Works}\label{subsec: related works}

\subsubsection{Multi-task and Transfer Learning}
\textbf{Representation MTL and TL:} \cite{baxter2000model} is among the earliest works to study the theory of representation MTL under general function classes, where all tasks are generated from the same distribution. \cite{maurer2016benefit} improved their results by using the analysis based on Rademacher complexity. \cite{ando2005framework} explored the case of semi-supervised learning. More recently, \cite{du2020few} and \cite{tripuraneni2021provable} studied linear model \eqref{eq: linear model intro} with $S = [T]$ and $h = 0$, i.e., under the assumption that there are no outlier tasks and all tasks share the same representation. They proposed the so-called task diversity condition, under which the learning rate can be significantly improved by a non-convex empirical risk minimization (ERM) algorithm. \cite{tripuraneni2020theory} extended the analysis to general non-linear models and provided general results. \cite{thekumparampil2021statistically} proposed a polynomial-time alternating gradient descent algorithm that achieves similar performance as ERM but avoids solving the non-convex optimization directly. \cite{meunier2023nonlinear} characterizes the shared representation via a mapping into a finite-dimensional subspace of a reproducing kernel Hilbert space (RKHS). Other related works include federated representation learning \citep{collins2021exploiting, duchi2022subspace}, tensor representation meta-learning \citep{deng2022learning},  conditional meta-learning \citep{denevi2020advantage}, and matrix completion via representation MTL \citep{zhou2021multi}. Note that MTL under the assumption that $\bbetaks{t}$'s in \eqref{eq: linear model intro} share the same or similar support sets \citep{lounici2009taking, lounici2011oracle, jalali2010dirty, li2021targeting, xu2021learning} can also be viewed as a special case of the general representation MTL.

\textbf{Distance-based MTL and TL:} There has been much literature in the statistics community studying model \eqref{eq: linear model intro} under the assumption that Euclidean distance or $\ell_1$-distance between $\bbetaks{t}$'s are small \citep{bastani2021predicting, li2022transfer, duan2023adaptive, gu2023commute}, which is called ``distance-based'' MTL and TL in \cite{gu2024robust}. Some extensions include high-dimensional GLMs \citep{tian2022transfer}, graphical models \citep{li2022transfer2}, functional regression \citep{lin2022transfer}, semi-supervised classification \citep{zhou2022doubly}, and unsupervised mixture models \citep{tian2022unsupervised, tian2024towards}. Recently, \cite{gu2024robust} proposed the ``angle-based'' TL where they assume the angle between every pair of $\bbetaks{t}$'s is small. As we will discuss in the next section, their setting is a special case of \eqref{eq: linear model intro} when $r = 1$.

\textbf{Other related literature:} Other relevant literature includes the non-parametric TL \citep{cai2021transfer, kpotufe2021marginal}, the hardness of MTL \citep{hanneke2019value, hanneke2022no}, adversarial robustness of MTL or distributed learning \citep{chen2017distributed,  alistarh2018byzantine, yin2018byzantine, qiao2018outliers, qiao2018learning, konstantinov2020sample, zhu2023byzantine, guerraoui2024byzantine}, gradient-based meta-learning \citep{finn2017model, nichol2018first, finn2019online}, and theory of MTL based on distributional measure \citep{ben2008notion, ben2010theory}.

\begin{figure}[!h]
     \centering
     \begin{subfigure}[b]{0.46\textwidth}
         \centering
         \includegraphics[width=\textwidth]{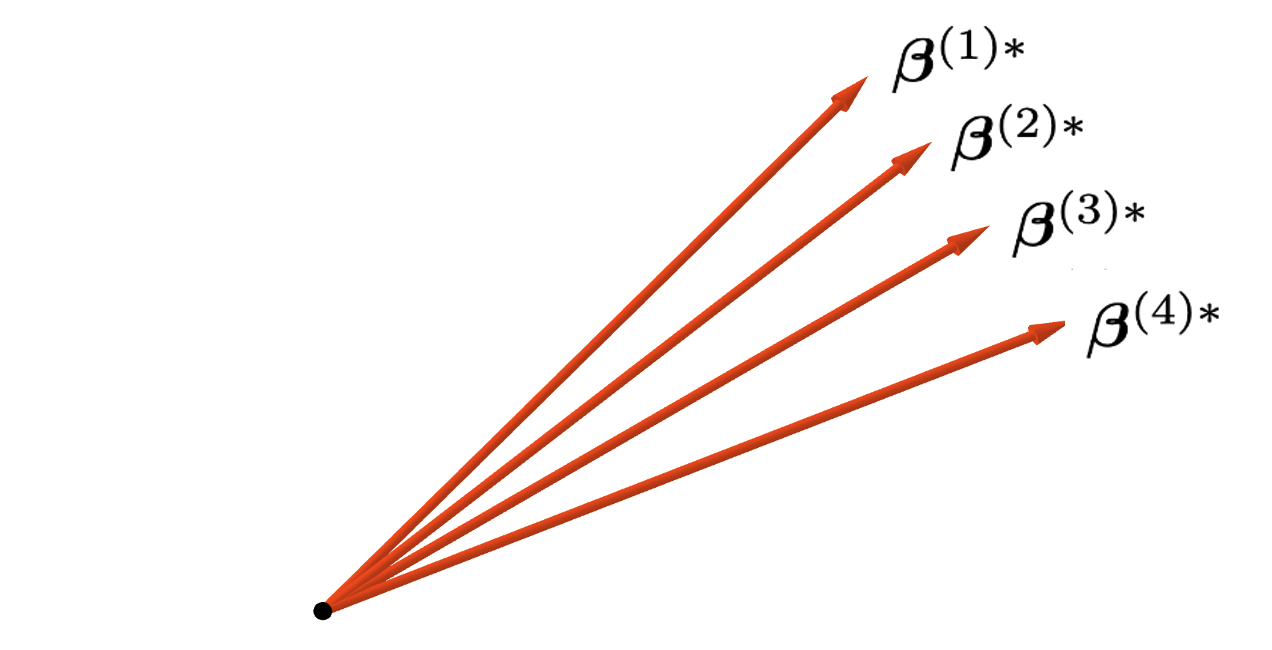}
         \caption{Distance-based similarity \citep{bastani2021predicting, li2022transfer, tian2022transfer, duan2023adaptive, tian2024towards}}
         \label{fig: diag-distance}
     \end{subfigure}
     \hfill
     \begin{subfigure}[b]{0.47\textwidth}
         \centering
         \includegraphics[width=\textwidth]{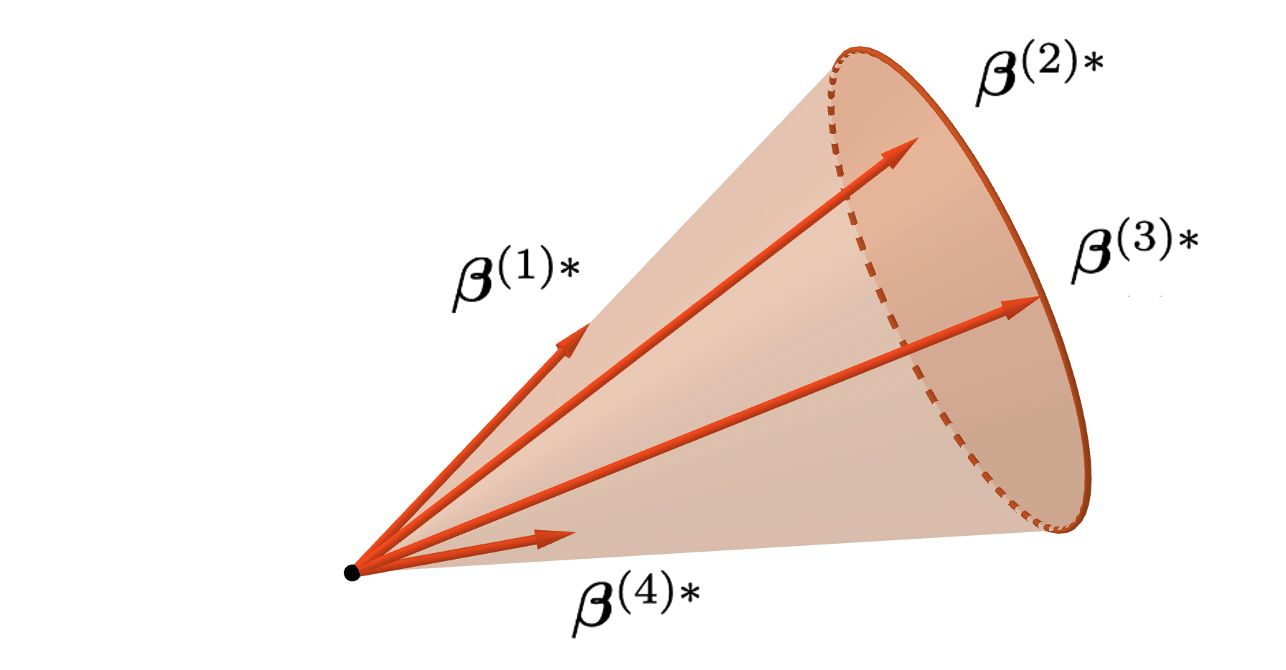}
         \caption{Angle-based similarity \citep{gu2024robust}\\\hspace{\textwidth}\\\hspace{\textwidth}}
         \label{fig: diag-angle}
     \end{subfigure}
     \\
     \begin{subfigure}[b]{0.45\textwidth}
         \centering
         \includegraphics[width=\textwidth]{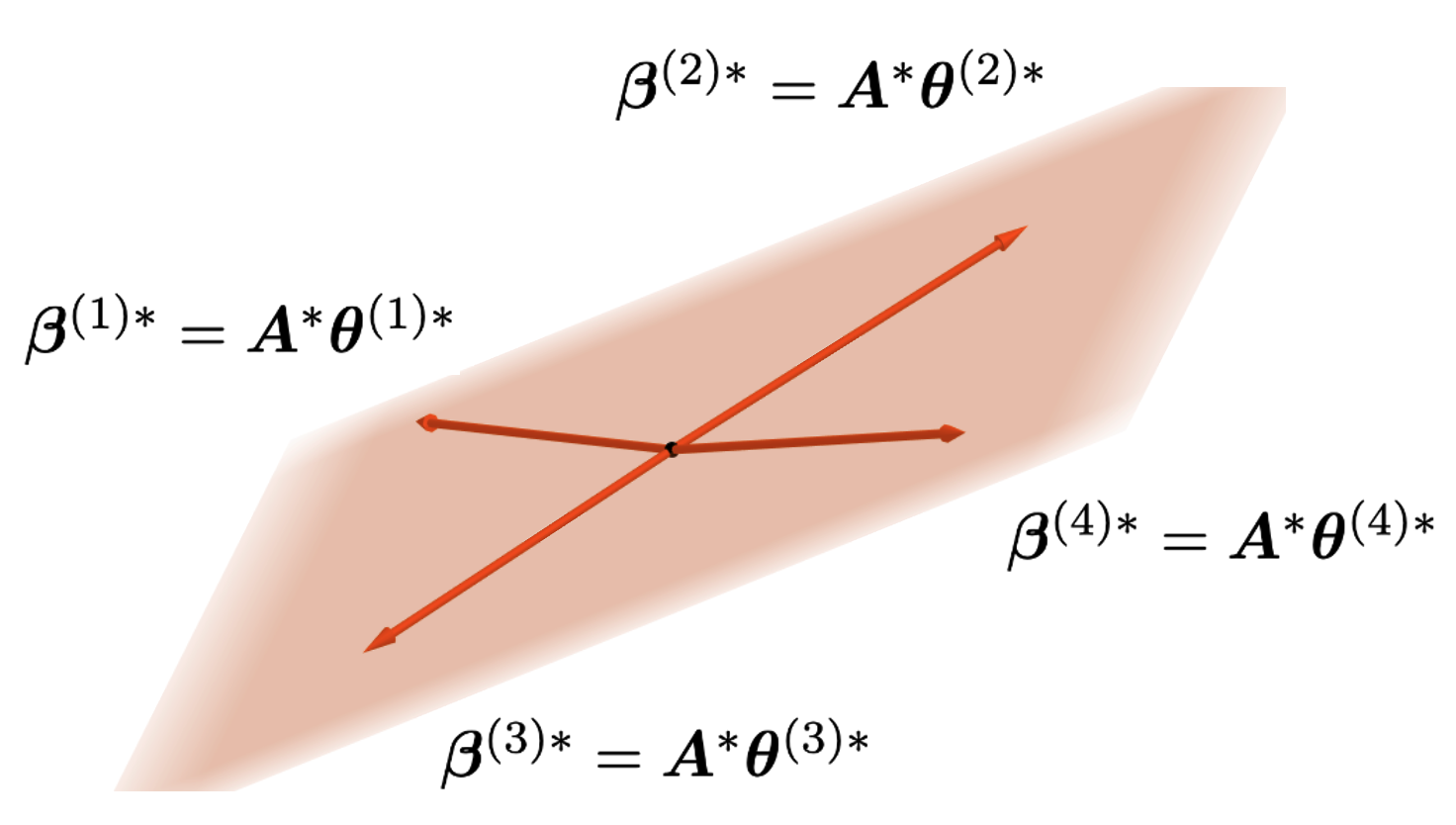}
         \caption{The same representation \citep{du2020few, tripuraneni2021provable, thekumparampil2021statistically}}
         \label{fig: diag-same-A}
     \end{subfigure}
     \hfill
     \begin{subfigure}[b]{0.54\textwidth}
         \centering
         \includegraphics[width=\textwidth]{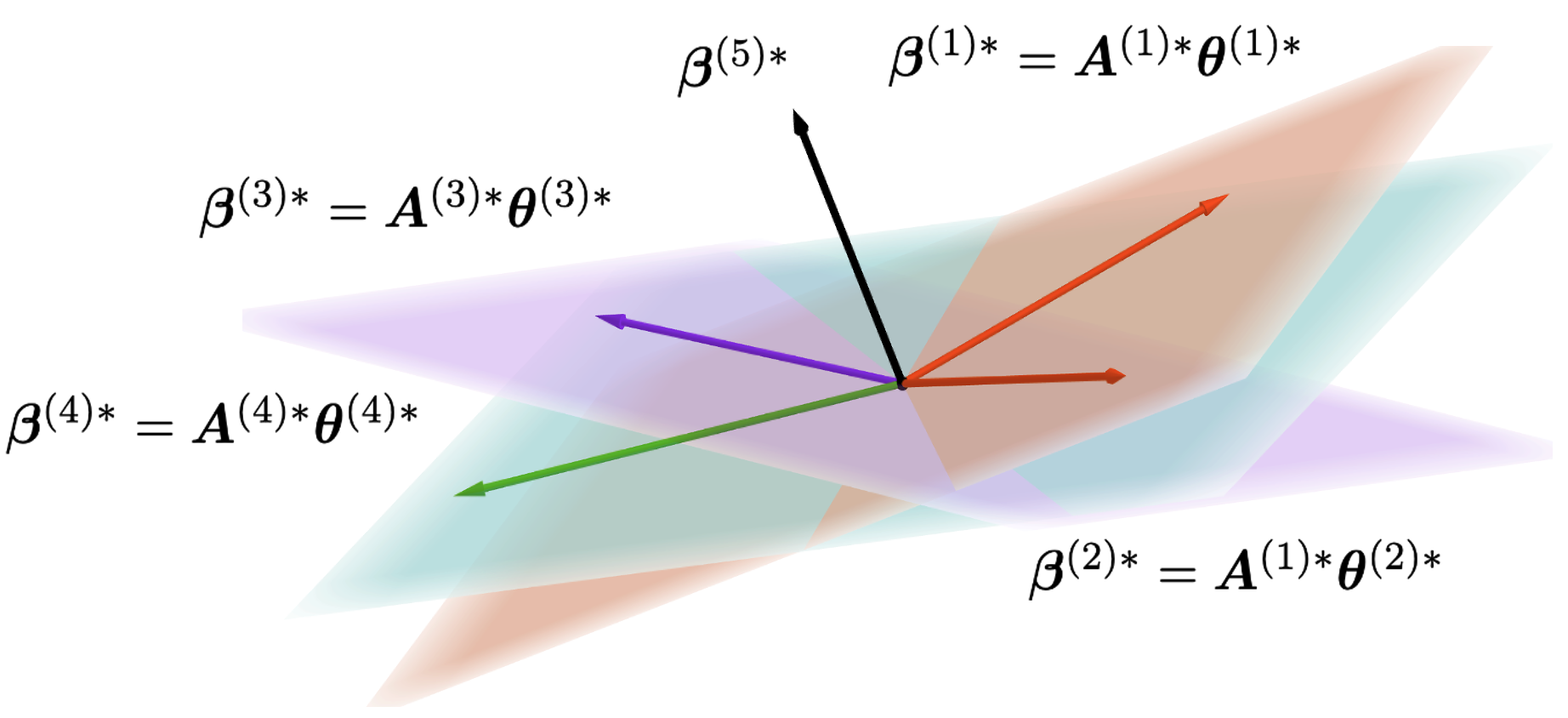}
         \caption{Similar representations with outliers (ours)\\\hspace{\textwidth}}
         \label{fig: diag-ours}
     \end{subfigure}
        \caption{A simple visualization of four different settings under the linear model \eqref{eq: linear model intro}.}
        \label{fig: visualization}
\end{figure}

To help readers better understand the difference between some settings in literature with our setting under the linear model \eqref{eq: linear model intro}, we drew Figure \ref{fig: visualization} as a simple visualization corresponding to the case where $p = 3$ and $r = 2$.

\subsubsection{Beyond the Assumption of the Same Representation}\label{subsubsec: beyong same repre}
Several works have studied similar problems to the current work, but in different formulations. \cite{chua2021fine} explored linear model \eqref{eq: linear model intro} but with the assumption that $\bAks{t} = \oA + \bm{\Delta}^{(t)*}$ with some $\oA \in \mathbb{R}^{p \times r}$ and $\fnorm{\bm{\Delta}^{(t)*}} \leq h''$ and $S = [T]$. \cite{duan2023adaptive} considered the same model with $\bbetaks{t} = \oA\barthetaks{t} + \bm{\delta}^{(t)*}$ with $\twonorm{\bm{\delta}^{(t)*}} \leq h'$ and $S = [T]$. The following theorem establishes the equivalence between these alternative formulations and our proposed setting.

\begin{theorem}\label{thm: equivalence of settings}
	Consider the following three settings: \footnote{$\oA \in \mO$ in the three settings are the same.}
	\begin{enumerate}
		\item $\bbetaks{t} = \bAks{t}\bthetaks{t}$, where $\twonorm{\bAks{t}(\bAks{t})^\top -\oA(\oA)^\top} \leq h$ for some $\oA \in \mO$;
		\item $\bbetaks{t} = \oA\barthetaks{t} + \bm{\delta}^{(t)*}$, with $\twonorm{\bm{\delta}^{(t)*}} \leq h'$, $(\oA)^\top \bdeltaks{t} = \bm{0}$;
		\item $\bbetaks{t} = (\oA + \bm{\Delta}^{(t)*})\tildethetaks{t}$, with $\fnorm{\bm{\Delta}^{(t)*}} \leq h''$.
	\end{enumerate}
	They are equivalent in the following sense:
	\begin{enumerate}[(i)]
		\item If Setting 1 holds, then there exist $\barthetaks{t} \in \mathbb{R}^r$, $\bm{\delta}^{(t)*} \in \mathbb{R}^p$ with $\twonorm{\bdeltaks{t}} \leq h\twonorm{\bthetaks{t}}$ satisfying Setting 2, and there exist $\bm{\Delta}^{(t)*} \in \mathbb{R}^{p \times r}$, $\tildethetaks{t} \in \mathbb{R}^p$ with $\fnorm{\bm{\Delta}^{(t)*}} \leq \frac{h}{\sqrt{1-h^2}}$ satisfying Setting 3.
		\item If Setting 2 holds, then there exists $\bAks{t} \in \mO$ satisfying Setting 1 with $\twonorm{\bAks{t}(\bAks{t})^\top -\oA(\oA)^\top} \leq (10\sqrt{2} + 4)\frac{h'}{\twonorm{\bbetaks{t}}}$;
		\item If Setting 3 holds, then there exists $\bAks{t} \in \mO$ and $\bthetaks{t} \in \mathbb{R}^r$ satisfying Setting 1 with $\twonorm{\bAks{t}(\bAks{t})^\top -\oA(\oA)^\top} \leq (10\sqrt{2} + 4)\frac{h''}{1-h''}$ when $h'' \in [0, 1)$.
	\end{enumerate}
\end{theorem}

It is worth emphasizing that, despite the equivalence of the three settings, both \cite{chua2021fine} and \cite{duan2023adaptive} require $S = [T]$ and impose the constraints $\twonorm{\barthetaks{t}} \lesssim 1$ and $\twonorm{\tildethetaks{t}} \lesssim 1$ for all $t \in [T]$, respectively. In contrast, our setting allows for $S \neq [T]$ and for $\twonorm{\bthetaks{t}}$ to diverge. Moreover, when $S = [T]$ and $h'=h'' = 0$, all three settings reduce to the shared representation case studied in \cite{du2020few, thekumparampil2021statistically, tripuraneni2021provable}. \footnote{\cite{du2020few} also studied the non-linear representations; here we are referring to their setting of linear representations.} 

Thus, our work addresses a more general scenario compared to these existing formulations. Despite the equivalence established in Theorem \ref{thm: equivalence of settings}, we focus on the setting defined in Section \ref{subsec: RMTL}, as this formulation naturally facilitates the development of algorithms that adapt to the unknown similarity between tasks and enjoy robustness against adversarial contamination. More concretely, compared to the settings in \cite{chua2021fine} and \cite{duan2023adaptive}, our formulation leads to simpler learning algorithms, which can adapt to the unknown similarity level $h$ (unlike \citealp{chua2021fine}, where the algorithm requires tuning parameters depending on the similarity level $h$), with stronger theoretical results. Our formulation also inspires algorithms that remain robust to a small fraction of outlier tasks, a merit not shared by the methods following the formulations in \cite{chua2021fine} and \cite{duan2023adaptive}.

Moreover, our framework is easier to generalize to an unsupervised learning setting (e.g., multiple linear/nonlinear latent factor models with similar factor loading matrices or similar factor score matrices).  Additionally, a representation-based reweighting strategy was proposed in \cite{chen2021weighted}, which is motivated by the concern of assuming the same representation. Their similarity metric between tasks depends on the weight assigned to the objective function of each task, while our similarity metric depends on the difference between representations explicitly, which is more intuitive. Moreover, their approach can suffer from a negative transfer in the worst case, while our approaches do not. Furthermore, none of  \cite{chen2021weighted, chua2021fine, duan2023adaptive} considered the presence of outlier tasks.



\begin{figure}[!h]
	\centering
	\includegraphics[width=1\textwidth]{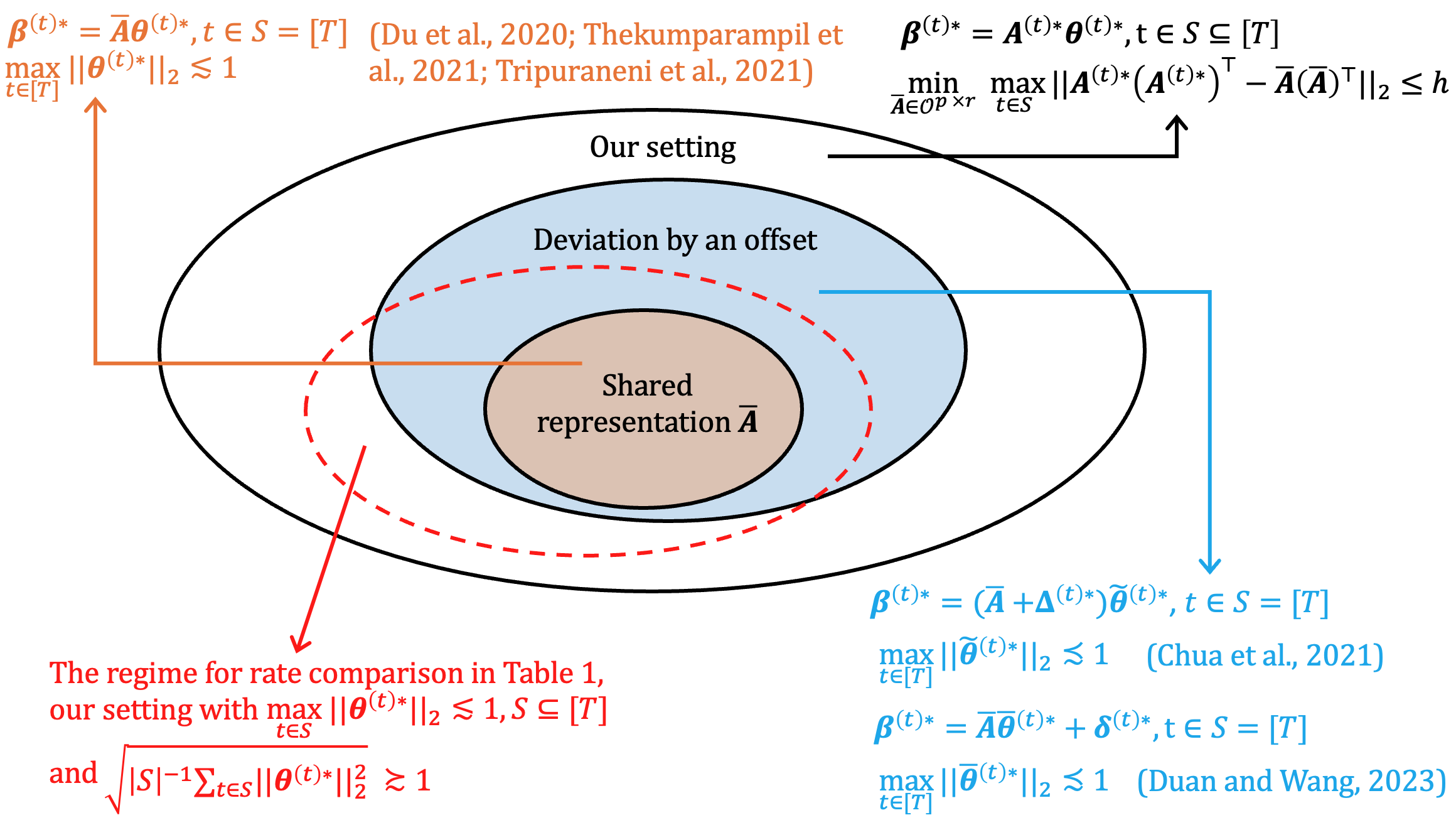}
	\caption{The diagram illustrating the relationship between different regimes studied in different papers.}
	\label{fig: rate-comparison}
\end{figure}

Finally, we present a diagram in Figure \ref{fig: rate-comparison} to summarize the relationship between the different regimes studied in various papers mentioned earlier. To compare algorithms across different settings more clearly, we focus on a specific regime, enclosed by a dashed line in Figure \ref{fig: rate-comparison}. This regime allows for simpler, more intuitive, and explicit results, making it easier to compare the estimation errors of $\bbetaks{t}$ across different approaches. In this regime, defined for rate comparison, we have our setting introduced in Section \ref{subsec: RMTL} with $\min_{\oA \in \mO}\max_{t \in S}\twonorm{\bAks{t}(\bAks{t})^\top - \oA(\oA)^\top} \leq h$ and $\max_{t \in S}\twonorm{\bthetaks{t}} \lesssim 1$, and $\sqrt{|S|^{-1}\sum_{t\in S}\twonorm{\bthetaks{t}}^2} \gtrsim 1$. We summarize the estimation errors of different approaches in Table \ref{table: rate summary}, highlighting our two proposed algorithms: penalized ERM (``pERM") and the spectral method (``Spectral"). We show that the spectral method is minimax optimal with computational efficiency when there is no contamination ($\epsilon = 0$), and pERM can handle the contaminated case more effectively.

\begin{table}[ht]
    \centering
    \begin{adjustbox}{max width=1\textwidth}
    {\renewcommand{\arraystretch}{1.5} 
    \begin{tabular}{ccccc}
        \toprule
        \textbf{Regime} &\textbf{Algorithm} & $\max_{t \in S}\twonorm{\hbetak{t} - \bbetaks{t}}$ & \textbf{Optimal ($\epsilon = 0$)?} & \textbf{Poly-time?} \\
        \midrule
        \multirow{3}{*}{$h =\epsilon = 0$} & ERM & $r\sqrt{\frac{p}{nT}} + r\sqrt{\frac{1}{n}}$ & No & No \\ \cline{2-5}
        & MoM  & $r\sqrt{\frac{p}{nT}} + \sqrt{\frac{r}{n}}$ &No & Yes\\ \cline{2-5}
        & AltMinGD & $r\sqrt{\frac{p}{nT}} + \sqrt{\frac{r}{n}}$ & No & Yes \\ \hline
        \multirow{2}{*}{\begin{tabular}[l]{@{}l@{}}$h\neq 0$ \\ $\epsilon = 0$\end{tabular}} &AdaptRep & $\Big[r\sqrt{\frac{p}{nT}} + r\sqrt{\frac{1}{n}} + \sqrt{rh}(\frac{p}{n})^{1/4}\Big]\wedge \sqrt{\frac{p}{n}}$ & No & No \\ \cline{2-5}
        &ARMUL & $\Big(r\sqrt{\frac{p}{nT}} + r\sqrt{\frac{1}{n}} + rh\Big)\wedge \Big(r\sqrt{\frac{p}{n}}\Big)$ & No & No \\ \hline
         \multirow{3}{*}{$h,\epsilon \neq 0$} & \cellcolor{green!20}pERM & \cellcolor{green!20}$\Big(r\sqrt{\frac{p}{nT}} + r\sqrt{\frac{1}{n}} + \sqrt{r}h + \epsilon \frac{r^{3/2}\sqrt{p}}{\sqrt{n}}\Big)\wedge \sqrt{\frac{p}{n}}$  & \cellcolor{green!20}No & \cellcolor{green!20}No \\ \hhline{|~----|}
         & \cellcolor{cyan!20}Spectral & \cellcolor{cyan!20} $\Big(\sqrt{\frac{pr}{nT}} + \sqrt{\frac{r}{n}} + h + \sqrt{\epsilon r}\Big)\wedge \sqrt{\frac{p}{n}}$ &\cellcolor{cyan!20} Yes & \cellcolor{cyan!20}Yes \\ \cline{2-5}
         & Single-task &$\sqrt{\frac{p}{n}}$ & No & Yes \\ \hline
        \multicolumn{2}{c}{Lower-bound} & $\Big(\sqrt{\frac{pr}{nT}} + \sqrt{\frac{r}{n}} + h + \epsilon \frac{r}{\sqrt{n}}\Big)\wedge \sqrt{\frac{p}{n}}$ & -- & -- \\
        \bottomrule
    \end{tabular}
    }
    \end{adjustbox}
    \caption{A summary of the high-probability estimation error of $\bbetaks{t}$'s for different methods in different regimes, where the two algorithms we proposed, penalized ERM (``pERM", Algorithm \ref{algo: mtl}) and the spectral method (``Spectral", Algorithm \ref{algo: spectral}), are highlighted. Denote the contamination proportion $\epsilon = |S^c|/T$. Except for the lower bound, all the other rates are upper bounds. All the rates are up to logarithmic factors. ``ERM": \cite{du2020few, tripuraneni2021provable}. ``MoM": method-of-moments, \cite{tripuraneni2021provable}. ``AltMinGD": alternating minimization gradient descent, \cite{thekumparampil2021statistically}. ``AdaptRep": adaptive representation learning, \cite{chua2021fine}. ``ARMUL": adaptive and robust multi-task learning, \cite{duan2023adaptive}. ``Single-task": single-task regression by only using the local data from each task.}
    \label{table: rate summary}
\end{table}

\subsection{Our Contributions}
Our contributions can be summarized below. 
\begin{enumerate}[(i)]
	\item Compared to most literature on representation MTL, we considered a more general framework. Here, the linear representations can vary across tasks, signal strengths may differ between tasks, and there can be a small fraction of unknown outlier tasks. 
	\item We proposed two algorithms, the penalized ERM and the spectral method, to learn the regression coefficients and the representations from multiple tasks. Our algorithms were shown to have the following properties:
		\begin{itemize}
			\item They outperform single-task learning when the representations of different tasks are sufficiently similar, and the proportion of outlier tasks is low.
			\item They guarantee no worse performance than single-task learning (safe-net guarantee), even when task representations are dissimilar.
		\end{itemize}
	\item We thoroughly analyzed the relationship between different regimes of representation MTL studied in the literature. Our derived upper bounds improve over existing rates, particularly in scenarios without task contamination. Furthermore, in the context of representation MTL, we are the first to examine the scenario where a small proportion of the tasks is contaminated.
	\item We derived the lower bounds for model \eqref{eq: linear model intro}. To our knowledge, these are the first lower bound results for regression coefficient estimation under the representation MTL. Prior works such as \cite{duchi2022subspace} and \cite{tripuraneni2021provable} provided analogous lower bounds for the subspace recovery problem, assuming identical representations without outlier tasks. In a TL setup, \cite{chua2021fine} showed that assuming the same representation can lead to worse performance than target-only learning when source representations differ from each other, but they did not provide a full lower bound that relates to the representation difference. Comparing the upper and lower bounds, we demonstrated that both proposed algorithms are nearly minimax optimal, with the spectral method being optimal in uncontaminated settings.
	\item We extended our analysis from linear model \eqref{eq: linear model intro} to generalized linear models (GLMs) and non-linear regression models, and obtained similar theoretical guarantees in these settings.
	\item We proposed a thresholding algorithm based on singular value decomposition (SVD) to estimate the unknown intrinsic dimension $r$. This adaptation enables our penalized ERM and spectral methods to handle cases where $r$ is unknown, addressing a common challenge where $r$ is not a priori known in most prior works.
\end{enumerate}

\subsection{Notations and Organization}
Throughout the paper, we use bold capitalized and lower-case letters to denote matrices and vectors, respectively. For a real number $a$, $|a|$ stands for its absolute value. For a vector $\bm{u}$, $\twonorm{\bm{u}}$ stands for its Euclidean norm. For a matrix $\bA$, $\twonorm{\bA}$ and $\fnorm{\bA}$ represent its spectral and Frobenius norm, respectively. $\bA^\top$ denotes its transpose. $\sigma_j(\bA)$, $\sigmamax(\bA)$, $\sigmamin(\bA)$ are its $j$-th largest singular value, maximum singular value, and minimum (non-zero) singular value, respectively. When $\bA$ is a square matrix, we denote its maximum and minimum eigenvalues as $\lambdamax(\bA)$ and $\lambdamin(\bA)$, respectively. For a function $\psi: \mathcal{X} \rightarrow \mathbb{R}$, $\infnorm{\psi}$ is defined to be $\max_{x \in \mathcal{X}}|\psi(x)|$. For two real numbers $a$ and $b$, we denote their minimum by $\min\{a,b\}$ or $a \wedge b$ and their maximum by $\max\{a, b\}$ or $a \vee b$, respectively. For two positive real sequences $\{a_n\}_{n=1}^{\infty}$ and $\{b_n\}_{n=1}^{\infty}$, $a_n \lesssim b_n$ or $b_n \gtrsim a_n$ means there exists a universal constant $C > 0$ such that $a_n \leq Cb_n$ for all $n$, and $a_n \ll b_n$ or $b_n \gg a_n$ means that $a_n/b_n \rightarrow 0$ as $n \rightarrow \infty$. $a_n \asymp b_n$ means $a_n \lesssim b_n$ and $a_n \gtrsim b_n$ hold simultaneously. Sometimes, we abbreviate ``with probability'' as ``w.p.'' and ``with respect to'' as ``w.r.t.''. For any $N \in \mathbb{N}_+$, $[N]$ and $1:N$ are defined to be $\{1,\ldots, N\}$. $\tP$ and $\tE$ are the probability measure and expectation taken over all randomness. We use $C, C', C'', C_1, C_2, C_3, c, c'$ to represent universal constants that could change from place to place.

The rest of this paper is organized as follows. In Section \ref{sec: mtl}, we first propose a penalized ERM algorithm for the linear model \eqref{eq: linear model intro}, establish the upper bound of estimation error for global minimizers of the ERM, study properties of the local minimizers, and discuss implementation details. Subsequently, we introduce a novel spectral method that is computationally more efficient and achieves sharper estimation error upper bounds than the penalized ERM when there is no contamination. Next, we present lower bound results for the representation MTL problem and conclude Section \ref{sec: mtl} with a brief discussion on extensions to generalized linear models (GLMs) and non-linear regression models. In Section \ref{sec: unknown r}, we propose a thresholding algorithm to estimate the intrinsic dimension $r$ and adapt our penalized ERM algorithm and the spectral method to the case where $r$ is unknown. We conduct extensive simulation studies and analyze a real-world dataset to demonstrate our theoretical findings in Section \ref{sec: numerical}. Finally, we summarize our contributions and outline a few potential avenues for future research in Section \ref{sec: discussions}. 

Due to space constraints, certain results are deferred to the appendix. In Section \ref{sec: extensions} of the appendix, we provide details on the extension of our methods and theory to GLMs and non-linear regression models. In Section \ref{sec: tl} of the appendix, we study how to transfer the knowledge to an unknown task, i.e., under the setting of transfer learning (TL) or learning-to-learn. We propose an algorithm that leverages outputs from MTL algorithms to adapt to a new target task, presenting corresponding upper and lower bounds. All the proofs are also provided in the appendix.

\section{Multi-task Learning with Similar Representations}\label{sec: mtl}

\subsection{Problem Set-up}\label{subsec: setup}

Let us describe the problem setting introduced in Section \ref{subsec: RMTL} in more detail. Suppose there are $T$ tasks, and we have collected sample $\{\bxk{t}_i, \yk{t}_i\}_{i=1}^n$ from the $t$-th task, where $\bxk{t}_i \in \mathbb{R}^p$, $\yk{t}_i \in \mathbb{R}$, and $t \in [T]$. There exists an unknown subset $S \subseteq [T]$, such that for all $t \in S$, 
\begin{equation}\label{eq: linear model}
	\yk{t}_i = (\bxk{t}_i)^\top\bbetaks{t} + \epsilonk{t}_i, \quad i = 1:n,
\end{equation}
where $\bbetaks{t}=\bAks{t}\bthetaks{t}$, $\bAks{t} \in \mO = \{\bA \in \mathbb{R}^{p \times r}: \bA^\top \bA = \bm{I}_r\}$, low-dimensional parameter $\bthetaks{t} \in \mathbb{R}^r$, $r \leq p$, and $\{\epsilonk{t}_i\}_{i=1}^n$ are i.i.d. zero-mean sub-Gaussian variables independent of $\{\bxk{t}_i\}_{i=1}^n$ \footnote{This is assumed for simplicity and can be relaxed. In fact, it suffices to require $\{\epsilonk{t}_i|\bxk{t}_i = \bx_i\}_{i=1}^n$ to be independent zero-mean sub-Gaussian variables for almost surely $\{\bx_i\}_{i=1}^n$ w.r.t. the product probability measure induced by the distribution of $\bxk{t}_i$'s.}. Throughout this and the next sections, we assume the intrinsic dimension $r$ is known. The case that $r$ is unknown will be addressed in Section \ref{sec: unknown r}.

Here the $T$ tasks are divided into two groups, $S$ and $S^c$. The tasks in $S$ have ``similar'' representations (similarity to be defined in the following), while the tasks in $S^c$ can be understood as \textit{outlier} tasks or \textit{contaminated} tasks with \textit{an arbitrary distribution}. Our goal is twofold: 
\begin{enumerate}
	\item Improve the learning performance simultaneously on the tasks in $S$, when they share ``similar'' representations and the proportion of outlier tasks in $S^c$ among all $T$ tasks is small;
	\item Maintain the single-task learning performance when the ``similarity'' between tasks in $S$ is low.
\end{enumerate}
It should be emphasized that if we allow the outlier tasks in $S^c$ to be arbitrarily distributed, no guarantee can be obtained for these tasks in the worst case. However, as we will discuss later, if these tasks still follow linear models \eqref{eq: linear model} (without a low-dimensional representation), then the single-task linear regression estimation rate can be achieved for $\{\bbetaks{t}\}_{t \in S^c}$. 

\begin{figure}[!h]
	\centering
	\includegraphics[width=\textwidth]{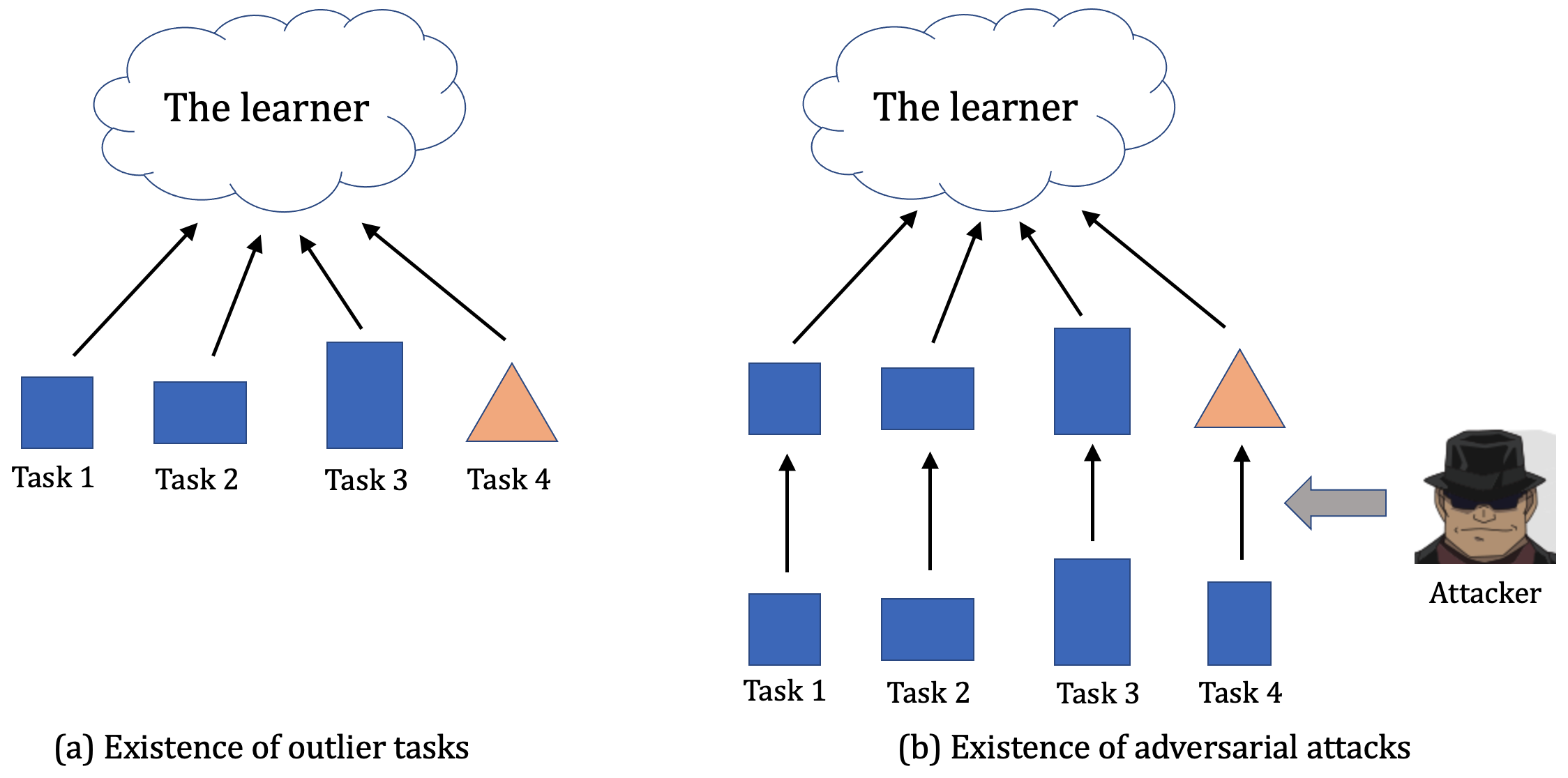}
	\caption{An illustration for two perspectives of viewing model \eqref{eq: linear model}. In both cases, $S^c = \{4\}$.}
	\label{fig: outlier-tasks}
\end{figure}

We also want to point out that the set $S$ is unknown. We will show that our penalized ERM algorithm and spectral method can perform well across all potential sets $S \subseteq [T]$ under certain conditions. This flexibility is crucial from both perspectives of outlier tasks and adversarial attacks. From the perspective of outlier tasks, we expect an algorithm to succeed for all possible outlier task index sets $S^c$ as long as $|S^c|/T$ is small. In other words, the algorithm should not only work for a specific $S^c$, but also not rely on task indices (otherwise, we can always drop the data from tasks in $S^c$ to avoid the impact of outliers). From the perspective of adversarial attacks, the attacker can choose to corrupt the data from any task, which usually happens after the release of the machine learning system. Therefore, a robust learner should achieve ideal performance for all possible sets $S$. See Figure \ref{fig: outlier-tasks} for an illustration of these two points of view.

The same setting when $\bAks{t}$'s are the same and $S = [T]$ (i.e. no outlier tasks) has been studied in \cite{du2020few}, where they argued that when $r$ is much smaller than $p$, a better estimation error rate of $\bbetaks{t}$ can be achieved compared to the single-task learning. Our framework is more general and realistic because it is difficult for all tasks to be embedded in precisely the same subspace as the number of tasks $T$ grows \citep{rostami2022transfer}, and the prevalence of outlier tasks is common \citep{zhang2018overview}.

To mathematically quantify the similarity between representations $\{\bAks{t}\}_{t \in S}$, we consider the maximum principal angle between subspaces spanned by the columns of these representation matrices. More specifically, we assume that there exists $h \in [0, 1]$ such that
\begin{equation}\label{eq: A similarity}
	\min_{\oA \in \mO}\max_{t \in S}\twonorm{\bAks{t}(\bAks{t})^\top - \oA  (\oA)^\top} \leq h. \quad \footnote{Note that the LHS of the following inequality is always less than or equal to 1, because $\bAks{t}$ and $\oA$ are orthonormal matrices. Here, the minimum is achievable because of the Weierstrass Theorem. Note that the objective function is continuous (over $\oA$) under $\twonorm{\cdot}$ norm and the space $(\mO, \twonorm{\cdot})$ is compact.}
\end{equation}
A small $h$ means the representations are more similar. The case when $h = 0$ reduces to the setting of the same representations in literature \citep{du2020few, tripuraneni2021provable}. In the literature, $\twonorm{\bAks{t}(\bAks{t})^\top - \oA  (\oA)^\top}$ is often referred to as the \textit{maximum principal angle} between column spaces of $\bAks{t}$ and $\oA$. This concept and its variations have been widely used to measure the difference between subspaces in perturbation theory (e.g., \citealp{wedin1972perturbation, cai2013sparse, kato2013perturbation, yu2015useful, chen2021spectral}). The case when $r = 1$ (where all representations are $p \times 1$ vectors and $\bthetaks{t}$'s are scalars) reduces to the setting of \cite{gu2024robust}. In this case, the principal angle between subspaces becomes the angle between regression coefficient vectors. 

We now make some assumptions. Without loss of generality, suppose $\bxk{t}$ is mean-zero. Denote the covariance matrix $\bSigmak{t} = \tE [\bxk{t}(\bxk{t})^\top]$ and the joint distribution of $\{\{\bxk{t}_i, \yk{t}_i\}_{i=1}^n\}_{t\in S^c}$ as $\mQ_{S^c}$. For the convenience of description, define a coefficient matrix $\bB_S^* \in \mathbb{R}^{p \times |S|}$, each column of which is a coefficient vector in $\{\bbetaks{t}\}_{t \in S}$. Denote the $\ell_2$-norm $\zetak{t} = \twonorm{\bthetaks{t}} = \twonorm{\bbetaks{t}}$, the average $\ell_2$-norm $\barzeta = \sqrt{|S|^{-1}\sum_{t \in S}(\zetak{t})^2}$, and assume $\min_{t \in S}\zetak{t} \gtrsim \sqrt{\frac{p+\log T}{n}}$. \footnote{In general, we have the same results hold by defining $\zetak{t} = \twonorm{\bthetaks{t}}\vee \sqrt{\frac{p+\log T}{n}}$. Here we define $\zetak{t} = \twonorm{\bthetaks{t}}$ and assume $\min_{t \in S}\zetak{t} \gtrsim \sqrt{\frac{p+\log T}{n}}$ for presentation simplicity.} $\zetak{t}$ can be viewed as the signal strength of the $t$-th task. In almost all the existing literature, it is assumed that $\zetak{t} \lesssim 1$ for all $t \in S = [T]$ and $\barzeta \gtrsim 1$, and the impact of $\zetak{t}$ on the model performance is ignored. In this work, however, we allow $\zetak{t}$'s to vary across tasks, and we will show later how the performance of representation MTL on each task depends on $\zetak{t}$.

\begin{assumption}\label{asmp: x}
	For any $t \in S$, $\bxk{t}$ is sub-Gaussian in the sense that for any $\bm{u} \in \mathbb{R}^p$ and $\lambda \in \mathbb{R}$, $\tE[e^{\lambda \bu^\top\bxk{t}}] \leq e^{C\lambda^2\twonorm{\bm{u}}^2}$ with some constant $C > 0$. And there exist constants $c, C$ such that $0<c \leq \lambda_{\min}(\bSigmak{t}) \leq \lambda_{\max}(\bSigmak{t}) \leq C < \infty$, for all $t \in S$.
\end{assumption}

\begin{assumption}\label{asmp: theta}
	There exists a constant $c > 0$ such that $\sigma_r(\bB_S^*/\allowbreak \sqrt{|S|}) \geq \frac{c}{\sqrt{r}}\barzeta$, where each column of $\bB_S^*$ is a coefficient vector in $\{\bbetaks{t}\}_{t \in S}$.
\end{assumption}

\begin{assumption}\label{asmp: n}
	$n \geq C(p+\log T)$ with a sufficiently large constant $C > 0$.
\end{assumption}

Assumptions \ref{asmp: x} and \ref{asmp: n} are standard conditions in literature \citep{du2020few, duan2023adaptive}. Assumption \ref{asmp: theta} is often called the task diversity condition. When $\bAks{t}$'s are the same, $\sigma_r(\bB_S^*/\sqrt{|S|}) = \sigma_r(\bm{\Theta}^*_S/\sqrt{|S|})$, where each column of $\bm{\Theta}^*_S$ is a coefficient vector in $\{\bthetaks{t}\}_{t \in S}$, which means that the low-dimensional task-specific parameters $\bthetaks{t}$'s are diverse. Such a task diversity condition has been adopted in other related studies \citep{du2020few, chua2021fine, tripuraneni2021provable, duchi2022subspace} \footnote{In \cite{duchi2022subspace} and \cite{tripuraneni2021provable}, the lower bound of $\sigma_r(\bB_S^*/\sqrt{|S|})$ is defined as a parameter and appears in the estimation error. Here we follow \cite{du2020few} and impose an explicit bound on it to obtain a cleaner result, but our analysis can carry over to the analysis where the lower bound of $\sigma_r(\bB_S^*/\sqrt{|S|})$ is denoted as a parameter.}, to obtain a parametric rate which is faster than the rate without this condition \citep{maurer2016benefit}. The benefit of this condition is intuitive, because a full exploration of all directions in the subspace is necessary to learn representations well, which is the key to representation MTL. 

It is important to note that the presented assumptions are imposed on both the set of non-outlier tasks $S$ and the model parameters. These assumptions are made from the perspective of outlier tasks. However, when considering adversarial attacks, we can replace $S$ with the set of all tasks $[T]$ and assume that each task follows the linear model \eqref{eq: linear model}. An attacker can then adversarially select a subset $S^c \subseteq [T]$ and distort the data distribution for these tasks. For simplicity, we do not distinguish between these two perspectives in the following parts of this paper.

In the following subsections, we will present two algorithms, study their properties, and derive a lower bound of the estimation error.

\subsection{The First Algorithm: Penalized ERM}

\subsubsection{Algorithm and Upper Bounds}\label{subsubsec: alg and upper bounds}

In a special case of our setting, when $\bAks{t}$'s are the same, $S = [T]$, $\max_{t \in S}\zetak{t} \lesssim 1$, and $\barzeta \gtrsim 1$, \cite{du2020few} proposed an algorithm by combining the objective functions of all tasks and solving the optimization problem. A more general version accommodating various loss functions has been explored in \cite{tripuraneni2020theory}. Under our setting, where the representations are similar but not exactly the same, and with the inclusion of potential outlier tasks, the objective function needs to be properly adjusted. When $h$ is large, learning $\{\bbetaks{t}\}_{t \in S}$ by presuming similar representations may lead to a negative transfer effect. Considering these differences, we proposed a two-step learning approach in Algorithm \ref{algo: mtl} that addresses these issues. Note that we will apply the same algorithm to some extended models in Section \ref{subsec: extension main text}, so for description convenience, we introduce the algorithm with generic loss functions $\fk{t}$ for the $t$-th task. For the linear model \eqref{eq: linear model}, specifically, we set $\fk{t}(\bbeta) = \frac{1}{2n}\twonorm{\bYk{t} - \bXk{t}\bbeta} = \frac{1}{2n}\sum_{i=1}^n [\yk{t}_i - (\bxk{t}_i)^\top\bbeta]^2$ for $\bbeta \in \mathbb{R}^p$, where $\bXk{t} \in \mathbb{R}^{n \times p}$ and $\bYk{t} \in \mathbb{R}^n$ are corresponding matrix representations of the data from the $t$-th task.

\begin{algorithm}[!h]
\caption{Penalized ERM}
\label{algo: mtl}
\KwIn{Data $\{\bXk{t}, \bYk{t}\}_{t=1}^T = \{\{\bxk{t}_i, \yk{t}_i\}_{i=1}^n\}_{t=1}^T$, penalty parameters $\lambda$ and $\gamma$, intrinsic dimension $r$}
\KwOut{Estimators $\{\hbetak{t}\}_{t=1}^T, \hbarA$}
\underline{Step 1:} (Aggregation) $\{\hAk{t}\}_{t=1}^T$, $\{\hthetak{t}\}_{t=1}^T$, $\hbarA \in$ $ \argmin_{\{\bAk{t}\}_{t=1}^T\subseteq \mO, \barA \in \mO, \{\bthetak{t}\}_{t=1}^T \subseteq \mathbb{R}^r} \big\{\frac{1}{T}\sum_{t=1}^T \big[\fk{t}(\bAk{t}\bthetak{t})$ \hspace*{8.5cm} $+ \frac{\lambda}{\sqrt{n}} \twonorm{\bAk{t}(\bAk{t})^\top - \barA(\barA)^\top}\big]\big\}$ \\
\underline{Step 2:} (Biased regularization) $\hbetak{t} = \argmin_{\bbeta \in \mathbb{R}^p} \big\{\fk{t}(\bbeta) + \frac{\gamma}{\sqrt{n}}\twonorm{\bbeta - \hAk{t}\hthetak{t}}\big\}$ for $t \in [T]$
\end{algorithm}

In Algorithm \ref{algo: mtl}, Step 1 aims to learn all tasks by aggregating the data, where the penalty $\twonorm{\bAk{t}(\bAk{t})^\top - \barA(\barA)^\top}$ is added to force the subspaces represented by $\{\hAk{t}\}_{t=1}^T$ to be similar. This penalty is motivated by the connection between penalized over-parameterized models and robustified empirical risk minimization (ERM) (e.g., \citealp{gannaz2007robust, she2011outlier, donoho2016high, duan2023adaptive}). Specifically, adding such a penalty is equivalent to employing a robustified loss function to estimate the central subspace $\oA$, which is then adapted to obtain individual estimators for each $\bAks{t}$. As discussed in Section \ref{subsubsec: beyong same repre}, formulating the problem in our setting naturally leads to this penalty structure, which provides an advantage over existing approaches such as those in \cite{chua2021fine} and \cite{duan2023adaptive}. Moreover, note that when $\twonorm{\bAk{t}(\bAk{t})^\top - \barA(\barA)^\top} = 0$, it can be shown that $\bAk{t}$ and $\oA$ are identical up to a rotation. And this distance is equivalent to other subspace distances like $\sin$-$\Theta$ distance or the distance between two matrices up to a rotation \citep{chen2021spectral}. Step 2 uses data from each task to make proper corrections to prevent negative transfer, which is often referred to as \textit{biased regularization} in the literature \citep{scholkopf2001generalized, kuzborskij2013stability, kuzborskij2017fast}. Such two-step methods are widely used in the distance-based MTL and TL literature to alleviate the adverse effect of negative transfer \citep{bastani2021predicting, li2022transfer, lin2022transfer, tian2022transfer}. It is important to point out that $\bAks{t}$'s and $\bthetaks{t}$'s are not uniquely identifiable in model \eqref{eq: linear model}, which does not pose an issue because our focus is on estimating $\bbetaks{t}$'s.

Next, we proceed to present the upper bound on estimation errors of $\{\bbetaks{t}\}_{t \in S}$ incurred by Algorithm \ref{algo: mtl}. 
\begin{theorem}[Upper bound for Algorithm \ref{algo: mtl}]\label{thm: mtl}
	Suppose Assumptions \ref{asmp: x}-\ref{asmp: n} hold with a subset $S \subseteq [T]$ satisfying $\epsilon = \frac{|S^c|}{T}\leq cr^{-3/2}\frac{\barzeta^2}{\max_{t \in S}(\zetak{t})^2}$, where $c > 0$ is a small constant. By setting $\lambda = C\frac{\max_{t \in S}(\zetak{t})^2}{\min_{t \in S}\zetak{t}}\sqrt{r(p+\log T)}$ and $\gamma = C'\sqrt{p+\log T}$ with sufficiently large positive constants $C$ and $C'$, for an arbitrary distribution $\mQ_{S^c}$ of $\{\{\bxk{t}_i, \yk{t}_i\}_{i=1}^n\}_{t\in S^c}$, w.p. at least $1-e^{-C''(r+\log T)}$, we have
	\begin{align}
		\twonorm{\hbetak{t}-\bbetaks{t}} &\lesssim \Bigg\{\frac{\zetak{t}}{\barzeta}r\sqrt{\frac{p}{nT}} + \bigg(\frac{\zetak{t}\sqrt{r}}{\barzeta}\vee 1\bigg) \sqrt{\frac{r+\log T}{n}} + \zetak{t}\sqrt{r}h \\
		&\quad\quad + \frac{\zetak{t}\max_{t \in S}(\zetak{t})^2}{\barzeta^2\min_{t \in S}\zetak{t}}r^{3/2}\sqrt{\frac{p+\log T}{n}}\epsilon\Bigg\} \wedge \sqrt{\frac{p + \log T}{n}}, \quad \forall t \in S.
	\end{align}
	Furthermore, if the data from tasks in $S^c$ satisfies the linear model \eqref{eq: linear model} (without any latent structure assumption) and Assumption \ref{asmp: x}, then w.p. at least $1-e^{-C'(p+\log T)}$, we also have
	\begin{equation}
		\max_{t \in S^c}\twonorm{\hbetak{t}-\bbetaks{t}} \lesssim \sqrt{\frac{p + \log T}{n}}.
	\end{equation}
\end{theorem}

In Theorem \ref{thm: mtl}, the upper bound of $\twonorm{\hbetak{t}-\bbetaks{t}}$ for $t \in S$ is the minimum of two terms, where the first term corresponds to the rate obtained through data aggregation and the second term corresponds to the single-task rate. This result shows that our algorithm is automatically \textbf{adaptive} to the optimal situation, whether or not aggregating data across tasks is beneficial. Furthermore, it demonstrates that Algorithm \ref{algo: mtl} is \textbf{robust} to a small fraction of outlier tasks, in the sense that representation MTL remains beneficial when the outlier proportion $\epsilon$ is small.

Analyzing the components of the upper bound of $\twonorm{\hbetak{t}-\bbetaks{t}}$ for $t \in S$ further elucidates their interpretability. The term $\frac{\zetak{t}}{\barzeta}r\sqrt{\frac{p}{nT}} + \zetak{t}\sqrt{r}h$ arises from learning similar representations, $\big(\frac{\zetak{t}\sqrt{r}}{\barzeta}\vee 1\big) \sqrt{\frac{r+\log T}{n}}$ is due to learning the representations and task-specific parameters, $\frac{\zetak{t}\max_{t \in S}(\zetak{t})^2}{\barzeta^2\min_{t \in S}\zetak{t}}r^{3/2}\sqrt{\frac{p+\log T}{n}} \epsilon$ accounts for outlier tasks, and $\sqrt{\frac{p + \log T}{n}}$ is the error rate of single-task learning.  Note that the term $\zetak{t}\sqrt{r}h$ does not explicitly depend on $n$ (although $\zetak{t}, r, h$ might depend on $n$). At first glance, this may appear overly restrictive, but the lower bound in Section \ref{subsec: lower bdd mtl} contains a similar term that does not involve $n$ explicitly. This phenomenon was also observed in many distance-based multi-task and transfer learning studies, such as \cite{li2022transfer, li2022transfer2, lin2022transfer, tian2022transfer, duan2023adaptive}.

To understand when Algorithm \ref{algo: mtl} improves upon single-task learning, consider the scenario $\max_{t \in S}\zetak{t} \lesssim 1$ and $\barzeta \gtrsim 1$, where the upper bound simplifies to $\big(r\sqrt{\frac{p}{nT}} + \sqrt{r}\sqrt{\frac{r+\log T}{n}} + \sqrt{r}h + r^{3/2}\sqrt{\frac{p+\log T}{n}}\epsilon\big) \wedge \sqrt{\frac{p+\log T}{n}}$. When $T \gg r^2$ (many tasks), $h \ll \sqrt{\frac{p+\log T}{nr}}$ (similar representations), $p \gg r(r\vee \log T)$ (low intrinsic dimension), and $\epsilon  \ll r^{-3/2}$ (a small fraction of outlier tasks), the rate is faster than the single-task error rate $\sqrt{\frac{p + \log T}{n}}$. 

Furthermore, contrary to the results in existing literature \citep{du2020few, chua2021fine, tripuraneni2021provable, thekumparampil2021statistically, duan2023adaptive}, our results indicate that the performance of representation MTL on each task depends critically on the norm $\zetak{t} = \twonorm{\bthetaks{t}}$. This insight suggests that tasks characterized by smaller coefficients may benefit more from representation MTL. We will verify this phenomenon through numerical experiments in Section \ref{subsubsec: sim theta}.

%

It is also worth noting that Assumption \ref{asmp: n} can be relaxed to $n \gtrsim r+\log T$ for penalized ERM, provided that we are willing to forego the safe-net guarantee $\sqrt{\frac{p+\log T}{n}}$. However, omitting the minimum with $\sqrt{\frac{p+\log T}{n}}$ in the estimation error can expose penalized ERM to negative transfer effects when either $h$ or $\epsilon$ is large.

Before closing this subsection, we present the following theorem, which demonstrates that the personalized estimators $\{\hAk{t}\hthetak{t}\}_{t = 1}^T$ obtained in Step 1 already achieve strong performance, albeit with a loss of a factor of $\sqrt{r}$ and other scaling factors compared to the estimation error of $\{\hbetak{t}\}_{t=1}^T$ in Theorem \ref{thm: mtl}. In this sense, Step 2 in Algorithm \ref{algo: mtl} can be seen as a refinement that further improves estimation accuracy. 

\begin{theorem}[Step 1 only] \label{thm: mtl step 1 only}
Under the same assumptions imposed in Theorem \ref{thm: mtl}, for any $S \subseteq [T]$ satisfying $\epsilon = \frac{|S^c|}{T}\leq cr^{-3/2}\frac{\barzeta^2}{\max_{t \in S}(\zetak{t})^2}$ with a small constant $c > 0$, and an arbitrary distribution $\mQ_{S^c}$ of $\{\{\bxk{t}_i, \yk{t}_i\}_{i=1}^n\}_{t\in S^c}$, w.p. at least $1-e^{-C'(r+\log T)}$, we have
	\begin{align}
		\twonorm{\hAk{t}\hthetak{t}-\bbetaks{t}} &\lesssim \Bigg\{\frac{\zetak{t}}{\barzeta}\sqrt{\frac{p}{nT}} + \bigg(\frac{\zetak{t}\sqrt{r}}{\barzeta}\vee 1\bigg) \sqrt{\frac{r+\log T}{n}} + \zetak{t}\sqrt{r}h   \\
		&\quad\quad  + \frac{\zetak{t}\max_{t \in S}(\zetak{t})^2}{\barzeta^2\min_{t \in S}\zetak{t}}\sqrt{\frac{p+\log T}{n}}\epsilon \Bigg\} \wedge \Bigg\{\sqrt{\frac{r(p + \log T)}{n}}\frac{\max_{t \in S}(\zetak{t})^2}{\zetak{t}\min_{t\in S}\zetak{t}}\Bigg\}, \\
		&\quad\quad\quad \forall t \in S.
	\end{align}
\end{theorem}

As pointed out in Section \ref{subsec: related works}, our setting reduces to the setting in \cite{gu2024robust} when $r = 1$, where they considered a ridge regression by penalizing the angle between different regression coefficients. The main idea of our Algorithm \ref{algo: mtl} resembles their approach, but they only considered the case when the angles between different regression coefficients are in $[0, \pi/2]$. Our result shows that representation MTL helps as long as the subspaces spanned by each coefficient (i.e., the straight line) are similar, where the angles can be either close to $0$ or $\pi$. Moreover, Theorem \ref{thm: mtl step 1 only} shows that when $r$ is a constant, the estimators obtained from Step 1 of Algorithm \ref{algo: mtl} already achieve a desired rate (despite an inflation term in the single-task rate $\sqrt{\frac{p+\log T}{n}}$), which means Step 2 may be omitted. 

\subsubsection{Discussions on Local Minimizers}\label{subsubsec: local minimizers}
It can be observed that the optimization problem in Step 1 of Algorithm \ref{algo: mtl} is non-convex. Therefore, in addition to studying the global minimizers as previously examined, it is also crucial to investigate the properties of local minimizers. Specifically, if only a local minimizer is found in Step 1, does Algorithm \ref{algo: mtl} still achieve the same upper bound in Theorem \ref{thm: mtl}?

Our first result below focuses on a special case where $S = [T]$ and $\lambda = +\infty$. In this scenario, Step 1 of Algorithm \ref{algo: mtl} reduces to the ERM algorithm introduced in \cite{du2020few}. The following theorem shows that even if we replace the global minimizer with a local minimizer in Step 1, Algorithm \ref{algo: mtl} still achieves the same upper bound as stated in Theorem \ref{thm: mtl}. In other words, there is no ``bad" local minimizer in this case.

\begin{theorem}[No bad local minimizer when {$S = [T]$} and $\lambda = +\infty$]\label{thm: no local easy}
	When $S = [T]$ and $\lambda = +\infty$, replacing the global minimizer in Step 1 of Algorithm \ref{algo: mtl} with any local minimizer \footnote{Here we say $\{\{\hAk{t}\}_{t=1}^T, \hoA, \{\hthetak{t}\}_{t=1}^T\}$ is a local minimizer of function $G(\{\bAk{t}\}_{t=1}^T, \oA, \{\bthetak{t}\}_{t =1}^T) = \sum_{t=1}^T \fk{t}(\bAk{t}\bthetak{t}) + \frac{\lambda}{\sqrt{n}} \twonorm{\bAk{t}(\bAk{t})^\top - \barA(\barA)^\top}$, if there exists a constant $\delta > 0$, such that for any $\{\{\bAk{t}\}_{t=1}^T, \bA, \{\bthetak{t}\}_{t=1}^T\}$ with $\max_{t \in [T]}\min_{\bm{R} \in \mathcal{O}^{p \times r}}\big\{\twonorm{\hAk{t}-\bAk{t}\bm{R}} + \twonorm{\hthetak{t} - \bm{R}^\top \bthetak{t}} \big\} + \min_{\bm{R} \in \mathcal{O}^{p \times r}}\twonorm{\hoA-\bm{A}\bm{R}} \leq \delta$, we must have $G(\{\hAk{t}\}_{t=1}^T, \hoA, \{\hthetak{t}\}_{t =1}^T) < G(\{\bAk{t}\}_{t=1}^T, \bA, \{\bthetak{t}\}_{t =1}^T)$. The same definition is used in Theorem \ref{thm: no local hard}.}  delivers the same upper bound as in Theorem \ref{thm: mtl}, i.e. \wprp, 
	\begin{equation}
		\twonorm{\hbetak{t} - \bbetaks{t}} \lesssim \Bigg\{\frac{\zetak{t}}{\barzeta}r\sqrt{\frac{p}{nT}} + \bigg(\frac{\zetak{t}\sqrt{r}}{\barzeta}\vee 1\bigg) \sqrt{\frac{r+\log T}{n}} + \zetak{t}\sqrt{r}h\Bigg\} \wedge \sqrt{\frac{p+\log T}{n}}, \,\,\forall t \in [T].
	\end{equation}
\end{theorem}

\cite{tripuraneni2021provable} proves a similar result for the ERM when $S = [T]$, $h = 0$, and $\lambda = +\infty$. However, their result only holds for ERM with an additional regularization term, and the optimization must be conducted within a constrained set. In contrast, Theorem \ref{thm: no local easy} shows that there is no bad local minimizer across the \textit{entire optimization landscape}. This result significantly supplements the existing literature for the case $S = [T]$.

Next, we present a more general result for the generic case $S \neq [T]$ with the same $\lambda$ value as in Theorem \ref{thm: mtl}. Unfortunately, similar to the result in \cite{tripuraneni2021provable}, in this general case, we can only establish that there is no bad local minimizer within a large regime instead of the entire landscape. This implies that we can replace the global minimizer in Step 1 with a local minimizer under some constraints to obtain the same upper bound.

\begin{theorem}[No bad local minimizer in a large regime, for general $S \neq {[T]}$]\label{thm: no local hard}
	Consider the same $\lambda$ value taken in Theorem \ref{thm: mtl}. Replacing the global minimizer in Step 1 of Algorithm \ref{algo: mtl} with any local minimizer satisfying
	\begin{align}
		4(1+\sqrt{2})\max_{t \in S}\twonorm{\hAk{t}(\hAk{t})^\top - \oA(\oA)^\top } + 4\frac{\epsilon}{1-\epsilon} &< 1-\gamma; \\
		\sqrt{2}\max_{t \in S}\bigg\{\frac{\sigmamax(\bSigmak{t})}{\sigmamin(\bSigmak{t})}\twonorm{\hAk{t}(\hAk{t})^\top - \bAks{t}(\bAks{t})^\top}\bigg\} &< 1-\gamma,
	\end{align}
	with a universal constant $\gamma > 0$ delivers the same upper bound as in Theorem \ref{thm: mtl}, i.e. \wprp, for all $t \in S$,
	\begin{align}
		\twonorm{\hbetak{t} - \bbetaks{t}} &\lesssim \Bigg\{\frac{\zetak{t}}{\barzeta}r\sqrt{\frac{p}{nT}} + \bigg(\frac{\zetak{t}\sqrt{r}}{\barzeta}\vee 1\bigg) \sqrt{\frac{r+\log T}{n}} + \zetak{t}\sqrt{r}h\\
		&\quad\quad + \frac{\zetak{t}\max_{t \in S}(\zetak{t})^2}{\barzeta^2\min_{t \in S}\zetak{t}}r^{3/2}\sqrt{\frac{p+\log T}{n}}\cdot \epsilon \Bigg\} \wedge \sqrt{\frac{p+\log T}{n}}.
	\end{align}
\end{theorem}

Theorem \ref{thm: no local hard} is similar to the results from other non-convex problems such as the EM algorithm for the mixture models (e.g., \citealp{balakrishnan2017statistical, cai2019chime}), where a contraction basin exists within which the only local minimizer is the global one. This suggests that a good initialization can be used in the optimization problem of Step 1 to avoid bad local minima. We conjecture a similar result in Theorem \ref{thm: no local easy} holds in the general case as well, i.e. there is no bad local minimizer in the entire optimization landscape. This conjecture remains a direction for future research.

The diagrams in Figure \ref{fig: opt-landscape} below summarize the results discussed in this subsection.

\begin{figure}[!h]
	\centering
	\includegraphics[width=\textwidth]{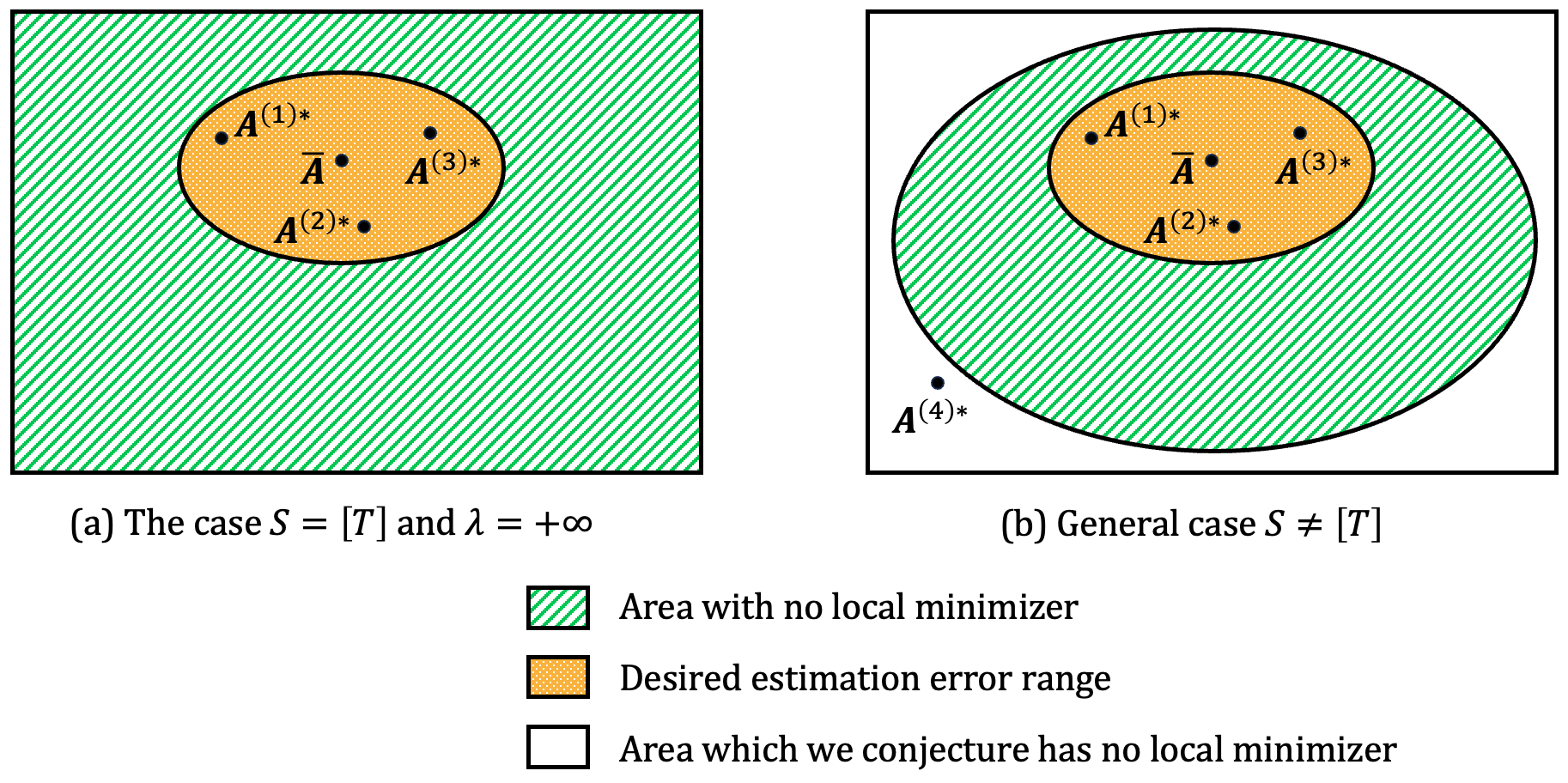}
	\caption{The optimization landscape of Algorithm \ref{algo: mtl} proved in Theorems \ref{thm: no local easy} and \ref{thm: no local hard}. The big rectangle is the full optimization landscape.}
	\label{fig: opt-landscape}
\end{figure}

\subsubsection{Implementation}
While we have provided desired upper bounds of estimation error with both local and global minimizers in Step 1 of Algorithm \ref{algo: mtl}, solving the optimization problem there is challenging due to the constraint that representation matrices $\{\hAk{t}\}_{t=1}^T$ and $\hoA$ must belong to the orthonormal space $\mO = \{\bA \in \mathbb{R}^{p \times r}: \bA^\top \bA = \bm{I}_r\}$. In this subsection, we want to point out that this restriction is adopted mainly for clarity of exposition. In practice, we can formulate the optimization problem in $\mathbb{R}^{p \times r}$ with a modified penalty term, achieving the same theoretical guarantees as with $\mO$.

Before discussing the practical formulation of the optimization problem in Step 1, we first introduce several new notations. For any non-zero matrix $\bA \in \mathbb{R}^{p \times r}$ with $p \geq r$, we define the projection matrix onto the column space of $\bA$ as $\bm{P}_{\bA} = \bm{U}\bm{U}^\top$, where $\bm{U} \in \mathcal{O}^{p \times r'}$ is the left singular matrix in the SVD of $\bA = \bm{U}_{p \times r'}\bm{\Lambda}_{r'\times r'}\bm{V}^\top_{r' \times r'}$,  $1 \leq r' \leq r$, and $\bm{\Lambda}_{r'\times r'}$ is diagonal with positive entries. When $\text{rank}(\bA) = r$, we have $\bP_{\bA} = \bA(\bA^\top \bA)^{-1}\bA^\top $. When $\bA$ is zero, we define $\bm{P}_{\bA} = \bm{0}_{p \times p}$. Note that while $\bm{U}$ in the SVD of $\bA$ may not be unique, $\bP_{\bA}$ is unique by the Hilbert projection theorem.

In practice, Step 1 of Algorithm \ref{algo: mtl}, i.e.
\begin{align}
	\{\hAk{t}\}_{t=1}^T, \{\hthetak{t}\}_{t=1}^T, \hbarA &\in  \argmin_{\{\bAk{t}\}_{t=1}^T, \barA \subseteq \mO, \{\bthetak{t}\}_{t=1}^T \subseteq \mathbb{R}^r} \Bigg\{\frac{1}{T}\sum_{t=1}^T \Bigg[\fk{t}(\bAk{t}\bthetak{t}) \\
	&\hspace{5.3cm}  + \frac{\lambda}{\sqrt{n}}\twonorm{\bAk{t}(\bAk{t})^\top - \barA(\barA)^\top}\Bigg]\Bigg\},
\end{align}
is equivalent to solving the following relaxed problem
\begin{equation}
	\{\hAk{t}\}_{t=1}^T, \{\hthetak{t}\}_{t=1}^T, \hbarA \in  \argmin_{\{\bAk{t}\}_{t=1}^T, \barA \subseteq \mathbb{R}^{p \times r}, \{\bthetak{t}\}_{t=1}^T \subseteq \mathbb{R}^r} \Bigg\{\frac{1}{T}\sum_{t=1}^T \Bigg[\fk{t}(\bAk{t}\bthetak{t})  + \frac{\lambda}{\sqrt{n}}\twonorm{\bm{P}_{\bAk{t}} - \bm{P}_{\barA}}\Bigg]\Bigg\}. \label{eq: replacement}
\end{equation}
This replacement relaxes the constraint that $\{\hAk{t}\}_{t=1}^T$ and $\hoA$ must belong to the orthonormal space $\mO = \{\bA \in \mathbb{R}^{p \times r}: \bA^\top \bA = \bm{I}_r\}$, allowing them to reside in $\mathbb{R}^{p\times r}$ with a modified penalty term. Intuitively, we can see why this relaxation preserves our previous theoretical guarantees. For any $\bAk{t} \in \mathbb{R}^{p\times r}$ and $\bthetak{t} \in \mathbb{R}^r$, it can be shown that there exist $\widetilde{\bA}^{(t)} \in \mO$ and $\widetilde{\btheta}^{(t)}\in \mathbb{R}^r$ such that $\bAk{t}\bthetak{t} = \widetilde{\bA}^{(t)}\widetilde{\btheta}^{(t)}$ and $\bP_{\bAk{t}} = \bP_{\widetilde{\bA}^{(t)}}$. Since we are interested in the product instead of $\bAk{t} \in \mathbb{R}^{p\times r}$ and $\bthetak{t} \in \mathbb{R}^r$ individually, the replacement does not affect the estimation error of $\bbetaks{t}$. The following theorem formally presents this result.

\begin{theorem}\label{thm: implementation}
	We replace the Step 1 of Algorithm \ref{algo: mtl} with the real-matrix optimization problem \eqref{eq: replacement}, and replace $\twonorm{\hAk{t}(\hAk{t})- \oA(\oA)^\top}$, $\twonorm{\hAk{t}(\hAk{t})- \bAks{t}(\bAks{t})^\top}$ in Theorem \ref{thm: no local hard} with $\twonorm{\bP_{\hAk{t}} - \bP_{\oA}}$ and $\twonorm{\bP_{\hAk{t}} - \bP_{\bAks{t}}}$. All theoretical results in Sections \ref{subsubsec: alg and upper bounds} and \ref{subsubsec: local minimizers} remain valid.
\end{theorem}

In all the numerical experiments, we used the automatic differentiation implemented in PyTorch \citep{paszke2019pytorch} along with the Adam optimizer \citep{kingma2015adam} to solve the optimization problem \eqref{eq: replacement}. Further implementation details will be discussed in Section \ref{sec: numerical}.

It is also possible to conduct optimization directly on the Stiefel manifold $\mO$ in Step 1, which is recently explored in \cite{chen2025geometric}.

\subsection{The Second Algorithm: Spectral Method}\label{subsec: spectral}
In this subsection, we propose another algorithm for the representation MTL, which is based on singular value decomposition (SVD), and we refer to this approach as the \emph{Spectral Method}.

The motivation of the spectral method arises from the special case $h = 0$ and $S = [T]$, implying $\bAks{t} \equiv \oA$ for all $t \in [T]$. In this scenario, $\bB^*_S = (\bbetaks{1}, \ldots, \bbetaks{T}) = \oA_{p \times r} \bTheta^*_{r \times T}$ with $\bTheta^* = (\bthetaks{1}, \ldots, \bthetaks{T})$ is a rank-$r$ matrix, and the column spaces of $\bB^*_S$ and $\oA$ are the same. Thus, the left singular matrix in the SVD of $\bB^*_S$ can exactly recover $\oA$. In practice, we can use the single-task estimator to estimate each column of $\bB^*_S$ and apply SVD to estimate $\oA$. With the estimated $\oA$, we can then perform a single-task regression to estimate $\bthetaks{t}$. When $h$ is not necessarily zero and $S \neq [T]$, the same idea can still be applied. However, to avoid negative transfer when $h$ is large, we can follow the SVD with a biased regularization step similar to Algorithm \ref{algo: mtl}. Additionally, we need to conduct SVD on a robust estimator of $\bB^*_S$ to accommodate outlier tasks. 

We use $\prod_R$ to denote the projection operator to an $\ell_2$-ball centered at zero of radius $R$ in $\mathbb{R}^p$, and use $\texttt{quantile}(\{a_t\}_{t=1}^T, 1-\bar{\epsilon})$ to denote the lower $(1-\bar{\epsilon})$-quantile of a sequence $\{a_t\}_{t=1}^T \subseteq \mathbb{R}$, where $\bar{\epsilon} \in [0, 1]$. We formalize the intuition into the spectral method in Algorithm \ref{algo: spectral}.

\begin{algorithm}[!h]
\caption{Spectral Method}
\label{algo: spectral}
\KwIn{Data $\{\bXk{t}, \bYk{t}\}_{t=1}^T = \{\{\bxk{t}_i, \yk{t}_i\}_{i=1}^n\}_{t=1}^T$, penalty parameter $\gamma$, an upper bound $\bar{\epsilon}$ (for $\epsilon$), intrinsic dimension $r$}
\KwOut{Estimators $\{\hbetak{t}\}_{t=1}^T, \hbarA$}
\underline{Step 1:} (Single-task regression) $\widetilde{\bbeta}^{(t)} = \argmin_{\bbeta \in \mathbb{R}^p}\big\{ \fk{t}(\bbeta)\big\}$ for $t \in [T]$ \\
\underline{Step 2:} (Projection and concatenation) Create a $p \times T$ matrix $\widehat{\bB}$ of which the $t$-th column is $\prod_R(\widetilde{\bbeta}^{(t)})$, where $R = \texttt{quantile}(\{\twonorm{\widetilde{\bbeta}^{(t)}}\}_{t=1}^T, 1-\bar{\epsilon})$\\
\underline{Step 3:} (SVD) Conduct SVD $\widehat{\bB} = \widehat{\bm{U}}\widehat{\bm{\Lambda}}\widehat{\bm{V}}^\top$ with $\widehat{\bm{U}} \in \mathcal{O}^{p \times T}$, let $\widehat{\oA}$ be the first $r$ columns of $\widehat{\bm{U}}$, and set $\hthetak{t} = \argmin_{\btheta \in \mathbb{R}^r}\fk{t}(\widehat{\oA}\btheta)$ \\
\underline{Step 4:} (Biased regularization) $\hbetak{t} = \argmin_{\bbeta \in \mathbb{R}^p} \big\{\fk{t}(\bbeta) + \frac{\gamma}{\sqrt{n}}\twonorm{\bbeta - \widehat{\oA}\hthetak{t}}\big\}$ for $t \in [T]$
\end{algorithm}

Steps 1 and 2 of Algorithm \ref{algo: spectral} construct a robust estimator of $\bm{B}^*_S$ by concatenating projected single-task estimators. The projection limits the impact of outlier tasks by setting the projection radius to a quantile of the $\ell_2$-norms of all single-task estimators. This projection technique, sometimes referred to as truncation or winsorization, is widely used in robust statistics (e.g., \citealp{lugosi2021robust}) and differential privacy (e.g., \citealp{dwork2006differential, dwork2014algorithmic}). To determine the quantile percentage, we need an upper bound $\bar{\epsilon}$ of $\epsilon$. If we have prior knowledge of $\epsilon$ (but not $S$ and $S^c$), we can set $\bar{\epsilon} = \epsilon$. Without this information, $\bar{\epsilon}$ can be chosen as a small constant, such as $0.05$. In the simulations, we set $\bar{\epsilon}$ as the true $\epsilon$ value, and in the real-data study, we set $\bar{\epsilon} = 0.05$. Step 3 performs the SVD to estimate the central representation $\oA$. Step 4 applies biased regularization as in Algorithm \ref{algo: mtl} to prevent negative transfer. The spectral method avoids the complicated non-convex optimization problems present in the penalized ERM algorithm, making it easy to implement in practice. In fact, the spectral method can be solved in polynomial time since the objective function in Step 4, which consists of a sum of a smooth and strongly convex component and an $\ell_2$-Lipschitz component, can be minimized in polynomial time (e.g., via stochastic gradient descent; see \citealp{shamir2013stochastic}). Moreover, SVD can also be computed in polynomial time. Besides its simplicity and efficiency, we can show that it achieves a better upper bound of estimation error for $\bbetaks{t}$'s when $S = [T]$. 

\begin{theorem}[Upper bound for spectral method]\label{thm: spectral mtl}
	Suppose Assumptions \ref{asmp: x}-\ref{asmp: n} hold with a subset $S \subseteq [T]$ satisfying $\epsilon = \frac{|S^c|}{T} \leq cr^{-1}\cdot \Big(\frac{\barzeta}{\max_{t \in S}\zetak{t}}\Big)^2$, where $c > 0$ is a small constant. By setting $\gamma = C'\sqrt{p+\log T}$ with a sufficiently large positive constant $C'$, w.p. at least $1-e^{-C''(r+\log T)}$ and $\bar{\epsilon}$ satisfying $\epsilon \leq \bar{\epsilon} \leq \frac{c''}{r}\cdot \Big(\frac{\barzeta}{\max_{t \in S}\zetak{t}}\Big)^2$ with a sufficiently small positive constant $c''$, we have \footnote{For convention, we define $0/0 = 0$, which is mainly for the case $(\oA^\perp)^\top \bm{B}^* = \bm{0}_{(p-r)\times r}$.}
	\begin{align}
		\twonorm{\hbetak{t}-\bbetaks{t}} &\lesssim \Bigg\{\frac{\zetak{t}}{\barzeta}\sqrt{\frac{pr}{nT}} + \bigg(\frac{\zetak{t}}{\barzeta}\vee 1\bigg) \sqrt{\frac{r}{n}} + \sqrt{\frac{\log T}{n}} + \zetak{t}h\cdot \bigg[\frac{\sigma_{\max}((\oA^\perp)^\top \bm{B}^*_S)}{\sigma_{\min}((\oA^\perp)^\top \bm{B}^*_S)} \wedge \sqrt{r}\bigg] \\
		&\quad\quad + \frac{\zetak{t}}{\barzeta}\max_{t \in S}\zetak{t}\cdot \sqrt{r\bar{\epsilon}} \Bigg\} \wedge \sqrt{\frac{p + \log T}{n}}, \hspace{2cm}\quad \forall t \in S,
	\end{align}
	where $\bm{B}^*_S \in \mathbb{R}^{p \times |S|}$ is the coefficient matrix whose columns are $\{\bbetaks{t}\}_{t \in S}$, $\oA \in \argmin_{\bm{A} \in \mathcal{O}^{p \times r}}\allowbreak \max_{t \in [T]}\twonorm{\bAks{t}(\bAks{t})^\top - \bA\bA^\top}$ is the central representation, and $\oA^\perp$ is orthogonal to $\oA$ in the sense that $\oA^\perp \in \mathcal{O}^{p\times (p-r)}$ and $(\oA^\perp)^\top \oA = \bm{0}_{(p-r)\times r}$.
	
	Furthermore, if we assume the data from tasks in $S^c$ satisfies the linear model \eqref{eq: linear model} (without any latent structure assumption) and Assumption \ref{asmp: x}, then w.p. at least $1-e^{-C'(p+\log T)}$, we also have
	\begin{equation}
		\max_{t \in [T]}\twonorm{\hbetak{t}-\bbetaks{t}} \lesssim \sqrt{\frac{p + \log T}{n}}.
	\end{equation}
\end{theorem}

We can similarly discuss when Algorithm \ref{algo: spectral} improves single-task learning as in Theorem \ref{thm: mtl}, which we do not repeat here. However, it is worth emphasizing that, like Algorithm \ref{algo: mtl}, Algorithm \ref{algo: spectral} is also \textbf{adaptive} to the unknown similarity structure and it is also \textbf{robust} to a small fraction of outlier tasks. Comparing the rates in Theorems \ref{thm: mtl} and \ref{thm: spectral mtl}, we can see that the estimation error rate of the spectral method has a better dependence on $r$ when $S = [T]$, although the last term related to outlier tasks is worse. In the next subsection, we will see that the spectral method is minimax optimal when $S = [T]$ and the condition number $\sigma_{\max}((\oA^\perp)^\top \bm{B}^*_S)/\sigma_{\min}((\oA^\perp)^\top \bm{B}^*_S)$ is bounded by a constant.

As mentioned, Algorithm \ref{algo: spectral} requires an upper bound $\bar{\epsilon}$ of $\epsilon$ for determining the quantile percentage, and $\bar{\epsilon}$ also appears in the upper bound of estimation error. If $\bar{\epsilon} \asymp \epsilon$, then we can replace $\bar{\epsilon}$ in the upper bound with the true $\epsilon$.

Besides the intuition provided at the beginning of this subsection, Algorithm \ref{algo: spectral} also has connections to average derivative estimation (ADE) and expected gradient outer product (EGOP) methods used to estimate the index space in single-index models and multi-index models (e.g., \citealp{hardle1989investigating, samarov1993exploring, hristache2001direct, yang2017learning, yuan2023efficient}). For instance, in the multi-index model $\tE[Y|X = \bx] \coloneqq g(\bx) = f((\bB^*)^\top \bx) \in \mathbb{R}$ with $\bB^* \in \mathbb{R}^{p \times T}$, $\bx \in \mathbb{R}^p$, $g$ continuously differentiable and unknown, and $\tE[\nabla g(X)(\nabla g(X))^\top]$ existing, we have:
\begin{equation}
	\tE[\nabla g(X)(\nabla g(X))^\top] = (\bB^*)^\top\tE\{\nabla f((\bB^*)^\top X)[\nabla f((\bB^*)^\top X)]^\top\}\bB^*.
\end{equation}
Therefore, it is possible to use SVD or PCA on some estimator of $\tE[\nabla g(X)(\nabla g(X))^\top]$ to recover the column space of $\bB^*_S$, and many estimators of the expected gradient outer product have been proposed in the literature. In the context of representation MTL, when $S = [T]$, we can view $g$ as a multivariate function $\bx \mapsto ((\bbetaks{1})^\top \bx, \ldots, (\bbetaks{T})^\top \bx)^\top$ with $f$ the identity function from $\mathbb{R}^T$ to $\mathbb{R}^T$, implying that $\tE[\nabla g(X)(\nabla g(X))^\top] = (\bB^*_S)^\top \bB^*_S$. Thus, our spectral method, which performs SVD on the estimated $\bB^*_S$, can be viewed as a multi-task variant of the EGOP framework.

The SVD step in our method is also closely related to the SVD-based approaches used in \cite{kong2020meta} and \cite{meunier2023nonlinear}, where the former considers a mixture model setting and the latter operates in an infinite-dimensional RKHS framework. Translated into our terminology, the key difference is that their methods rely on data splitting to estimate two instances of $\bm{B}_S^*$'s separately in the product $(\bm{B}_S^*)^\top \bm{B}_S^*$. In contrast, our approach applies SVD directly to the unbiased estimator $\widehat{\bB}$ rather than to the potentially biased $\widehat{\bB}^\top \widehat{\bB}$ and thereby eliminates the need for sample splitting. 

When there are outlier tasks or contaminations, i.e., $S \neq [T]$, besides the simple projection technique, we can also borrow ideas from robust principal component analysis (PCA) literature (e.g., \citealp{wright2009robust, candes2011robust, vidal2016principal}) to robustify the SVD procedure in Algorithm \ref{algo: spectral}. For example, instead of conducting SVD on $\widehat{\bB}$ in Step 3 of Algorithm \ref{algo: spectral}, we may conduct SVD on another $p \times T$ matrix $\widehat{\bm{L}}$ which is the solution of the convex optimization \citep{liu2012robust, xu2012robust}
\begin{equation}
	\min_{\bm{L}, \bm{Z}} \|\bm{L}\|_* + \lambda\|\bm{Z}\|_{2,1} \quad \text{s.t. } \widehat{\bB} = \bm{L} + \bm{Z},
\end{equation} 
where $\|\bm{L}\|_* = \sum_{j=1}^{p \wedge T} \sigma_j(\bm{L})$ is the nuclear norm, and $\|\bm{Z}\|_{2,1} = \sum_{t=1}^T \twonorm{\bm{z}_t}$ with $\bm{z}_t$ the $t$-th column of $\bm{Z}$ is the $L_{2,1}$-norm. This approach might lead to better performance when $S \neq [T]$. Given the extensive scope of the current paper, we leave the study of this approach for future research.

\subsection{Lower Bound}\label{subsec: lower bdd mtl}
In this subsection, we derive lower bounds to explore the information-theoretic hardness of the representation MTL problem. Consider a collection of all subsets $S \subseteq [T]$ as
\begin{equation}
	\mS = \{S \subseteq [T]: |S^c|/T \leq \epsilon\}.
\end{equation}
Given the subset $S$, define a coefficient matrix $\bB_S \in \mathbb{R}^{p \times |S|}$, where each column corresponds to a coefficient vector in $\{\bbetak{t}\}_{t \in S}$. Consider the parameter space for the coefficient vectors $\{\bbetak{t}\}_{t \in S}$ as
\begin{align}
	\mathscr{B}(S, h) =& \Bigg\{\{\bbetak{t}\}_{t\in S}: \bbetak{t} = \bAk{t}\bthetak{t} \text{ for all } t \in S, \{\bAk{t}\}_{t\in S} \subseteq \mO, \twonorm{\bthetak{t}} \leq \zetak{t},\\
	&\min_{\oA \in \mO}\max_{t \in S}\twonorm{\bAk{t}(\bAk{t})^\top - \oA  \oA^\top} \leq h,  \sigma_r\Big(|S|^{-1/2}\bB_S\Big) \geq \frac{c}{\sqrt{r}}\sqrt{\frac{1}{|S|}\sum_{t \in S}\twonorm{\bthetak{t}}^2}\Bigg\}
\end{align}
where $c$ can be any fixed positive constant such that $\mathscr{B}(S, h) \neq \emptyset$. Given a set $S \subseteq [T]$, denote $\barzeta \coloneqq \barzeta(S) = \sqrt{\frac{1}{|S|}\sum_{t \in S} (\zetak{t})^2}$. We have the following lower bound. 

\begin{theorem}[Lower bound for MTL]\label{thm: mtl lower bdd}
	Suppose $p \geq 2r$, $T\geq r^{1.01}$, $\min_{t \in [T]}\zetak{t} \geq C$, and $\epsilon \leq c/r$ where $C$ and $c$ are some positive constants. We have the following lower bound hold:
	\begin{align}
		&\inf_{\{\hbetak{t}\}_{t=1}^T}\sup_{S \subseteq \mS}\sup_{\substack{\{\bbetak{t}\}_{t\in S} \in \mathscr{B}(S, h)  \\ \mQ_{S^c}}}\tP \Bigg(\bigcup_{t \in S}\Bigg\{\twonorm{\hbetak{t}-\bbetaks{t}} \gtrsim \bigg[\frac{\zetak{t}}{\barzeta}\sqrt{\frac{pr}{nT}} + \bigg(\frac{\zetak{t}}{\barzeta} \vee 1\bigg)\sqrt{\frac{r}{n}} + \frac{\log T}{n} \\
		&\quad  + \zetak{t}h + \frac{\zetak{t}}{\barzeta}\frac{\epsilon r}{\sqrt{n}}\bigg]\wedge \sqrt{\frac{p+\log T}{n}}\Bigg\}\Bigg) \geq \frac{1}{10},
	\end{align}
	where for any given $S \subseteq \mS$, $\tP = \tP_S \otimes \mQ_{S^c}$, and $\tP_S$, $\mQ_{S^c}$ are the probability measures on sample space of tasks in $S$, $S^c$, respectively. Furthermore, if tasks in $S^c$ also follow the linear model \eqref{eq: linear model}, then we have the following lower bound, where $\tP$ is the probability measure on sample space of all tasks:
	\begin{equation}
		\inf_{\{\hbetak{t}\}_{t=1}^T}\sup_{S \subseteq \mS}\sup_{\substack{\{\bbetak{t}\}_{t\in S} \in \mathscr{B}(S, h)  \\ \{\bbetak{t}\}_{t \in S^c}}}\tP \bigg(\max_{t \in [T]}\twonorm{\hbetak{t}-\bbetaks{t}} \gtrsim \sqrt{\frac{p+\log T}{n}}\bigg) \geq \frac{1}{10}.
	\end{equation}
\end{theorem}
 
Similar to the upper bound in Theorem \ref{thm: mtl}, the lower bound of $\max_{t \in S}\twonorm{\hbetak{t}-\bbetaks{t}}$ contains several terms reflecting the difficulty of learning different components. For example, $\frac{\zetak{t}}{\barzeta}\sqrt{\frac{pr}{nT}}+ \zetak{t}h$ arises from learning the similar representations in $S$; $\big(\frac{\zetak{t}}{\barzeta} \vee 1\big)\sqrt{\frac{r}{n}} + \sqrt{\frac{\log T}{n}}$ is due to learning the task-specific parameters; $\frac{\zetak{t}}{\barzeta}\frac{\epsilon r}{\sqrt{n}}$ is caused by outlier tasks; and $\sqrt{\frac{p+\log T}{n}}$ is the single-task rate.

To our knowledge, this is the first lower bound for learning regression parameters in the context of representation MTL. \cite{tripuraneni2021provable} and \cite{duchi2022subspace} derived lower bounds for the subspace recovery when $\zetak{t} \lesssim 1$ for all $t \in S= [T]$, with no outlier tasks ($\epsilon = 0$) and all tasks sharing the same representation ($h = 0$). 

Comparing the upper bound for penalized ERM in Theorem \ref{thm: mtl} with the lower bound, the upper bound exhibits a sub-optimal dependence on $r$ compared to the information-theoretic lower bound. This phenomenon has been noted in \cite{du2020few} and \cite{tripuraneni2021provable} for both the ERM estimator and a method-of-moments estimator. The upper bounds of estimation errors for both estimators have sub-optimal dependence on $r$. They related this to a similar phenomenon observed in other works on linear regression models \citep{raskutti2011minimax}, where the upper bounds of estimation errors have sub-optimal dependence on eigenvalues of design matrices. In addition, the last term in our lower bound, $\frac{\zetak{t}}{\barzeta}\frac{\epsilon r}{\sqrt{n}}$, does not depend on the full dimension $p$, whereas the counterpart in the upper bound does. A similar phenomenon has been noted in several papers (e.g. \citealp{tian2022unsupervised, duan2023adaptive}). An open question is whether the dependence on $p$ can be removed. Unfortunately, our ongoing work shows that this is impossible for the penalized ERM method like Algorithm \ref{algo: mtl} for a broad class of commonly used regularizers. As pointed out by \cite{tian2022unsupervised}, estimators based on techniques in robust statistics like Tukey's depth function have been shown to achieve minimax rate under Huber's contamination model for location and covariance estimation \citep{chen2018robust}, which might help improve the upper bound in our setting. In summary, when $r$ is bounded, $\epsilon = 0$, and $\zetak{t} \lesssim \barzeta$ for all $t \in S = [T]$, the penalized ERM is optimal. An adaptation of techniques in robust statistics to improve our estimation algorithm will be an interesting future research direction.

On the other hand, the upper bound for the spectral method in Theorem \ref{thm: spectral mtl} matches the lower bound when $\epsilon = 0$ and the condition number $\sigma_{\max}((\oA^\perp)^\top \bm{B}^*_S)/\sigma_{\min}((\oA^\perp)^\top \bm{B}^*_S)$ in Theorem \ref{thm: spectral mtl} is bounded by a constant. To our knowledge, when there is no contamination ($\epsilon = 0$), no algorithm in the literature could achieve the optimal estimation error rate of $\bbetaks{t}$'s before our work, even in the special case $h = 0$. This demonstrates the power of the spectral method in the representation MTL. However, the term related to the outlier tasks does not involve $n$ and is not optimal. It remains unknown what algorithm can achieve the optimal estimation error rate when $\epsilon \neq 0$.

Finally, we would like to emphasize that the condition $\epsilon r \leq c$ for some constant $c > 0$ is necessary for representation multi-task learning. As discussed at the end of the proof of Theorem \ref{thm: mtl lower bdd} in Section \ref{subsec: proof lower bound mtl supp} of the appendix, when $\epsilon r > 1$, the lower bound immediately deteriorates to $\sqrt{p/n}$. This reduces the multi-task rate to the single-task rate, thereby eliminating the benefits of data integration in the worst-case scenario. 

\subsection{Extensions to Generalized Linear Models and Non-linear Regression}\label{subsec: extension main text}
We can extend the proposed methods and theoretical framework beyond the linear model \eqref{eq: linear model}.  For example, we can consider generalized linear models (GLMs) \citep{mccullagh1989generalized}, where the conditional distribution of $Y$ given $X = \bx$ for task $t$ has density
\begin{equation}\label{eq: GLM main text}
	p(\yk{t}_i = y|\bxk{t}_i = \bx) = \rho(y)\exp\big\{y\cdot \bx^\top\bbetaks{t} - \psi(\bx^\top\bbetaks{t})\big\}, \quad i = 1:n,
\end{equation} 
for $t \in S$, w.r.t. some measure $\mu$ on a subset of $\mathbb{R}$, where $\psi$ is second-order continuously differentiable on $\mathbb{R}$, and $\psi'$ is often called the inverse link function.

In addition to GLMs, we can also extend the linear model \eqref{eq: linear model} to a non-linear regression model \citep{yang2015sparse}, where
\begin{equation}\label{eq: non-linear model main text}
	\yk{t}_i = g\big((\bxk{t}_i)^\top\bbetaks{t}\big) + \epsilonk{t}_i, \quad i = 1:n,
\end{equation}
for $t \in S$, where $g$ is a monotone function with a continuous second-order derivative on $\mathbb{R}$ and $\{\epsilonk{t}_i\}_{i=1}^n$ are i.i.d. zero-mean sub-Gaussian variables independent of $\{\bxk{t}_i\}_{i=1}^n$. In literature, $g$ is often referred to as the link function. 

Given the extensive coverage in the main text, we defer the details of these extensions to Section \ref{sec: extensions} of the appendix.

\section{Adaptation to Unknown Intrinsic Dimension $r$}\label{sec: unknown r}
In Sections \ref{sec: mtl} and \ref{sec: extensions}, the intrinsic dimension $r$ is assumed to be known a priori. To our knowledge, this assumption is standard in almost all related theoretical literature on representation multi-task and transfer learning (e.g., \citealp{ando2005framework, maurer2016benefit, du2020few, thekumparampil2021statistically, tripuraneni2021provable, chua2021fine, collins2021exploiting, deng2022learning, duchi2022subspace, duan2023adaptive}), despite being potentially unrealistic in practice. In this section, we propose a simple yet effective algorithm to adapt the previous MTL algorithms to the case of an unknown $r$. Recall the notations $\zetak{t} = \twonorm{\bthetaks{t}}$ and $\barzeta = \sqrt{|S|^{-1}\sum_{t \in S}(\zetak{t})^2}$. Similar to the notations used in Algorithm \ref{algo: spectral}, we use $\prod_R$ to denote the projection operator to an $\ell_2$-ball centered at zero of radius $R$ in $\mathbb{R}^p$, and use $\texttt{quantile}(\{a_t\}_{t=1}^T, 1-\bar{\epsilon})$ to denote the lower $(1-\bar{\epsilon})$-quantile of $\{a_t\}_{t=1}^T \subseteq \mathbb{R}$, where $\bar{\epsilon} \in [0, 1]$.

The algorithm is based on SVD, and the details are summarized in Algorithm \ref{algo: adaptation unknown r}.

\begin{algorithm}[!h]
\caption{Adaptation to unknown intrinsic dimension $r$}
\label{algo: adaptation unknown r}
\KwIn{Data from tasks $\{\bXk{t}, \bYk{t}\}_{t=1}^T = \{\{\bxk{t}_i, \yk{t}_i\}_{i=1}^n\}_{t=1}^T$, threshold parameters $T_1, T_2 > 0$, an upper bound $\bar{\epsilon}$ (for $\epsilon$)}
\KwOut{An estimate $\hat{r}$}
\underline{Step 1:} (Single-task regression) $\widetilde{\bbeta}^{(t)} = \argmin_{\bbeta \in \mathbb{R}^p}\big\{ \fk{t}(\bbeta)\big\}$ for $t \in [T]$ \\
\underline{Step 2:} (Projection and concatenation) Create a $p \times T$ matrix $\widehat{\bB}$ whose $t$-th column is $\prod_R(\widetilde{\bbeta}^{(t)})$, where $R = \texttt{quantile}(\{\twonorm{\widetilde{\bbeta}^{(t)}}\}_{t=1}^T, 1-\bar{\epsilon})$\\
\underline{Step 3:} (Thresholding) Set $\hat{r} = \max\Big\{r' \in [T]: \sigma_{r'}(\widehat{\bB}/\sqrt{T}) \geq T_1\sqrt{\frac{p+\log T}{n}} + T_2R\sqrt{\bar{\epsilon}}\Big\}$
\end{algorithm}

Algorithm \ref{algo: adaptation unknown r} leverages Assumption \ref{asmp: theta} to determine an appropriate value for $r$. The underlying rationale is that when $h$ is small, a significant spectral gap often exists between the $r$-th largest singular value and the $(r+1)$-th largest singular value of $\bB^*_S/\sqrt{T}$. For example, when $h=0$, $\bB^*_S$ is a rank-$r$ matrix, implying that $\sigma_{r'}(\bB^*_S/\sqrt{T}) = 0$ for $r' \geq r+1$. Therefore, thresholding on singular values of an empirical version of $\bB^*_S$ can be an effective strategy to estimate $r$. In Section \ref{subsec: spectral}, a similar approach based on SVD of the estimated $\bB^*$ was used to develop the spectral method. In fact, Steps 1-2 of Algorithms \ref{algo: spectral} and \ref{algo: adaptation unknown r} are the same, which construct a robust estimate of $\bB^*_S$ through projected single-task estimators. Algorithm \ref{algo: adaptation unknown r} is also conceptually similar to the thresholding method often used to determine the intrinsic dimension in principal component analysis \citep{onatski2010determining, fan2021robust}. 

Under almost identical assumptions imposed in previous sections, with proper choices of tuning parameters, Algorithm \ref{algo: adaptation unknown r} is shown to be consistent in estimating the true intrinsic dimension $r$, when representation matrices are similar (i.e., $h$ is small). As we will elaborate, this suffices to ensure the same upper bounds of estimation error for Algorithms \ref{algo: mtl} and \ref{algo: spectral} when $r$ is unknown.

\begin{theorem}[Consistency of the intrinsic dimension estimation]\label{thm: adaptation unknown r}
	Suppose we choose an $\bar{\epsilon}$ such that $\epsilon \leq \bar{\epsilon} \leq \frac{c}{r}\Big(\frac{\barzeta}{\max_{t \in S}\zetak{t}}\Big)^2$ with a small constant $c > 0$. Assume $\min_{t \in S}\zetak{t}\cdot h \Big[\frac{\sigma_{\max}((\oA^\perp)^\top \bm{B}^*_S)}{\sigma_{\min}((\oA^\perp)^\top \bm{B}^*_S)} \wedge \sqrt{r}\Big] \leq c'\sqrt{\frac{p+\log T}{n}}$ with a small constant $c' > 0$, where $\bm{B}^*_S \in \mathbb{R}^{p \times |S|}$ is the coefficient matrix whose columns are $\{\bbetaks{t}\}_{t \in S}$, $\oA \in \argmin_{\bm{A} \in \mathcal{O}^{p \times r}}\allowbreak \max_{t \in [T]}\twonorm{\bAks{t}(\bAks{t})^\top - \bA\bA^\top}$ is the central representation, and $\oA^\perp$ is orthogonal to $\oA$ in the sense that $\oA^\perp \in \mathcal{O}^{p\times (p-r)}$ and $(\oA^\perp)^\top \oA = \bm{0}_{(p-r)\times r}$. Further assume that: \footnote{Assumptions \ref{asmp: glm}, \ref{asmp: n glm mtl}, \ref{asmp: non-linear}, and \ref{asmp: non-linear n} for GLMs and non-linear regression models are presented in Section \ref{sec: extensions} of the appendix.}
	\begin{enumerate}[(i)]
		\item For the linear model \eqref{eq: linear model}, Assumptions \ref{asmp: x}, \ref{asmp: theta}, and \ref{asmp: n glm mtl} hold;
		\item For the GLM \eqref{eq: GLM main text}, Assumptions \ref{asmp: x}, \ref{asmp: theta}, \ref{asmp: glm}, and \ref{asmp: n glm mtl} hold;
		\item For the non-linear regression model \eqref{eq: non-linear model main text}, Assumptions \ref{asmp: x}, \ref{asmp: theta}, \ref{asmp: non-linear}, \ref{asmp: non-linear n} hold;
	\end{enumerate}
	Then there exist constants $T_1, T_2 > 0$ such that the output of Algorithm \ref{algo: adaptation unknown r} satisfies $\hat{r} = r$ \wppp\, with some constant $C' > 0$.
\end{theorem}

\begin{remark}\label{rmk: adaptation}
	When $\min_{t \in S}\zetak{t}\cdot h \Big[\frac{\sigma_{\max}((\oA^\perp)^\top \bm{B}^*_S)}{\sigma_{\min}((\oA^\perp)^\top \bm{B}^*_S)} \wedge \sqrt{r}\Big] \gtrsim \sqrt{\frac{p+\log T}{n}}$ in the MTL problem, estimating $r$ becomes unnecessary as the upper bounds of MTL estimation errors are dominated by the single-task rate $\sqrt{\frac{p+\log T}{n}}$ for both Algorithms \ref{algo: mtl} and \ref{algo: spectral}. In such cases, single-task learning is sufficient to achieve the minimax rate. Specifically, by the proofs of these upper bounds, the single-task rate is always guaranteed by biased regularization in Step 2 of Algorithm \ref{algo: mtl} and Step 4 of Algorithm \ref{algo: spectral}, regardless of the performance achieved in other steps.
\end{remark}

According to Theorem \ref{thm: adaptation unknown r} and Remark \ref{rmk: adaptation}, when $r$ is unknown, we can first run Algorithm \ref{algo: adaptation unknown r} to obtain an estimate $\hat{r}$, then run Algorithms \ref{algo: mtl} and \ref{algo: spectral} with $\hat{r}$. All the previous results remain valid. This confirms that our full procedure is \textbf{adaptive} to an unknown intrinsic dimension $r$.

Similar to Algorithm \ref{algo: spectral}, Algorithm \ref{algo: adaptation unknown r} also requires $\bar{\epsilon}$ as an upper bound of the proportion of outlier tasks $\epsilon$. If we have prior knowledge of $\epsilon$ (but not $S$ and $S^c$), we can set $\bar{\epsilon} = \epsilon$. Without such information, a small constant such as $0.05$ can be chosen for $\bar{\epsilon}$. Similar to Algorithm \ref{algo: spectral}, we set $\bar{\epsilon}$ to the true $\epsilon$ value in simulations, while setting $\bar{\epsilon} = 0.05$ in the real-data study. We set the tuning parameters to $T_1 = 0.5$ and $T_2 = 0.25$ in the numerical experiments. Generally, $T_1, T_2$ can be chosen through cross-validation.

Before concluding this section, we would like to highlight that, in addition to the thresholding method, other approaches have been proposed for selecting the number of factors in factor models, which may also be beneficial in our context. For example, information criterion-based methods \citep{bai2002determining, bunea2011optimal} provide an alternative approach. Algorithm \ref{algo: adaptation unknown r} is naturally motivated by the singular value gap condition in Assumption \ref{asmp: theta}. However, we believe that an information criterion can also be developed to consistently select $r$, leveraging the estimation error bounds we have established. Furthermore, this selection process can be framed as a hypothesis testing problem \citep{onatski2009testing}. While all these methods appear promising, it remains unclear how to effectively address heterogeneity across tasks and mitigate task contamination. We leave these challenges for future exploration.

\section{Numerical Experiments}\label{sec: numerical}
To validate the theoretical insights discussed in previous sections, we conducted extensive simulations and one real-data study, and the results are presented in this section. 

All the experiments were implemented in Python. For penalized ERM (``pERM", Algorithm \ref{algo: mtl}), we used the automatic differentiation implemented in PyTorch \citep{paszke2019pytorch} along with the Adam optimizer \citep{kingma2015adam} to solve the optimization problem \eqref{eq: replacement} in Step 1. We set the learning rate equal to $0.01$ in \texttt{torch.optim.Adam} function and kept all the other parameter choices as in default. Step 2 of pERM and Step 4 of the spectral method (``Spectral", Algorithm \ref{algo: spectral}) were also solved by the Adam optimizer with a learning rate $0.01$. Consistent with our theory, we set penalty parameters $\lambda = \sqrt{r(p+\log T)}$ and $\gamma = \sqrt{p+\log T}$ in pERM, and $\gamma = 0.5\sqrt{p +\log T}$ in the spectral method. As mentioned in Section \ref{subsec: spectral}, Spectral requires an upper bound $\bar{\epsilon}$ of the contamination proportion $\epsilon$. We set $\bar{\epsilon} = \epsilon$ in simulations and $\bar{\epsilon} = 0.05$ in the real-data study. Besides pERM and Spectral, we included the following approaches as benchmarks. 
\begin{itemize}
	\item Empirical risk minimization (``ERM") in \cite{du2020few, tripuraneni2021provable}: The optimization was also solved by the Adam solver in PyTorch with a learning rate $0.01$;
	\item Method-of-moments (``MoM") in \cite{tripuraneni2021provable};
	\item Adaptive representation learning (``AdaptRep") in \cite{chua2021fine}: We used the code included in the original paper and kept all the parameter settings as default.
	\item Adaptive and robust multi-task learning (``ARMUL") in \cite{duan2023adaptive}: We used the code included in their paper (\url{https://github.com/kw2934/ARMUL}) and retained all default parameter settings. The tuning parameters were chosen using 5-fold cross-validation as default.
	\item Group Lasso (``GLasso") in \cite{yuan2006model, lounici2009taking, lounici2011oracle}: We grouped the same coordinate of coefficients from different tasks and applied an $L_{2,1}$-matrix penalty. The method was implemented in an R package \texttt{RMTL} \citep{cao2019rmtl}, and we used the Python package \texttt{rpy2} to call functions \texttt{cv.MTL} and \texttt{MTL} in the R package \texttt{RMTL}. The penalty parameter was chosen by a 5-fold cross-validation as default.
	\item Data pooling or pooled regression (``Pooled") in \cite{crammer2008learning, ben2010theory}: We fitted the linear regression and logistic regression models on the pooled data from all tasks. Both models were implemented in the Python module \texttt{sklearn.linear\_models}.  
	\item Single-task regression (Single-task): In simulations and the real-data study, we run single-task linear regression and logistic regression on the data of each task, respectively. 
\end{itemize}

In Section \ref{subsec: simulation}, we present the performance of different approaches under different simulation settings, such as different heterogeneity parameter $h$ (Section \ref{subsubsec: sim h}), different contamination proportion $\epsilon$ (Section \ref{subsubsec: sim epsilon}), and different number of tasks $T$ (Section \ref{subsubsec: sim T}). We also change the full dimension $p$, the intrinsic dimension $r$, and the per-task sample size $n$ from setting to setting. To verify the intuition we obtained in Section \ref{sec: mtl}, where we mentioned that the performance of each task in ERM, pERM, and Spectral depends on the signal strength $\zetak{t}=\twonorm{\bthetaks{t}}$, we conduct a simulation with different $\twonorm{\bthetaks{t}}$ values across tasks, and the results are presented in Section \ref{subsubsec: sim theta}. Finally, in Section \ref{subsubsec: sim r}, we evaluate Algorithm \ref{algo: adaptation unknown r} for estimating $r$, compare the performance of pERM and Spectral with the estimated $r$ and the true $r$, and demonstrate the effectiveness of Algorithm \ref{algo: adaptation unknown r}.

In Section \ref{subsec: real data}, we compare the performance of different approaches on a real dataset.

The code to reproduce the results is available at \url{https://github.com/ytstat/RL-MTL-TL}.

\subsection{Simulations}\label{subsec: simulation}
\subsubsection{Simulation with Different Heterogeneity Parameter $h$}\label{subsubsec: sim h}
In this subsection, we investigate the linear model \eqref{eq: linear model intro} with various values of $h$. We explore four distinct settings with different combinations of $(n, p, r, T)$:
\begin{enumerate}[(i)]
	\item $n = 100, p = 30, r = 5, T = 50$;
	\item $n = 100, p = 50, r = 5, T = 50$;
	\item $n = 100, p = 80, r = 5, T = 50$;
	\item $n = 150, p = 80, r = 10, T = 50$.
\end{enumerate}
In all settings, no outlier tasks are included, implying $S = [T]$ and $\epsilon = |S^c|/T = 0$. Given each $(n, p, r, T)$ combination, we generated $\bxk{t}_i$ i.i.d. from $N(\bm{0}_p, \bm{I}_p)$, $\epsilonk{t}_i$ i.i.d. from $N(0, 1)$, and a random $p \times r$ matrix $\bm{C}$ with i.i.d. standard normal entries. We defined $\barA$ as the first $r$ columns of the left singular matrix of $\bm{C}$, $\widetilde{\bA}^{(t)} = \barA + a^{(t)}(\bm{I}_{r \times r}, \bm{0}_{r \times (p-r)})^\top$, and $\bAks{t} = \widetilde{\bA}^{(t)}[(\widetilde{\bA}^{(t)})^\top \widetilde{\bA}^{(t)}]^{-1}(\widetilde{\bA}^{(t)})^\top$ for $t \in [T]$, where $a^{(t)}$'s are i.i.d. sampled from $\textup{Unif}([-h, h])$. We generated each coordinate of $\bthetaks{t} \in \mathbb{R}^r$ from $\textup{Unif}([-2, 2])$ independently. We considered $h$ from 0 to 0.8 in increments of 0.1 and replicated each setting 100 times.

\begin{figure}[!h]
	\centering
	\includegraphics[width=\textwidth]{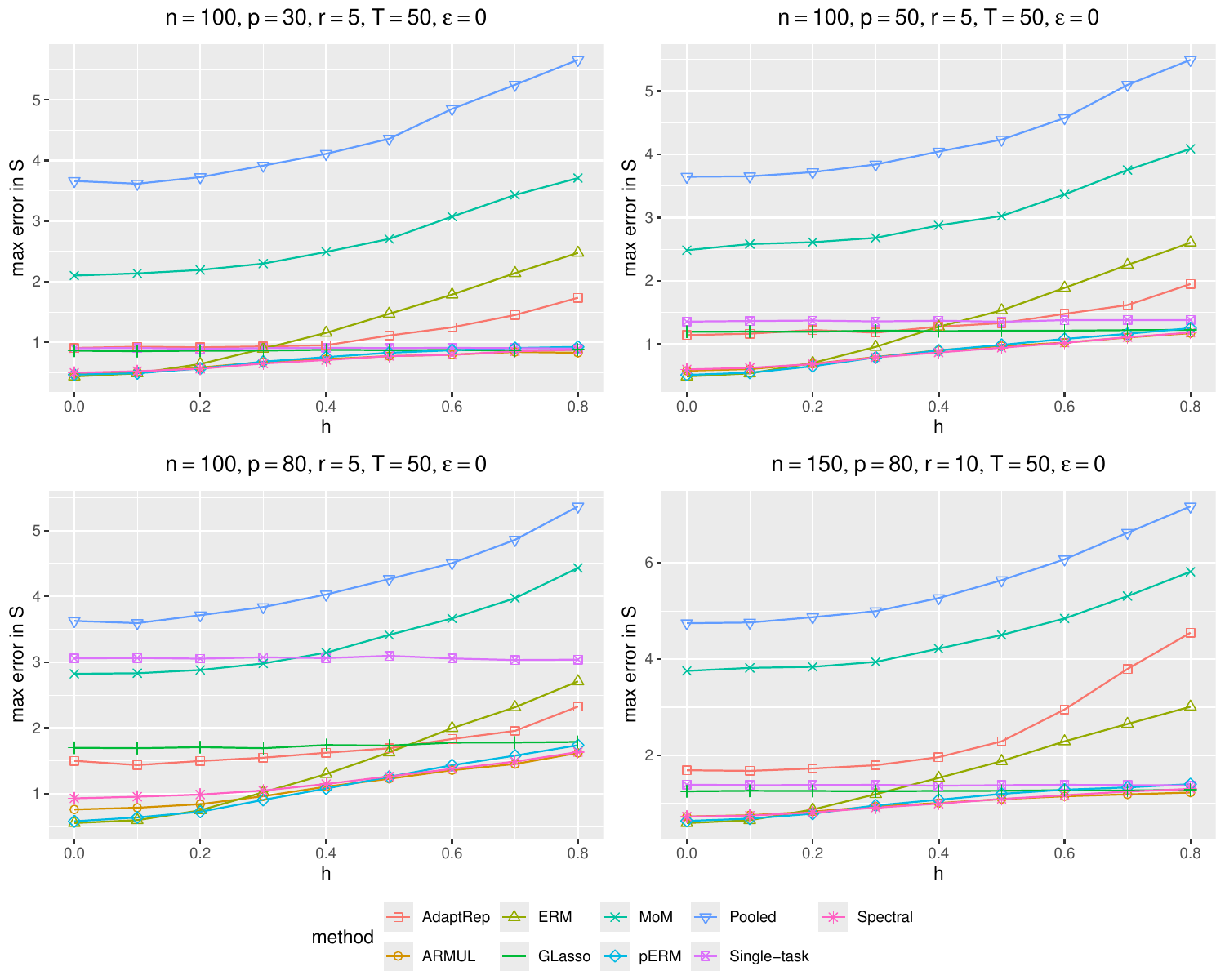}
	\caption{Simulation with different heterogeneity parameters $h$: estimation error $\max_{t \in [T]}\twonorm{\hbetak{t}-\bbetaks{t}}$ of different algorithms with different $(n, p, r, T)$ settings. ``max error in $S$" in the $y$-axis stands for $\max_{t \in S}\twonorm{\hbetak{t}-\bbetaks{t}}$ with $S = [T]$. Each point represents the average over 100 replications.}
	\label{fig: sim_h}
\end{figure}

Figure \ref{fig: sim_h} presents the simulation results. Across all settings, pooled regression and MoM perform worse than or on par with single-task regression, with their performance deteriorating as $h$ increases. AdaptRep exhibits slightly better performance than single-task regression initially, but declines as $h$ increases. GLasso performs similarly to single-task regression in settings (\rom{1}), (\rom{2}), and (\rom{4}), while notably improving upon it in setting (\rom{3}). ERM significantly enhances single-task regression performance when $h$ is small, but performs worse than single-task regression for large $h$. In contrast, ARMUL, pERM, and Spectral can improve the performance of single-task regression when $h$ is small, and their performance will be comparable to single-task regression for large $h$.

As one reviewer pointed out, Spectral performs worse than pERM when $h$ is small, particularly when $p/n$ is large, despite having a sharper theoretical estimation upper bound. There are several possible explanations for this phenomenon. First, comparing the estimation errors in Theorems \ref{thm: mtl} and \ref{thm: spectral mtl}, Spectral outperforms pERM only in the term that depends on the similarity level $h$. When $h$ is small, this advantage may be overshadowed by other terms, making the benefit of Spectral less apparent. However, as $h$ increases, Spectral can catch up and eventually surpass pERM in performance. Second, our theoretical analysis primarily focuses on convergence rates while ignoring constant factors, which can significantly impact practical performance. Third, our analysis does not explicitly account for the effect of $p/n$ on Spectral's performance, although we believe it plays a crucial role, particularly when $p/n$ is large. In such cases, replacing OLS with single-task ridge regression may improve performance. On the other hand, as discussed at the end of Section \ref{subsubsec: alg and upper bounds}, Spectral requires a more stringent sample size condition, namely $n \gtrsim p + \log T$, compared to $n \gtrsim r + \log T$ for pERM when $h$ is small. An even stronger requirement, $n \gtrsim r^2(p + \log T)$, is needed for ARMUL. A more refined theoretical analysis in the proportional regime $p/n \rightarrow \gamma$ with some constant $\gamma>0$ could provide deeper insights into the behaviors of Spectral and ARMUL. We leave these investigations for future work.

\subsubsection{Simulation with Different Contamination Proportion $\epsilon$}\label{subsubsec: sim epsilon}

In this subsection, we considered the linear model \eqref{eq: linear model intro} with varying contamination proportions $\epsilon = |S^c|/T$. We explored two settings with different values of $(n, p, r, T)$:
\begin{enumerate}[(i)]
	\item $n = 100, p = 50, r = 5, T = 100$;
	\item $n = 150, p = 80, r = 10, T = 100$.
\end{enumerate}
For each replication in each setting, we randomly selected a subset of size $T(1-\epsilon)$ from $[T]$ without replacement to form $S$. We set $h = 0$ and generated tasks in $S$ using the same mechanism as in Section \ref{subsubsec: sim h}. The outlier tasks in $S^c$ were generated by the linear model with $\bxk{t}_i$ i.i.d. from $N(\bm{0}_p, 2\bm{I}_p)$ and each coordinate of the coefficient $\bbetaks{t}$ was generated i.i.d. from $\textup{Unif}([-3, 3])$. We varied $\epsilon$ from 0 to $10\%$ in increments of $2\%$ and replicated each setting 100 times.

\begin{figure}[!h]
	\centering
	\includegraphics[width=\textwidth]{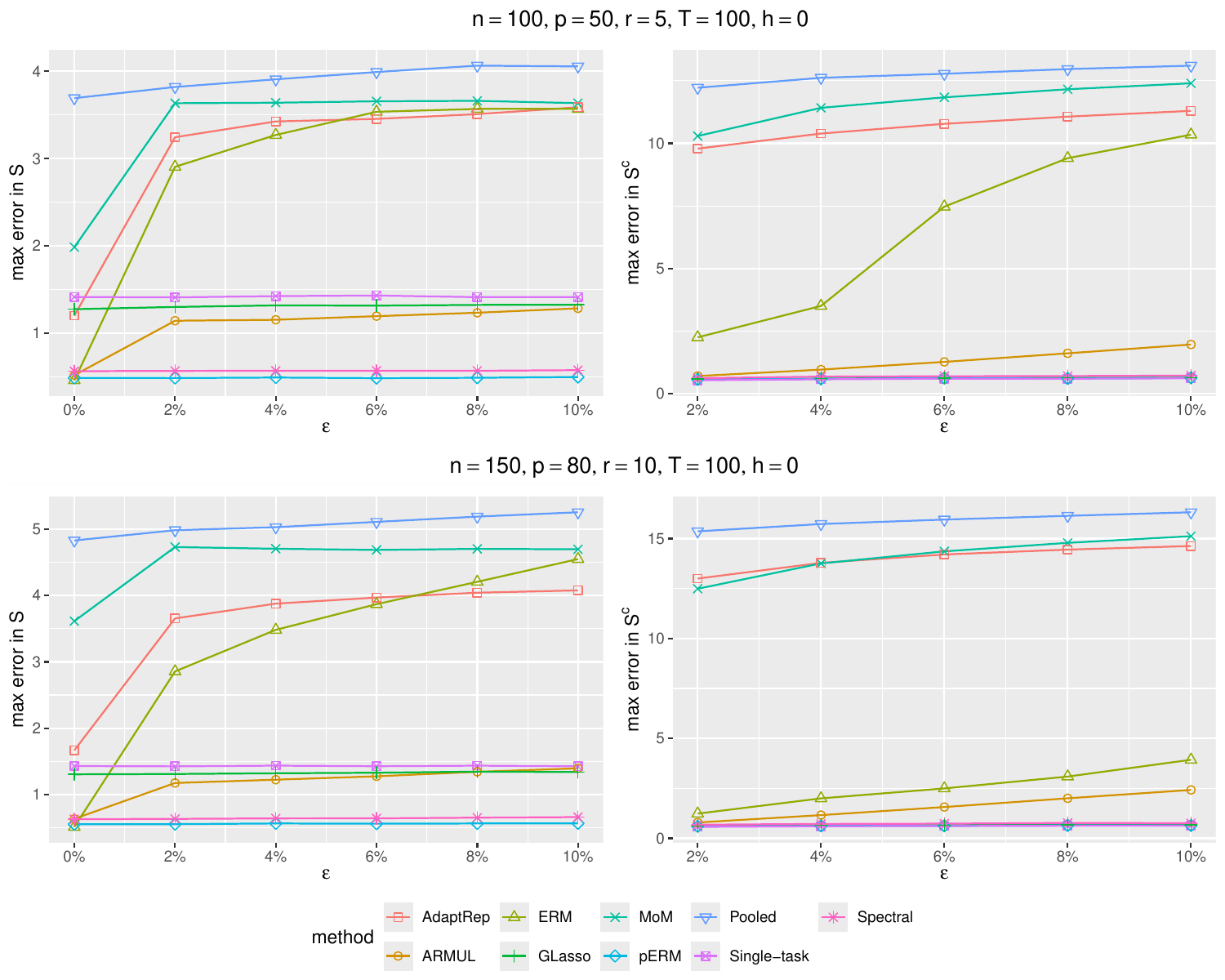}
	\caption{Simulation with different contamination proportions $\epsilon$: within-$S$ estimation error $\max_{t \in S}\twonorm{\hbetak{t}-\bbetaks{t}}$ and outlier estimation error $\max_{t \in S^c}\twonorm{\hbetak{t}-\bbetaks{t}}$ of different algorithms with different $(n, p, r, T)$ settings. ``max error in $S$" and ``max error in $S^c$" in the $y$-axis stand for $\max_{t \in S}\twonorm{\hbetak{t}-\bbetaks{t}}$ and $\max_{t \in S^c}\twonorm{\hbetak{t}-\bbetaks{t}}$, respectively. Each point represents the average over 100 replications.}
	\label{fig: sim_epsilon}
\end{figure}

The results are summarized in Figure \ref{fig: sim_epsilon}, where we evaluated each method by both $\max_{t \in S}\twonorm{\hbetak{t} - \bbetaks{t}}$ and $\max_{t \in S^c}\twonorm{\hbetak{t} - \bbetaks{t}}$. As the outlier proportion $\epsilon$ increases, the performance of most algorithms deteriorates rapidly. Even with just $2\%$ outlier tasks, all methods perform similarly to or worse than single-task regression on tasks in $S$, except for pERM and Spectral. This demonstrates the robustness of pERM and Spectral against outlier tasks. On the other hand, ARMUL does perform worse than single-task regression, even when $\epsilon$ is large. Since the outlier tasks were also generated from the linear model, our theory guarantees that pERM and Spectral can match the single-task performance on outlier tasks in $S^c$, which is indeed observed. In contrast, ARMUL performs much worse on outlier tasks in $S^c$ compared to single-task regression.

\subsubsection{Simulation with Different Number of Tasks $T$}\label{subsubsec: sim T}
\begin{figure}[!h]
	\centering
	\includegraphics[width=\textwidth]{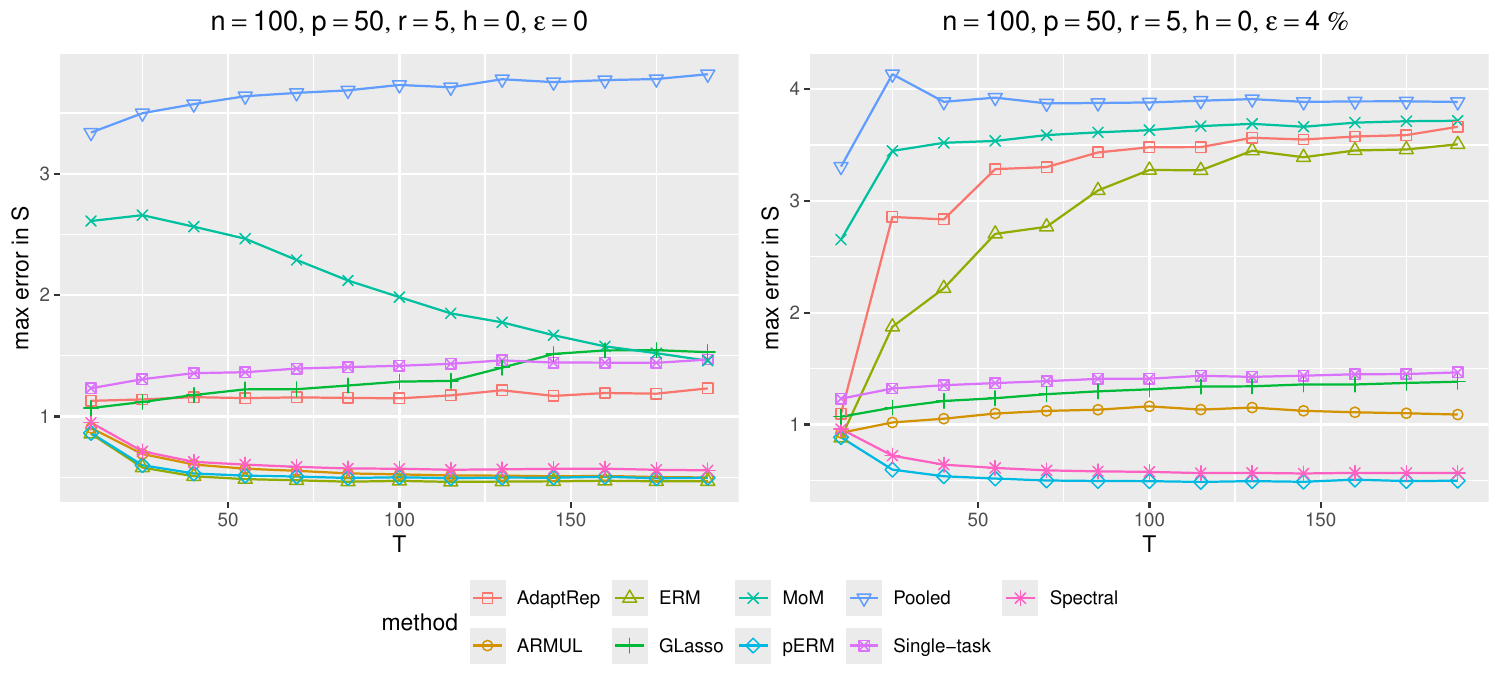}
	\caption{Simulation with different number of tasks $T$: estimation error $\max_{t \in S}\twonorm{\hbetak{t}-\bbetaks{t}}$ of different algorithms with different $(n, p, r, T)$ settings. ``max error in $S$" in the $y$-axis stands for $\max_{t \in S}\twonorm{\hbetak{t}-\bbetaks{t}}$. Each point represents the average over 100 replications.}
	\label{fig: sim_T}
\end{figure}

In this subsection, we explore the impact of the number of tasks $T$ on the performance of different methods in this subsection. We considered the linear model \eqref{eq: linear model intro} in two settings, with different values of $(n, p, r, \epsilon)$:
\begin{enumerate}[(i)]
	\item $n = 100, p = 50, r = 5, \epsilon = 0$;
	\item $n = 100, p = 50, r = 5, \epsilon = 4\%$.
\end{enumerate}
For each $(n, p, r, \epsilon)$ setting and $T$ value, we set $h = 0$ and generated data in the same way as in Section \ref{subsubsec: sim epsilon}. We increased $T$ from 10 to 190 with increments of 15. The performance of different methods on tasks in $S$ is summarized in Figure \ref{fig: sim_T}.

In the absence of outlier tasks, ERM, pERM, ARMUL, and Spectral exhibit comparable performance. As $T$ increases, their estimation errors first decrease gradually and then stabilize, aligning with the theoretical result. For example, when $\zetak{t} = \twonorm{\bthetaks{t}} \lesssim 1$ for all $t \in S=[T]$, $p \gtrsim r^2$, and $h = \epsilon = 0$, Theorem \ref{thm: mtl} implies that $\max_{t \in [T]}\twonorm{\hbetak{t} - \bbetaks{t}} \lesssim r\sqrt{\frac{p}{nT}} + r\sqrt{\frac{1}{n}}$, up to logarithmic factors with high probability. When $T$ becomes large, the error is dominated by the second term $r\sqrt{\frac{1}{n}}$, which is independent of $T$. Besides ERM, pERM, ARMUL, and Spectral, the performance of MoM also improves as $T$ increases. Similar to our findings in Sections \ref{subsubsec: sim h} and \ref{subsubsec: sim epsilon}, MoM's performance improves only when $T$ is sufficiently large. This aligns with the empirical observations in \cite{tripuraneni2021provable}, where they found that ERM always outperforms MoM until $T$ is very large, and the underlying reason is unclear.

When $\epsilon = 4\%$ tasks are contaminated, only the performance of pERM and Spectral improves as $T$ increases, and they outperform all other methods, demonstrating their robustness against outlier tasks. ARMUL and GLasso perform slightly better than single-task regression, while the other benchmark methods suffer from severe negative transfer and are significantly impacted by the outliers.

\begin{figure}[!h]
	\centering
	\includegraphics[width=\textwidth]{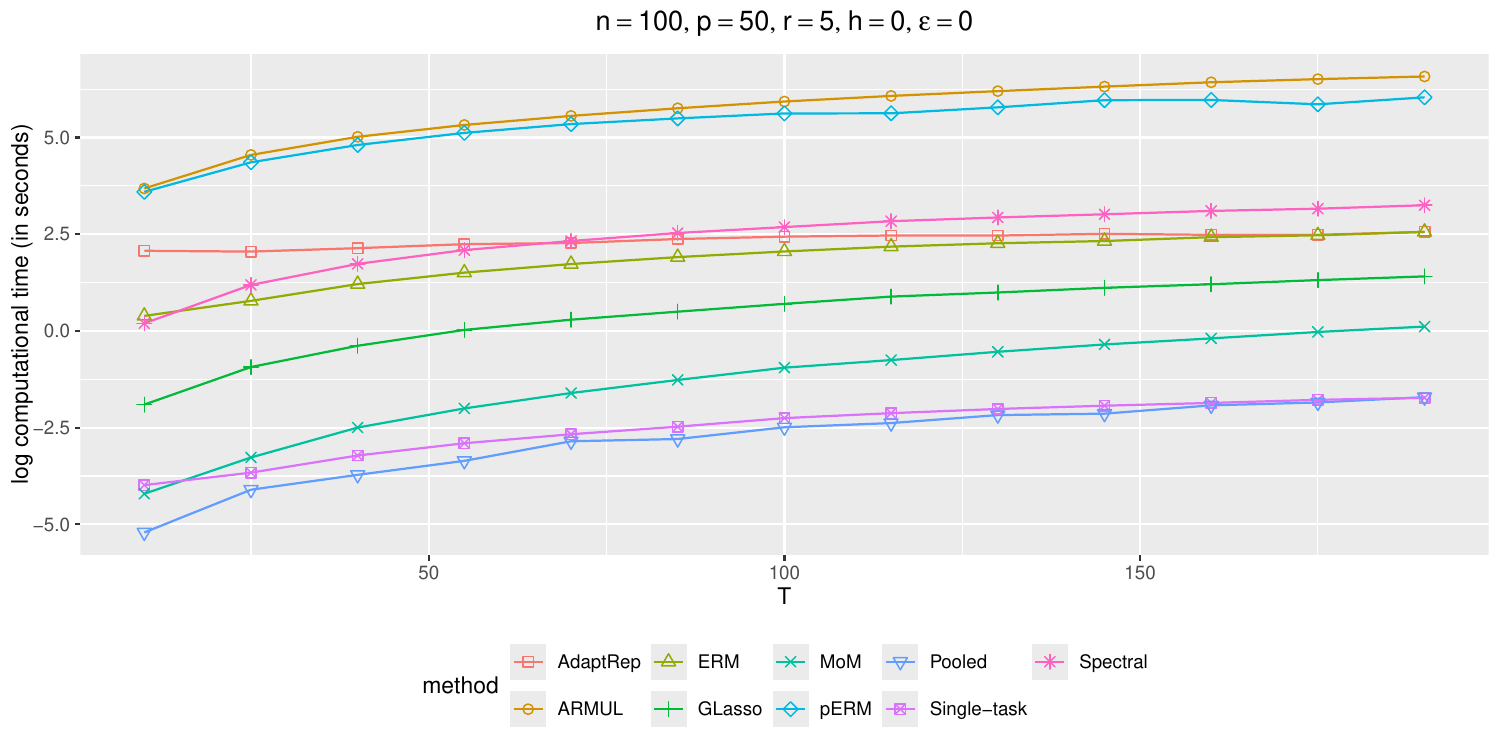}
	\caption{Computational time of different methods with various numbers of tasks $T$. The $y$-axis is in the $\log$-scale, and the unit is seconds. Each point represents the average over 100 replications.}
	\label{fig: sim_T_time}
\end{figure}

Finally, we recorded the computational time for different methods in setting (\rom{1}). The experiments were run on the Terremoto HPC Cluster of Columbia University with a CPU Intel Xeon Gold 6126 2.6 GHz. We used a single core with 3 GB of memory when running each method. The computational time for different methods with different $T$ values is plotted on a logarithmic scale in Figure \ref{fig: sim_T_time}. We can see that ARMUL and pERM are the most time-consuming methods, with pERM slightly faster than ARMUL, taking approximately $e^{6.25} \approx 500$ seconds for $T = 190$. In contrast, all the other methods can be run within $e^{3.5} \approx 30$ seconds for all values of $T$. This demonstrates the computational efficiency of the spectral method.

\subsubsection{Relationship between Task Performance and Signal Strength \texorpdfstring{$\twonorm{\bm{\theta}^{(t)*}}$}{p}}\label{subsubsec: sim theta}
In this subsection, we aim to verify our theoretical findings that the performance of ERM, pERM, and Spectral on each task can depend on the signal strength in terms of $\twonorm{\bthetaks{t}}$, a relationship not previously discussed in the literature.

\begin{figure}[!h]
	\centering
	\includegraphics[width=\textwidth]{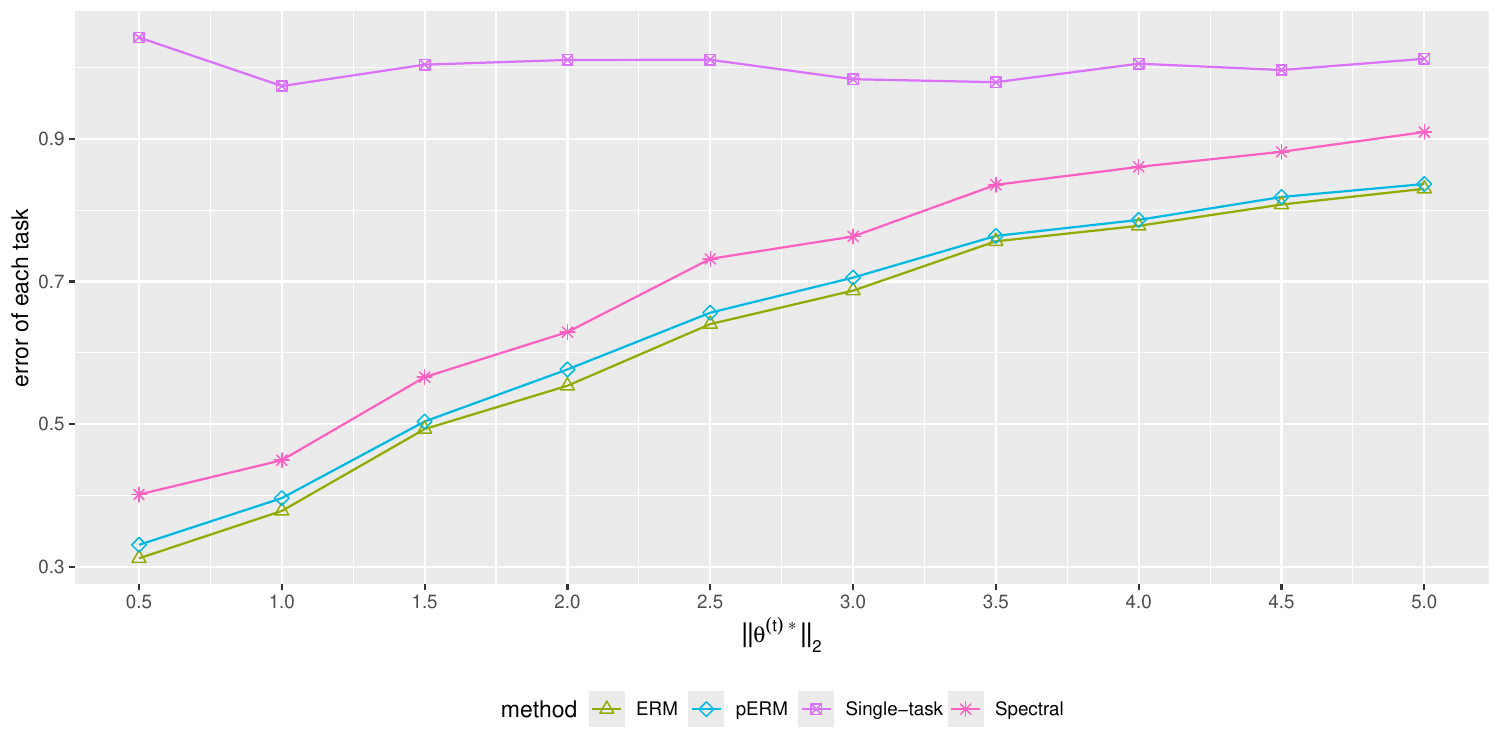}
	\caption{Simulation verifying the relationship between the estimation error $\twonorm{\hbetak{t} - \bbetaks{t}}$ and signal strength $\twonorm{\bthetaks{t}}$ on each task for different methods. ``error of each task" in the $y$-axis stands for $\twonorm{\hbetak{t} - \bbetaks{t}}$. Each point represents the average over 100 replications.}
	\label{fig: sim_theta}
\end{figure}

We generated data by a mechanism similar to that in Section \ref{subsubsec: sim h}, with $n = 100$, $p = 50$, $r = 5$, $T = 10$, $h = \epsilon = 0$. The only change in this section's data generation mechanism is that each $\bthetaks{t}$ is uniformly generated from $0.5t\mathcal{S}^{r-1}$, i.e., the sphere centered at $\bm{0}$ in $\mathbb{R}^r$ with a radius of $0.5t$. We replicated this setting 100 times and summarized the average estimation error of the single-task regression, ERM, pERM, and Spectral on each task with different $\twonorm{\bthetaks{t}}$ values in Figure \ref{fig: sim_theta}.

We can observe that the estimation error of ERM, pERM, and Spectral is approximately proportional to $\twonorm{\bthetaks{t}}$, which matches our theoretical findings. Specifically, the Pearson correlation coefficients between $\twonorm{\bthetaks{t}}$ and the average $\twonorm{\hbetak{t}-\bbetaks{t}}$ for ERM, pERM, and Spectral are 0.977, 0.975, and 0.976, respectively. In contrast, single-task regression has comparable performance across different tasks, regardless of the signal strength $\twonorm{\bthetaks{t}}$. This indicates that the benefit each individual task derives from representation MTL is highly dependent on the signal strength in terms of $\twonorm{\bthetaks{t}}$, unlike in single-task linear regression where the $\ell_2$-estimation error does not depend on the scale of the coefficient $\twonorm{\bbetaks{t}}$.

\subsubsection{Adaptivity to the Intrinsic Dimension $r$}\label{subsubsec: sim r}
In the previous simulations, we used the true intrinsic dimension $r$ in different representation MTL methods. In this subsection, we want to test the performance of our Algorithm \ref{algo: adaptation unknown r} for estimating $r$ and how it enables pERM and Spectral to adapt to unknown $r$ in practice.

\begin{figure}[!h]
	\centering
	\includegraphics[width=\textwidth]{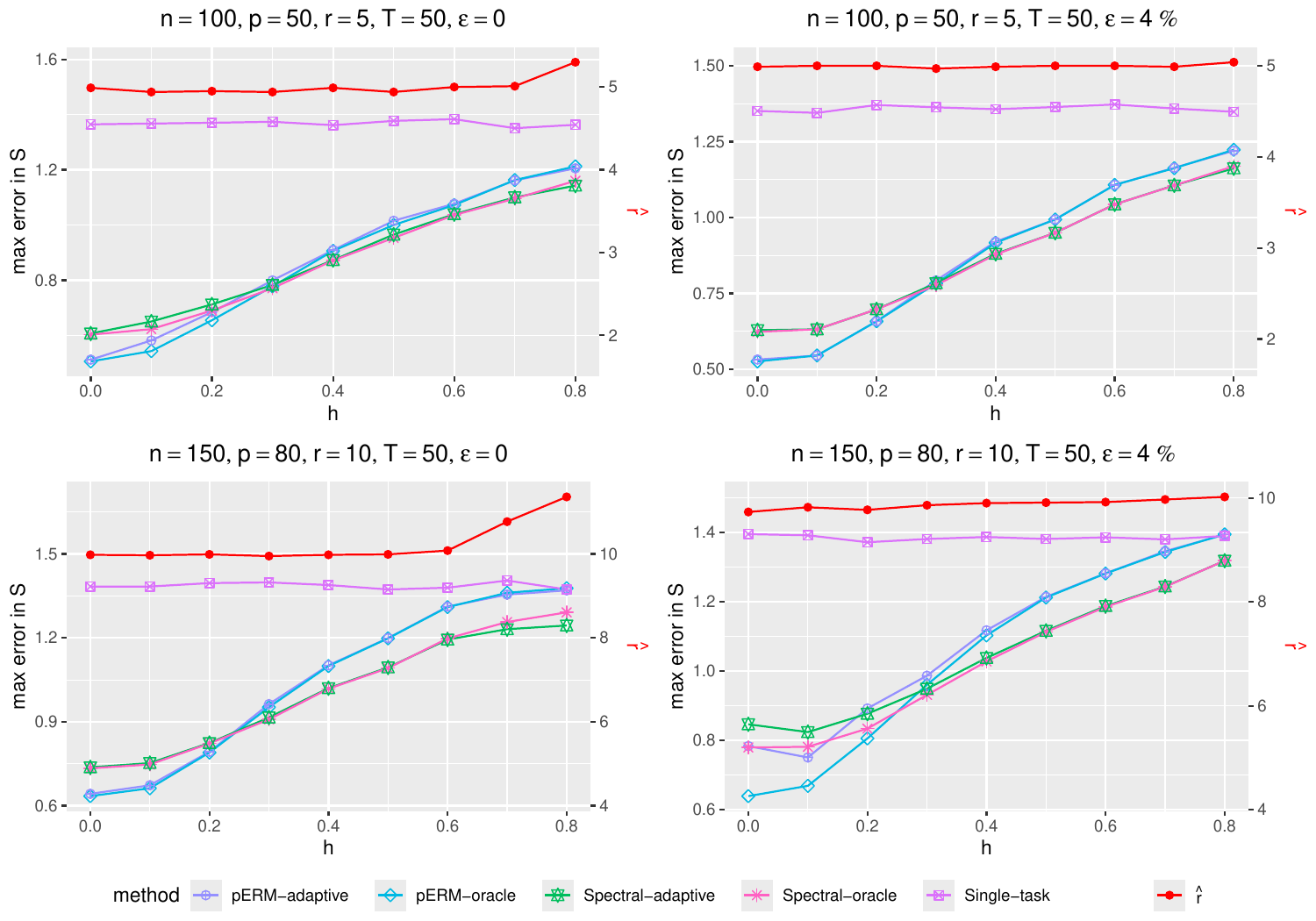}
	\caption{Simulation verifying the performance of Algorithm \ref{algo: adaptation unknown r} for estimating $r$ and its impact on helping pERM and Spectral adapt to unknown $r$. ``max error in $S$" in the left $y$-axis stands for $\max_{t \in S}\twonorm{\hbetak{t}-\bbetaks{t}}$. The red $\hat{r}$ in the right $y$-axis stands for the estimated $r$ value. Each point represents the average over 100 replications.}
	\label{fig: sim_r_adaptive}
\end{figure}

We consider the linear model \eqref{eq: linear model intro} under four settings, with different $(n, p, r, T, \epsilon)$ values:
\begin{enumerate}[(i)]
	\item $n = 100, p = 50, r = 5, T = 50, \epsilon = 0$;
	\item $n = 100, p = 50, r = 5, T = 50, \epsilon = 4\%$;
	\item $n = 150, p = 80, r = 10, T = 50, \epsilon = 0$;
	\item $n = 150, p = 80, r = 10, T = 50, \epsilon = 4\%$.
\end{enumerate}
Given each $(n, p, r, T, \epsilon)$, we increased $h$ from 0 to 0.8 in increments of 0.1, generated data for tasks in $S$ following the same mechanism used in Section \ref{subsubsec: sim h}, and generated data for tasks in $S^c$ following the same mechanism used in Section \ref{subsubsec: sim epsilon}. We first ran Algorithm \ref{algo: adaptation unknown r} to obtain an estimate $\hat{r}$ of $r$, then ran pERM and Spectral with $\hat{r}$. We denote these versions as pERM-adaptive and Spectral-adaptive, respectively. We also ran pERM and Spectral with the true $r$ value as benchmarks, and we call them pERM-oracle and Spectral-oracle, respectively. Each setting was replicated 100 times, and the average estimation error on tasks in $S$ of different approaches, as well as the average $\hat{r}$, was plotted in Figure \ref{fig: sim_r_adaptive}.

We can see that the estimate $\hat{r}$ is very close to the true $r$ in all four settings when $h$ is small, ensuring that pERM and Spectral using $\hat{r}$ perform comparably to their oracle counterparts using the true $r$. When $h$ becomes large, the estimate $\hat{r}$ may deviate from the true $r$. However, as we mentioned in Remark \ref{rmk: adaptation} of Section \ref{sec: unknown r}, there is no need to estimate $r$ precisely when $h$ is large, and biased regularization in Step 2 of pERM and Step 4 of Spectral can always guarantee the single-task performance.

\subsection{A Real-data Study}\label{subsec: real data}
In this subsection, we applied different approaches to a real data set, Human Activity Recognition (HAR) Using a Smartphones Data Set. This data set includes data collected from 30 volunteers performing six activities (walking, walking upstairs, walking downstairs, sitting, standing, and laying) with a smartphone \citep{anguita2013public}. Each observation has $p = 561$ time and frequency domain variables. We treated each volunteer as a task, with the sample size per task ranging from 281 to 409. The original data set is available at UCI Machine Learning Repository: \url{https://archive.ics.uci.edu/ml/datasets/human+activity+recognition+using+smartphones}.

We focused on the binary classification problem of discriminating between walking and standing postures (i.e., walking, walking upstairs, walking downstairs, and standing) and the others (i.e., sitting and laying). We standardized the data of each task before training different algorithms. For each task, in each replication, we used 50\% of the samples as training data and held 50\% of the sample as test data.

We ran single-task logistic regression, pooled logistic regression, ERM, ARMUL, pERM, the spectral method, and GLasso on this problem, replicated it 100 times, and summarized the average misclassification test error rates of different methods over $T = 30$ tasks with different $r$ values in Table \ref{table: har}. We can see that pERM consistently achieved the lowest error rate (tied with ERM when $r=15$) among all the methods for different $r$ values. 

\begin{table}[!h]
\centering
\begin{adjustbox}{width=1\textwidth}
\begin{tabular}{c| c c c c c c c }
\toprule
$r$/Method & Single-task & Pooled & ERM & ARMUL & pERM & Spectral & GLasso\\ \midrule
$r = 5$  & \rule{0pt}{2ex} 1.66 (0.20) & 1.79 (0.21) & 1.62 (1.42) & 2.12 (0.27) & 1.33 (0.23) & 1.85 (0.27) & 1.44 (0.25) \\ \hline
$r = 10$  & \rule{0pt}{2.5ex} 1.66 (0.20) & 1.79 (0.21) & 1.42 (0.23) & 1.77 (0.23) & 1.25 (0.20) & 1.47 (0.18) & 1.44 (0.25) \\ \hline
$r = 15$  & \rule{0pt}{2.5ex} 1.66 (0.20) & 1.79 (0.21) & 1.36 (0.23) & 1.68 (0.21) & 1.36 (1.07) & 1.50 (0.19) & 1.44 (0.25) \\ \bottomrule
\end{tabular}
\end{adjustbox}
\caption{The average mis-classification test error rates (standard deviations) of different methods over $T = 30$ tasks with different $r$ values. All values are in percentages.}
\label{table: har}
\end{table}

\section{Discussions}\label{sec: discussions}
In this work, we investigated the representation multi-task learning (MTL) problem, where most tasks share \textit{similar} linear representations, and a small fraction of tasks can be \textit{arbitrarily} contaminated. To address this problem, we proposed a penalized empirical risk minimization (ERM) method and a spectral method, and derived upper bounds for the estimation error. Our theory demonstrated that both algorithms are \textit{adaptive} to the unknown similarity level between tasks and \textit{robust} to a small fraction of outlier tasks. Additionally, the spectral method achieves a sharper estimation error bound than the penalized ERM when there is no contamination. Our theory also reveals the relationship between the performance of representation MTL methods on each task and the signal strength, which is usually ignored in the literature. We also presented the first lower bound results for estimating regression coefficients in the context of representation MTL. Our new spectral method is minimax optimal when there is no outlier task, and our penalized ERM is nearly optimal in a large regime with little impact from outlier tasks. We extended the algorithms and theory to generalized linear models and non-linear regression models. We also proposed a simple thresholding algorithm to adapt our MTL algorithms to the case of an unknown intrinsic dimension $r$. Finally, we conducted extensive numerical experiments to empirically validate our theoretical findings.

A recent paper \citep{niu2024collaborative} conducts a sophisticated and sharp analysis for the case where $h = \epsilon = 0$, accurately characterizing how the estimation error could depend on the singular values of $\bB_S^* = (\bbetaks{1} \,\, \ldots \,\, \bbetaks{T}) \in \mathbb{R}^{p \times T}$. It would be interesting to explore how their analysis extends to two of our proposed algorithms in the more general regime considered in our study.

\acks
Ye Tian is grateful to Gan Yuan (City University of Hong Kong) and Yasaman Mahdaviyeh (Columbia University) for their valuable discussions, which greatly improved the quality of this paper. He also extends his gratitude to Prof. Linjun Zhang (Rutgers University) for his insightful discussions that initially inspired this work. Additionally, Ye Tian appreciates the valuable feedback received following his presentations at the 2024 IMS-China International Conference on Statistics and Probability, at the Department of Statistics,  Iowa State University, and at the Workshop in Operations Research and Data Science (WORDS 2024) hosted by the Fuqua School of Business, Duke University. All numerical experiments were conducted on Ginsburg HPC Cluster and Terremoto HPC Cluster of Columbia University. Yuqi Gu acknowledges the support of the NSF Grant DMS-2210796. Yang Feng's research is partially supported by NIH grant 1R21AG074205-01, NSF Grant DMS-2324489, NYU University Research Challenge Fund, and a grant from NYU School of Global Public Health.
The authors are grateful to the Action Editor and four reviewers for their helpful and constructive comments.

%
%
%
%
%
%
%
%
%
%
%

\bibliographystyle{plain}
\bibliography{references.bib} 

\renewcommand{\cftsecnumwidth}{20pt}
\renewcommand{\cftsubsecnumwidth}{30pt}
\renewcommand{\cftsubsubsecnumwidth}{45pt}

\newpage
\renewcommand{\contentsname}{Content of Appendices}
\appendix

\tableofcontents
\addtocontents{toc}{\protect\setcounter{tocdepth}{5}}

\section{Extensions to More General Models}\label{sec: extensions}
In this section, we consider two extensions to the linear model \eqref{eq: linear model}.

\subsection{Generalized Linear Models}\label{subsec: glm}
A generalization to generalized linear models (GLMs) from the linear model \eqref{eq: linear model} is as follows. Suppose that the conditional distribution of $Y$ given $X = \bx$ for task $t$ is 
\begin{equation}\label{eq: GLM}
	p(\yk{t}_i = y|\bxk{t}_i = \bx) = \rho(y)\exp\big\{y\cdot \bx^\top\bbetaks{t} - \psi(\bx^\top\bbetaks{t})\big\}, \quad i = 1:n,
\end{equation} 
for $t \in S$, w.r.t. some measure $\mu$ on a subset of $\mathbb{R}$, where $\psi$ is second-order continuously differentiable on $\mathbb{R}$, and $\psi'$ is often called the inverse link function. More discussions on GLMs can be found in \cite{mccullagh1989generalized}.

\begin{example}
	Some canonical examples of GLMs include:
	\begin{enumerate}[(i)]
		\item Linear models: $\psi(u) = \frac{1}{2}u^2$, $\psi'(u) = u$, and $\mu$ is Lebesgue measure;
		\item Logistic regression models: $\psi(u) = u+\log(1+e^{-u})$, $\psi'(u) = \frac{1}{1+e^{-u}}$, and $\mu$ is the counting measure on $\{0, 1\}$;
		\item Poisson regression models: $\psi(u) = e^u$, $\psi'(u) = e^u$, and $\mu$ is the counting measure on $\mathbb{N}=\{0, 1, 2, \ldots\}$;
	\end{enumerate}
\end{example}

We replace the linear model \eqref{eq: linear model} with the GLM \eqref{eq: GLM} and keep all the other settings the same as in Section \ref{sec: mtl}. Moreover, we impose the following extra conditions for GLMs.

\begin{assumption}\label{asmp: glm}	
$\psi$ satisfies the following three conditions:
\begin{enumerate}[(i)]
	\item $\psi$ is strictly convex;
	\item $\max_{t \in S}\max_{\twonorm{\bDelta} \leq 1}\psi''((\bbetaks{t})^\top \bxk{t} + \bDelta^\top \bxk{t}) \leq C$ a.s. with a constant $C > 0$;
	\item One of the following conditions holds:
		\begin{enumerate}
			\item There exists a large constant $C' > 0$ such that $\min_{t \in S}\min_{|u|\leq C'\zetak{t}}\psi''(u) \geq C'$, where $C'' > 0$ is another universal constant.
			\item $\min_{t \in S}\min_{\twonorm{\bDelta} \leq 1}\psi''((\bbetaks{t})^\top \bxk{t} + \bDelta^\top \bxk{t}) \geq C'''$ a.s., where $C''' > 0$ is a constant.
		\end{enumerate}
\end{enumerate}
\end{assumption}

\begin{assumption}\label{asmp: n glm mtl}	
	$n \geq Cr(p+\log T)$ with a sufficiently large constant $C > 0$.
\end{assumption}

Assumption \ref{asmp: glm} is commonly used in the non-asymptotic analysis of GLMs (e.g., see \citealp{negahban2012unified, loh2015regularized}). The sample size requirement in Assumption \ref{asmp: n glm mtl} is more stringent than that for linear regression, serving as a technical condition needed for our proof. Specifically, the Hessian matrices of GLMs may not exhibit good spectral controls when evaluated far from the true parameter value. To ensure that our estimators avoid these problematic regions, we require the penalty term in Step 1 of the penalized ERM method in Algorithm \ref{algo: mtl} (which is upper bounded by $\lambda/\sqrt{n}$) to remain reasonably small. For more details, see the proof of Lemma \ref{lem: glm a theta not far} in Appendix. Note that for the spectral method in Algorithm \ref{algo: spectral}, $n \gtrsim p +\log T$ in Assumption \ref{asmp: n} is sufficient. For simplicity, we use the stronger condition in Assumption \ref{asmp: glm} for both methods. 

It is noteworthy that the function $\psi$ in \eqref{eq: GLM} is allowed to be different across tasks. Here we assume $\psi$ is the same for different tasks for simplicity. Additionally, we assume $\sqrt{r}\gtrsim \zetak{t} \gtrsim 1$ for all $t \in S$. For the GLM \eqref{eq: GLM}, we apply Algorithms \ref{algo: mtl} and \ref{algo: spectral} with $\fk{t}(\bbeta) = \frac{1}{n}\sum_{i=1}^n[-\yk{t}_i\cdot (\bxk{t}_i)^\top\bbeta + \psi((\bxk{t}_i)^\top\bbeta)]$ for $\bbeta \in \mathbb{R}^p$. With these GLM assumptions in place, Algorithms \ref{algo: mtl} and \ref{algo: spectral} achieve the same upper bounds of estimation error as in linear regression models.

\begin{theorem}[Upper bound for MTL under GLMs]\label{thm: glm mtl}
	Suppose Assumptions \ref{asmp: glm}, \ref{asmp: n glm mtl} and other conditions imposed in Theorems \ref{thm: mtl} and \ref{thm: spectral mtl} hold, the same high-probability upper bounds in Theorems \ref{thm: mtl} and \ref{thm: spectral mtl} hold for GLMs.
\end{theorem}

\subsection{Non-linear Regression}
In addition to GLMs, we can extend the linear model \eqref{eq: linear model} to a non-linear regression model as follows. Suppose 
\begin{equation}\label{eq: non-linear model}
	\yk{t}_i = g\big((\bxk{t}_i)^\top\bbetaks{t}\big) + \epsilonk{t}_i, \quad i = 1:n,
\end{equation}
for $t \in S$, where $g$ is a monotone function with a continuous second-order derivative on $\mathbb{R}$ and $\{\epsilonk{t}_i\}_{i=1}^n$ are i.i.d. zero-mean sub-Gaussian variables independent of $\{\bxk{t}_i\}_{i=1}^n$. In literature, $g$ is often referred to as the link function. More discussions for this model under a single-task learning setting can be found in \cite{yang2015sparse}. Note that \cite{yang2015sparse} considered the case of a fixed design while we considered the random design case, which is more challenging. Hence, stronger conditions are necessary to guarantee the desired rate.

We replace the linear model \eqref{eq: linear model} with the non-linear regression model \eqref{eq: non-linear model}, and keep all the other settings the same as in Section \ref{sec: mtl}. Furthermore, we impose the following assumptions for non-linear regression models. Recall the notation $\bSigmak{t} = \tE[\bxk{t}(\bxk{t})^{T}]$ and denote $\bSigmak{t}_{\epsilon} = \tE[\bepsilonk{t}(\bepsilonk{t})^{T}]$.

\begin{assumption}\label{asmp: non-linear}
	$0 < C_1 \leq [g'(u)]^2 \leq C_2 < \infty$, for all $u \in \mathbb{R}$, and $\infnorm{g''} \leq C_3 <\infty$, where $C_3 \leq cC_1^2 \cdot \min_{t \in S}\{\lambda_{\min}(\bSigmak{t})/\lambda_{\max}(\bSigmak{t})\} \min_{t \in S}\{\lambda_{\max}^{-1/2}(\bSigmak{t}_{\epsilon})\}$ with a small constant $c$. 
\end{assumption}

\begin{assumption}\label{asmp: non-linear n}
	$n \geq C[(p+\log T)^{3/2}]\vee [r(p+\log T)]$ with a sufficiently large constant $C$. 
\end{assumption}

The sample size requirement is stronger than that for linear regression. The first term $(p+\log T)^{3/2}$ arises due to the heavy-tailed distributions appearing in the analysis. A similar requirement for the high-dimensional sparse non-linear regression can be found in \cite{yang2015sparse}. The reason for the term $r(p+\log T)$ is the same as in Section \ref{subsec: glm} for GLMs. More details can be found in the proof of Lemma \ref{lem: non-linear a theta not far} in Appendix.

Similar to the case of GLMs, the link function $g$ in \eqref{eq: non-linear model} is also allowed to vary across tasks. Here we assume $g$ is the same for different tasks for simplicity. We further assume $\sqrt{r}\gtrsim \zetak{t} \gtrsim 1$ for all $t \in S$. We apply Algorithms \ref{algo: mtl} and \ref{algo: spectral} for the non-linear regression model \eqref{eq: non-linear model}, by setting $\fk{t}(\bbeta) = \frac{1}{2n}\sum_{i=1}^n[\yk{t}_i - g((\bxk{t}_i)^\top\bbeta)]^2$ for $\bbeta \in \mathbb{R}^p$. We have the same upper bounds of estimation error for Algorithms \ref{algo: mtl} and \ref{algo: spectral} as in linear regression models.

\begin{theorem}[Upper bound for MTL under non-linear regression models]\label{thm: non-linear mtl}
	Suppose Assumptions \ref{asmp: non-linear}, \ref{asmp: non-linear n} and other conditions imposed in Theorems \ref{thm: mtl} and \ref{thm: spectral mtl} hold, then the same high-probability upper bounds in Theorems \ref{thm: mtl} and \ref{thm: spectral mtl} hold for non-linear regression models.
\end{theorem}

\section{Transferring to New Tasks (Learning-to-learn)}\label{sec: tl}
In this section, we extend the MTL framework discussed in the main text to a transfer learning (TL) setting.

\subsection{Problem Set-up}

In this section, in addition to data from the $T$ tasks, suppose we also observe data $\{(\bxk{0}_i, \yk{0}_i)\}_{i=1}^{n_0}$ from a new task
\begin{equation}\label{eq: linear model tl}
	\yk{0}_i = (\bxk{0}_i)^\top\bbetaks{0} + \epsilonk{0}_i, \quad i = 1:n_0,
\end{equation}
where $\bbetaks{0}=\bAks{0}\bthetaks{0}$, $\bAks{0} \in \mO = \{\bA \in \mathbb{R}^{p \times r}: \bA^\top \bA = \bm{I}_r\}$, $\bthetaks{0} \in \mathbb{R}^r$, and $\{\epsilonk{0}_i\}_{i=1}^{n_0}$ are i.i.d. zero-mean sub-Gaussian variables independent of $\{\bxk{0}_i\}_{i=1}^{n_0}$. Similar to Section \ref{sec: mtl}, the intrinsic dimension $r$ is assumed to be known. Under such a transfer learning (TL) or learning-to-learn setting, the new task is often called the \textit{target} task, and the $T$ tasks are called \textit{source} tasks. Our goal is two-fold:
\begin{enumerate}
	\item Transfer knowledge from source tasks to improve the learning performance on the \textit{target} task, when the source and target share ``similar'' representations and the number of outlier source tasks is small;
	\item Ensure the learning performance is no worse than the target-only learning performance to avoid the negative transfer.
\end{enumerate} 
To describe the similarity between target and source representations, we assume
\begin{equation}
	\max_{t \in S}\twonorm{\bAks{t}(\bAks{t})^\top - \bAks{0}(\bAks{0})^\top} \leq h,
\end{equation}
where $S$ is a subset of $[T]$. Similar to the setting in the last section, the joint distribution $\mQ_{S^c}$ of data from source tasks in $S^c$, i.e., $\{\{\bxk{t}_i, \yk{t}_i\}_{i=1}^n\}_{t\in S^c}$, is allowed to be arbitrary. For simplicity, we focus on the regime that $\twonorm{\bthetaks{t}} \leq C < \infty$ for all $t \in S$. Denote $\bSigmak{0} = \tE [\bxk{0}(\bxk{0})^\top]$. We impose the following assumptions on the target task.

\begin{assumption}\label{asmp: x tl}
	For $\bm{u} \in \mathbb{R}^p$, $\bu^\top\bxk{0}$ is sub-Gaussian in the sense that $\tE[e^{\lambda \bu^\top\bxk{0}}] \leq e^{C\lambda^2\twonorm{\bm{u}}^2}$ for any $\lambda \in \mathbb{R}$ with some constant $C > 0$. And there exist constants $c, C$ such that $0<c \leq \lambda_{\min}(\bSigmak{0}) \leq \lambda_{\max}(\bSigmak{0}) \leq C < \infty$.
\end{assumption}

\begin{assumption}\label{asmp: n tl}
	$n_0 \geq Cp$ with a sufficiently large constant $C > 0$. 
\end{assumption}

\begin{remark}\label{rmt: few shot learning}
	Similar to our discussion in Section \ref{subsubsec: alg and upper bounds} for the penalized ERM in MTL, Assumption \ref{asmp: n tl} is imposed to guarantee the target-only rate $\sqrt{p/n_0}$. If we do not care about this safe-net guarantee, then it suffices to require $n_0 \geq Cr$, the same as the condition imposed in literature \citep{tripuraneni2021provable, thekumparampil2021statistically}. The RHS of the following Theorem \ref{thm: tl} shall be replaced by $r\sqrt{\frac{p}{nT}} + \sqrt{r}h + \sqrt{r}\sqrt{\frac{r+\log T}{n}} + \sqrt{\frac{p}{n}}\cdot \frac{|S^c|}{T}r^{3/2} + \sqrt{\frac{r}{n_0}}$, which allows us for a few-shot learning when both $h$ and $|S^c|/T$ are sufficiently small. See more details in the proof of Theorem \ref{thm: tl} in Appendix.
\end{remark}

\subsection{Upper Bounds}\label{subsec: upper bdd tl}
Similar to the Algorithm \ref{algo: mtl}, a two-step transfer learning method is proposed in Algorithm \ref{algo: tl}. We introduce the algorithm with a general loss function $\fk{0}$ for the target since the same algorithm with different losses will be extended to other models later. For linear model \eqref{eq: linear model tl}, define $\fk{0}(\bbeta) = \frac{1}{2n_0}\twonorm{\bYk{0} - \bXk{0}\bbeta} = \frac{1}{2n_0}\sum_{i=1}^{n_0} [\yk{0}_i - (\bxk{0}_i)^\top\bbeta]^2$ for $\bbeta \in \mathbb{R}^p$, where $\bXk{0} \in \mathbb{R}^{n_0 \times p}$ and $\bYk{0} \in \mathbb{R}^{n_0}$ are corresponding matrix/vector representations of target data.

\begin{algorithm}[!h]
\caption{Transferring to new tasks}
\label{algo: tl}
\KwIn{Data from a new task $(\bXk{0}, \bYk{0}) = \{\bxk{0}_i, \yk{0}_i\}_{i=1}^{n_0}$, estimator $\hbarA$ from Algorithm \ref{algo: mtl} or \ref{algo: spectral},  penalty parameter $\gamma$}
\KwOut{Estimator $\hbetak{0}$}
\underline{Step 1:}  $\hthetak{0} = \argmin_{\btheta \in \mathbb{R}^r}\big\{ \fk{0}(\hbarA\btheta)\big\}$ \\
\underline{Step 2:} $\hbetak{0} = \argmin_{\bbeta \in \mathbb{R}^p} \big\{\fk{0}(\bbeta) + \frac{\gamma}{\sqrt{n_0}}\twonorm{\bbeta - \hbarA\hthetak{0}}\big\}$
\end{algorithm}

In Algorithm \ref{algo: tl}, the ``central representation" $\hbarA$ learned by Algorithm \ref{algo: mtl} is passed to Step 1 to obtain the estimator $\hthetak{0}$ of the target-specific low-dimensional parameter. The same step has appeared in literature when there are no outliers, and the target and source share the same representations \citep{du2020few, tripuraneni2021provable}. Step 2 is similar to Step 2 of Algorithm \ref{algo: mtl}, which guarantees the target-only rate even when the representations of the target and sources are dissimilar.

We have the following upper bounds of target estimation error for Algorithm \ref{algo: tl}.

\begin{theorem}[Upper bound for TL]\label{thm: tl}
	Suppose Assumptions \ref{asmp: x}, \ref{asmp: theta}, \ref{asmp: n}, \ref{asmp: x tl}, and \ref{asmp: n tl} hold. 
	\begin{enumerate}[(i)]
		\item (TL with penalized ERM) By setting $\lambda = Cr^{3/4}\sqrt{p+\log T}$ and $\gamma = C'\sqrt{p+\log T}$ with sufficiently large constants $C, C' > 0$ in Algorithm \ref{algo: tl} along with Algorithm \ref{algo: mtl} to learn $\hoA$, for all $S \subseteq [T]$ satisfying $\epsilon = \frac{|S^c|}{T} \leq cr^{-3/2}$ with a small constant $c > 0$ and an arbitrary distribution $\mQ_{S^c}$ of $\{\{\bxk{t}_i, \yk{t}_i\}_{i=1}^n\}_{t\in S^c}$, w.p. at least $1-e^{-C''(r+\log T)}$, we have
			\begin{equation}
				\twonorm{\hbetak{0}-\bbetaks{0}} \lesssim \left(r\sqrt{\frac{p}{nT}} + \sqrt{r}h + \sqrt{r}\sqrt{\frac{r+\log T}{n}} + \sqrt{\frac{p}{n}}\cdot \epsilon r^{3/2}\right) \wedge \sqrt{\frac{p}{n_0}}  + \sqrt{\frac{r}{n_0}}.
			\end{equation}
		\item (TL with the spectral method) By setting $\gamma = C'\sqrt{p+\log T}$ with a sufficiently large positive constant $C'$, for any subset $S \subseteq [T]$ satisfying $\epsilon = \frac{|S^c|}{T} \leq cr^{-1}\cdot \Big(\frac{\barzeta}{\max_{t \in S}\zetak{t}}\Big)^2$ with $c > 0$ a small constant, w.p. at least $1-e^{-C''(r+\log T)}$ and $\bar{\epsilon}$ satisfying $\epsilon \leq \bar{\epsilon} \leq \frac{c''}{r}\cdot \Big(\frac{\barzeta}{\max_{t \in S}\zetak{t}}\Big)^2$ with a sufficiently small positive constant $c''$, we have
		\begin{equation}
			\twonorm{\hbetak{0}-\bbetaks{0}} \lesssim \Bigg\{\sqrt{\frac{pr}{nT}} + \sqrt{\frac{r}{n}}  + h\cdot \bigg[\frac{\sigma_{\max}((\oA^\perp)^\top \bm{B}^*_S)}{\sigma_{\min}((\oA^\perp)^\top \bm{B}^*_S)} \wedge \sqrt{r}\bigg] + \sqrt{r\bar{\epsilon}} \Bigg\} \wedge \sqrt{\frac{p}{n_0}} + \sqrt{\frac{r}{n_0}},		\end{equation}
		where $\bm{B}^*_S \in \mathbb{R}^{p \times |S|}$ is the coefficient matrix whose columns are $\{\bbetaks{t}\}_{t \in S}$, $\oA \in \argmin_{\bm{A} \in \mathcal{O}^{p \times r}}\allowbreak \max_{t \in [T]}\twonorm{\bAks{t}(\bAks{t})^\top - \bA\bA^\top}$ is the central representation, and $\oA^\perp$ is orthogonal to $\oA$ in the sense that $\oA^\perp \in \mathcal{O}^{p\times (p-r)}$ and $(\oA^\perp)^\top \oA = \bm{0}_{(p-r)\times r}$.
	\end{enumerate}
\end{theorem}

Both upper bounds can be seen as the minimum of two terms which represent the rate of learning target model via data aggregation and the target-only rate $\sqrt{\frac{p}{n_0}}$, respectively. This rate entails that our algorithm is \textbf{adaptive} to the optimal situation regardless of whether transferring from source to target is beneficial. Moreover, it is \textbf{robust} to a small fraction of outlier source tasks, in the sense that TL is still helpful when the outlier proportion $\epsilon$ is sufficiently small.

%

\subsection{Lower Bounds}\label{subsec: lower bdd tl}
In this subsection, we explore the lower bound of the TL problem. Consider the space for all subsets $S \subseteq [T]$ as
\begin{equation}
	\mS = \{S \subseteq [T]: |S^c|/T \leq \epsilon\}.
\end{equation}
Given the subset $S$, consider the parameter spaces for the coefficient vectors $\{\bbetak{t}\}_{t\in \{0\}\cup S}$ as
\begin{align}
	\mathscr{B}_0(S, h) =& \bigg\{\{\bbetak{t}\}_{t\in \{0\}\cup S}: \bbetak{t} = \bAk{t}\bthetak{t} \text{ for all } t \in \{0\}\cup S, \{\bAk{t}\}_{t\in \{0\}\cup S} \subseteq \mO,\\
	& \max_{t \in \{0\}\cup S}\twonorm{\bthetak{t}} \leq C, \max_{t \in S}\twonorm{\bAk{t}(\bAk{t})^\top - \bAk{0} (\bAk{0})^\top} \leq h, \sigma_r\Big(|S|^{-1/2}\bm{B}^*_S\Big) \geq \frac{c}{\sqrt{r}}\bigg\}
\end{align}
where $C$ and $c$ can be any fixed positive constants such that $\mathscr{B}_0(S, h) \neq \emptyset$. 

\begin{theorem}[Lower bound for TL]\label{thm: tl lower bdd}
	Suppose $p \geq 2r$ and $\epsilon \leq c/r$ where $c$ is a small constant. We have the following lower bound:
	\begin{align}
		\inf_{\hbetak{0}}\sup_{S \subseteq \mathcal{S}} \sup_{\{\bbetak{t}\}_{t\in \{0\}\cup S} \in \mathscr{B}_0(S, h)}\tP \bigg(\twonorm{\hbetak{0}-\bbetaks{0}} \gtrsim \bigg(\sqrt{\frac{pr}{nT}} + h\bigg)\wedge \sqrt{\frac{p}{n_0}}  + \sqrt{\frac{r}{n_0}} + \frac{\epsilon r}{\sqrt{n}} \wedge \sqrt{\frac{1}{n_0}}\bigg) \geq \frac{1}{10}.
	\end{align}
\end{theorem}

To our knowledge, this is the first lower bound for learning regression parameters under representation transfer learning. Comparing the upper and lower bounds of the representation TL problem, we can see that the upper bound of penalized ERM has suboptimal dependence on $r$ and $\log T$. The spectral method has a sharper upper bound when $S = [T]$ (i.e. $\epsilon = 0$) with $\bar{\epsilon}=0$, and is minimax optimal when the condition number $\frac{\sigma_{\max}((\oA^\perp)^\top \bm{B}^*_S)}{\sigma_{\min}((\oA^\perp)^\top \bm{B}^*_S)}$ is bounded and $n \gtrsim n_0$.

Similar to the MTL case, we can extend the upper bound from the linear model to the GLMs and non-linear regression models.

\begin{theorem}[Upper bound for TL under GLMs]\label{thm: glm tl}
	Suppose Assumptions \ref{asmp: glm}, \ref{asmp: n glm mtl} and other conditions imposed in Theorem \ref{thm: tl} hold, then the same high-probability upper bounds in Theorem \ref{thm: tl} hold for GLMs.
\end{theorem}

\begin{theorem}[Upper bound for TL under non-linear regression models]\label{thm: non-linear tl}
	Suppose Assumptions \ref{asmp: non-linear}, \ref{asmp: non-linear n}, and other conditions imposed in Theorem \ref{thm: tl} hold. Let $n_0 \geq Cp^{3/2}$ with $C>0$ a sufficiently large constant. The same high-probability upper bounds in Theorem \ref{thm: tl} hold for non-linear regression models.
\end{theorem}

\section{General Lemmas}

\subsection{Lemmas}

\begin{lemma}[Theorem 6.5 in \cite{wainwright2019high}]\label{lem: cov hat}
	Suppose that Assumptions \ref{asmp: x} and \ref{asmp: n} hold. Then for any $\delta > 0$ and any $t \in [T]$, w.p. at least $1-C_1e^{-nC_2(\delta\wedge \delta^2)}$,
	\begin{equation}
		\twonorm{\hSigmak{t}-\bSigmak{t}} \leq C_3\sqrt{\frac{p}{n}} + \delta,
	\end{equation}
	with some constants $C_1, C_2, C_3 > 0$. Note that $C_1, C_2, C_3$ are universal in the sense that they do not depend on $t$ or $\delta$. As a consequence, we have
	\begin{equation}
		\max_{t\in [T]}\twonorm{\hSigmak{t}-\bSigmak{t}} \lesssim \sqrt{\frac{p+\log T}{n}},
	\end{equation}
	w.p. at least $1-e^{-C(p+\log T)}$.
\end{lemma}

\begin{lemma}[A variant of Theorem 6.5 in \cite{wainwright2019high}]\label{lem: cov hat r}
	Suppose that Assumptions \ref{asmp: x tl} and \ref{asmp: n tl} hold. Then for any $\delta > 0$, for any fixed $\bA, \bm{B} \in \mO$, w.p. at least $1-C_1e^{-n_0C_2(\delta\wedge \delta)}$,
	\begin{equation}
		\twonorm{\bA^\top\hSigmak{t}\bm{B}-\bA^\top\bSigmak{t}\bm{B}} \leq C_3\sqrt{\frac{r}{n_0}} + \delta,
	\end{equation}
	with some constants $C_1, C_2, C_3 > 0$. Note that $C_1, C_2, C_3$ are universal in the sense that they do not depend on $\delta$ or $t$. As a consequence, for any fixed $\bA, \bm{B} \in \mO$, we have
	\begin{equation}
		\twonorm{\bA^\top\hSigmak{0}\bm{B}-\bA^\top\bSigmak{0}\bm{B}} \lesssim \sqrt{\frac{r}{n_0}},
	\end{equation}
	w.p. at least $1-e^{-C(r+\log T)}$.
\end{lemma}

\begin{lemma}[Lemmas 2.5 and 2.6 in \cite{chen2021spectral}]\label{lem: distance equivalence}
	Suppose $p \geq r$. Consider two matrics $\bA, \widetilde{\bA} \in \mO$. Suppose that $(\bA, \bA^{\perp}), (\widetilde{\bA}, \widetilde{\bA}^{\perp}) \in \mathbb{R}^{p \times p}$ are both orthonormal matrices, which means that  $\bA^{\perp}$ and $\widetilde{\bA}^{\perp}$ are orthonormal complements of $\bA$ and $\widetilde{\bA}$, respectively. Then
	\begin{align}
		\twonorm{\bA\bA^\top - \widetilde{\bA}(\widetilde{\bA})^\top} &= \twonorm{\bA^\top(\widetilde{\bA}^{\perp})} = \twonorm{(\widetilde{\bA})^\top\bA^{\perp}},\\
		\frac{1}{\sqrt{2}}\fnorm{\bA\bA^\top - \widetilde{\bA}(\widetilde{\bA})^\top} &= \fnorm{\bA^\top(\widetilde{\bA}^{\perp})} = \fnorm{(\widetilde{\bA})^\top\bA^{\perp}},\\
		\twonorm{\bA\bA^\top - \widetilde{\bA}(\widetilde{\bA})^\top} &\leq \min_{\bR \in \mOr}\twonorm{\bA - \widetilde{\bA}\bR} \leq \sqrt{2}\twonorm{\bA\bA^\top - \widetilde{\bA}(\widetilde{\bA})^\top},\\
		\frac{1}{\sqrt{2}}\fnorm{\bA\bA^\top - \widetilde{\bA}(\widetilde{\bA})^\top} &\leq \min_{\bR \in \mOr}\fnorm{\bA - \widetilde{\bA}\bR} \leq \fnorm{\bA\bA^\top - \widetilde{\bA}(\widetilde{\bA})^\top}.
	\end{align}
\end{lemma}

\begin{lemma}[Proposition 8 in \cite{pajor1998metric}]\label{lem: A covering}
	When $p \geq 2r$, for any $\delta > 0$, we have
	\begin{enumerate}[(i)]
		\item $(C1\frac{\sqrt{r}}{\delta})^{r(p-r)} \leq N(\mO, \distf, \delta) \leq C_2(\frac{\sqrt{r}}{8})^{r(p-r)}$;
		\item $(C_1\frac{1}{\delta})^{r(p-r)} \leq N(\mO, \disttwo, \delta) \leq (C_2\frac{1}{\delta})^{r(p-r)}$;
		\item (Consequence of (\rom{1})) $M(\mO, \distf, \delta) \geq C(\frac{\sqrt{r}}{\delta})^{r(p-r)}$.
	\end{enumerate}
\end{lemma}

\begin{lemma}\label{lem: ball packing}
	When $p \geq 2r$, for any $\delta > 0$ and $\alpha \in (0, 1)$, $\exists \barA \in \mO$, such that 
	\begin{equation}
		M(\mB_{\delta}(\barA, \mO, \disttwo), \distf, \alpha\delta) \geq \left(\frac{C\sqrt{r}}{\alpha}\right)^{r(p-r)}.
	\end{equation}
\end{lemma}

\begin{lemma}[Fano's lemma, see \cite{tsybakov2009introduction} and \cite{wainwright2019high}]\label{lem: fano}
	Suppose $(\Theta, d)$ is a metric space and each $\theta$ in this space is associated with a probability measure $\tP_{\theta}$. If $\{\theta_j\}_{j = 1}^N$ is an $s$-separated set (i.e. $d(\theta_j, \theta_k) \geq s$ for any $j \neq k$), and $\textup{KL}(\tP_{\theta_j}, \tP_{\theta_k}) \leq \alpha \log N$, then
	\begin{equation}
		\inf_{\widehat{\theta}}\sup_{\theta \in \Theta}\tP_{\theta}(d(\widehat{\theta}, \theta) \geq s/2) \geq 1-\alpha -\frac{\log 2}{\log N}.
	\end{equation}
\end{lemma}

\begin{lemma}\label{lem: KL}
	Consider the following generative model:
	\begin{equation}
		y|\bx \sim \tP_{y|\bx, \bbeta} = N(\bx^\top\bbeta, 1), \quad \bx \sim \tP_{\bx},
	\end{equation}
	where $\tP_{\bx}$ is sub-Gaussian with $\bSigma = \tE(\bx\bx^\top)$. Suppose there exist constants $c$ and $C$ such that $0<c \leq \lambda_{\min}(\bSigma) \leq \lambda_{\max}(\bSigma) \leq C < \infty$.
	Define two distributions $\tP$ and $\widetilde{\tP}$ of $(\bx, y)$ as $\tP_{\bx, y} = \tP_{y|\bx; \bbeta}\cdot \tP_{\bx}$ and $\widetilde{\tP}_{\bx, y} = \tP_{y|\bx; \widetilde{\bbeta}}\cdot \tP_{\bx}$. Then, the KL divergence between $\tP_{\bx, y}$ and $\widetilde{\tP}_{\bx, y}$ can be bounded as
	\begin{equation}
		\KL(\tP_{\bx, y}\|\widetilde{\tP}_{\bx, y}) \leq C\twonorm{\bbeta - \widetilde{\bbeta}}^2.
	\end{equation}
\end{lemma}

\begin{lemma}[Theorem 5.1 in \cite{chen2018robust}]\label{lem: from chen}
	Given a family of distributions $\{\tP_{\theta}: \theta \in  \Theta\}$,  which is indexed by a parameter $\theta \in  \Theta$. Consider $\bxk{t} \sim (1-\epsilon')\tP_{\theta} + \epsilon'\mathbb{Q}$ independently for $t \in [T]$, and $\widetilde{\bx} \sim \tP_{\widetilde{\bx}}$. Denote the joint distribution of $\{\bxk{t}\}_{t=1}^T$ and $\widetilde{x}$ as $\tP_{(\epsilon', \theta, \mQ)}\cdot \tP_{\widetilde{\bx}}$. Then
	\begin{equation}
		\inf_{\widehat{\theta}} \sup_{\substack{\theta \in \Theta \\ \mQ}} (\tP_{(\epsilon', \theta, \mQ)}\cdot \tP_{\widetilde{\bx}})\left(\|\widehat{\theta}-\theta\| \geq C\varpi(\epsilon', \Theta)\right) \geq \frac{1}{2},
	\end{equation}
	where $\varpi(\epsilon', \Theta) \coloneqq \sup\{\|\theta_1-\theta_2\|: \textup{TV}\big(\tP_{\theta_1}, \tP_{\theta_2}\big) \leq \epsilon'/(1-\epsilon'), \theta_1, \theta_2 \in \Theta\}$.
\end{lemma}

\begin{lemma}[Lemma 22 in \cite{tian2022unsupervised}]\label{lem: binomial lower bound}
	Consider two data generating mechanisms:
		\begin{enumerate}[(i)]
			\item $\bxk{t} \sim (1-\epsilon')\tP_{\theta} + \epsilon' \mathbb{Q}$ independently for $t \in [T]$, where $\epsilon' = \frac{|S^c|}{T}$, and $\widetilde{x} \sim \tP_{\widetilde{\bx}}$;
			\item With a preserved set $S \subseteq [T]$, generate $\{\bxk{t}\}_{t \in S^c} \sim \mathbb{Q}_{S^c}$ and $\bxk{t} \sim \tP_{\theta}$ independently for $t \in S$, and $\widetilde{x} \sim \tP_{\widetilde{\bx}}$.
		\end{enumerate}
	Denote the joint distributions of $\{\bxk{t}\}_{t=1}^T$ and $\widetilde{x}$ in (\rom{1}) and (\rom{2}) as $\tP_{(\epsilon, \theta, \mathbb{Q})}\cdot \tP_{\widetilde{\bx}}$ and $\tP_{(S, \theta, \mathbb{Q}_{S^c})}\cdot \tP_{\widetilde{\bx}}$, respectively. We claim that if
	\begin{equation}
		\inf_{\widehat{\theta}} \sup_{\substack{\theta \in \Theta \\ \mQ}} (\tP_{(\epsilon'/50, \theta, \mQ)}\cdot \tP_{\widetilde{\bx}})\left(\|\widehat{\theta}-\theta\| \geq C\varpi(\epsilon'/50, \Theta)\right) \geq \frac{1}{2}
	\end{equation} 
	then
	\begin{equation}\label{eq: conclusion binomial lemma}
		\inf_{\widehat{\theta}} \sup_{S: |S| \geq T(1-\epsilon')}\sup_{\substack{\theta \in \Theta \\ \mQ_{S^c}}} (\tP_{(S, \theta, \mQ_{S^c})}\cdot \tP_{\widetilde{\bx}})\left(\|\widehat{\theta}-\theta \| \geq C\varpi\left(\epsilon'/50, \Theta \right)\right) \geq \frac{1}{10},
	\end{equation}
	where $\varpi(\epsilon', \Theta) \coloneqq \sup\{\|\theta_1-\theta_2\|: \textup{TV}\big(\tP_{\theta_1}, \tP_{\theta_2}\big) \leq \epsilon'/(1-\epsilon'), \theta_1, \theta_2 \in \Theta\}$.
\end{lemma}

\begin{lemma}[Lemma 33 in \cite{tian2022unsupervised}]\label{lem: from chen second}
	Given a family of distributions $\{\tP_{\theta}: \theta \in  \Theta\}$,  which is indexed by a parameter $\theta \in  \Theta$. Consider $\bxk{t} \sim (1-\epsilon')\tP_{\theta} + \epsilon'\mathbb{Q}$ independently for $t \in [T]$, and $\widetilde{x} \sim \tP_{\widetilde{\bx}}$. Consider another family of distributions $\{\tP_{\theta}^{(0)}: \theta \in \Theta\}$ indexed by the same parameter set, and $\bxk{0} \sim \tP_{\theta}^{(0)}$. Denote the joint distribution of $\{\bxk{t}\}_{t=0}^T$ and $\widetilde{x}$ as $\tP_{(\epsilon', \theta, \mQ)}\cdot \tP_{\widetilde{\bx}}$. Then
	\begin{equation}
		\inf_{\widehat{\theta}} \sup_{\substack{\theta \in \Theta \\ \mathbb{Q}}} (\tP_{(\epsilon', \theta, \mathbb{Q})}\cdot \tP_{\widetilde{\bx}})\left(\|\widehat{\theta}-\theta\| \geq C\varpi(\epsilon', \Theta)\right) \geq \frac{9}{20},
	\end{equation}
	where $\varpi(\epsilon', \Theta) \coloneqq \sup\big\{\|\theta_1-\theta_2\|: \textup{TV}\big(\tP_{\theta_1}, \tP_{\theta_2}\big) \leq \epsilon'/(1-\epsilon'), \textup{TV}\big(\tP^{(0)}_{\theta_1}, \tP^{(0)}_{\theta_2}\big) \leq 1/20, \theta_1, \theta_2 \in \Theta\big\}$.
\end{lemma}

\begin{lemma}[Lemma 34 in \cite{tian2022unsupervised}]\label{lem: transfer binomial lower bound}
	Consider two data generating mechanisms:
		\begin{enumerate}[(i)]
			\item $\bxk{t} \sim (1-\epsilon')\tP_{\theta} + \epsilon' \mathbb{Q}$ independently for $t \in [T]$, $\bxk{0} \sim \tP^{(0)}_{\theta}$, and $\widetilde{x} \sim \tP_{\widetilde{\bx}}$, where $\epsilon' = \frac{|S^c|}{T}$;
			\item With a preserved set $S \subseteq [T]$, generate $\{\bxk{t}\}_{t \in S^c} \sim \mathbb{Q}_{S^c}$ and $\bxk{t} \sim \tP_{\theta}$ independently for $t \in S$, $\bxk{0} \sim \tP^{(0)}_{\theta}$, and $\widetilde{x} \sim \tP_{\widetilde{\bx}}$.
		\end{enumerate}
	Denote the joint distributions of $\{\bxk{t}\}_{t=0}^T$ and $\widetilde{x}$ in (\rom{1}) and (\rom{2}) as $\tP_{(\epsilon, \theta, \mathbb{Q})}\cdot \tP_{\widetilde{\bx}}$ and $\tP_{(S, \theta, \mathbb{Q}_{S^c})}\cdot \tP_{\widetilde{\bx}}$, respectively. We claim that if
	\begin{equation}
		\inf_{\widehat{\theta}} \sup_{\substack{\theta \in \Theta \\ \mathbb{Q}}} (\tP_{(\epsilon'/50, \theta, \mathbb{Q})}\cdot \tP_{\widetilde{\bx}})\left(\|\widehat{\theta}-\theta\| \geq C\varpi'\left(\epsilon'/50, \Theta \right)\right) \geq \frac{9}{20},
	\end{equation} 
	then
	\begin{equation}
		\inf_{\widehat{\theta}} \sup_{S: |S| \geq T(1-\epsilon')}\sup_{\substack{\theta \in \Theta \\ \mathbb{Q}_S}} (\tP_{(S, \theta, \mathbb{Q}_S)}\cdot \tP_{\widetilde{\bx}})\left(\|\widehat{\theta}-\theta\| \geq C\varpi'\left(\epsilon'/50, \Theta \right)\right) \geq \frac{1}{10},
	\end{equation}
	where $\varpi(\epsilon', \Theta) \coloneqq \sup\big\{\|\theta_1-\theta_2\|: \textup{TV}\big(\tP_{\theta_1}\| \tP_{\theta_2}\big) \leq \epsilon'/(1-\epsilon'), \textup{TV}\big(\tP^{(0)}_{\theta_1}\| \tP^{(0)}_{\theta_2}\big) \leq 1/20, \theta_1, \theta_2 \in \Theta\big\}$.
\end{lemma}

\subsection{Proofs of Lemmas}
Denote $\hSigmak{t} = \sum_{i=1}^n\frac{1}{n}\bxk{t}_i(\bxk{t}_i)^\top$.

\subsubsection{Proof of Lemma \ref{lem: cov hat r}}
Note that for any fixed $\bA, \bm{B} \in \mO$,
\begin{equation}
	\twonorm{\bA^\top(\hSigmak{0}-\bSigmak{0})\bm{B}} = \sup_{\twonorm{\bu},\twonorm{\bm{v}} \leq 1} (\bA \bu)^\top(\hSigmak{0}-\bSigmak{0})(\bm{B}\bm{v}), 
\end{equation}
where $\bA \bu, \bm{B}\bm{v} \in \mathbb{R}^p$ and $\twonorm{\bA\bu} = \twonorm{\bm{B}\bm{v}} = 1$. Note that both $\bA \bu$ and $\bm{B}\bm{v}$ live in $r$-dimensional space (isomorphic to the unit ball in $\mathbb{R}^r$). Therefore, by Example 5.8 in \cite{wainwright2019high}, there exist two $1/8$-covers (whose components are inside the set to be covered) of $\{\bbeta \in \mathbb{R}: \bbeta = \bA \bu, \twonorm{\bu} \leq  1\}$ and $\{\bbeta \in \mathbb{R}: \bbeta = \bm{B} \bu, \twonorm{\bu} \leq  1\}$ under Euclidean norm, denoted as $\{\bbeta_j^{\bu}\}_{j=1}^{N_1}$ and $\{\bbeta_j^{\bu}\}_{j=1}^{N_2}$, respectively, such that $|N_1| = |N_2| \leq 17^r$. Then we can proceed in the same steps as in the proof of Theorem 6.5 in \cite{wainwright2019high} to finish the proof.

\subsubsection{Proof of Lemma \ref{lem: ball packing}}
We apply the same trick as in the proof of Proposition 3 in \cite{cai2013sparse}. Suppose the minimum cover of $\mO$ under $\disttwo(\cdot )$ corresponding to $N(\mO, \disttwo, \delta)$ is $\mathcal{N}$. By pigeonhole theorem, for any $\alpha \in (0,1)$,
\begin{align}
	N(\mO, \disttwo, \alpha\delta) &\leq N(\cup_{\bA \in \mathcal{N}}\mB_{\delta}(\bA, \mO, \disttwo), \distf, \alpha\delta) \\
	&\leq |\mathcal{N}|\max_{\bA \in \mathcal{N}} N(\mB_{\delta}(\bA, \mO, \disttwo), \distf, \alpha\delta).
\end{align}
Then applying Lemma \ref{lem: A covering} leads to
\begin{equation}
	\max_{\bA \in \mathcal{N}} N(\mB_{\delta}(\bA, \mO, \disttwo), \distf, \alpha\delta) \geq \frac{N(\mO, \disttwo, \alpha\delta)}{N(\mO, \disttwo, \delta)} \geq \left(\frac{C\sqrt{r}}{\alpha}\right)^{r(p-r)}.
\end{equation}
Since $\mathcal{N}$ has only finite elements, there must exist one $\bA \in \mathcal{N}$ achieving the maximum of LHS. Because the packing number is always larger than or equal to the covering number, the proof is done.

\subsubsection{Proof of Lemma \ref{lem: KL}}
By the form of Gaussian density function, it is straightfoward to see that
\begin{align}
	\KL(\tP_{\bx, y}\|\widetilde{\tP}_{\bx, y}) &= \int \log\left(\frac{\tP_{y|\bx, \bbeta}}{\tP_{y|\bx, \widetilde{\bbeta}}}\right)\dint\tP_{y|\bx, \bbeta}\dint\tP_{\bx} \\
	&\leq \underbrace{\int (y-\bx^\top\bbeta)\cdot \bx^\top(\bbeta - \widetilde{\bbeta})\dint\tP_{y|\bx, \bbeta}\dint\tP_{\bx}}_{=0} + \int [\bx^\top(\bbeta - \widetilde{\bbeta})]^2 \dint\tP_{\bx} \\
	&\lesssim \twonorm{\bbeta - \widetilde{\bbeta}}^2.
\end{align}

\section{Proofs for Linear Regression Models}
Denote $\bthetak{t}_{\bA} \in \argmin_{\btheta \in \mathbb{R}^r} \fk{t}(\bA \btheta)$ and $\fk{t}(\bbeta) = \frac{1}{n}\twonorm{\bYk{t} - \bXk{t}\bbeta}^2$. For $\bA, \widetilde{\bA} \in \mO$, define metrics $\disttwo(\bA, \widetilde{\bA}) = \twonorm{\bA \bA^\top - \widetilde{\bA}(\widetilde{\bA})^\top}$, $\distf(\bA, \widetilde{\bA}) = \fnorm{\bA \bA^\top - \widetilde{\bA}(\widetilde{\bA})^\top}$. In any metric space $(\mathcal{X}, \rho)$, denote the ball of radius $\delta$ with center $x$ under metric $\rho$ as $\mB_{\delta}(x, \mathcal{X}, \rho)$. Denote $\zetak{t} = \twonorm{\bthetaks{t}} \vee \sqrt{\frac{p+\log T}{n}}$ and $\barzeta = \sqrt{|S|^{-1}\sum_{t\in S}(\zetak{t})^2}$.

\subsection{Lemmas}

\begin{lemma}\label{lem: a theta a}
	Suppose Assumptions \ref{asmp: x}-\ref{asmp: n} hold. Then w.p. at least $1-e^{-C(r+\log T)}$, for all $t \in [T]$ and $\bA \in \mO$, 
	\begin{equation}
		\twonorm{\bA\bthetak{t}_{\bA}-\bAks{t}\bthetaks{t}} \lesssim \twonorm{\bA\bA^\top - \bAks{t}(\bAks{t})^\top}\cdot \zetak{t} + \sqrt{\frac{r+\log T}{n}}.
	\end{equation}
\end{lemma}

\begin{lemma}\label{lem: theta bdded}
	Suppose Assumptions \ref{asmp: x}-\ref{asmp: n} hold. Then w.p. at least $1-e^{-C(r+\log T)}$, for all $t \in S$ and $\bA \in \mO$, there exists a universal constant $C' > 0$ (which does not depend on $t$ or $\bA$), such that $\twonorm{\bA\bthetak{t}_{\bA}} = \twonorm{\bthetak{t}_{\bA}} \leq C'\zetak{t}$.
\end{lemma}

\begin{lemma}\label{lem: a theta a extention}
	Suppose Assumptions \ref{asmp: x}-\ref{asmp: n} hold. Then w.p. at least $1-e^{-C(r+\log T)}$, for all $t \in [T]$ and $\bA, \widetilde{\bA} \in \mO$, 
	\begin{equation}
		\twonorm{\bA\bthetak{t}_{\bA}-\widetilde{\bA}\bthetak{t}_{\widetilde{\bA}}} \lesssim \twonorm{\bA\bA^\top - \widetilde{\bA}\widetilde{\bA}^\top}(\twonorm{\bthetak{t}_{\widetilde{\bA}}}+\twonorm{\nabla \fk{t}(\widetilde{\bA}\bthetak{t}_{\widetilde{\bA}})}).
	\end{equation}
\end{lemma}

\begin{lemma}\label{lem: gradient 2 norm}
	Suppose Assumptions \ref{asmp: x}-\ref{asmp: n} hold. For any $\barA \in \argmin_{\barA \in \mO}\max_{t \in S}\{\twonorm{\bAk{t}(\bAk{t})^\top - \barA\barA^\top}\}$, denote $\widebar{\bm{G}} = \big\{\nabla \fk{t}(\barA\bthetak{t}_{\barA})\big\}_{t \in S}$, then w.p. at least $1-e^{-C(p+\log T)}$,
	\begin{equation}
		\twonorm{\widebar{\bm{G}}} \lesssim \sqrt{\frac{p}{n}} + \sqrt{T}\left(h\barzeta + \sqrt{\frac{r+\log T}{n}}\right).
	\end{equation}
\end{lemma}

\begin{lemma}\label{lem: matrix 2 norm}
	Suppose $\bm{G} = \{\bm{g}_j\}_{j=1}^d \in \mathbb{R}^{p \times d}$ and $\widebar{\bm{G}} = \{\widebar{\bm{g}}_j\}_{j=1}^d \in \mathbb{R}^{p \times d}$ with each $\bm{g}_j, \widebar{\bm{g}}_j \in \mathbb{R}^{p}$ and $\max_{j \in [d]}\twonorm{\bm{g}_j - \widebar{\bm{g}}_j} \leq \Delta$. Then 
	\begin{equation}
		\twonorm{\widebar{\bm{G}} - \bm{G}} \leq \sqrt{d}\Delta.
	\end{equation}
\end{lemma}

\begin{lemma}\label{lem: the same A}
	Suppose Assumptions \ref{asmp: x}-\ref{asmp: n} hold. Consider any $\hbarA \in \mO$. 
	\begin{enumerate}[(i)]
		\item When $\frac{\lambda}{\sqrt{n}} \geq C\zetak{t}\Big[\twonorm{\hbarA \bthetak{t}_{\hbarA} - \bAks{t}\bthetaks{t}} + \twonorm{\nabla \fk{t}(\bAks{t}\bthetaks{t})}\Big]$ with a sufficiently large $C$ and the minimizer $(\bAk{t}, \bthetak{t}) = \argmin_{\bA \in \mO, \btheta \in \mathbb{R}^r}\{\frac{1}{T}\fk{t}(\bA\btheta) + \frac{\sqrt{n}}{nT}\lambda \twonorm{\bA\bA^\top - \hbarA(\hbarA)^\top}\}$, we must have $\bAk{t}(\bAk{t})^\top = \hbarA(\hbarA)^\top$. 
		\item As a consequence, by Lemma \ref{lem: a theta a}, when $\frac{\lambda}{\sqrt{n}} \geq C\zetak{t}\Big[\zetak{t}\twonorm{\hbarA(\hbarA)^\top - \bAks{t}(\bAks{t})^\top} + \sqrt{\frac{p+\log T}{n}}\Big]$, w.p. at least $1-e^{-C(r+\log T)}$, we have $\bAk{t}(\bAk{t})^\top = \hbarA(\hbarA)^\top$. 
	\end{enumerate}
\end{lemma}

\begin{lemma}\label{lem: small fraction diversity}
	Suppose $\{\bthetak{t}\}_{t \in S} \subseteq \mathbb{R}^r$ satisfying $|S|^{-1}\sum_{t \in S}\bthetak{t}(\bthetak{t})^\top \succeq \frac{c}{r}\barzeta \bm{I}_r$, where $c$ is a positive constant. For any subset $S' \subseteq S$ with $|S'| \geq (1-\alpha)|S|$ and 
	\begin{equation}
		\alpha \leq \frac{c}{r}\cdot \frac{\barzeta^2}{\max_{t \in S}(\zetak{t})^2},
	\end{equation}
	where $c$ is the same positive constant above and $\zetak{t} = \twonorm{\bthetak{t}}$, we have
	\begin{equation}
		\frac{1}{|S'|}\sum_{t\in S'} \bthetak{t}(\bthetak{t})^\top \succeq \frac{c}{2r}\barzeta^2\bm{I}_r.
	\end{equation}
\end{lemma}

\begin{lemma}\label{lem: step 1 control}
	Under Assumptions \ref{asmp: x}-\ref{asmp: n}, w.p. at least $1-e^{-C(r+\log T)}$, for all $t \in S$,
	\begin{equation}
		\twonorm{\hAk{t}\hthetak{t}-\bAks{t}\bthetaks{t}} \lesssim \sqrt{\frac{p+\log T}{n}} + \frac{\lambda}{\sqrt{n}\zetak{t}}.
	\end{equation}
\end{lemma}

\begin{lemma}\label{lem: no step 2 when r = 1}
	Consider the case that $r = 1$. Suppose that $\min_{t \in S}|\bthetaks{t}| \geq c > 0$ with some constant $c > 0$. Under Assumptions \ref{asmp: x}-\ref{asmp: n}, w.p. at least $1-e^{-C'(1+\log T)}$, for all $\bA \in \mO$ and $t \in S$,
	\begin{equation}
		\twonorm{\bA \bthetak{t}_{\bA} - \bAks{t}\bthetaks{t}} \geq C'\zetak{t}\twonorm{\bA \bA^\top - \bAks{t}(\bAks{t})^\top} - C\sqrt{\frac{p+\log T}{n}}.
	\end{equation}
\end{lemma}

\begin{lemma}[For MTL]\label{lem: safe net mtl}
	Under Assumptions \ref{asmp: x}-\ref{asmp: n}, we have:
	\begin{enumerate}[(i)]
		\item For all $t \in S$, when $\frac{\gamma}{\sqrt{n}} \geq \twonorm{\nabla \fk{t}(\bAks{t}\bthetaks{t})} + C\twonorm{\hAk{t}\hthetak{t} - \bAks{t}\bthetaks{t}}$, it holds that $\hbetak{t} = \hAk{t}\hthetak{t}$ \wpp;
		\item $\max_{t \in S}\twonorm{\hbetak{t} - \bbetaks{t}} \leq C\frac{\gamma}{\sqrt{n}} + \max_{t \in S}\twonorm{\widetilde{\bbeta}^{(t)} - \bbetaks{t}}$ \wpp, where $\widetilde{\bbeta}^{(t)} \in \argmin_{\bbeta \in \mathbb{R}^p}\fk{t}(\bbeta)$;
		\item If the data from tasks in $S^c$ satisfies the linear model \eqref{eq: linear model} (without any latent structure assumption) and Assumption \ref{asmp: x}, then \wpp, $\max_{t \in S^c}\twonorm{\hbetak{t} - \bbetaks{t}} \leq C\frac{\gamma}{\sqrt{n}} + \max_{t \in S^c}\twonorm{\widetilde{\bbeta}^{(t)} - \bbetaks{t}}$, where $\widetilde{\bbeta}^{(t)} \in \argmin_{\bbeta \in \mathbb{R}^p}\fk{t}(\bbeta)$.
	\end{enumerate}
\end{lemma}

\begin{lemma}\label{lem: a theta a r tl}
	Suppose Assumption \ref{asmp: x tl} hold and $n \geq Cr$. Then for \textbf{any fixed} $\bA \in \mO$, w.p. at least $1-e^{-C(r+\log T)}$, 
	\begin{equation}
		\twonorm{\bA\bthetak{0}_{\bA}-\bAks{0}\bthetaks{0}} \lesssim \twonorm{\bA\bA^\top - \bAks{0}(\bAks{0})^\top} + \sqrt{\frac{r}{n}}.
	\end{equation}
\end{lemma}

\begin{lemma}\label{lem: a theta a tl}
	Suppose Assumptions \ref{asmp: x tl}-\ref{asmp: n tl} hold. Then w.p. at least $1-e^{-C(r+\log T)}$, for \textbf{all} $\bA \in \mO$, 
	\begin{equation}
		\twonorm{\bA\bthetak{0}_{\bA}-\bAks{0}\bthetaks{0}} \lesssim \twonorm{\bA\bA^\top - \bAks{0}(\bAks{0})^\top} + \sqrt{\frac{r}{n}}.
	\end{equation}
\end{lemma}

\begin{lemma}[For TL]\label{lem: safe net tl}
	Under Assumptions \ref{asmp: x tl}-\ref{asmp: n tl}, we have:
	\begin{enumerate}[(i)]
		\item When $\frac{\gamma}{\sqrt{n_0}} \geq \twonorm{\nabla \fk{t}(\bAks{0}\bthetaks{0})} + C\twonorm{\hSigmak{0}}\cdot\twonorm{\hbarA\bthetak{0}_{\hbarA} - \bAks{0}\bthetaks{0}}$, it holds that $\hbetak{0} = \hbarA\bthetak{0}_{\hbarA}$ \wpp; \footnote{For this result to hold, it suffices to require $n_0 \geq Cr$, which is important for Remark \ref{rmt: few shot learning} to be true.}
		\item $\twonorm{\hbetak{0} - \bbetaks{0}} \leq C\frac{\gamma}{\sqrt{n_0}} + \twonorm{\widetilde{\bbeta}^{(0)} - \bbetaks{0}}$ \wpp, where $\widetilde{\bbeta}^{(0)} \in \argmin_{\bbeta \in \mathbb{R}^p}\fk{0}(\bbeta)$.
	\end{enumerate}
\end{lemma}

\subsection{Proof of Theorem \ref{thm: equivalence of settings}}
(\rom{1}) $\bbetaks{t} = \oA(\oA)^\top \bAks{t}\bthetaks{t} + \oA^\perp (\oA^\perp)^\top \bAks{t}\bthetaks{t} = \oA \barthetaks{t} + \bdeltaks{t}$, where $\barthetaks{t} = (\oA)^\top \bAks{t}\bthetaks{t}$ and $\bdeltaks{t} = \oA^\perp (\oA^\perp)^\top \bAks{t}\bthetaks{t}$. Then $\twonorm{\bdeltaks{t}} \leq \twonorm{\bthetaks{t}}\twonorm{(\oA)^\top \bAks{t}} \leq h\zetak{t}$. This shows how we can transform Setting 1 to Setting 2. 

On the other hand, if we denote $\bm{\Delta}^{(t)*} = \bdeltaks{t}(\barthetaks{t})^\top \frac{1}{\twonorm{\barthetaks{t}}}$, then $\bm{\Delta}^{(t)*} \barthetaks{t} = \bdeltaks{t}$, hence $\bbetaks{t} = (\oA + \bm{\Delta}^{(t)*}) \barthetaks{t}$. Note that we have shown that $\twonorm{\bdeltaks{t}} \leq h\zetak{t}$. Also, $\twonorm{\barthetaks{t}} \geq \sigma_{\min}((\oA)^\top \bAks{t})\twonorm{\bthetaks{t}} \geq \sqrt{1-h^2}\zetak{t}$. Therefore $\fnorm{\bm{\Delta}^{(t)*}} = \twonorm{\bm{\Delta}^{(t)*}} \leq \twonorm{\bdeltaks{t}}/\twonorm{\barthetaks{t}} \leq h/\sqrt{1-h^2}$. This shows how we can transform Setting 1 to Setting 3.

\noindent(\rom{2}) Let $\bar{\bbeta}^{(t)*} = \oA\barthetaks{t}$, $\widetilde{\bA} = (\frac{\bbetaks{t}}{\twonorm{\bbetaks{t}}}, \widetilde{\bA}_{-1})$ with $(\bbetaks{t})^\top \widetilde{\bA}_{-1} = \bm{0}$, $\widetilde{\bA}_{-1}^\top \widetilde{\bA}_{-1} = \bm{I}$. WLOG, consider $\oA = (\frac{\bar{\bbeta}^{(t)*}}{\twonorm{\bar{\bbeta}^{(t)*}}}, \oA_{-1})$ with $(\bar{\bbeta}^{(t)*})^\top \oA_{-1} = \bm{0}$, $\oA_{-1}^\top \oA_{-1} = \bm{I}$. Also, we have
\begin{equation}
	\widetilde{\bA}\widetilde{\bA}^\top = \frac{\bbetaks{t}(\bbetaks{t})^\top}{\twonorm{\bbetaks{t}}^2} + \widetilde{\bA}_{-1}\widetilde{\bA}_{-1}^\top, \quad \oA(\oA)^\top = \frac{\bar{\bbeta}^{(t)*}(\bar{\bbeta}^{(t)*})^\top}{\twonorm{\bar{\bbeta}^{(t)*}}^2} + \oA_{-1}\oA_{-1}^\top. 
\end{equation}
And
\begin{align}
	&\frac{\bbetaks{t}(\bbetaks{t})^\top}{\twonorm{\bbetaks{t}}^2} - \frac{\bar{\bbeta}^{(t)*}(\bar{\bbeta}^{(t)*})^\top}{\twonorm{\bar{\bbeta}^{(t)*}}^2} \\
	&= \frac{\bar{\bbeta}^{(t)*}(\bar{\bbeta}^{(t)*})^\top + \bdeltaks{t}(\bdeltaks{t})^\top + \bar{\bbeta}^{(t)*}(\bdeltaks{t})^\top + \bdeltaks{t}(\bar{\bbeta}^{(t)*})^\top}{\twonorm{\bar{\bbeta}^{(t)*}}^2+\twonorm{\bdeltaks{t}}^2}- \frac{\bar{\bbeta}^{(t)*}(\bar{\bbeta}^{(t)*})^\top}{\twonorm{\bar{\bbeta}^{(t)*}}^2} \\
	&= \frac{-\twonorm{\bdeltaks{t}}^2\cdot \bar{\bbeta}^{(t)*}(\bar{\bbeta}^{(t)*})^\top + \twonorm{\bar{\bbeta}^{(t)*}}^2[\bdeltaks{t}(\bdeltaks{t})^\top + \bar{\bbeta}^{(t)*}(\bdeltaks{t})^\top + \bdeltaks{t}(\bar{\bbeta}^{(t)*})^\top]}{(\twonorm{\bar{\bbeta}^{(t)*}}^2+\twonorm{\bdeltaks{t}}^2)\twonorm{\bar{\bbeta}^{(t)*}}^2},
\end{align}
which implies that
\begin{align}
	\twonorma{\frac{\bbetaks{t}(\bbetaks{t})^\top}{\twonorm{\bbetaks{t}}^2} - \frac{\bar{\bbeta}^{(t)*}(\bar{\bbeta}^{(t)*})^\top}{\twonorm{\bar{\bbeta}^{(t)*}}^2}} &\leq \max\bigg\{\frac{\twonorm{\bdeltaks{t}}^2}{\twonorm{\bbetaks{t}}^2} + 2\frac{\twonorm{\bar{\bbeta}^{(t)*}}\twonorm{\bdeltaks{t}}}{\twonorm{\bbetaks{t}}^2}, \frac{\twonorm{\bdeltaks{t}}^2}{\twonorm{\bbetaks{t}}^2}\bigg\} \\
	&\leq \frac{\twonorm{\bdeltaks{t}}^2}{\twonorm{\bbetaks{t}}^2} + 2\frac{\twonorm{\bdeltaks{t}}}{\twonorm{\bbetaks{t}}}\cdot \sqrt{1-\bigg(\frac{\twonorm{\bdeltaks{t}}}{\twonorm{\bbetaks{t}}}\bigg)^2} \\
	&= \xi^2 + 2\xi\sqrt{1-\xi^2} \\
	&\leq 2\sqrt{2}\xi,
\end{align}
where $\xi \coloneqq \twonorm{\bdeltaks{t}}/\twonorm{\bbetaks{t}}$. 

Let $\bm{P}_{\bbeta} = \frac{\bbeta \bbeta^\top }{\twonorm{\bbeta}^2}$ as a projection matrix onto the linear space spanned by $\bbeta \in \mathbb{R}^p \backslash \{\bm{0}\}$, and $\widetilde{\bA}_{-1} = (\bm{I} - \bm{P}_{\bbetaks{t}})\oA_{-1}[(\oA_{-1})^\top (\bm{I} - \bm{P}_{\bbetaks{t}})\oA_{-1}]^{-1/2}$. Then
\begin{equation}
	\widetilde{\bA}_{-1}\widetilde{\bA}_{-1}^\top - \oA_{-1}(\oA_{-1})^\top = (\bm{I} - \bm{P}_{\bbetaks{t}})\oA_{-1}[(\oA_{-1})^\top (\bm{I} - \bm{P}_{\bbetaks{t}})\oA_{-1}]^{-1}(\oA_{-1})^\top (\bm{I} - \bm{P}_{\bbetaks{t}}) - \oA_{-1}(\oA_{-1})^\top.
\end{equation}
By noticing that $\oA_{-1} = (\bm{I} - \bm{P}_{\bar{\bbeta}^{(t)*}})\oA$, after some simplifications, we have
\begin{align}
	\twonorm{\widetilde{\bA}_{-1}\widetilde{\bA}_{-1}^\top - \oA_{-1}(\oA_{-1})^\top} &\leq 2\twonorm{\bm{P}_{\bbeta^{(t)*}}(\bm{I} - \bm{P}_{\bar{\bbeta}^{(t)*}})}\twonorm{[(\oA_{-1})^\top (\bm{I} - \bm{P}_{\bbetaks{t}})\oA_{-1}]^{-1}} \\
	&\quad + \twonorm{\bm{I} - [(\oA_{-1})^\top (\bm{I} - \bm{P}_{\bbetaks{t}})\oA_{-1}]^{-1}},
\end{align}
where
\begin{align}
	\twonorm{\bm{P}_{\bbeta^{(t)*}}(\bm{I} - \bm{P}_{\bar{\bbeta}^{(t)*}})} &= \twonorm{\bm{P}_{\bbeta^{(t)*}} - \bm{P}_{\bar{\bbeta}^{(t)*}}} \leq 2\sqrt{2}\cdot \frac{\twonorm{\bdeltaks{t}}}{\twonorm{\bbetaks{t}}} = 2\sqrt{2}\xi, \\
	 \twonorm{\bm{I} - [(\oA_{-1})^\top (\bm{I} - \bm{P}_{\bbetaks{t}})\oA_{-1}]^{-1}} &= \twonorm{\bm{I} - (\bm{I} - (\oA_{-1})^\top\bm{P}_{\bbetaks{t}})\oA_{-1})^{-1}} \\
	 &= \twonorma{\bm{I} - \bigg(\bm{I} + \frac{(\oA_{-1})^\top\bm{P}_{\bbetaks{t}})\oA_{-1}}{1-\frac{(\bbetaks{t})^\top \oA_{-1}(\oA_{-1})^\top \bbetaks{t}}{\twonorm{\bbetaks{t}}^2}}\bigg)} \\
	 &= \twonorma{\frac{(\oA_{-1})^\top\bm{P}_{\bbetaks{t}}\oA_{-1}}{1-\frac{(\bbetaks{t})^\top \oA_{-1}(\oA_{-1})^\top \bbetaks{t}}{\twonorm{\bbetaks{t}}^2}}}, \\
	 \twonorm{(\oA_{-1})^\top\bm{P}_{\bbetaks{t}}\oA_{-1}} &= \twonorm{(\oA_{-1})^\top(\bm{I} - \bm{P}_{\bar{\bbeta}^{(t)*}})\bm{P}_{\bbetaks{t}})(\bm{I} - \bm{P}_{\bar{\bbeta}^{(t)*}})\oA_{-1}} \\
	 &\leq \twonorm{\bm{P}_{\bbetaks{t}}(\bm{I} - \bm{P}_{\bar{\bbeta}^{(t)*}})}^2, \\
	 \norma{\frac{(\bbetaks{t})^\top \oA_{-1}(\oA_{-1})^\top \bbetaks{t}}{\twonorm{\bbetaks{t}}^2}} &= \twonorm{(\oA_{-1})^\top\bm{P}_{\bbetaks{t}}\oA_{-1}}.
\end{align}
Therefore, we have
\begin{align}
	\twonorm{\bm{I} - [(\oA_{-1})^\top (\bm{I} - \bm{P}_{\bbetaks{t}})\oA_{-1}]^{-1}} &\leq \frac{\twonorm{\bm{P}_{\bbeta^{(t)*}}(\bm{I} - \bm{P}_{\bar{\bbeta}^{(t)*}})}^2}{1-\twonorm{\bm{P}_{\bbeta^{(t)*}}(\bm{I} - \bm{P}_{\bar{\bbeta}^{(t)*}})}^2}, \\
	\twonorm{[(\oA_{-1})^\top (\bm{I} - \bm{P}_{\bbetaks{t}})\oA_{-1}]^{-1}} &= 1 + \twonorma{\frac{(\oA_{-1})^\top\bm{P}_{\bbetaks{t}}\oA_{-1}}{1-\frac{(\bbetaks{t})^\top \oA_{-1}(\oA_{-1})^\top \bbetaks{t}}{\twonorm{\bbetaks{t}}^2}}} \leq 1 + \frac{\twonorm{\bm{P}_{\bbeta^{(t)*}}(\bm{I} - \bm{P}_{\bar{\bbeta}^{(t)*}})}^2}{1-\twonorm{\bm{P}_{\bbeta^{(t)*}}(\bm{I} - \bm{P}_{\bar{\bbeta}^{(t)*}})}^2}.
\end{align}
Combining all of the results, we have
\begin{align}
	\twonorm{\widetilde{\bA}_{-1}\widetilde{\bA}_{-1}^\top - \oA_{-1}(\oA_{-1})^\top} &\leq 2 \cdot 2\sqrt{2}\xi \cdot \bigg[1+\frac{(2\sqrt{2}\xi)^2}{1-(2\sqrt{2}\xi)^2}\bigg] + \frac{(2\sqrt{2}\xi)^2}{1-(2\sqrt{2}\xi)^2} \\
	&= 4\sqrt{2}\xi \bigg(1+\frac{8\xi^2}{1-8\xi^2}\bigg) + \frac{8\xi^2}{1-8\xi^2},
\end{align}
which implies that
\begin{align}
	\twonorm{\widetilde{\bA}\widetilde{\bA}^\top - \oA(\oA)^\top} &\leq \twonorm{\bm{P}_{\bbetaks{t}} \bm{P}_{\bar{\bbeta}^{(t)*}}} + \twonorm{\widetilde{\bA}_{-1}\widetilde{\bA}_{-1}^\top - \oA_{-1}(\oA_{-1})^\top} \\
	&\leq 2\sqrt{2}\xi + 4\sqrt{2}\xi \bigg(1+\frac{8\xi^2}{1-8\xi^2}\bigg) + \frac{8\xi^2}{1-8\xi^2}. \quad\quad (*)
\end{align}
\begin{itemize}
	\item When $\xi \leq 1/4$: $(*) \leq 2\sqrt{2}\xi + 4\sqrt{2}\xi\cdot (1+1) + 8\xi \cdot \frac{1/4}{1-1/2} = (10\sqrt{2} + 4)\xi$.
	\item When $\xi > 1/4$: $\twonorm{\widetilde{\bA}\widetilde{\bA}^\top - \oA(\oA)^\top} \leq 1 \leq 4\xi$.
\end{itemize}
Therefore, $\twonorm{\widetilde{\bA}\widetilde{\bA}^\top - \oA(\oA)^\top} \leq (10\sqrt{2}+4)\xi$.

\noindent (\rom{3}) $\bbetaks{t} = (\oA+\bDelta^{(t)*})\tildethetaks{t} = \oA(\bm{I}+(\oA)^\top \bDelta^{(t)*})\tildethetaks{t} + \oA^\perp(\oA^\perp)^\top \bDelta^{(t)*}\tildethetaks{t}$. Let $\barthetaks{t} = (\bm{I}+(\oA)^\top \bDelta^{(t)*})\tildethetaks{t}$ and $\bdeltaks{t} = \oA^\perp(\oA^\perp)^\top \bDelta^{(t)*}\tildethetaks{t}$. By (\rom{2}), there exists $\bAks{t} \in \mO$ such that $\twonorm{\bAks{t}(\bAks{t})^\top - \oA(\oA)^\top} \leq (10\sqrt{2}+4)\frac{\twonorm{\bdeltaks{t}}}{\twonorm{\bbetaks{t}}}$.

Since 
\begin{align}
	\twonorm{\bdeltaks{t}} &\leq \twonorm{(\oA^\perp)^\top \bDelta^{(t)*}}\twonorm{\tildethetaks{t}}, \\
	\twonorm{\bbetaks{t}} &= \twonorm{\tildethetaks{t}}^2 + \twonorm{\bDelta^{(t)*}\tildethetaks{t}}^2 + 2(\tildethetaks{t})^\top \oA^\top \bDelta^{(t)*}\tildethetaks{t} \\
	&\geq \twonorm{\tildethetaks{t}}^2 + \twonorm{\oA^\perp(\oA^\perp)^\top \bDelta^{(t)*} \tildethetaks{t}}^2 - 2\twonorm{\tildethetaks{t}}^2\twonorm{\oA^\top \bDelta^{(t)*}} \\
	&= \twonorm{\tildethetaks{t}}^2 + \twonorm{\bdeltaks{t}}^2 - 2\twonorm{\tildethetaks{t}}^2\twonorm{\oA^\top \bDelta^{(t)*}}, 
\end{align}
we have
\begin{equation}
	\frac{\twonorm{\bdeltaks{t}}}{\twonorm{\bbetaks{t}}}\leq \frac{\twonorm{(\oA^\perp)^\top \bDelta^{(t)*}}}{\sqrt{1 + \twonorm{(\oA^\perp)^\top \bDelta^{(t)*}}^2 - 2\twonorm{\bDelta^{(t)*}}}} \leq \frac{\twonorm{\bDelta^{(t)*}}}{1-\twonorm{\bDelta^{(t)*}}} \leq \frac{\fnorm{\bDelta^{(t)*}}}{1-\fnorm{\bDelta^{(t)*}}}, 
\end{equation}
when $\fnorm{\bDelta^{(t)*}} < 1$. Hence there exists $\bAks{t} \in \mO$ such that $\twonorm{\bAks{t}(\bAks{t})^\top - \oA(\oA)^\top} \leq (10\sqrt{2}+4)\frac{\twonorm{\bdeltaks{t}}}{\twonorm{\bbetaks{t}}} \leq (10\sqrt{2}+4)\frac{\fnorm{\bDelta^{(t)*}}}{1-\fnorm{\bDelta^{(t)*}}}$ when $\fnorm{\bDelta^{(t)*}} < 1$.

\subsection{Proof of Theorem \ref{thm: mtl}}
First, we have the following proposition holds.
\begin{proposition}\label{prop: mtl}
	Suppose Assumptions \ref{asmp: x}-\ref{asmp: n} hold. Further assume
	\begin{enumerate}[(i)]
		\item $\min_{t \in S}\Big\{\frac{\zetak{t}}{\barzeta}r\sqrt{\frac{p+\log T}{nT}} + \frac{\zetak{t}}{\barzeta}\sqrt{r}\sqrt{\frac{r+\log T}{n}} + \zetak{t}\sqrt{r}h\Big\} \lesssim \sqrt{\frac{p+\log T}{n}}$;
		\item $|S^c|/T \leq cr^{-3/2}\frac{\barzeta}{\max_{t \in S}\zetak{t}}$ with a small constant $c > 0$.
	\end{enumerate}
	Let $\lambda \geq C\frac{\max_{t \in S}(\zetak{t})^2}{\min_{t \in S}\zetak{t}}\sqrt{r(p+\log T)}$ with a sufficiently large constant $C$ and $\lambda \leq C'\frac{\max_{t \in S}(\zetak{t})^2}{\min_{t \in S}\zetak{t}}\sqrt{r(p+\log T)}$ with another constant $C'$. Then w.p. at least $1-e^{-C'(r+\log T)}$,
	\begin{equation}
		\twonorm{\hbarA(\hbarA)^\top - \barA \barA^\top} \lesssim r\barzeta^{-1}\sqrt{\frac{p}{nT}} + \sqrt{r}h + \barzeta^{-1}\sqrt{r}\sqrt{\frac{r + \log T}{n}} +  \frac{\lambda r}{\sqrt{n}\barzeta^2}\epsilon. 
	\end{equation}
\end{proposition}

Proposition \ref{prop: mtl} provides the bound when the first term is faster than the second term in the upper bound of Theorem \ref{thm: mtl}. In the other case, a direct application of Lemma \ref{lem: safe net mtl} leads to the second term in the upper bound. Combining two situations gives us the disired result for tasks in $S$.

Denote $\eta^{(t)} = \frac{\zetak{t}}{\barzeta}r\sqrt{\frac{p+\log T}{nT}} + \frac{\zetak{t}}{\barzeta}\sqrt{r}\sqrt{\frac{r+\log T}{n}} + \zetak{t}\sqrt{r}h + \frac{\lambda r \zetak{t}}{\sqrt{n}\barzeta^2}\epsilon$.

\noindent(\rom{1}) For any $t$ satisfying $\eta^{(t)} \leq C\sqrt{\frac{p + \log T}{n}}$: By Proposition \ref{prop: mtl}, \wprp,
\begin{equation}
	\twonorm{\hbarA(\hbarA)^\top - \bAks{t} (\bAks{t})^\top} \lesssim r\barzeta^{-1}\sqrt{\frac{p}{nT}} + \sqrt{r}h + \barzeta^{-1}\sqrt{r}\sqrt{\frac{r + \log T}{n}} +  \frac{\lambda r}{\sqrt{n}\barzeta^2}\epsilon. 
\end{equation}
By Lemma \ref{lem: the same A}, since $\frac{\lambda}{\sqrt{n}} \geq C\frac{\max_{t \in S}(\zetak{t})^2}{\min_{t \in S}\zetak{t}}\sqrt{\frac{r(p+\log T)}{n}} \geq C\max_{t \in S}\zetak{t}\eta^{(t)} \geq C\zetak{t}\Big[\zetak{t}\twonorm{\hbarA(\hbarA)^\top - \bAks{t}(\bAks{t})^\top} + \sqrt{\frac{p+\log T}{n}}\Big]$, we have $\hAk{t}(\hAk{t})^\top = \hbarA(\hbarA)^\top$ \wprp, which combining with Lemma \ref{lem: a theta a} implies that
\begin{align}
	\twonorm{\hAk{t}\hthetak{t} - \bAks{t}\bthetaks{t}} &= \twonorm{\hbarA\bthetak{t}_{\hbarA} - \bAks{t}\bthetaks{t}} \\
	&\lesssim \twonorm{\hbarA(\hbarA)^\top - \bAks{t} (\bAks{t})^\top}\zetak{t} + \sqrt{\frac{r+\log T}{n}} \\
	&\lesssim \eta^{(t)} + \sqrt{\frac{r+\log T}{n}},
\end{align}
\wprp. Therefore by our choice $\gamma = C'\sqrt{p+\log T}$ with a large constant $C' > 0$:

\begin{equation}
	\twonorm{\nabla \fk{t}(\bAks{t}\bthetaks{t})} + C\twonorm{\hAk{t}\hthetak{t}-\bAks{t}\bthetaks{t}} \lesssim \frac{\gamma}{\sqrt{n}}.
\end{equation}
Then by Lemma \ref{lem: safe net mtl}.(\rom{1}) and Proposition \ref{prop: mtl}, \wprp, $\max_{t \in S}\twonorm{\hbetak{t} - \bbetaks{t}} \lesssim \eta^{(t)}+ \sqrt{\frac{r+\log T}{n}}$.

\noindent(\rom{2}) For any $t$ satisfying $\eta^{(t)} > C\sqrt{\frac{p + \log T}{n}}$: by Lemma \ref{lem: safe net mtl}.(\rom{2}), \wppp, $\max_{t \in S}\twonorm{\hbetak{t} - \bbetaks{t}} \lesssim \frac{\gamma}{\sqrt{n}} + \max_{t \in S}\twonorm{\widetilde{\bbeta}^{(t)} - \bbetaks{t}} \lesssim \sqrt{\frac{p+\log T}{n}}$, where $\widetilde{\bbeta}^{(t)} \in \argmin_{\bbeta \in \mathbb{R}^p}\fk{t}(\bbeta) $. The fact that $\max_{t \in S}\twonorm{\widetilde{\bbeta}^{(t)} - \bbetaks{t}} \lesssim \sqrt{\frac{p+\log T}{n}}$ \wppp \, is the standard linear regression result.

\noindent(\rom{3}) When data from tasks in $S^c$ also satisfies the linear model \eqref{eq: linear model}, the result comes from the same argument as in (\rom{2}).

\subsection{Proof of Theorem \ref{thm: mtl step 1 only}}
The conclusion follows directly by combining Lemma 36 with the upper bound on $\twonorm{\hAk{t}\hthetak{t} - \bAks{t}\bthetaks{t}}$ established in the proof of Theorem \ref{thm: mtl}.

\subsection{Proof of Theorem \ref{thm: no local easy}}
First, note that when $\lambda = +\infty$, we have $\hAk{t} \equiv \hoA$ for all $t \in S=[T]$. We will prove that any local minimizer $(\hoA, \{\hthetak{t}\}_{t=1}^T)$ which minimizes
\begin{equation}
	f(\bA, \bTheta) \coloneqq \frac{1}{T}\sum_{t=1}^T \fk{t}(\bA\bthetak{t}).
\end{equation}
over $\bA \in \mathcal{O}^{p \times r}$ and $\bTheta = (\bthetak{1}, \ldots, \bthetak{T}) \in \mathbb{R}^{r \times T}$ must satisfy
\begin{equation}
	\fnorm{\hoA(\hoA)^\top - \oA(\oA)^\top} \lesssim (\bar{\zeta}^{-1})r\sqrt{\frac{p}{nT}} + (\bar{\zeta}^{-1})\sqrt{r}\sqrt{\frac{r+\log T}{n}} + \sqrt{r}h,
\end{equation}
simulteneously \wprp. The following the argument in the proof of Theorem \ref{thm: mtl}, we will obtain the desired result. 

We prove the claim above by construction. Consider a local minimizer $(\hoA, \{\hthetak{t}\}_{t=1}^T)$ of $f$ and $\hoA$'s neightbor $\widetilde{\bm{A}} = a_1\oA + a_2\hoA\bm{D}$, where $a_1 + a_2 = 1$, $a_1, a_2 \geq 0$, and $\bm{D} \in \mathbb{R}^{r \times r}$ is invertible. Note that
\begin{equation}\label{eq: A disc small}
	\twonorm{((\hoA)^\perp)^\top \widetilde{\bm{A}}} = a_1\twonorm{((\hoA)^\perp)^\top \oA} \leq a_1.
\end{equation}
Define $\bthetak{t}_{\bA} = \argmin_{\btheta \in \mathbb{R}^r}\fk{t}(\bA\btheta)$, $\widehat{\bTheta} = \{\bthetak{t}_{\hoA}\}_{t=1}^T$, $\widetilde{\bm{R}} \in \mathcal{O}^{r \times r}$ by 
\begin{equation}
	\widetilde{\bm{A}}^\top \widetilde{\bm{A}} = a_1^2\bm{I}_p + a_2^2\bm{D}^\top\bm{D} + a_1a_2((\oA)^\top \hoA\bm{D} + \bm{D}^\top (\hoA)^\top \oA) = (\widetilde{\bm{R}}^{-1})^\top \widetilde{\bm{R}}^{-1}.
\end{equation}
Then define $\bA = \widetilde{\bm{A}}\widetilde{\bm{R}}$, hence $\bA^\top \bA = \bm{I}_r$. Let $\bm{D}$ satisfy $\bm{D}\widetilde{\bm{R}}\widebar{\bm{\Theta}} = \widehat{\bm{\Theta}}$, where $\widebar{\bTheta} = \{\bthetak{t}_{\oA}\}_{t =1}^T \in \mathbb{R}^{r \times T}$. Such $\bm{D}$ must exist. For example, we can take $\bm{D} = (\widetilde{\bm{R}}\widehat{\bm{\Theta}}\widehat{\bm{\Theta}}^\top)^{-1}\widehat{\bm{\Theta}}\widebar{\bm{\Theta}}^\top$. Consider $\bTheta$ s.t. $\widetilde{\bm{R}}\bTheta = \widebar{\bTheta}$, then 
\begin{equation}\label{eq: A Theta decomp}
	\bA\bTheta = a_1\oA\widetilde{\bm{R}}\bTheta + a_2\hoA\bm{D}\widetilde{\bm{R}}\bTheta = a_1\oA\widebar{\bTheta} + a_2\hoA\widehat{\bTheta}.
\end{equation}
Since $\fk{t}$ is defined through square loss, we have
\begin{align}
	f(\bA, \bTheta) - f(\hoA, \widehat{\bTheta}) &= \frac{1}{2T}\sum_{t=1}^T (\bA\bthetak{t} - \hoA\hthetak{t})^\top \hSigmak{t}(\bA\bthetak{t} - \hoA\hthetak{t}) \\
	&\quad + \underbrace{\frac{1}{T}\sum_{t=1}^T \big[\nabla \fk{t}(\hoA\hthetak{t})\big]^\top (\bA\bthetak{t} - \hoA\hthetak{t})}_{(*)}.
\end{align}
Therefore, \eqref{eq: A Theta decomp} implies that
\begin{align}
	(*) &\leq \frac{1}{T}\sum_{t=1}^T \big[\nabla \fk{t}(\oA\bthetak{t}_{\oA})\big]^\top (\bA\bthetak{t} - \hoA\hthetak{t}) - \frac{1}{T}a_1 \sum_{t=1}^T (\hoA\hthetak{t} - \oA \bthetak{t}_{\oA})^\top \hSigmak{t}(\hoA\hthetak{t} - \oA \bthetak{t}_{\oA}) \\
	&\leq \frac{1}{T}\twonorm{\{\nabla \fk{t}(\bAks{t}\bthetaks{t})\}_{t=1}^T}\cdot \sqrt{2r}\fnorm{\bA\bTheta - \hoA \widehat{\bTheta}} \\
	&\quad + \frac{1}{T}\fnorm{\{\nabla \fk{t}(\bAks{t}\bthetaks{t})- \nabla \fk{t}(\oA\bthetak{t}_{\oA})\}_{t=1}^T}\fnorm{\bA\bTheta - \hoA\widehat{\bTheta}} - \frac{C}{T}a_1\fnorm{\hoA\widehat{\bTheta} - \oA\widebar{\bTheta}}^2 \\
	&\leq \frac{1}{T}\sqrt{2r}\sqrt{\frac{p+T}{n}}\fnorm{\bA\bTheta - \hoA \widehat{\bTheta}} + \frac{C}{T}\Bigg(\sqrt{T}\sqrt{\frac{r+\log T}{n}} + h\sqrt{\sum_{t=1}^T (\zetak{t})^2}\Bigg)\fnorm{\bA\bTheta - \hoA\widehat{\bTheta}} \\
	&\quad - \frac{C}{T}a_1\fnorm{\hoA\widehat{\bTheta} - \oA\widebar{\bTheta}}^2 \\
	&\leq \frac{C}{T}a_1\Bigg(\sqrt{\frac{pr}{n}} + \sqrt{\frac{r+\log T}{n}}\sqrt{T} + h\sqrt{T}\bar{\zeta}\Bigg)\fnorm{\hoA\widehat{\bTheta} - \oA\widebar{\bTheta}} - \frac{C}{T}a_1\fnorm{\hoA\widehat{\bTheta} - \oA\widebar{\bTheta}}^2,
\end{align}
\wprp. Hence \wprp,
\begin{align}
	f(\bA, \bTheta) - f(\hoA, \widehat{\bTheta}) &\leq \frac{C''}{T}a_1^2 + \frac{C}{T}a_1\Bigg(\sqrt{\frac{pr}{n}} + \sqrt{\frac{r+\log T}{n}}\sqrt{T} + h\sqrt{T}\bar{\zeta}\Bigg)\fnorm{\hoA\widehat{\bTheta} - \oA\widebar{\bTheta}} \\
	&\quad - \frac{C}{T}a_1\fnorm{\hoA\widehat{\bTheta} - \oA\widebar{\bTheta}}^2. \label{eq: final f A}
\end{align}
By \eqref{eq: A Theta decomp}, as $a_1 \rightarrow 0+$,
\begin{equation}
\bTheta = a_1\bA^\top \oA\widebar{\bm{\Theta}} + a_2\bA^\top \hoA\widehat{\bm{\Theta}} \rightarrow \widetilde{\bm{R}}^\top \bm{D}^\top  (\hoA)^\top \hoA\widehat{\bm{\Theta}} = \widetilde{\bm{R}}^\top \bm{D}^\top\widehat{\bm{\Theta}},
\end{equation}
where ``$\rightarrow$" is in the sense of $\maxnorm{\cdot}$ (similar below). On the other hand, as $a_1 \rightarrow 0+$,
\begin{equation}
	\bA = a_1\oA\widetilde{\bm{R}} + a_2\hoA\bm{D}\widetilde{\bm{R}} \rightarrow \hoA\bm{D}\widetilde{\bm{R}},
\end{equation}
and $\widetilde{\bm{R}}^\top \bm{D}^\top \bm{D}\widetilde{\bm{R}} \rightarrow \bm{I}_r$.
Therefore, by continuity, as $a_1 \rightarrow 0+$,
\begin{equation}
	\max_{t \in [T]}\min_{\bm{R} \in \mathcal{O}^{r \times r}}\Big\{\twonorm{\bA - \hoA\bm{R}}+ \twonorm{\bthetak{t} - \hthetak{t}\bm{R}}\Big\} \rightarrow 0.
\end{equation}
Due to the local optimality of $(\hoA, \widehat{\bTheta})$, when $a_1$ is very close to $0$, we must have $f(\bA, \bTheta) - f(\hoA, \widehat{\bTheta}) \geq 0$, hence the RHS of \eqref{eq: final f A} is non-negative. Let $a_1 \lesssim \sqrt{\frac{pr}{n}} + \sqrt{\frac{r+\log T}{n}}\sqrt{T} + h\sqrt{T}\bar{\zeta}$, we have \wprp,
\begin{equation}
	\fnorm{\hoA\widehat{\bTheta} - \oA\widebar{\bTheta}} \lesssim a_1 +  \sqrt{\frac{pr}{n}} + \sqrt{\frac{r+\log T}{n}}\sqrt{T} + h\sqrt{T}\bar{\zeta} \lesssim   \sqrt{\frac{pr}{n}} + \sqrt{\frac{r+\log T}{n}}\sqrt{T} + h\sqrt{T}\bar{\zeta}.
\end{equation}
By Wedin's Theorem, \wprp,
\begin{equation}
	\fnorm{\hoA(\hoA)^\top - \oA(\oA)^\top} \lesssim \sqrt{\frac{T}{r}}\bar{\zeta}\fnorm{\hoA\widehat{\bTheta} - \oA\widebar{\bTheta}} \lesssim \bar{\zeta}^{-1}r\sqrt{\frac{p}{nT}} + \bar{\zeta}^{-1}\sqrt{r}\sqrt{\frac{r+\log T}{n}} + \sqrt{r}h.
\end{equation}
This proves our previous claim. The remaining argument is the same as in the proof of Theorem \ref{thm: mtl}.

\subsection{Proof of Theorem \ref{thm: no local hard}}
We prove a slightly stronger version of Theorem \ref{thm: no local hard} by replacing the constraint set
\begin{align}
	&4(1+\sqrt{2})\max_{t \in S}\twonorm{\hAk{t}(\hAk{t}) - \oA(\oA)^\top} + 4\frac{\epsilon}{1-\epsilon} < 1-\gamma, \\
	&\sqrt{2}\max_{t \in S}\bigg\{\frac{\sigmamax(\bSigmak{t})}{\sigmamin(\bSigmak{t})}\twonorm{\hAk{t}(\hAk{t})^\top  - \bAks{t}(\bAks{t})^\top}\bigg\} < 1-\gamma. \label{eq: current constraint}
\end{align}
with
\begin{align}
	&2(2+\sqrt{2})\max_{t \in S}\min_{\bm{R} \in \mathcal{O}^{r \times r}}\twonorm{\hAk{t} - \oA\bm{R}} + 4\frac{\epsilon}{1-\epsilon} < 1-\gamma, \\
	&\max_{t \in S}\bigg\{\frac{\sigmamax(\bSigmak{t})}{\sigmamin(\bSigmak{t})}\min_{\bm{R} \in \mathcal{O}^{r \times r}} \twonorm{\hAk{t}-\bAks{t}\bm{R}}\bigg\} < 1-\gamma. \label{eq: main text constraint}
\end{align}
Note that for any matrices $\bA, \bA' \in \mO$, Lemma 2.6 in \cite{chen2021spectral} tells us that
\begin{equation}
	\twonorm{\bA\bA^\top - \bA'(\bA')^\top} \leq \min_{\bm{R} \in \mathcal{O}^{r \times r}}\twonorm{\bA - \bA'\bm{R}} \leq \sqrt{2}\twonorm{\bA\bA^\top - \bA'(\bA')^\top}.
\end{equation} 
Therefore the current constraint set \eqref{eq: current constraint} is indeed weaker than the constraint set \eqref{eq: main text constraint} in the main text.

The proof follows the construction idea in the proof of Theorem \ref{thm: no local easy} and the argument in the proof of Proposition \ref{prop: mtl}. The difference is that here we will first consider a relaxed optimization problem where we require $\hAk{t}$ and $\hoA$ to satisfy $\twonorm{\hAk{t}(\hAk{t})^\top - \bm{I}_r}$, $\twonorm{\hoA(\hoA)^\top - \bm{I}_r} \leq \delta$ with a small constant $\delta > 0$ instead of $\hAk{t}, \hoA \in \mathcal{O}^{p \times r}$. Finally, we will let $\delta \rightarrow 0+$ to obtain the desired original result.

Denote
\begin{equation}
	G(\{\bAk{t}\}_{t=1}^T, \oA, \bTheta) = \frac{1}{T}\sum_{t=1}^T \fk{t}(\bAk{t}\bthetak{t}) + \sum_{t=1}^T \frac{\lambda}{\sqrt{n}T}\twonorm{\bAk{t}(\bAk{t})^\top -\oA(\oA)^\top},
\end{equation}
with $\bTheta = (\bthetak{1}, \ldots, \bthetak{T}) \in \mathbb{R}^{r \times T}$. First, consider any local minimizer $\{\hAk{t}\}_{t=1}^T$ and $\hoA$ of
\begin{equation}
	\min_{\{\bthetak{t}\}_{t=1}^T \subseteq \mathbb{R}^r}G(\{\bAk{t}\}_{t=1}^T, \oA, \bTheta),
\end{equation}
where $\twonorm{\hAk{t}(\hAk{t})^\top - \bm{I}_r}$, $\twonorm{\hoA(\hoA)^\top - \bm{I}_r} \leq \delta$ with a small constant $\delta > 0$. Define $\bA^{(t)} = a_1\oA\bm{D}^{(t)} + a_2\hAk{t}$ with $a_1 + a_2 = 1$ and $a_1, a_2 \geq 0$, $\bm{D}^{(t)} \in \mathcal{O}^{p \times r}$. Define a ``normalization" operator $\mathcal{N}(\cdot)$ such that $\mathcal{N}(\bA) \coloneqq \bA(\bA^\top \bA)^{-1/2} \in \mathcal{O}^{p \times r}$ for any $\bA \in \mathbb{R}^{p \times r}$ with full column rank. Define two distances between subspaces spanned by columns of $\bA$ and $\bm{B} \in \mathbb{R}^{p \times r}$ as
\begin{align}
	d_1(\bA, \bm{B}) &= \twonorm{\mathcal{N}(\bA)(\mathcal{N}(\bA))^\top - \mathcal{N}(\bm{B})(\mathcal{N}(\bm{B}))^\top}, \\
	d_2(\bA, \bm{B}) &= \min_{\bm{R} \in \mathcal{O}^{r \times r}}\twonorm{\mathcal{N}(\bA) - \mathcal{N}(\bm{B})\bm{R}}.
\end{align}
By Lemma 2.6 of \cite{chen2021spectral}, $d_1(\bA, \bm{B})\leq d_2(\bA, \bm{B}) \leq \sqrt{2}d_1(\bA, \bm{B})$.

Note that
\begin{equation}
	\bm{A}^{(t)}(\bm{A}^{(t)})^\top = a_1^2\bm{I}_r + a_2^2\bm{I}_r + a_1a_2[(\bm{D}^{(t)})^\top (\oA)^\top \hAk{t} + (\hAk{t})^\top \oA\bm{D}^{(t)}].
\end{equation}
In addition, there exists a rotation matrix $\bm{R}^{(t)} \in \mathcal{O}^{r \times r}$ s.t.
\begin{equation}
	\hthetak{t} = \bm{R}^{(t)}\frac{\twonorm{\hthetak{t}}}{\twonorm{\bthetak{t}_{\oA}}}\bthetak{t}_{\oA}.
\end{equation}
Consider $\bthetak{t} \in \mathbb{R}^r$ s.t.
\begin{equation}
	\bthetak{t} = \alpha \cdot \frac{\hthetak{t}}{\twonorm{\hthetak{t}}},
\end{equation}
then by taking $\bm{D}^{(t)} = (\bm{R}^{(t)})^{-1}$, we have
\begin{equation}
	\bAk{t}\bthetak{t} = a_1\oA(\bm{R}^{(t)})^{-1}\bthetak{t} + a_2\hAk{t}\bthetak{t} = a_1\frac{\alpha}{\twonorm{\bthetak{t}_{\oA}}}\oA\bthetak{t}_{\oA} + a_2\frac{\alpha}{\twonorm{\hthetak{t}}}\hAk{t}\hthetak{t}.
\end{equation}
Let $a_1\frac{\alpha}{\twonorm{\bthetak{t}_{\oA}}} + a_2\frac{\alpha}{\twonorm{\hthetak{t}}} = 1$, we obtain that $\alpha = \big(\frac{a_1}{\twonorm{\bthetak{t}_{\oA}}} + \frac{a_2}{\twonorm{\hthetak{t}}}\big)^{-1}$, which implies that
\begin{equation}
	\bAk{t}\bthetak{t} - \hAk{t}\hthetak{t} = \underbrace{\frac{a_1/\twonorm{\bthetak{t}_{\oA}}}{a_1/\twonorm{\bthetak{t}_{\oA}} + a_2/\twonorm{\hthetak{t}}}}_{\coloneqq a_1^{(t)}}(\oA\bthetak{t}_{\oA} - \hAk{t}\hthetak{t}).
\end{equation}
Also define $a_2^{(t)} = 1-a_1^{(t)}$. We claim that
\begin{equation}\label{eq: claim a1 t}
	C \geq \frac{a_1^{(t)}}{a_1} = \frac{1}{a_1 + a_2\cdot \frac{\twonorm{\bthetak{t}_{\oA}}}{\twonorm{\hthetak{t}}}} \gtrsim \frac{\twonorm{\hthetak{t}}}{\twonorm{\bthetak{t}_{\oA}}} \geq C' > 0, \quad \forall t \in S,
\end{equation}
with some constants $C, C' > 0$. To see this, note that 
\begin{equation}
	\hthetak{t} = ((\hAk{t})^\top \hSigmak{t} \hAk{t})^{-1} (\hAk{t})^\top \hSigmak{t}\bAks{t}\bthetaks{t} -  ((\hAk{t})^\top \hSigmak{t} \hAk{t})^{-1}(\hAk{t})^\top\nabla \fk{t}(\bAks{t}\bthetaks{t}).
\end{equation}
Also note that $((\hAk{t})^\top \hSigmak{t} \hAk{t})^{-1}$ and $(\hSigmak{t})^{-1}$ share the same eigenvalues. Since $\sigmamax((\hSigmak{t})^{-1}) \asymp \sigmamin((\hSigmak{t})^{-1}) \asymp 1$ \wppp\, by Lemma \ref{lem: cov hat}, we have 
\begin{equation}
	\sigmamax\big(((\hAk{t})^\top \hSigmak{t} \hAk{t})^{-1}\big) \asymp \sigmamin\big(((\hAk{t})^\top \hSigmak{t} \hAk{t})^{-1}\big) \asymp 1,
\end{equation}
\wppp. Furthermore, by Weyl's inequality and Lemma \ref{lem: cov hat}, with $\bm{R} \in \argmin_{\bm{R} \in \mathcal{O}^{r\times r}}\twonorm{\mathcal{N}(\hAk{t}) - \bAks{t}\bm{R}}$,
\begin{align}
	&\sigma_r((\hAk{t})^\top \hSigmak{t}\bAks{t}) \\
	&\geq \sigma_r((\hAk{t})^\top \bSigmak{t}\bAks{t}) - \twonorm{\hSigmak{t} - \bSigmak{t}}\\
	&\geq \sigma_r(\bm{R}^\top (\bAks{t})^\top \bSigmak{t}\bAks{t}) - \sigmamax((\bm{R}^\top (\bAks{t})^\top - \mathcal{N}(\hAk{t})^\top ) \bSigmak{t}\bAks{t}) -C\delta - C\sqrt{\frac{p+\log T}{n}} \\
	&\geq \sigma_r((\bAks{t})^\top \bSigmak{t}\bAks{t}) - \twonorm{\mathcal{N}(\hAk{t}) - \bAks{t}\bm{R}}\twonorm{\bSigmak{t}} -C\delta - C\sqrt{\frac{p+\log T}{n}} \\
	&\geq \sigmamin(\bSigmak{t}) - \twonorm{\mathcal{N}(\hAk{t}) - \bAks{t}\bm{R}}\cdot \sigmamax(\bSigmak{t})-C\delta - C\sqrt{\frac{p+\log T}{n}} \\
	&> 0,
\end{align}
\wppp, if $d_2(\hAk{t}, \bAks{t}) \leq \frac{\sigmamin(\bSigmak{t})}{\sigmamax(\bSigmak{t})}(1-\gamma)$ with some $\gamma > 0$. On the other hand, by Lemma \ref{lem: cov hat}, \wppp, for all $t \in S$,
\begin{equation}
	\sigmamax((\hAk{t})^\top \hSigmak{t}\bAks{t}) \lesssim 1.
\end{equation}
Hence \wppp,
\begin{align}
	\twonorm{\hthetak{t}} &\geq C\sigma_r((\hAk{t})^\top \hSigmak{t}\bAks{t})\twonorm{\bthetaks{t}} - C'\twonorm{\nabla \fk{t}(\bAks{t}\bthetaks{t})} \\
	&\geq C''\twonorm{\bthetaks{t}} -  C'''\sqrt{\frac{p+\log T}{n}} \\
	&\geq \frac{C''}{2}\twonorm{\bthetaks{t}}.
\end{align}
In addition, note that \wprp,
\begin{equation}
	\norm{\twonorm{\bthetak{t}_{\oA}} - \twonorm{\bthetaks{t}}} \leq \twonorm{\bthetak{t}_{\oA} - \bthetaks{t}} \lesssim h\zetak{t} + \sqrt{\frac{r+\log T}{n}}, 
\end{equation}
implying that \wprp,
\begin{align}
	\twonorm{\bthetak{t}_{\oA}} &\leq \twonorm{\bthetaks{t}} + C\bigg(h\zetak{t} + \sqrt{\frac{r+\log T}{n}}\bigg) \leq \twonorm{\bthetaks{t}} + C'\bigg(r^{-1/2}\zetak{t} + \sqrt{\frac{r+\log T}{n}}\bigg) \lesssim \twonorm{\bthetaks{t}},\\
	&\twonorm{\bthetaks{t}} \lesssim \twonorm{\bthetaks{t}} - C\bigg(h\zetak{t} + \sqrt{\frac{r+\log T}{n}}\bigg) \leq \twonorm{\bthetak{t}_{\oA}}.
\end{align}
Hence \wppp,
\begin{equation}
	\twonorm{\hthetak{t}} \geq \frac{C''}{2}\twonorm{\bthetaks{t}} \gtrsim \twonorm{\bthetak{t}_{\oA}}.
\end{equation}
This proves the second half of \eqref{eq: claim a1 t}. On the other hand, when $a_1 \leq 1/2$, we must have
\begin{equation}
	\frac{a_1^{(t)}}{a_1} \leq 2\frac{\twonorm{\hthetak{t}}}{\twonorm{\bthetak{t}_{\oA}}},
\end{equation}
and it is easy to see that \wprp,
\begin{equation}
	\twonorm{\hthetak{t}} \leq C\twonorm{\bthetaks{t}} + C\sqrt{\frac{p+\log T}{n}} \lesssim \twonorm{\bthetaks{t}} \lesssim \twonorm{\bthetak{t}_{\oA}},
\end{equation}
which proves the first half of \eqref{eq: claim a1 t}. Therefore, our claim \eqref{eq: claim a1 t} holds, and we will use it later in our proof.

Now similar to the proof of Proposition \ref{prop: mtl}, we divide $S$ into the following two index sets
\begin{align}
	\mA_1 &= \left\{t \in S: \twonorm{\bAk{t}(\bAk{t})^\top - \hbarA(\hbarA)^\top} \geq c\twonorm{\hbarA(\hbarA)^\top - \barA \barA^\top}\cdot \frac{1}{\sqrt{r}}\right\}, \\
	\mA_2 &= \left\{t \in S: \twonorm{\bAk{t}(\bAk{t})^\top - \hbarA(\hbarA)^\top} < c\twonorm{\hbarA(\hbarA)^\top - \barA \barA^\top}\cdot \frac{1}{\sqrt{r}}\right\},
\end{align}
where $c > 0$ is a small constant. For any index set $\mA \subseteq [T]$, define 
\begin{align}
	G_{\mA}(\{\bAk{t}\}_{t=1}^T, \oA, \bTheta) &= \frac{1}{T}\sum_{t\in \mA} \fk{t}(\bAk{t}\bthetak{t}) + \sum_{t\in \mA} \frac{\lambda}{\sqrt{n}T}\twonorm{\bAk{t}(\bAk{t})^\top -\oA(\oA)^\top} \\
	&= \frac{1}{2T}\sum_{t \in \mA} \twonorm{\bYk{t} - \bXk{t}\bAk{t}\bthetak{t}}^2 + \frac{\lambda}{T\sqrt{n}}\sum_{t \in \mA}\twonorm{\bAk{t}(\bAk{t})^\top - \bA \bA^\top}.
\end{align}
Note that by basic algebra,
\begin{align}
	&\frac{1}{2T}\sum_{t \in \mA} \twonorm{\bYk{t} - \bXk{t}\bAk{t}\bthetak{t}}^2  - \frac{1}{2T}\sum_{t \in \mA} \twonorm{\bYk{t} - \bXk{t}\hAk{t}\hthetak{t}}^2 \\
	&= \frac{1}{2T}(\bAk{t}\bthetak{t} - \hAk{t}\hthetak{t})^\top \hSigmak{t}(\bAk{t}\bthetak{t} - \hAk{t}\hthetak{t}) + \frac{1}{T}\sum_{t \in \mA}[\nabla \fk{t}(\hAk{t}\hthetak{t})]^\top (\bAk{t}\bthetak{t} - \hAk{t}\hthetak{t}) \\
	&=  \frac{1}{2T}(\bAk{t}\bthetak{t} - \hAk{t}\hthetak{t})^\top \hSigmak{t}(\bAk{t}\bthetak{t} - \hAk{t}\hthetak{t}) + \frac{1}{T}\sum_{t \in \mA}[\nabla \fk{t}(\oA\bthetak{t}_{\oA})]^\top (\bAk{t}\bthetak{t} - \hAk{t}\hthetak{t}) \\
	&\quad + \frac{1}{T}\sum_{t \in \mA}(\hAk{t}\hthetak{t} - \oA\bthetak{t}_{\oA})^\top \hSigmak{t} (\bAk{t}\bthetak{t}-\hAk{t}\hthetak{t}).
\end{align}
Hence
\begin{align}
	&G_{\mA_1}(\{\bAk{t}\}_{t=1}^T, \bA, \bTheta) - G_{\mA_1}(\{\hAk{t}\}_{t=1}^T, \hoA, \widehat{\bTheta}) \\
	&\leq \frac{1}{2T}(\bAk{t}\bthetak{t} - \hAk{t}\hthetak{t})^\top \hSigmak{t}(\bAk{t}\bthetak{t} - \hAk{t}\hthetak{t}) \\
	&\quad + \frac{1}{T}\sum_{t \in \mA_1}\twonorm{\nabla \fk{t}(\bAks{t}\bthetaks{t})}a_1^{(t)}(\twonorm{\oA\bthetak{t}_{\oA} - \hoA\bthetak{t}_{\hoA}} + \twonorm{\hoA\bthetak{t}_{\hoA} - \hAk{t}\hthetak{t}}) \\
	&\quad + \frac{1}{T}\fnorma{\Big\{\sqrt{a_1^{(t)}}[\nabla \fk{t}(\oA\bthetak{t}_{\oA}) - \nabla \fk{t}(\bAks{t}\bthetaks{t})]\Big\}_{t \in \mA_1}} \cdot \sqrt{\sum_{t \in \mA_1}a_1^{(t)}\twonorm{\oA\bthetak{t}_{\oA} - \hAk{t}\hthetak{t}}^2} \\
	&\quad - \frac{C}{T}\sum_{t \in \mA_1}a_1^{(t)}\twonorm{\oA\bthetak{t}_{\oA} - \hAk{t}\hthetak{t}}^2 + \frac{\lambda}{T\sqrt{n}}\sum_{t \in \mA_1}\twonorm{\bAk{t}(\bAk{t})^\top - \bA\bA^\top} \\
	&\quad - \frac{\lambda}{T\sqrt{n}}\sum_{t \in \mA_1}\twonorm{\hAk{t}(\hAk{t})^\top - \hoA(\hoA)^\top}, \label{eq: G A1}
\end{align}
with
\begin{align}
	[1] &=\frac{1}{2T}(\bAk{t}\bthetak{t} - \hAk{t}\hthetak{t})^\top \hSigmak{t}(\bAk{t}\bthetak{t} - \hAk{t}\hthetak{t}) \\
	&\quad + \frac{1}{T}\sum_{t \in \mA_1}\twonorm{\nabla \fk{t}(\bAks{t}\bthetaks{t})}a_1^{(t)}(\twonorm{\oA\bthetak{t}_{\oA} - \hoA\bthetak{t}_{\hoA}} + \twonorm{\hoA\bthetak{t}_{\hoA} - \hAk{t}\hthetak{t}}) \\
	&\quad + \frac{1}{T}\fnorma{\Big\{\sqrt{a_1^{(t)}}[\nabla \fk{t}(\oA\bthetak{t}_{\oA}) - \nabla \fk{t}(\bAks{t}\bthetaks{t})]\Big\}_{t \in \mA_1}} \cdot \sqrt{\sum_{t \in \mA_1}a_1^{(t)}\twonorm{\oA\bthetak{t}_{\oA} - \hAk{t}\hthetak{t}}^2} \\
	&\quad - \frac{C}{T}\sum_{t \in \mA_1}a_1^{(t)}\twonorm{\oA\bthetak{t}_{\oA} - \hAk{t}\hthetak{t}}^2,
\end{align}
and
\begin{equation}\label{eq: term [2] decomp}
	[2] = \frac{\lambda}{T\sqrt{n}}\sum_{t \in \mA_1}\twonorm{\bAk{t}(\bAk{t})^\top - \bA\bA^\top} - \frac{\lambda}{T\sqrt{n}}\sum_{t \in \mA_1}\twonorm{\hAk{t}(\hAk{t})^\top - \hoA(\hoA)^\top}.
\end{equation}
Recall that $\bAk{t}\bthetak{t} - \hAk{t}\hthetak{t} = a_1^{(t)}(\oA\bthetak{t}_{\oA} - \hAk{t}\hthetak{t})$ and our previous conclusion that \wprp, $a_1^{(t)}/a_1 \asymp 1$ for all $t \in S$, hence \wprp,
\begin{align}
	[1] &\leq \frac{C}{T}\sum_{t \in \mA_1}a_1^2 \twonorm{\oA\bthetak{t}_{\oA} - \hAk{t}\hthetak{t}}^2 \\
	&\quad + \frac{1}{T}\sum_{t \in \mA_1}\twonorm{\nabla \fk{t}(\bAks{t}\bthetaks{t})}a_1\zetak{t}\Big[d_1(\hoA, \oA) + d_1(\hoA, \hAk{t})\Big] \\
	&\quad + \frac{1}{T}\bigg(\sqrt{T}\sqrt{a_1}\sqrt{\frac{r+\log T}{n}} + h\sqrt{\sum_{t \in \mA_1} (\zetak{t})^2 a_1}\bigg)\sqrt{\sum_{t \in \mA_1}a_1\twonorm{\oA\bthetak{t}_{\oA} - \hAk{t}\hthetak{t}}^2} \\
	&\quad - \frac{C}{T}\sum_{t \in \mA_1}a_1\twonorm{\oA\bthetak{t}_{\oA} - \hAk{t}\hthetak{t}}^2 \\
	&\leq \frac{C}{T}\sum_{t \in \mA_1}a_1^2 \twonorm{\oA\bthetak{t}_{\oA} - \hAk{t}\hthetak{t}}^2 \\
	&\quad + \frac{1}{T}\sum_{t \in \mA_1}\twonorm{\nabla \fk{t}(\bAks{t}\bthetaks{t})}a_1\zetak{t}(\twonorm{\hoA(\hoA)^\top - \oA(\oA)^\top} + \twonorm{\hoA(\hoA)^\top - \hAk{t}(\hAk{t})^\top} + \delta) \\
	&\quad + \frac{1}{T}\bigg(\sqrt{T}\sqrt{a_1}\sqrt{\frac{r+\log T}{n}} + h\sqrt{\sum_{t \in \mA_1} (\zetak{t})^2 a_1}\bigg)\sqrt{\sum_{t \in \mA_1}a_1\twonorm{\oA\bthetak{t}_{\oA} - \hAk{t}\hthetak{t}}^2} \\
	&\quad - \frac{C}{T}\sum_{t \in \mA_1}a_1\twonorm{\oA\bthetak{t}_{\oA} - \hAk{t}\hthetak{t}}^2 \\
	&\leq \frac{C}{T}\sum_{t \in \mA_1}\sqrt{\frac{p+\log T}{n}}a_1\zetak{t}\sqrt{r}\cdot \twonorm{\hAk{t}(\hAk{t})^\top - \hoA(\hoA)^\top} + Ca_1\bigg(\frac{r+\log T}{n} + h^2 \cdot \frac{1}{T}\sum_{t \in \mA_1}(\zetak{t})^2\bigg) \\
	&\quad + C\frac{a_1}{T}\delta\sqrt{\frac{p+\log T}{n}}\sum_{t \in \mA_1}\zetak{t}
\end{align}
To bound $[2]$, we need to do some preparations. Note that
\begin{align}
	\bAk{t}(\bAk{t})^\top &= (a_1\oA\bm{D}^{(t)} + a_2\hAk{t})(a_1\oA\bm{D}^{(t)} + a_2\hAk{t})^\top \\
	&= a_2^2 \hAk{t}(\hAk{t})^\top + a_1a_2[\oA\bm{D}^{(t)}(\hAk{t})^\top + \hAk{t}(\bm{D}^{(t)})^\top (\oA)^\top] + \mathcal{O}(a_1^2) \\
	&= (1-2a_1)\hAk{t}(\hAk{t})^\top + a_1[\oA\bm{D}^{(t)}(\hAk{t})^\top + \hAk{t}(\bm{D}^{(t)})^\top (\oA)^\top] + \mathcal{O}(a_1^2),
\end{align}
where $\mathcal{O}(a_1^2)$ here refers to a $p \times p$ matrix with each entry of order $\mathcal{O}(a_1^2)$. Let $\bA = a_1\oA + a_2\hoA$, then similarly we have
\begin{equation}\label{eq: AAT}
	\bA\bA^\top = (1-2a_1)\hoA(\hoA)^\top  + a_1[\oA(\hoA)^\top + \hoA(\oA)^\top]+ \mathcal{O}(a_1^2).
\end{equation}
This implies that
\begin{align}
	&\bAk{t}(\bAk{t})^\top - \bA\bA^\top \\
	&= (1-a_1)[\hAk{t}(\hAk{t})^\top - \hoA(\hoA)^\top] -\frac{a_1}{2}(\hAk{t}(\bm{D}^{(t)})^\top + \hoA)(\hAk{t}(\bm{D}^{(t)})^\top - \hoA)^\top \\
	&\quad  - \frac{a_1}{2}(\hAk{t}(\bm{D}^{(t)})^\top - \hoA)(\hAk{t}(\bm{D}^{(t)})^\top + \hoA)^\top + a_1\oA[\bm{D}^{(t)}(\hAk{t})^\top - (\hoA)^\top] \\
	&\quad + a_1[\hAk{t}(\bm{D}^{(t)})^\top - \hoA](\oA)^\top + \mathcal{O}(a_1^2) \\
	&=  (1-a_1)[\hAk{t}(\hAk{t})^\top - \hoA(\hoA)^\top] + a_1\bigg[\oA - \frac{1}{2}(\hAk{t}(\bm{D}^{(t)})^\top + \hoA)\bigg](\hAk{t}(\bm{D}^{(t)})^\top - \hoA)^\top \\
	&\quad + a_1[\hAk{t}(\bm{D}^{(t)})^\top - \hoA]\bigg[\oA - \frac{1}{2}(\hAk{t}(\bm{D}^{(t)})^\top + \hoA)\bigg]^\top + \mathcal{O}(a_1^2) \\
	&= (1-a_1)[\hAk{t}(\hAk{t})^\top - \hoA(\hoA)^\top]+ a_1(\oA - \hoA)[\hAk{t}(\bm{D}^{(t)})^\top - \hoA]^\top \\
	&\quad - a_1[\hAk{t}(\bm{D}^{(t)})^\top - \hoA][\hAk{t}(\bm{D}^{(t)})^\top - \hoA]^\top + a_1[\hAk{t}(\bm{D}^{(t)})^\top - \hoA](\oA - \hoA)^\top  + \mathcal{O}(a_1^2).
\end{align}
Note that we can multiply $\oA$, $\hAk{t}$, and $\hoA$ by a rotation matrix from the right without changing their definitions
\begin{align}
	\oA &\in \argmin_{\bA \in \mathcal{O}^{p \times r}}\twonorm{\bAks{t}(\bAks{t})^\top - \bA\bA^\top}, \\
	\{\hAk{t}\}_{t=1}^T, \hoA &\in \argmin_{\substack{\{\bAk{t}\}_{t=1}^T, \oA \in \mathbb{R}^{p \times r}\\ \twonorm{(\bAk{t})^\top \bAk{t} - \bm{I}_r}, \twonorm{(\oA)^\top \oA - \bm{I}_r}\leq \delta}}  \min_{\bTheta \in \mathbb{R}^{r \times T}}G_{[T]}(\{\bAk{t}\}_{t=1}^T, \oA, \bTheta).
\end{align}
Therefore, WLOG, we assume
\begin{align}
	\twonorm{\oA - \mathcal{N}(\hoA)} &= \min_{\bm{R} \in \mathcal{O}^{r \times r}}\twonorm{\oA - \mathcal{N}(\hoA)\bm{R}} = d_2(\oA, \hoA), \label{eq: wlog asmp}\\
	\twonorm{\mathcal{N}(\hAk{t}(\bm{D}^{(t)})^\top) - \mathcal{N}(\hoA)} &= \min_{\bm{R} \in \mathcal{O}^{r \times r}}\twonorm{\mathcal{N}(\hAk{t}) - \mathcal{N}(\hoA)\bm{R}} = d_2(\hAk{t}, \hoA). 
\end{align}
Therefore, by the triangle inequality,
\begin{align}
	\twonorm{\bAk{t}(\bAk{t})^\top - \bA\bA^\top} &\leq (1-a_1)\twonorm{\hAk{t}(\hAk{t})^\top - \hoA(\hoA)^\top} + \max\Big\{2a_1\twonorm{\oA - \hoA}\twonorm{\hAk{t}(\bm{D}^{(t)})^\top - \hoA}, \\
	&\quad\,\, a_1\twonorm{\hAk{t}(\bm{D}^{(t)})^\top - \hoA}^2\Big\} + C\sqrt{p}a_1^2 \\
	&\leq (1-a_1)\twonorm{\hAk{t}(\hAk{t})^\top - \hoA(\hoA)^\top} + \max\Big\{2\sqrt{2}a_1\cdot d_2(\oA, \hoA)d_1(\hAk{t}, \hoA), \\
	&\quad\,\, \sqrt{2}a_1\cdot d_2(\hAk{t}, \hoA)d_1(\hAk{t}, \hoA)\Big\} + Ca_1\delta +  C\sqrt{p}a_1^2 \\
	&\leq \twonorm{\hAk{t}(\hAk{t})^\top - \hoA(\hoA)^\top}+ \Big[-1 + \big(2\sqrt{2}d_2(\oA, \hoA)\big) \vee \big(\sqrt{2}d_2(\hAk{t}, \hoA)\big)\Big] \\
	&\quad \cdot a_1d_1(\hAk{t}, \hoA) + Ca_1\delta +  C\sqrt{p}a_1^2 \\
	&\leq \twonorm{\hAk{t}(\hAk{t})^\top - \hoA(\hoA)^\top}+ \Big[-2^{-\frac{1}{2}} + \big(2d_2(\oA, \hoA)\big) \vee \big(d_2(\oA, \hoA) + d_2(\hAk{t}, \oA)\big)\Big] \\
	&\quad \cdot \sqrt{2}a_1d_1(\hAk{t}, \hoA) + Ca_1\delta +  C\sqrt{p}a_1^2 \\
	&\leq \twonorm{\hAk{t}(\hAk{t})^\top - \hoA(\hoA)^\top} + \Big[-2^{-\frac{1}{2}} + \big(2d_2(\oA, \hoA)\big) \vee \big(d_2(\oA, \hoA) + d_2(\hAk{t}, \oA)\big)\Big]\\
	&\quad \cdot \sqrt{2}a_1\twonorm{\hAk{t}(\hAk{t})^\top - \hoA(\hoA)^\top} + Ca_1\delta +  C\sqrt{p}a_1^2.
\end{align}
To further bound the RHS, we claim that $d_2(\oA, \hoA) \leq (1 + \sqrt{2})\max_{t \in S}d_2(\hAk{t}, \oA) + \sqrt{2}\frac{\epsilon}{1-\epsilon} + C'\delta$ with some constant $C' > 0$. To see this, denote $q = \max_{t \in S}d_2(\hAk{t}, \oA)$, and suppose $d_2(\hoA, \oA) \geq q+b$ with some $b \geq 0$, then $\min_{t \in S}d_2(\hoA, \hAk{t}) \geq b$. This leads to
\begin{equation}
	\sum_{t=1}^T \twonorm{\hoA(\hoA)^\top - \hAk{t}(\hAk{t})^\top} \geq \frac{1}{\sqrt{2}}\sum_{t \in S}d_2(\hoA, \hAk{t}) - C\delta \geq \frac{|S|}{\sqrt{2}}b - CT\delta.
\end{equation}
On the other hand, we have
\begin{equation}
	\sum_{t \in S}d_2(\hAk{t}, \oA) \leq |S|q, \quad \sum_{t \in S^c}d_2(\hAk{t}, \oA) \leq |S^c|,
\end{equation}
which implies that
\begin{equation}
	\sum_{t=1}^T \twonorm{\oA(\oA)^\top - \hAk{t}(\hAk{t})^\top} \leq |S|q + |S^c| + CT\delta.
\end{equation}
By the optimization procedure, we know that $\hoA \in \argmin\limits_{\bA \in \mathbb{R}^{p \times r}: \twonorm{\bA^\top \bA - \bm{I}_r} \leq \delta} \sum_{t=1}^T \twonorm{\bA \bA^\top - \hAk{t}(\hAk{t})^\top}$. Therefore we must have $|S|q + |S^c| + CT\delta \geq \frac{|S|}{\sqrt{2}}b - CT\delta$, i.e.
\begin{equation}
	b \leq \sqrt{2}q + \sqrt{2}\frac{|S^c|}{|S|} + C'\delta \leq \sqrt{2}q + \sqrt{2}\frac{\epsilon}{1-\epsilon} + C'\delta.
\end{equation}
In other words,
\begin{equation}
	d_2(\oA, \hoA) \leq q+b \leq (1 + \sqrt{2})\max_{t \in S}d_2(\hAk{t}, \oA) + \sqrt{2}\frac{\epsilon}{1-\epsilon} + C'\delta,
\end{equation}
which proves the claim. Therefore, using this claim, we have
\begin{align}
	&\twonorm{\bAk{t}(\bAk{t})^\top - \bA\bA^\top} \\
	&\leq \twonorm{\hAk{t}(\hAk{t})^\top - \hoA(\hoA)^\top} + \Big[-2^{-\frac{1}{2}} + 2(1 + \sqrt{2})\max_{t \in S}d_2(\hAk{t}, \oA) + 2\sqrt{2}\frac{\epsilon}{1-\epsilon}\Big]\\
	&\quad \cdot \sqrt{2}a_1\twonorm{\hAk{t}(\hAk{t})^\top - \hoA(\hoA)^\top} + Ca_1\delta +  C\sqrt{p}a_1^2\\
	&\leq \twonorm{\hAk{t}(\hAk{t})^\top - \hoA(\hoA)^\top} - \gamma a_1\twonorm{\hAk{t}(\hAk{t})^\top - \hoA(\hoA)^\top} + Ca_1\delta +  C\sqrt{p}a_1^2,
\end{align}
when $2(2+\sqrt{2})\max_{t \in S}d_2(\hAk{t}, \oA) + 4\frac{\epsilon}{1-\epsilon} < 1-\gamma$. 

Therefore, we can upper bound the term in \eqref{eq: term [2] decomp} as
\begin{equation}
	[2] \leq -\frac{\lambda}{T\sqrt{n}}\cdot \gamma\cdot a_1 \cdot \sum_{t \in \mA_1}\twonorm{\hAk{t}(\hAk{t})^\top - \hoA(\hoA)^\top} + C\frac{\lambda}{\sqrt{n}}\gamma\cdot a_1\delta +  C\frac{\lambda}{\sqrt{n}}\cdot \gamma\cdot \sqrt{p}a_1^2.
\end{equation}
Therefore, using the bounds we have proved for $[1]$ and $[2]$, plugging them back in \eqref{eq: G A1}, we have
Hence
\begin{align}
	&G_{\mA_1}(\{\bAk{t}\}_{t=1}^T, \bA, \bTheta) - G_{\mA_1}(\{\hAk{t}\}_{t=1}^T, \hoA, \widehat{\bTheta}) \\
	&\leq \frac{C}{T}\sum_{t \in \mA_1}\sqrt{\frac{p+\log T}{n}}a_1\zetak{t}\sqrt{r}\cdot \twonorm{\hAk{t}(\hAk{t})^\top - \hoA(\hoA)^\top} + Ca_1\bigg(\frac{r+\log T}{n} + h^2 \cdot \frac{1}{T}\sum_{t \in \mA_1}(\zetak{t})^2\bigg) \\
	&\quad + C\frac{a_1}{T}\delta\sqrt{\frac{p+\log T}{n}}\sum_{t \in \mA_1}\zetak{t}  -\frac{\lambda}{T\sqrt{n}}\cdot \gamma\cdot a_1 \cdot \sum_{t \in \mA_1}\twonorm{\hAk{t}(\hAk{t})^\top - \hoA(\hoA)^\top} \\
	&\quad + C\frac{\lambda}{\sqrt{n}}\gamma\cdot a_1\delta +  C\frac{\lambda}{\sqrt{n}}\cdot \gamma\cdot \sqrt{p}a_1^2 \\
	&\leq Ca_1\bigg(\frac{r+\log T}{n} + h^2 \cdot \frac{1}{T}\sum_{t \in \mA_1}(\zetak{t})^2\bigg) + C\frac{a_1}{T}\delta\sqrt{\frac{p+\log T}{n}}\sum_{t \in \mA_1}\zetak{t} \\
	&\quad - \frac{\lambda}{T\sqrt{n}}\cdot \gamma\cdot a_1 \cdot \sum_{t \in \mA_1}\twonorm{\hAk{t}(\hAk{t})^\top - \hoA(\hoA)^\top} + C\frac{\lambda}{\sqrt{n}}\gamma\cdot a_1\delta +  C\frac{\lambda}{\sqrt{n}}\cdot \gamma\cdot \sqrt{p}a_1^2.
\end{align}

Considering $G_{\mA_2}(\{\bAk{t}\}_{t=1}^T, \oA, \bTheta) - G_{\mA_2}(\{\hAk{t}\}_{t=1}^T, \hoA, \widehat{\bTheta})$, first we have
\begin{align}
	&\frac{1}{T}\sum_{t \in \mA_2}[\nabla \fk{t}(\bAks{t}\bthetaks{t})]^\top a_1^{(t)}\cdot (\oA\bthetak{t}_{\oA} - \hoA\bthetak{t}_{\hoA} + \hoA\bthetak{t}_{\hoA} - \hAk{t}\hthetak{t}) \\
	&\leq \frac{1}{T}\twonorma{\Big\{\sqrt{a_1^{(t)}}\nabla \fk{t}(\bAks{t}\bthetaks{t})\Big\}_{t \in \mA_2}}\sqrt{2r}\cdot \fnorma{\Big\{\sqrt{a_1^{(t)}}(\oA\bthetak{t}_{\oA} - \hoA\bthetak{t}_{\hoA})\Big\}_{t \in \mA_2}} \\
	&\quad + \frac{1}{T}\sum_{t \in \mA_2}\twonorm{\nabla \fk{t}(\bAks{t}\bthetaks{t})}\cdot a_1^{(t)}\cdot \twonorm{\hoA\bthetak{t}_{\hoA} - \hAk{t}\hthetak{t}} \\
	&\leq \frac{1}{T}\sqrt{2r}C\cdot \sqrt{\frac{p+T}{n}}a_1 \fnorma{\big\{(\oA\bthetak{t}_{\oA} - \hoA\bthetak{t}_{\hoA})\big\}_{t \in \mA_2}} + \frac{1}{T}\sum_{t \in \mA_2}\sqrt{\frac{p+\log T}{n}}a_1\cdot \twonorm{\hAk{t}(\hAk{t})^\top - \hoA(\hoA)^\top} \\
	&\leq C\frac{r(p+T)}{nT}a_1 + \frac{C}{T}a_1\fnorma{\big\{(\oA\bthetak{t}_{\oA} - \hoA\bthetak{t}_{\hoA})\big\}_{t \in \mA_2}}^2 + \frac{1}{T}\sum_{t \in \mA_2}\sqrt{\frac{p+\log T}{n}}a_1\cdot \twonorm{\hAk{t}(\hAk{t})^\top - \hoA(\hoA)^\top},
\end{align}
where the second inequality holds because
\begin{equation}
	\twonorma{\Big\{\sqrt{a_1^{(t)}}\nabla \fk{t}(\bAks{t}\bthetaks{t})\Big\}_{t \in \mA_2}} \lesssim \max_{t \in \mA_2}\sqrt{a_1^{(t)}}\cdot \sqrt{\frac{p+T}{n}}, 
\end{equation}
\wpp, which is due to Lemma 5.39 in \cite{vershynin2010introduction}. Therefore, by using the same arguments to bound $G_{\mA_1}(\{\bAk{t}\}_{t=1}^T, \oA, \bTheta) - G_{\mA_1}(\{\hAk{t}\}_{t=1}^T, \hoA, \widehat{\bTheta})$, we can prove that
\begin{align}
	&G_{\mA_2}(\{\bAk{t}\}_{t=1}^T,\bA, \bTheta) - G_{\mA_2}(\{\hAk{t}\}_{t=1}^T, \hoA, \widehat{\bTheta}) \\
	&\leq \frac{1}{2T}(\bAk{t}\bthetak{t} - \hAk{t}\hthetak{t})^\top \hSigmak{t}(\bAk{t}\bthetak{t} - \hAk{t}\hthetak{t}) \\
	&\quad + C\frac{r(p+T)}{nT}a_1^2 + \frac{C}{T}\fnorma{\big\{(\oA\bthetak{t}_{\oA} - \hoA\bthetak{t}_{\hoA})\big\}_{t \in \mA_2}}^2 + \frac{1}{T}\sum_{t \in \mA_2}\sqrt{\frac{p+\log T}{n}}a_1\cdot \twonorm{\hAk{t}(\hAk{t})^\top - \hoA(\hoA)^\top}\\
	&\quad + \frac{C}{T}\bigg(\sqrt{T}\sqrt{a_1}\sqrt{\frac{r+\log T}{n}} + h\sqrt{\sum_{t \in \mA_2} (\zetak{t})^2 a_1}\bigg)\sqrt{\sum_{t \in \mA_2}a_1\twonorm{\oA\bthetak{t}_{\oA} - \hAk{t}\hthetak{t}}^2} \\
	&\quad  - \frac{C}{T}\sum_{t \in \mA_2}a_1\twonorm{\hAk{t}\hthetak{t} - \oA\bthetak{t}_{\oA}}^2 - \frac{\lambda}{T\sqrt{n}}\cdot \gamma a_1\cdot \sum_{t \in \mA_2}\twonorm{\hAk{t}(\hAk{t})^\top - \hoA(\hoA)^\top} \label{eq: A 1}\\
	&\leq - \frac{C}{T}\sum_{t \in \mA_2}a_1\twonorm{\hAk{t}\hthetak{t} - \oA\bthetak{t}_{\oA}}^2 - \frac{\lambda}{T\sqrt{n}}\cdot \gamma a_1\cdot \sum_{t \in \mA_2}\twonorm{\hAk{t}(\hAk{t})^\top - \hoA(\hoA)^\top} \\
	&\quad + Ca_1\bigg(\frac{pr}{nT} + \frac{r+\log T}{n} + h^2\cdot \frac{1}{T}\sum_{t \in \mA_2}(\zetak{t})^2\bigg).
\end{align}
Since $\twonorm{\hAk{t}\hthetak{t} - \hbarA\bthetak{t}_{\hbarA}} \leq \twonorm{\hAk{t}(\hAk{t})^\top - \hbarA(\hbarA)^\top}(\twonorm{\bthetak{t}_{\hbarA}}+\twonorm{\nabla \fk{t}(\hbarA \bthetak{t}_{\hbarA})}) \leq C\zetak{t}\twonorm{\hAk{t}(\hAk{t})^\top-\hbarA(\hbarA)^\top}\leq C\zetak{t}\twonorm{\oA(\oA)^\top-\hbarA(\hbarA)^\top}\frac{c}{\sqrt{r}}$, we have
\begin{align}
	\twonorm{\hAk{t}\hthetak{t} - \oA\bthetak{t}_{\oA}}^2 &\geq \frac{1}{2}\twonorm{\hoA\bthetak{t}_{\hoA} - \oA\bthetak{t}_{\oA}}^2 - 2\twonorm{\hAk{t}\hthetak{t} - \hoA\bthetak{t}_{\hoA}}^2 \\
	&\geq \frac{1}{2}\twonorm{\hoA\bthetak{t}_{\hoA} - \oA\bthetak{t}_{\oA}}^2 - \frac{2C^2c}{r}(\zetak{t})^2\twonorm{\oA(\oA)^\top-\hbarA(\hbarA)^\top}^2.
\end{align}
This implies that
\begin{align}
	&G_{\mA_2}(\{\bAk{t}\}_{t=1}^T, \bA, \bTheta) - G_{\mA_2}(\{\hAk{t}\}_{t=1}^T, \hoA, \widehat{\bTheta}) \\
	&\leq -\frac{C'}{T}a_1 \fnorm{\hoA\widehat{\bTheta} - \oA\widebar{\bTheta}}^2 + \frac{2C^2c}{r}a_1(\zetak{t})^2\twonorm{\oA(\oA)^\top-\hbarA(\hbarA)^\top}^2\\
	&\quad + Ca_1\bigg(\frac{pr}{nT} + \frac{r+\log T}{n} + h^2\cdot \frac{1}{T}\sum_{t \in \mA_2}(\zetak{t})^2\bigg).
\end{align}
In addition, by triangle inequality,
\begin{equation}
	G_{S^c}(\{\bAk{t}\}_{t=1}^T, \bA, \bTheta) - G_{S^c}(\{\hAk{t}\}_{t=1}^T, \hoA, \widehat{\bTheta}) \leq \frac{\lambda}{\sqrt{n}T}|S^c|\twonorm{\hoA(\hoA)^\top - \bA\bA^\top}.
\end{equation}
Recall \eqref{eq: AAT}:
\begin{equation}
	\bA\bA^\top = (1-2a_1)\hoA(\hoA)^\top  + a_1[\oA(\hoA)^\top + \hoA(\oA)^\top]+ \mathcal{O}(a_1^2).
\end{equation}
Hecne we have
\begin{align}
	\twonorm{\bA\bA^\top - \hoA(\hoA)^\top} &\leq a_1\twonorm{\hoA(\oA - \hoA)^\top} + a_1\twonorm{\oA(\hoA - \oA)^\top}+ Ca_1^2 \\
	&\leq 2a_1d_2(\oA, \hoA) + Ca_1\delta + Ca_1^2\\
	&\leq 2\sqrt{2}a_1\twonorm{\oA(\oA)^\top - \hoA(\hoA)^\top} + Ca_1\delta + Ca_1^2,
\end{align}
where the second inequality is due to \eqref{eq: wlog asmp}. Therefore,
\begin{equation}
	G_{S^c}(\{\bAk{t}\}_{t=1}^T, \bA, \bTheta) - G_{S^c}(\{\hAk{t}\}_{t=1}^T, \hoA, \widehat{\bTheta}) \leq 2\sqrt{2}a_1\frac{\lambda}{\sqrt{n}T}|S^c|\twonorm{\oA(\oA)^\top - \hoA(\hoA)^\top} + Ca_1\delta + Ca_1^2.
\end{equation}
\underline{\textbf{Case 1:}} If $|\mA_2| \geq \Big(1-cr^{-1}\frac{\bar{\zeta}^2}{\max_{t \in S}(\zetak{t})^2}\Big)|S|$:

By Wedin's $\sin\Theta$-theorem, we have
By Assumption \ref{asmp: theta}, Wedin's $\sin \Theta$-Theorem, and applying Lemma \ref{lem: small fraction diversity} on $\{\bthetak{t}_{\barA}\}_{t \in \mathcal{A}_2}$,
\begin{equation}\label{eq: thm y eq x}
	\twonorm{\mathcal{N}(\hbarA)(\mathcal{N}(\hbarA))^\top - \barA(\barA)^\top} \leq \fnorm{\mathcal{N}(\hbarA)(\mathcal{N}(\hbarA)^\top - \barA(\barA)^\top} \lesssim \sqrt{\frac{r}{|S|}}\barzeta^{-1} \cdot \fnorm{\{\hbarA\bthetak{t}_{\hbarA} - \barA\bthetak{t}_{\barA}\}_{t \in \mA_2}}.
\end{equation}
Here we used the fact that $\frac{1}{|S|}\sum_{t \in S}\bthetak{t}_{\barA}(\bthetak{t}_{\barA})^\top \succeq \frac{c}{r}\barzeta^2 \bm{I}_p$. To see this, notice that
\begin{align}
	\sigma_r\left(\frac{1}{\sqrt{|S|}}\{\barA\bthetak{t}_{\barA}\}_{t \in S}\right) &= \sigma_r\left(\frac{1}{\sqrt{|S|}}\{\bbetaks{t}\}_{t \in S}\right) - \frac{1}{\sqrt{|S|}}\twonorm{\{\barA\bthetak{t}_{\barA}\}_{t \in S}-\{\bbetaks{t}\}_{t \in S}} \\
	&\geq \frac{c}{\sqrt{r}}\barzeta - \left(\barzeta h+\sqrt{\frac{r+\log T}{n}}\right) \\
	&\geq \frac{c'}{\sqrt{r}}\barzeta,
\end{align}
where by Lemmas \ref{lem: a theta a} and \ref{lem: matrix 2 norm},
\begin{equation}
	\frac{1}{\sqrt{|S|}}\twonorm{\{\barA\bthetak{t}_{\barA}\}_{t \in S}-\{\bbetaks{t}\}_{t \in S}} \leq \sqrt{\sum_{t \in S} \twonorm{\barA\bthetak{t}_{\barA} - \bAks{t}\bthetaks{t}}^2} \lesssim h\barzeta + \sqrt{\frac{r+\log T}{n}},
\end{equation}
and the last step comes from conditions (\rom{1}) and (\rom{2}) by noticing that
\begin{equation}
	\sqrt{\frac{r+\log T}{n}} \lesssim \frac{\barzeta}{\sqrt{r}}\cdot \sqrt{\frac{p+\log T}{n}}\cdot \frac{1}{\zetak{t}} \lesssim \frac{\barzeta}{\sqrt{r}}.
\end{equation}
This leads to
\begin{equation}
	\sigma_r\left(\frac{1}{|S|}\sum_{t \in S}\bthetak{t}_{\barA}(\bthetak{t}_{\barA})^\top\right) = \sigma_r\left(\frac{1}{|S|}\sum_{t \in S}\bthetak{t}_{\barA}(\barA)^\top\barA(\bthetak{t}_{\barA})^\top\right) \geq \frac{c'}{r}\barzeta^2.
\end{equation}
Hence
\begin{align}
	&G_{[T]}(\{\bAk{t}\}_{t=1}^T, \bA, \bTheta) - G_{[T]}(\{\hAk{t}\}_{t=1}^T, \hoA, \widehat{\bTheta}) \\
	&\leq [G_{\mA_1}(\{\bAk{t}\}_{t=1}^T, \bA, \bTheta) - G_{\mA_1}(\{\hAk{t}\}_{t=1}^T, \hoA, \widehat{\bTheta})] + [G_{\mA_2}(\{\bAk{t}\}_{t=1}^T, \bA, \bTheta) - G_{\mA_2}(\{\hAk{t}\}_{t=1}^T, \hoA, \widehat{\bTheta})] \\
	&\quad + [G_{S^c}(\{\bAk{t}\}_{t=1}^T, \bA, \bTheta) - G_{S^c}(\{\hAk{t}\}_{t=1}^T, \hoA, \widehat{\bTheta})] \\
	&\leq Ca_1\bigg(\frac{r+\log T}{n} + h^2 \cdot \frac{1}{T}\sum_{t \in \mA_1}(\zetak{t})^2\bigg) + C\frac{a_1}{T}\delta\sqrt{\frac{p+\log T}{n}}\sum_{t \in \mA_1}\zetak{t} + C\frac{\lambda}{\sqrt{n}}\gamma\cdot a_1\delta +  C\frac{\lambda}{\sqrt{n}}\cdot \gamma\cdot \sqrt{p}a_1^2 \\
	&\quad - \frac{C}{T}\sum_{t \in \mA_2}a_1\twonorm{\hAk{t}\hthetak{t} - \oA\bthetak{t}_{\oA}}^2 + Ca_1\bigg(\frac{pr}{nT} + \frac{r+\log T}{n} + h^2\cdot \frac{1}{T}\sum_{t \in \mA_2}(\zetak{t})^2\bigg)  \\
	&\quad + 2\sqrt{2}a_1\frac{\lambda}{\sqrt{n}T}|S^c|\twonorm{\oA(\oA)^\top - \hoA(\hoA)^\top} \\
	&\leq Ca_1\bigg(\frac{pr}{nT} + \frac{r+\log T}{n} + h^2\bar{\zeta}^2\bigg) - Ca_1\frac{\bar{\zeta}^2}{r}\twonorm{\hbarA(\hbarA)^\top - \barA(\barA)^\top}^2 + 2\sqrt{2}a_1\frac{\lambda}{\sqrt{n}T}|S^c|\twonorm{\oA(\oA)^\top - \hoA(\hoA)^\top}\\
	&\quad + Ca_1\delta^2 + C\frac{a_1}{T}\delta\sqrt{\frac{p+\log T}{n}}\sum_{t \in S}\zetak{t} + C\frac{\lambda}{\sqrt{n}}\gamma\cdot a_1\delta +  C\frac{\lambda}{\sqrt{n}}\cdot \gamma\cdot \sqrt{p}a_1^2\\
	&< 0, \label{eq: contradiction local 1}
\end{align}
\wprp, if $\twonorm{\hbarA(\hbarA)^\top - \barA(\barA)^\top} > C\bar{\zeta}^{-1}\big(\sqrt{\frac{pr}{nT}} + \sqrt{\frac{r+\log T}{n}}\big) + Ch + C\frac{r\lambda}{\bar{\zeta}^2\sqrt{n}}\frac{|S^c|}{T}$ and $Ca_1\delta^2 + C\frac{a_1}{T}\delta\sqrt{\frac{p+\log T}{n}}\sum_{t \in S}\zetak{t} + C\frac{\lambda}{\sqrt{n}}\gamma\cdot a_1\delta +  C\frac{\lambda}{\sqrt{n}}\cdot \gamma\cdot \sqrt{p}a_1^2 < Ca_1\frac{pr}{nT} + Ca_1\frac{r+\log T}{n}$. Note that the latter can easily hold because we are free to choose an arbitrary small $a_1$ and $\delta$. And as $a_1 \rightarrow 0+$, we have $\bAk{t} \rightarrow \hAk{t}$, $\bA \rightarrow \hoA$, $\bthetak{t} \rightarrow \hthetak{t}$, in the sense of $\maxnorm{\cdot}$. Therefore \eqref{eq: contradiction local 1} contradicts with the local optimality of $(\{\hAk{t}\}_{t =1}^T, \hoA, \{\hthetak{t}\}_{t=1}^T)$ in terms of $G_{[T]}(\{\hAk{t}\}_{t=1}^T, \hoA, \widehat{\bTheta})$. Therefore, \wprp, we must have
\begin{equation}
	\twonorm{\hbarA(\hbarA)^\top - \barA(\barA)^\top} \leq C\bar{\zeta}^{-1}\bigg(r\sqrt{\frac{p}{nT}} + \sqrt{r}\sqrt{\frac{r+\log T}{n}}\bigg) + C\sqrt{r}h + C\frac{r\lambda}{\bar{\zeta}^2\sqrt{n}}\frac{|S^c|}{T},
\end{equation}
when $Ca_1\delta^2 + C\frac{a_1}{T}\delta\sqrt{\frac{p+\log T}{n}}\sum_{t \in S}\zetak{t} + C\frac{\lambda}{\sqrt{n}}\gamma\cdot a_1\delta +  C\frac{\lambda}{\sqrt{n}}\cdot \gamma\cdot \sqrt{p}a_1^2 < Ca_1\frac{pr}{nT} + Ca_1\frac{r+\log T}{n}$.

\underline{\textbf{Case 2:}} If $|\mA_1| \geq cr^{-1}\frac{\bar{\zeta}^2}{\max_{t \in S}(\zetak{t})^2}|S|$:
Note that by \eqref{eq: A 1}, we have
\begin{align}
	&G_{\mA_2}(\{\bAk{t}\}_{t=1}^T,\bA, \bTheta) - G_{\mA_2}(\{\hAk{t}\}_{t=1}^T, \hoA, \widehat{\bTheta}) \\
	&\leq \frac{1}{2T}(\bAk{t}\bthetak{t} - \hAk{t}\hthetak{t})^\top \hSigmak{t}(\bAk{t}\bthetak{t} - \hAk{t}\hthetak{t}) \\
	&\quad + \frac{C}{T}a_1\sqrt{\frac{p+T}{n}}\sqrt{r}\fnorma{\big\{\oA\bthetak{t}_{\oA} - \hoA\bthetak{t}_{\hoA}\big\}_{t \in \mA_2}} + \frac{1}{T}\sum_{t \in \mA_2}\sqrt{\frac{p+\log T}{n}}a_1\cdot \twonorm{\hAk{t}(\hAk{t})^\top - \hoA(\hoA)^\top}\\
	&\quad + \frac{C}{T}\bigg(\sqrt{T}\sqrt{a_1}\sqrt{\frac{r+\log T}{n}} + h\sqrt{\sum_{t \in \mA_2} (\zetak{t})^2 a_1}\bigg)\sqrt{\sum_{t \in \mA_2}a_1\twonorm{\oA\bthetak{t}_{\oA} - \hAk{t}\hthetak{t}}^2} \\
	&\quad  - \frac{C}{T}\sum_{t \in \mA_2}a_1\twonorm{\hAk{t}\hthetak{t} - \oA\bthetak{t}_{\oA}}^2 - \frac{\lambda}{T\sqrt{n}}\cdot \gamma a_1\cdot \sum_{t \in \mA_2}\twonorm{\hAk{t}(\hAk{t})^\top - \hoA(\hoA)^\top} \\
	&\leq - \frac{C}{T}\sum_{t \in \mA_2}a_1\twonorm{\hAk{t}\hthetak{t} - \oA\bthetak{t}_{\oA}}^2 - \frac{\lambda}{T\sqrt{n}}\cdot \gamma a_1\cdot \sum_{t \in \mA_2}\twonorm{\hAk{t}(\hAk{t})^\top - \hoA(\hoA)^\top}\\
	&\quad + \frac{C}{T}\bigg(\sqrt{T}\sqrt{a_1}\sqrt{\frac{r+\log T}{n}} + h\sqrt{\sum_{t \in \mA_2} (\zetak{t})^2 a_1}\bigg)\sqrt{\sum_{t \in \mA_2}a_1\twonorm{\oA\bthetak{t}_{\oA} - \hAk{t}\hthetak{t}}^2} \\
	&\quad + \frac{C}{T}a_1\sqrt{\frac{p+T}{n}}\sqrt{r}\fnorma{\big\{\oA\bthetak{t}_{\oA} - \hAk{t}\hthetak{t}\big\}_{t \in \mA_2}} +  \frac{C}{T}a_1\sqrt{\frac{p+T}{n}}\sqrt{r}\fnorma{\big\{\hAk{t}\hthetak{t} -\hoA\bthetak{t}_{\hoA}\big\}_{t \in \mA_2}} \\
	&\leq Ca_1\bigg(\frac{pr}{nT}+ \frac{r+\log T}{n} + h^2 \barzeta^2\bigg) + \frac{C}{T}a_1\sqrt{\frac{p+T}{n}}\sqrt{r}\sqrt{\sum_{t \in \mA_2}\twonorm{\hAk{t}\hthetak{t} -\hoA\bthetak{t}_{\hoA}}^2} \\
	&\leq Ca_1\bigg(\frac{pr}{nT}+ \frac{r+\log T}{n} + h^2 \barzeta^2\bigg) + Ca_1\sqrt{\frac{p+T}{nT}}\barzeta \cdot \twonorm{\hoA(\hoA)^\top - \oA(\oA)^\top}.
\end{align}
Therefore,
\begin{align}
	&G_{[T]}(\{\bAk{t}\}_{t=1}^T, \bA, \bTheta) - G_{[T]}(\{\hAk{t}\}_{t=1}^T, \hoA, \widehat{\bTheta}) \\
	&\leq [G_{\mA_1}(\{\bAk{t}\}_{t=1}^T, \bA, \bTheta) - G_{\mA_1}(\{\hAk{t}\}_{t=1}^T, \hoA, \widehat{\bTheta})] + [G_{\mA_2}(\{\bAk{t}\}_{t=1}^T, \bA, \bTheta) - G_{\mA_2}(\{\hAk{t}\}_{t=1}^T, \hoA, \widehat{\bTheta})] \\
	&\quad + [G_{S^c}(\{\bAk{t}\}_{t=1}^T, \bA, \bTheta) - G_{S^c}(\{\hAk{t}\}_{t=1}^T, \hoA, \widehat{\bTheta})] \\
	&\leq Ca_1\bigg(\frac{r+\log T}{n} + h^2 \cdot \frac{1}{T}\sum_{t \in \mA_1}(\zetak{t})^2\bigg) + C\frac{a_1}{T}\delta\sqrt{\frac{p+\log T}{n}}\sum_{t \in \mA_1}\zetak{t} + C\frac{\lambda}{\sqrt{n}}\gamma\cdot a_1\delta +  C\frac{\lambda}{\sqrt{n}}\cdot \gamma\cdot \sqrt{p}a_1^2 \\
	&\quad - \frac{\lambda}{T\sqrt{n}}\cdot \gamma\cdot a_1 \cdot \sum_{t \in \mA_1}\twonorm{\hAk{t}(\hAk{t})^\top - \hoA(\hoA)^\top} + Ca_1\bigg(\frac{pr}{nT}+ \frac{r+\log T}{n} + h^2 \barzeta^2\bigg) \\
	&\quad + 2\sqrt{2}a_1\frac{\lambda}{\sqrt{n}T}|S^c|\twonorm{\oA(\oA)^\top - \hoA(\hoA)^\top} + Ca_1\sqrt{\frac{p+T}{nT}}\barzeta \cdot \twonorm{\hoA(\hoA)^\top - \oA(\oA)^\top} \\
	&\leq Ca_1\bigg(\frac{pr}{nT} + \frac{r+\log T}{n} + h^2\bar{\zeta}^2\bigg) + 2\sqrt{2}a_1\frac{\lambda}{\sqrt{n}T}|S^c|\twonorm{\oA(\oA)^\top - \hoA(\hoA)^\top}\\
	&\quad  - \frac{\lambda}{T\sqrt{n}}\cdot \gamma\cdot a_1 \cdot \sum_{t \in \mA_1}\twonorm{\hAk{t}(\hAk{t})^\top - \hoA(\hoA)^\top} + C\frac{a_1}{T}\delta\sqrt{\frac{p+\log T}{n}}\sum_{t \in S}\zetak{t} \\
	&\quad + C\frac{\lambda}{\sqrt{n}}\gamma\cdot a_1\delta +  C\frac{\lambda}{\sqrt{n}}\cdot \gamma\cdot \sqrt{p}a_1^2 + Ca_1\sqrt{\frac{p+T}{nT}}\barzeta \cdot \twonorm{\hoA(\hoA)^\top - \oA(\oA)^\top}\\
	&\leq Ca_1\bigg(\frac{pr}{nT} + \frac{r+\log T}{n} + h^2\bar{\zeta}^2\bigg) + \frac{\lambda}{\sqrt{n}T}a_1\bigg(2\sqrt{2}|S^c| - |\mA_1|\gamma\bigg)\cdot \twonorm{\oA(\oA)^\top - \hoA(\hoA)^\top} \\
	&\quad + C\frac{\lambda}{\sqrt{n}}\gamma\cdot a_1\delta +  C\frac{\lambda}{\sqrt{n}}\cdot \gamma\cdot \sqrt{p}a_1^2 + C\frac{a_1}{T}\delta\sqrt{\frac{p+\log T}{n}}\sum_{t \in S}\zetak{t} + Ca_1\sqrt{\frac{p+T}{nT}}\barzeta \cdot \twonorm{\hoA(\hoA)^\top - \oA(\oA)^\top}\\
	&\leq Ca_1\frac{p+\log T}{n} - C'\frac{\lambda}{\sqrt{n}Tr}\cdot \gamma \frac{\bar{\zeta}^2}{\max_{t \in S}(\zetak{t})^2}\cdot \twonorm{\oA(\oA)^\top - \hoA(\hoA)^\top}+ C\frac{\lambda}{\sqrt{n}}\gamma\cdot a_1\delta +  C\frac{\lambda}{\sqrt{n}}\cdot \gamma\cdot \sqrt{p}a_1^2 \\
	&\quad + C\frac{a_1}{T}\delta\sqrt{\frac{p+\log T}{n}}\sum_{t \in S}\zetak{t}+ Ca_1\sqrt{\frac{p+T}{nT}}\barzeta \cdot \twonorm{\hoA(\hoA)^\top - \oA(\oA)^\top}\\
	&\leq Ca_1\frac{p+\log T}{n} - \frac{C'}{2}\frac{\lambda}{\sqrt{n}Tr}\cdot \gamma \frac{\bar{\zeta}^2}{\max_{t \in S}(\zetak{t})^2}\cdot \twonorm{\oA(\oA)^\top - \hoA(\hoA)^\top}+ C\frac{\lambda}{\sqrt{n}}\gamma\cdot a_1\delta +  C\frac{\lambda}{\sqrt{n}}\cdot \gamma\cdot \sqrt{p}a_1^2 \\
	&\quad + C\frac{a_1}{T}\delta\sqrt{\frac{p+\log T}{n}}\sum_{t \in S}\zetak{t} \label{eq: local number 1}\\
	&< 0,
\end{align}
\wprp, when $\twonorm{\oA(\oA)^\top - \hoA(\hoA)^\top} \geq C\eta$, $C\frac{\lambda}{\sqrt{n}}\gamma\cdot a_1\delta +  C\frac{\lambda}{\sqrt{n}}\cdot \gamma\cdot \sqrt{p}a_1^2 < Ca_1\frac{p+\log T}{n}$, with a large $C'' > 0$, and $\min_{t \in S}\zetak{t}\gtrsim \sqrt{\frac{p+\log T}{n}}$. Note that 
\begin{equation}\label{eq: regime local}
	\frac{\min_{t \in S}\zetak{t}}{\barzeta}\cdot r \lesssim \sqrt{p+\log T}, \quad \frac{\min_{t \in S}\zetak{t}}{\barzeta}\cdot \frac{r}{\sqrt{T}} \lesssim 1.
\end{equation}
We used the fact that 
\begin{align}
		\lambda &\asymp \sqrt{r(p+\log T)}\cdot \frac{\max_{t \in S}(\zetak{t})^2}{\min_{t \in S}\zetak{t}} \\
		&\gtrsim \bigg[\sqrt{pr}\frac{\barzeta}{\min_{t \in S}\zetak{t}} + \sqrt{r(p+\log T)}\cdot \frac{\barzeta}{\min_{t \in S}\zetak{t}}\bigg] \cdot \frac{\max_{t \in S}(\zetak{t})^2}{\barzeta},
	\end{align}
where we used \eqref{eq: regime local} in the second inequality. This contradicts with the \textbf{local} optimality of $(\{\hAk{t}\}_{t =1}^T, \hoA, \{\hthetak{t}\}_{t=1}^T)$ in terms of $G_{[T]}(\{\hAk{t}\}_{t=1}^T, \hoA, \widehat{\bTheta})$. 

Therefore, \wprp, we must have
\begin{equation}
	\twonorm{\hbarA(\hbarA)^\top - \barA(\barA)^\top} \leq C\bar{\zeta}^{-1}\bigg(r\sqrt{\frac{p}{nT}} + \sqrt{r}\sqrt{\frac{r+\log T}{n}}\bigg) + C\sqrt{r}h + C\frac{r\lambda}{\bar{\zeta}^2\sqrt{n}}\frac{|S^c|}{T},
\end{equation}
when $C\frac{a_1}{T}\delta\sqrt{\frac{p+\log T}{n}}\sum_{t \in S}\zetak{t} + C\frac{\lambda}{\sqrt{n}}\gamma\cdot a_1\delta +  C\frac{\lambda}{\sqrt{n}}\cdot \gamma\cdot \sqrt{p}a_1^2 < Ca_1\frac{pr}{nT} + Ca_1\frac{r+\log T}{n}$.

Until now, we have proved that
\begin{equation}
	\twonorm{\hbarA(\hbarA)^\top - \barA(\barA)^\top} \leq C\bar{\zeta}^{-1}\bigg(r\sqrt{\frac{p}{nT}} + \sqrt{r}\sqrt{\frac{r+\log T}{n}}\bigg) + C\sqrt{r}h + C\frac{r\lambda}{\bar{\zeta}^2\sqrt{n}}\frac{|S^c|}{T},
\end{equation}
for any local minimizers $(\{\hAk{t}\}_{t =1}^T, \hoA, \{\hthetak{t}\}_{t=1}^T)$ of $G_{[T]}(\{\hAk{t}\}_{t=1}^T, \hoA, \widehat{\bTheta})$ satisfying
\begin{align}
	&2(2+\sqrt{2})\max_{t \in S}d_2(\hAk{t}, \oA) + 4\frac{\epsilon}{1-\epsilon} < 1-\gamma, \\
	&\max_{t \in S}\bigg\{\frac{\sigmamax(\bSigmak{t})}{\sigmamin(\bSigmak{t})}d_2(\hAk{t}, \bAks{t})\bigg\} < 1-\gamma.
\end{align}
The remaining proof follows the arguments in the proof of Theorem \ref{thm: mtl} and pushing $\delta \rightarrow 0$.

\subsection{Proof of Theorem \ref{thm: implementation}}
We briefly point out how the arguments in the proofs of theorems in Section \ref{subsubsec: alg and upper bounds} can be modified to prove the same results for the revised Algorithm \ref{algo: mtl} which solves Step 1 in the entire $\mathbb{R}^{p \times r}$ space with the penalty on projection matrices. 

We first discuss the results of global minimizers (Theorems \ref{thm: mtl} and \ref{thm: mtl step 1 only}). 
\begin{enumerate}[(i)]
	\item If $\text{rank}(\hAk{t}) = \text{rank}(\hoA) = r$ for all $t \in S$, then we can replace $\hAk{t}\bthetak{t}_{\hAk{t}}$ and $\hoA\bthetak{t}_{\hoA}$ with $\bthetak{t}_{\bA} \coloneqq \argmin_{\btheta \in \mathbb{R}^r}\fk{t}(\bA\btheta)$ used in the proofs by $\mathcal{N}(\hAk{t})\bthetak{t}_{\mathcal{N}(\hAk{t})}$ and $\mathcal{N}(\hoA)\bthetak{t}_{\mathcal{N}(\hoA)}$, where $\mathcal{N}(\hAk{t}) \coloneqq \hAk{t}[(\hAk{t})^\top \hAk{t}]^{-1}(\hAk{t})^\top$, and $\mathcal{N}(\hoA) = \hoA[(\hoA)^\top \hoA]^{-1}(\hoA)^\top$. It is straightforward to see that $\mathcal{N}(\hAk{t})\bthetak{t}_{\mathcal{N}(\hAk{t})} = \hAk{t}\bthetak{t}_{\hAk{t}}$ and $\mathcal{N}(\hoA)\bthetak{t}_{\mathcal{N}(\hoA)} = \hoA\bthetak{t}_{\hoA}$. We can also see that $\bP_{\hAk{t}} = \mathcal{N}(\hAk{t})(\mathcal{N}(\hAk{t}))^\top$ and $\bP_{\hoA} = \mathcal{N}(\hoA)(\mathcal{N}(\hoA))^\top$. We can proceed the same proofs of Theorems \ref{thm: mtl} and \ref{thm: mtl step 1 only}, with $\mathcal{N}(\hAk{t})$, $\mathcal{N}(\hoA)$, $\bthetak{t}_{\hAk{t}}$, and $\bthetak{t}_{\hoA}$.
	\item If $\text{rank}(\hAk{t}) < r$ or $\text{rank}(\hoA) < r$ for some $t \in S$, then for those $t$, we can define $\mathcal{N}(\hAk{t})$ and $\mathcal{N}(\hoA)$ in a different way. Consider a matrix $\bA \in \mathbb{R}^{p\times r}$ with $\text{rank}(\bA) = r' < r$ and its SVD $\bA_{p \times r} = \bm{U}_{p \times r'}\bm{\Lambda}_{r' \times r'}\bm{V}^\top_{r'\times r'}$, where $\bm{\Lambda}_{r' \times r'}$ is diagonal with positive entries and $\bm{U} \in \mathcal{O}^{p\times r'}$. Then we can define $\mathcal{N}(\bA) = (\bm{U}_{p \times r'} \quad \bm{0}_{p \times (r-r')}) \in \mathbb{R}^{p \times r}$. Note that $\mathcal{N}(\bA)(\mathcal{N}(\bA))^\top = \bm{U}\bm{U}^\top = \bP_{\bA} = \bP_{\mathcal{N}(\bA)}$, and this definition of $\mathcal{N}(\bA)$ can be seen as an extension of the full-rank case. Then we can proceed the same proofs of Theorems \ref{thm: mtl} and \ref{thm: mtl step 1 only}, with $\mathcal{N}(\hAk{t})$, $\mathcal{N}(\hoA)$, $\bthetak{t}_{\hAk{t}}$, and $\bthetak{t}_{\hoA}$.
\end{enumerate}
Then let us discuss the results of local minimizers (Theorems \ref{thm: no local easy} and \ref{thm: no local hard}). Recall that we proved  Theorems \ref{thm: no local easy} and \ref{thm: no local hard} by construction based on the local minimizers $\hoA$ and $\hAk{t}$. And it suffices to replace $\hoA$ and $\hAk{t}$ in the construction with $\mathcal{N}(\hoA)$ and $\mathcal{N}\hAk{t})$ as defined above. In addition, if we want to show the stronger version of Theorem \ref{thm: no local hard} we proved in the proof, we need to rewrite the constraint set by replacing 
\begin{align}
	&2(2+\sqrt{2})\max_{t \in S}\min_{\bm{R} \in \mathcal{O}^{r \times r}}\twonorm{\hAk{t} - \oA\bm{R}} + 4\frac{\epsilon}{1-\epsilon} < 1-\gamma, \\
	&\max_{t \in S}\bigg\{\frac{\sigmamax(\bSigmak{t})}{\sigmamin(\bSigmak{t})}\min_{\bm{R} \in \mathcal{O}^{r \times r}} \twonorm{\hAk{t}-\bAks{t}\bm{R}}\bigg\} < 1-\gamma.
\end{align}
with
\begin{align}
	&2(2+\sqrt{2})\max_{t \in S}\min_{\bm{R} \in \mathcal{O}^{r \times r}}\twonorm{\mathcal{N}(\hAk{t}) - \oA\bm{R}} + 4\frac{\epsilon}{1-\epsilon} < 1-\gamma, \\
	&\max_{t \in S}\bigg\{\frac{\sigmamax(\bSigmak{t})}{\sigmamin(\bSigmak{t})}\min_{\bm{R} \in \mathcal{O}^{r \times r}} \twonorm{\mathcal{N}(\hAk{t})-\bAks{t}\bm{R}}\bigg\} < 1-\gamma.
\end{align}
All the remaining arguments still hold.

\subsection{Proof of Theorem \ref{thm: spectral mtl}}
Let us rewrite $\bbetaks{t} = \oA\barthetaks{t} + \bdeltaks{t}$, where $\barthetaks{t} = \oA(\oA)^\top \bAks{t}\bthetaks{t}$ and $\bdeltaks{t} = \oA^\perp(\oA^\perp)^\top \bAks{t}\bthetaks{t}$.  Denote $D_S^* = \{\bdeltaks{t}\}_{t \in S} \in \mathbb{R}^{d \times |S|}$.

First, let us prove a useful result: When $nT \geq Cpr$, $n \geq Cp$, $\frac{\min_{t \in S}\zetak{t}}{\barzeta}\sqrt{\frac{r}{T}}\twonorm{\bm{D}_S^*}\leq c'\sqrt{\frac{p+\log T}{n}}$, and $\min_{t \in S}\twonorm{\bdeltaks{t}} \leq c'\sqrt{\frac{p+\log T}{n}}$, with a large constant $C$ and a small constant $c'$, we have
\begin{equation}\label{eq: useful result spectral}
	\twonorm{\hoA (\hoA)^\top - \oA(\oA)^\top} \lesssim \bar{\zeta}^{-1} \sqrt{\frac{pr}{nT}} + \bar{\zeta}^{-1}\sqrt{\frac{r}{n}} + \bar{\zeta}^{-1}\sqrt{\frac{r}{T}}\twonorm{\bm{D}^*_S} + \frac{\max_{t \in S}\zetak{t}}{\barzeta}\sqrt{r\bar{\epsilon}},
\end{equation}
\wpp, and
\begin{align}
	\twonorm{\hoA\bthetak{t}_{\hoA} - \bbetaks{t}} &\lesssim \frac{\zetak{t}}{\barzeta}\sqrt{\frac{pr}{nT}} + \bigg(\frac{\zetak{t}}{\barzeta} \vee 1\bigg)\sqrt{\frac{r}{n}} + \frac{\zetak{t}}{\barzeta}\sqrt{\frac{r}{T}}\twonorm{\bm{D}^*_S} + \frac{\zetak{t}}{\barzeta}\max_{t \in S}\zetak{t}\sqrt{r\bar{\epsilon}} \\
	&\quad + \twonorm{\bdeltaks{t}} + \sqrt{\frac{\log T}{n}}, \quad \forall t \in S, \label{eq: useful result spectral 2}
\end{align}
\wpr.

Denote $\widebar{\bm{B}}_S = \{\oA\barthetaks{t}\}_{t \in S} \in \mathbb{R}^{p \times |S|}$ and $\widetilde{\bB}_S = \{\widetilde{\bbeta}^{(t)}\}_{t \in S}$. 

Note that when $\frac{\min_{t \in S}\zetak{t}}{\barzeta}\sqrt{\frac{r}{T}}\twonorm{\bm{D}_S^*}\leq c'\sqrt{\frac{p+\log T}{n}}$, and $\min_{t \in S}\twonorm{\bdeltaks{t}} \leq c'\sqrt{\frac{p+\log T}{n}}$, we have
\begin{align}
	\frac{1}{|S|}\sigma_r(\widebar{\bm{B}}_S)^2 &= \frac{1}{|S|}\sigma_r(\sum_{t \in S}\bbetaks{t}\oA(\oA)^\top (\bbetaks{t})^\top) \\
	&= \frac{1}{|S|}\sigma_r(\sum_{t \in S}\bbetaks{t}(\bbetaks{t})^\top) - \frac{1}{|S|}\sigma_r(\sum_{t \in S}\bbetaks{t}\oA^\perp(\oA^\perp)^\top (\bbetaks{t})^\top) \\
	&\geq c\frac{\barzeta^2}{r} - \frac{1}{T(1-\epsilon)}\twonorm{\bm{D}_S^*(\bm{D}_S^*)^\top} \\
	&>  \frac{c}{2}\frac{\barzeta^2}{r},
\end{align}
because $\sqrt{\frac{p+\log T}{n}} \frac{1}{\barzeta} \sqrt{\frac{r}{T}} \twonorm{\bm{D}_S^*} \lesssim  \frac{\min_{t \in S}\zetak{t}}{\barzeta}\sqrt{\frac{r}{T}}\twonorm{\bm{D}_S^*}\leq c'\sqrt{\frac{p+\log T}{n}}$. 

Also, $\widetilde{\bbeta}^{(t)} - \bbetaks{t} = (\hSigmak{t})^{-1}\frac{1}{n}(\bXk{t})^\top \bepsilonk{t}$ is a sub-Gaussian vector with variance proxy $n^{-1}\sigmamax((\hSigmak{t})^{-1})$ given $\bXk{t}$, and $\{\widetilde{\bbeta}^{(t)} - \bbetaks{t}\}_{t \in [T]}$ are independent of each other. By Lemma 5.39 in \cite{vershynin2010introduction},  conditioned on $\{\bXk{t}\}_{t \in S}$, we have
\begin{equation}\label{eq: Delta two norm general}
	\twonorm{\widetilde{\bB}_S - \bm{B}^*_S} \lesssim \sqrt{\frac{p+T}{n}}\max_{t \in S}\sigmamax((\hSigmak{t})^{-1}),
\end{equation}
\wpp. Then combining
\begin{equation}
	\twonorm{(\hSigmak{t})^{-1}} \leq \twonorm{(\bSigmak{t})^{-1}} + \twonorm{(\hSigmak{t})^{-1}}\twonorm{\hSigmak{t} - \bSigmak{t}}\twonorm{(\bSigmak{t})^{-1}}
\end{equation}
with Lemma \ref{lem: cov hat}, we have $\sigmamax((\hSigmak{t})^{-1}) = \twonorm{(\hSigmak{t})^{-1}} \lesssim 1$. Therefore, \wpp,
\begin{equation}\label{eq: regular Delta}
	\twonorm{\widetilde{\bB}_S - \bm{B}^*_S} \lesssim \sqrt{\frac{p+T}{n}}.
\end{equation}
Note that $\widehat{\bm{B}}_S - \widebar{\bm{B}}_S = (\widehat{\bm{B}}_S - \widetilde{\bm{B}}_S) + (\widetilde{\bm{B}}_S - \bm{B}^*_S) + (\bm{B}^*_S - \widebar{\bm{B}}_S)$. By triangle inequality,
\begin{align}
	\twonorm{\widehat{\bm{B}} - \begin{pmatrix}
		\widebar{\bm{B}}_S & \bm{0}
	\end{pmatrix}_{p \times T}} &\leq \twonorm{\widehat{\bm{B}}_{S^c}} +  \twonorm{\widehat{\bm{B}}_S - \widetilde{\bm{B}}_S} + \twonorm{\widetilde{\bm{B}}_S - \bm{B}^*_S} + \twonorm{\bm{B}^*_S - \widebar{\bm{B}}_S} \\
	&\leq \fnorm{\widehat{\bm{B}}_{S^c}} +  \fnorm{\widehat{\bm{B}}_S - \widetilde{\bm{B}}_S} + \twonorm{\widetilde{\bm{B}}_S - \bm{B}^*_S} + \twonorm{\bm{B}^*_S - \widebar{\bm{B}}_S} \\
	&\leq R\sqrt{\epsilon T} + \sqrt{\sum_{t \in S}\twonorm{\Pi_R(\widetilde{\bbeta}^{(t)}) - \widetilde{\bbeta}^{(t)}}^2} + C\sqrt{\frac{p+T}{n}} + \twonorm{\bm{D}_S^*},
\end{align}
\wppp. Since $R = \texttt{quantile}(\{\twonorm{\widetilde{\bbeta}^{(t)}}\}_{t=1}^T, 1-\bar{\epsilon})$, $\bar{\epsilon} \geq \epsilon$, and 
\begin{equation}
	\max_{t \in S}\twonorm{\widetilde{\bbeta}^{(t)}} \leq \max_{t \in S}\twonorm{\widetilde{\bbeta}^{(t)} - \bbetaks{t}} + \max_{t \in S}\twonorm{\bbetaks{t}} \leq C\sqrt{\frac{p+\log T}{n}} + \max_{t \in S}\twonorm{\bbetaks{t}} \leq C\max_{t \in S}\twonorm{\bbetaks{t}},
\end{equation}
\wppp, we have
\begin{equation}
	R \leq C\max_{t \in S}\twonorm{\bbetaks{t}},
\end{equation}
\wppp. And there are at most $\lceil\bar{\epsilon}T\rceil$ number of $t \in S$ satisfying $\prod_R(\widetilde{\bbeta}^{(t)}) \neq \widetilde{\bbeta}^{(t)}$. Therefore, \wppp,
\begin{align}
	\twonorm{\widehat{\bm{B}} - \begin{pmatrix}
		\widebar{\bm{B}}_S & \bm{0}
	\end{pmatrix}_{p \times T}} &\leq R\sqrt{\epsilon T} + C\max_{t \in S}\twonorm{\bbetaks{t}}\sqrt{\bar{\epsilon} T} + C\sqrt{\frac{p+T}{n}} + \twonorm{\bm{D}_S^*} \\
	&\leq 2C\max_{t \in S}\twonorm{\bbetaks{t}}\sqrt{\bar{\epsilon} T}+ C\sqrt{\frac{p+T}{n}} + \twonorm{\bm{D}_S^*}.
\end{align}
Similarly, by Weyl's inequality, \wppp,
\begin{align}
	\sigma_r(\widehat{\bm{B}}) - \sigma_{r+1}(\begin{pmatrix}
		\widebar{\bm{B}}_S & \bm{0}
	\end{pmatrix}) &= \sigma_r(\widehat{\bm{B}}) \\
	&\geq \sigma_r(\widebar{\bB}_S) -  \twonorm{\widehat{\bm{B}}_{S^c}} -  \twonorm{\widehat{\bm{B}}_S - \widetilde{\bm{B}}_S} - \twonorm{\widetilde{\bm{B}}_S - \bm{B}^*_S} - \twonorm{\bm{B}^*_S - \widebar{\bm{B}}_S} \\
	&\geq c\barzeta\sqrt{\frac{T}{r}} - 2C\max_{t \in S}\twonorm{\bbetaks{t}}\sqrt{\bar{\epsilon} T} - C\sqrt{\frac{p+T}{n}} - \twonorm{\bm{D}_S^*} \\
	&\geq \frac{c}{2}\barzeta\sqrt{\frac{T}{r}},
\end{align}
since $\max_{t \in S}\twonorm{\bbetaks{t}}\sqrt{\bar{\epsilon}} \leq \frac{c}{6C}\frac{\barzeta}{\sqrt{r}}$, $\sqrt{\frac{p+T}{n}} \leq \frac{c}{6C}\sqrt{\frac{p+\log T}{n}}\cdot  \barzeta \sqrt{\frac{T}{r}} \leq \barzeta \frac{c}{6C}\sqrt{\frac{T}{r}}$, and $\twonorm{\bm{D}_S^*} \leq \frac{c}{6}\barzeta\sqrt{\frac{T}{r}}$, because $\bar{\epsilon} \leq \frac{c''}{r}\cdot \Big(\frac{\barzeta}{\max_{t \in S}\twonorm{\bbetaks{t}}}\Big)^2$, $nT \geq C''pr$, $p \geq r$, $n \geq C''r$, and $\frac{1}{\barzeta}\sqrt{\frac{r}{T}}\twonorm{\bm{D}_S^*}\leq \frac{1}{\barzeta}\sqrt{\frac{r}{T}}\twonorm{\bm{D}_S^*} \cdot \frac{\min_{t \in S}\zetak{t}}{\sqrt{(p+\log T)/n}} \leq c''$, where $c'' > 0$ is a small constant and $C'' > 0$ is a large constant.

Then by the SVD and Wedin's $\sin \Theta$-Theorem:
\begin{align}
	\twonorm{\hoA (\hoA)^\top - \oA(\oA)^\top} &\lesssim \frac{\twonorm{\widehat{\bm{B}} - \begin{pmatrix}
		\widebar{\bm{B}}_S & \bm{0}
	\end{pmatrix}_{p \times T}}}{\sigma_r(\widehat{\bm{B}}) - \sigma_{r+1}(\begin{pmatrix}
		\widebar{\bm{B}}_S & \bm{0}
	\end{pmatrix})} \\
	&\lesssim \frac{\max_{t \in S}\twonorm{\bbetaks{t}}\sqrt{\bar{\epsilon} T}+ \sqrt{\frac{p+T}{n}} + \twonorm{\bm{D}_S^*}}{\bar{\zeta}\sqrt{\frac{T}{r}}} \\
	&\lesssim \bar{\zeta}^{-1} \sqrt{\frac{pr}{nT}} + \bar{\zeta}^{-1}\sqrt{\frac{r}{n}} + \bar{\zeta}^{-1}\sqrt{\frac{r}{T}}\twonorm{\bm{D}^*_S} + \frac{\max_{t \in S}\twonorm{\bbetaks{t}}}{\barzeta}\sqrt{r\bar{\epsilon}},
\end{align}
\wpp. Hence \eqref{eq: useful result spectral} holds. 

Next, let us prove \eqref{eq: useful result spectral 2}. By noticing that 
$$\bthetak{t}_{\hoA} = ((\hoA)^\top \hSigmak{t}\hoA)^{-1}(\hoA)^\top \hSigmak{t} \bbetaks{t} - ((\hoA)^\top \hSigmak{t}\hoA)^{-1}(\hoA)^\top \nabla \fk{t}(\bbetaks{t}),$$
we have
\begin{align}
	\hoA\bthetak{t}_{\hoA} - \bbetaks{t} &= \Big[\hoA((\hoA)^\top \hSigmak{t}\hoA)^{-1}(\hoA)^\top \hSigmak{t} - \bm{I}\Big]\bbetaks{t} - ((\hoA)^\top \hSigmak{t}\hoA)^{-1}(\hoA)^\top \nabla \fk{t}(\bbetaks{t}) \\
	&= (\hSigmak{t})^{-1/2}\Big[(\hSigmak{t})^{1/2}\hoA((\hoA)^\top \hSigmak{t}\hoA)^{-1}(\hoA)^\top(\hSigmak{t})^{-1/2} - \bm{I}\Big](\hSigmak{t})^{1/2}\bbetaks{t} \\
	&\quad - ((\hoA)^\top \hSigmak{t}\hoA)^{-1}(\hoA)^\top \nabla \fk{t}(\bbetaks{t}).
\end{align}
Denote the projection matrix of a matrix $\bA \in \mathbb{R}^{p_1\times p_2}$ (projection onto the column space of $\bA$) with $\textup{rank}(\bA) = p_2 \leq p_1$ as
\begin{equation}
	\bm{P}_{\bA} = \bA(\bA^\top \bA)^{-1}\bA^\top. 
\end{equation}
Denote $\widehat{\widebar{\bm{B}}} = (\hSigmak{t})^{1/2}\hoA \in \mathbb{R}^{p\times r}$ and $(\widehat{\widebar{\bm{B}}})^{\perp} = (\hSigmak{t})^{-1/2}(\hoA)^{\perp} \in \mathbb{R}^{p \times (p-r)}$. Notice that $(\widehat{\widebar{\bm{B}}})^\top (\widehat{\widebar{\bm{B}}})^{\perp} = \bm{0}_{r \times (p-r)}$. By definition $\bm{P}_{\widehat{\widebar{\bm{B}}}} + \bm{P}_{(\widehat{\widebar{\bm{B}}})^{\perp}} = \bm{I}_p$. Hence
\begin{align}
	\hoA\bthetak{t}_{\hoA} - \bbetaks{t} &= (\hSigmak{t})^{-1/2}(\bm{P}_{\widehat{\widebar{\bm{B}}}} - \bm{I})(\hSigmak{t})^{1/2}(\oA\bthetaks{t} + \bdeltaks{t}) - ((\hoA)^\top \hSigmak{t}\hoA)^{-1}(\hoA)^\top \nabla \fk{t}(\bbetaks{t})\\
	&= -(\hSigmak{t})^{-1/2}\bm{P}_{(\widehat{\widebar{\bm{B}}})^{\perp}}(\hSigmak{t})^{1/2}\oA\bthetaks{t} - (\hSigmak{t})^{-1/2}\bm{P}_{(\widehat{\widebar{\bm{B}}})^{\perp}}(\hSigmak{t})^{1/2}\bdeltaks{t} \\
	&\quad - ((\hoA)^\top \hSigmak{t}\hoA)^{-1}(\hoA)^\top \nabla \fk{t}(\bbetaks{t}). 
\end{align}
Note that since $\twonorm{\hSigmak{t}}$ and $\twonorm{(\hSigmak{t})^{-1}}$ are bounded from above by constants \wppp, we have
\begin{align}
	\twonorm{(\hSigmak{t})^{-1/2}\bm{P}_{(\widehat{\widebar{\bm{B}}})^{\perp}}(\hSigmak{t})^{1/2}\oA\bthetaks{t}} &\lesssim \twonorm{((\hoA)^{\perp})^\top  \oA\bthetaks{t}} + \twonorm{\bdeltaks{t}} \lesssim \twonorm{\hoA(\hoA)^\top - \oA(\oA)^\top}\zetak{t} + \twonorm{\bdeltaks{t}}, \\
	\twonorm{(\hSigmak{t})^{-1/2}\bm{P}_{(\widehat{\widebar{\bm{B}}})^{\perp}}(\hSigmak{t})^{1/2}\bdeltaks{t}} &\lesssim \twonorm{\bdeltaks{t}},
\end{align}
\wppp. Therefore \wppp,
\begin{equation}
	\twonorm{\hoA\bthetak{t}_{\hoA} - \bbetaks{t}} \lesssim \twonorm{\hoA(\hoA)^\top - \oA(\oA)^\top}\zetak{t} + \twonorm{\bdeltaks{t}} + \twonorm{((\hoA)^\top \hSigmak{t}\hoA)^{-1}(\hoA)^\top \nabla \fk{t}(\bbetaks{t})}.
\end{equation}
Let us handle the last term. Denote $\widetilde{\bm{R}} = \argmin_{\bm{R}\in \mathcal{O}^{r \times r}}\twonorm{\hoA - \oA\bm{R}}$. Therefore
\begin{align}
	&\twonorm{((\hoA)^\top \hSigmak{t}\hoA)^{-1}(\hoA)^\top \nabla \fk{t}(\bbetaks{t})} \\
	&\leq \twonorm{((\hoA)^\top \hSigmak{t}\hoA)^{-1}(\hoA  - \oA\widetilde{\bm{R}})^\top \nabla \fk{t}(\bbetaks{t})} + \twonorm{((\hoA)^\top \hSigmak{t}\hoA)^{-1}(\oA\widetilde{\bm{R}})^\top \nabla \fk{t}(\bbetaks{t})} \\
	&\leq \twonorm{\hoA  - \oA\widetilde{\bm{R}}}\twonorm{\nabla \fk{t}(\bbetaks{t})} + \twonorm{(\oA\widetilde{\bm{R}})^\top \nabla \fk{t}(\bbetaks{t})} \\
	&\lesssim \twonorm{\hoA(\hoA)^\top - \oA(\oA)^\top} \sqrt{\frac{p+\log T}{n}} + \sqrt{\frac{r+\log T}{n}},
\end{align}
\wprp. Hence, \wprp,
\begin{align}
	\twonorm{\hoA\bthetak{t}_{\hoA} - \bbetaks{t}} &\lesssim \twonorm{\hoA(\hoA)^\top - \oA(\oA)^\top}\zetak{t} + \twonorm{\bdeltaks{t}} + \sqrt{\frac{r+\log T}{n}} \\
	&\lesssim \bigg(\bar{\zeta}^{-1} \sqrt{\frac{pr}{nT}} + \bar{\zeta}^{-1}\sqrt{\frac{r}{n}} + \bar{\zeta}^{-1}\sqrt{\frac{r}{T}}\twonorm{\bm{D}^*_S} + \frac{\max_{t \in S}\twonorm{\bbetaks{t}}}{\barzeta}\sqrt{r\bar{\epsilon}}\bigg)\zetak{t} \\
	&\quad + \twonorm{\bdeltaks{t}} + \sqrt{\frac{r+\log T}{n}} \\
	&\lesssim \frac{\zetak{t}}{\barzeta}\sqrt{\frac{pr}{nT}} + \bigg(\frac{\zetak{t}}{\barzeta} \vee 1\bigg)\sqrt{\frac{r}{n}} + \frac{\zetak{t}}{\barzeta}\sqrt{\frac{r}{T}}\twonorm{\bm{D}^*_S} + \frac{\zetak{t}}{\barzeta}\max_{t \in S}\zetak{t}\sqrt{r\bar{\epsilon}} \\
	&\quad + \twonorm{\bdeltaks{t}} + \sqrt{\frac{\log T}{n}} \\
	&\coloneqq \eta^{(t)}.
\end{align}
Therefore, \eqref{eq: useful result spectral 2} holds. Now let us use \eqref{eq: useful result spectral} and \eqref{eq: useful result spectral 2} to complete the proof of the theorem.

\noindent (\rom{1}) For $t \in S$ with $\eta^{(t)} \leq c'\sqrt{\frac{p+\log T}{n}}$ with a small $c' > 0$: we must have
\begin{equation}
	\twonorm{\nabla \fk{t}(\bbetaks{t})} + \twonorm{\hoA\bthetak{t}_{\hoA} - \bbetaks{t}} \lesssim \sqrt{\frac{p+\log T}{n}} + \eta^{(t)} \lesssim \frac{\gamma}{\sqrt{n}},
\end{equation}
hence by Lemma \ref{lem: safe net mtl}.(\rom{1}), $\hbetak{t} = \hoA\bthetak{t}_{\hoA}$ and $\twonorm{\hbetak{t} - \bbetaks{t}} \lesssim \eta^{(t)}$ \wprp. Note that we used \eqref{eq: useful result spectral 2} to obtain the bound $\twonorm{\hoA\bthetak{t}_{\hoA} - \bbetaks{t}} \lesssim \eta^{(t)}$, \wprp. We can use it because $\eta^{(t)} \leq c'\sqrt{\frac{p+\log T}{n}}$ implies that $\frac{\min_{t \in S}\zetak{t}}{\barzeta}\sqrt{\frac{r}{T}}\twonorm{\bm{D}^*_S}\leq c'\sqrt{\frac{p+\log T}{n}}$, hence the condition of \eqref{eq: useful result spectral 2} applies and \eqref{eq: useful result spectral 2} holds. Next, let us explain why \eqref{eq: useful result spectral 2} implies the bound in our theorem.

Note that
\begin{align}
	\twonorm{\bdeltaks{t}} &\leq \twonorm{(\oA)^\perp\bAks{t}\bthetaks{t}} \lesssim \twonorm{\bAks{t}(\bAks{t})^\top - \oA(\oA)^\top}\zetak{t} \lesssim h\zetak{t}, \\
	\frac{1}{\sqrt{T}}\twonorm{\bm{D}^*_S} &\leq \frac{1}{\sqrt{T}}\fnorm{\bm{D}^*_S}\cdot \frac{\sigmamax(\bm{D}^*_S)}{\sqrt{r}\sigmamin(\bm{D}^*_S)} \lesssim \barzeta\cdot \frac{\sigmamax(\bm{D}^*_S)}{\sqrt{r}\sigmamin(\bm{D}^*_S)} \cdot h, \label{eq: ineq last}
\end{align}
where $\bm{D}^*_S = (\oA)^\perp\bm{B}^*_S$ and $\fnorm{\bm{D}^*_S} \leq h\sqrt{\sum_{t \in S} \twonorm{\bdeltaks{t}}^2} \leq h\sqrt{\sum_{t \in S} (\zetak{t})^2}$.
Regarding the last inequality \eqref{eq: ineq last}, there is an alternative way to bound the LHS:
\begin{equation}
	\frac{1}{\sqrt{T}}\twonorm{\bm{D}^*_S} \leq \frac{1}{\sqrt{T}}\fnorm{\bm{D}^*_S} \leq \barzeta h.
\end{equation}
Therefore,
\begin{equation}
	\frac{1}{\sqrt{T}}\twonorm{\bm{D}^*_S} \lesssim \barzeta h \cdot \bigg[\frac{\sigmamax(\bm{D}^*_S)}{\sqrt{r}\sigmamin(\bm{D}^*_S)} \wedge 1\bigg].
\end{equation}
Combing all these facts, we obtain the high-probability bound  $\frac{\zetak{t}}{\barzeta}\sqrt{\frac{pr}{nT}} + \bigg(\frac{\zetak{t}}{\barzeta}\vee 1\bigg) \sqrt{\frac{r}{n}} + \sqrt{\frac{\log T}{n}} + \zetak{t}h\cdot \bigg[\frac{\sigma_{\max}((\oA^\perp)^\top \bm{B}^*_S)}{\sigma_{\min}((\oA^\perp)^\top \bm{B}^*_S)} \wedge \sqrt{r}\bigg] + \frac{\zetak{t}}{\barzeta}\max_{t \in S}\zetak{t}\cdot \sqrt{r\bar{\epsilon}}$ for $\twonorm{\hbetak{t}-\bbetaks{t}}$.

\noindent (\rom{2}) For any $t \in S$, by Lemma \ref{lem: safe net mtl}, $\max_{t \in S}\twonorm{\hbetak{t} - \bbetaks{t}} \leq C\frac{\gamma}{\sqrt{n}} + \max_{t \in S}\twonorm{\widetilde{\bbeta}^{(t)} - \bbetaks{t}}$ \wpp, where $\widetilde{\bbeta}^{(t)} \in \argmin_{\bbeta \in \mathbb{R}^p}\fk{t}(\bbeta)$. Since $\gamma \asymp \sqrt{p+\log T}$, this implies that 
\begin{equation}
	\max_{t \in S}\twonorm{\hbetak{t} - \bbetaks{t}} \lesssim \sqrt{\frac{p +\log T}{n}},
\end{equation}
\wppp.
 
Combining (\rom{1}) and (\rom{2}), we get the desired bound:
\begin{align}
	\twonorm{\hbetak{t}-\bbetaks{t}} &\lesssim \Bigg\{\frac{\zetak{t}}{\barzeta}\sqrt{\frac{pr}{nT}} + \bigg(\frac{\zetak{t}}{\barzeta}\vee 1\bigg) \sqrt{\frac{r}{n}} + \sqrt{\frac{\log T}{n}} + \zetak{t}h\cdot \bigg[\frac{\sigma_{\max}((\oA^\perp)^\top \bm{B}^*_S)}{\sigma_{\min}((\oA^\perp)^\top \bm{B}^*_S)} \wedge \sqrt{r}\bigg]\\
	&\quad\quad + \frac{\zetak{t}}{\barzeta}\max_{t \in S}\zetak{t}\sqrt{r\bar{\epsilon}}\Bigg\} \wedge \sqrt{\frac{p + \log T}{n}},
\end{align}
for all $t \in S$, \wprp.

When the tasks in $S^c$ are generated by the linear model, we can get
\begin{equation}
	\max_{t \in S^c}\twonorm{\hbetak{t}-\bbetaks{t}} \lesssim \sqrt{\frac{p+\log T}{n}},
\end{equation}
\wppp, similar to the argument in the proof of Theorem \ref{thm: mtl}.

\subsection{Proof of Theorem \ref{thm: mtl lower bdd}}\label{subsec: proof lower bound mtl supp}
First, when tasks in $S^c$ come from linear models, the term $\sqrt{p/n}$ in the lower bound of $\max_{t \in [T]}\twonorm{\hbetak{t}-\bbetaks{t}}$ is the standard result in linear regression. The other part $\sqrt{\log T/n}$ can be shown using the same arguments as in  \cite{tian2022unsupervised} and \cite{duan2023adaptive}. See the proof of Theorem 4.3 in \cite{duan2023adaptive} and the proof of Theorem 2 in \cite{tian2022unsupervised}. We omit the proof of lower bound for $\max_{t \in [T]}\twonorm{\hbetak{t}-\bbetaks{t}}$ when tasks in $S^c$ come from linear models, and only show the lower bound of $\max_{t \in S}\twonorm{\hbetak{t}-\bbetaks{t}}$ when tasks in $S^c$ can be arbitrarily distributed.

Throughout this subsection, we assume the following generative model for tasks in $S$:
	\begin{equation}
		y|\bx \sim \tP_{y|\bx, \bbeta} = N(\bx^\top\bbeta, 1), \quad \bx \sim \tP_{\bx},
	\end{equation}
	where $\tP_{\bx}$ is sub-Gaussian with $\bSigma = \tE(\bx\bx^\top)$. Suppose there exist constants $c, C$ such that $0<c \leq \lambda_{\min}(\bSigma) \leq \lambda_{\max}(\bSigma) \leq C < \infty$.
	Then any joint distribution $\tP$ of $(\bx, y)$ can be written as $\tP_{\bx, y} = \tP_{y|\bx; \bbeta}\cdot \tP_{\bx}$.

Recall the parameter space for the coefficient vectors $\{\bbetak{t}\}_{t \in S}$ as
\begin{align}
	\mathscr{B}(S, h) =& \Bigg\{\{\bbetak{t}\}_{t\in S}: \bbetak{t} = \bAk{t}\bthetak{t} \text{ for all } t \in S, \{\bAk{t}\}_{t\in S} \subseteq \mO, \twonorm{\bthetak{t}} \leq \zetak{t},\\
	&\min_{\oA \in \mO}\max_{t \in S}\twonorm{\bAk{t}(\bAk{t})^\top - \oA  \oA^\top} \leq h,  \sigma_r\Big(|S|^{-1/2}\bB_S\Big) \geq \frac{c}{\sqrt{r}}\sqrt{\frac{1}{|S|}\sum_{t \in S}\twonorm{\bthetak{t}}^2}\Bigg\}
\end{align}
where $c$ can be any fixed positive constants such that $\mathscr{B}(S, h) \neq \emptyset$.

Given $S \subseteq [T]$, define
\begin{align}
	\mA(S, h) &= \Big\{\{\bAk{t}\}_{t\in S} \subseteq \mO: \min_{\oA \in \mO}\max_{t \in S}\twonorm{\bAk{t}(\bAk{t})^\top - \oA  \oA^\top} \leq h\Big\}, \\
	\Theta(S) &= \Bigg\{\{\bthetak{t}\}_{t\in S} \subseteq \mathbb{R}^r:  \twonorm{\bthetak{t}} \leq \zetak{t}, \frac{1}{|S|}\sum_{t \in S}\bthetak{t}(\bthetak{t})^\top \succeq \frac{c}{r}\frac{1}{|S|}\sum_{t \in S}\twonorm{\bthetak{t}}^2\cdot \bm{I}_r \Bigg\},\\
	\barzeta_S &= \sqrt{\frac{1}{|S|}\sum_{t \in S}(\zetak{t})^2}.
\end{align}

The proof proceeds as follows. We will first show
\begin{align}
	&\inf_{\{\hbetak{t}\}_{t=1}^T}\sup_{S \subseteq \mS}\sup_{\substack{\{\bbetak{t}\}_{t\in S} \in \mathscr{B}(S, h)  \\ \mQ_{S^c}}}\tP \Bigg(\bigcup_{t \in S}\Bigg\{\twonorm{\hbetak{t}-\bbetaks{t}} \gtrsim \bigg[\frac{\zetak{t}}{\barzeta_S}\sqrt{\frac{pr}{nT}} + \bigg(\frac{\zetak{t}}{\barzeta_S} \vee 1\bigg)\sqrt{\frac{r}{n}} + \zetak{t}h\\
	&\quad  + \frac{\zetak{t}}{\barzeta_S}\frac{\epsilon r}{\sqrt{n}}\bigg]\wedge \sqrt{\frac{p}{n}}\Bigg\}\Bigg) \geq \frac{1}{10}, \label{eq: lower bound pre 1}\\
	&\inf_{\{\hbetak{t}\}_{t=1}^T}\sup_{S \subseteq \mS}\sup_{\substack{\{\bbetak{t}\}_{t\in S} \in \mathscr{B}([T], 0)  \\ \mQ_{S^c}}}\tP \Bigg(\max_{t \in S}\twonorm{\hbetak{t}-\bbetaks{t}} \gtrsim \sqrt{\frac{\log T}{n}}\Bigg) \geq \frac{1}{10}. \label{eq: lower bound pre 2}
\end{align} 
Then we will discuss how to get the final lower bound by the same arguments. 

\noindent \underline{\textbf{Part 1:}} First, let us prove \eqref{eq: lower bound pre 1} first. 
\begin{itemize}
	\item If $h \geq \frac{1}{\barzeta_{[T]}}\sqrt{\frac{pr}{nT}} + \frac{1}{\barzeta_{[T]}}\sqrt{\frac{r}{n}} + \frac{1}{\barzeta_{[T]}}\frac{\epsilon r}{\sqrt{n}}$: then $\zetak{t}h  \geq \frac{\zetak{t}}{\barzeta_{[T]}}\sqrt{\frac{pr}{nT}} + \frac{\zetak{t}}{\barzeta_{[T]}}\sqrt{\frac{r}{n}} + \frac{\zetak{t}}{\barzeta_{[T]}}\frac{\epsilon r}{\sqrt{n}}$ for all $t \in [T]$.
		\begin{itemize}
			\item[$\circ$] If $\sqrt{\frac{r}{n}} \leq (\zetak{t_0}h) \wedge \sqrt{\frac{p}{n}}$ for some $t_0 \in [T]$: then we take $S = [T]$ and prove the lower bound
				\begin{equation}
					\inf_{\{\hbetak{t}\}_{t=1}^T}\sup_{\substack{\{\bbetak{t}\}_{t\in [T]} \in \mathscr{B}([T], h)}}\tP \bigg(\twonorm{\hbetak{t_0}-\bbetak{t_0}} \gtrsim (\zetak{t_0}h) \wedge \sqrt{\frac{p}{n}}\bigg) \geq \frac{1}{10}. \label{eq: additional part 1}
				\end{equation}
				WLOG, assume $t_0 = 1$. For all $t \in [T]$, fix $\bAk{t} =$ an arbitrary $\barA \in \mO$, and fix $\{\bthetak{t}\}_{t=2}^T$ s.t. $\twonorm{\bthetak{t}} \leq \zetak{t}$. Let $\bthetak{1} = \frac{1}{\sqrt{r}}\zetak{1}\bm{1}_r$. Set $\delta = \frac{1}{12}\big[\big(\frac{1}{\zetak{1}}\sqrt{\frac{p}{n}}\big) \wedge h\big]$. Consider a $\delta/2$-packing of the ball $\mB_{\delta}(\bm{0}, \mathbb{R}^p, \twonorm{\cdot})$ (denoted as $\mathcal{M}$). By Example 5.8 in \cite{wainwright2019high}, $|\mathcal{M}| \geq 5^p$. Consider 
				\begin{equation}
					\bA = \barA + \frac{c'}{\sqrt{r}}\cdot \bu\bm{1}_r^\top, \quad \widetilde{\bA} = \barA + \frac{c'}{\sqrt{r}}\cdot \widetilde{\bu}\bm{1}_r^\top,
				\end{equation}
				where $\bu \neq \widetilde{\bu} \in \mathcal{M}$ and $c'$ is a small constant, then
				\begin{align}
					\twonorm{(\bA - \widetilde{\bA})\bthetak{1}} &= \frac{1}{r}\zetak{t}\twonorm{(\bu - \widetilde{\bu})\bm{1}_r^\top\bm{1}_r} = \twonorm{\bu - \widetilde{\bu}} \in \left[\frac{c'}{2}\zetak{1}\delta, 2c'\zetak{1}\delta\right], \label{eq: thm 3 eq u2}\\
					\twonorm{(\bA - \widetilde{\bA})\bthetak{t}} &=\twonorm{(\bA - \widetilde{\bA})\bm{e}_{j_k}} = \frac{1}{\sqrt{r}}\zetak{1}\twonorm{\bu - \widetilde{\bu}} \leq \frac{\delta}{\sqrt{r}}\zetak{1}, \label{eq: thm 3 eq u3}
				\end{align}
				where $\bthetak{t} = \bm{e}_{j_k}$. Hence $\mathscr{B} = \{\bbeta \in \mathbb{R}^p: \bbeta = \bA\bthetak{1}, \bA = \barA + \frac{c'}{\sqrt{r}}\cdot \bu\bm{1}_r^\top, \bu \in \mathcal{M}\}$ is a $c'\delta/2$-packing in $\mathbb{R}^p$ with $|\mathscr{B}| = |\mathcal{M}| \geq 5^p$. On the other hand, for any $\bA \in \mathbb{R}^{p \times r}$ with $1+2c'\delta \geq \sigma_{1}(\bA) \geq \sigma_{2}(\bA) \geq \cdots \geq\sigma_{r}(\bA)  \geq 1-2c'\delta > 0$ and $\twonorm{\bA - \barA} \leq \delta$, consider its SVD where $\bA = \bm{Q}\bm{\Lambda}\bm{V}$ where $\bm{Q} \in \mO$ and $\bm{V} \in \mathcal{O}^{r \times r}$. Denote $\bm{R} = \bm{\Lambda}\bm{V}$. Then $0 < 1-2c'\delta \leq \sigma_{\min}(\bm{R}) \leq \sigmamax(\bm{R}) \leq 1+2c'\delta < \infty$, and $\bA\bthetak{1} = \bm{QR}\bthetak{1} = \bm{Q}\widetilde{\btheta}^{(1)}$. Note that $\bm{Q} \in \mO$ and $\widetilde{\btheta}^{(1)} = \bm{R}\bthetak{1}$ with $\twonorm{\widetilde{\btheta}^{(1)}} \leq \twonorm{\bm{R}}\twonorm{\bthetak{1}} \leq C[1+2c']$. And $\twonorm{\bm{Q}\bm{Q}^\top - \barA(\barA)^\top} \leq \twonorm{(\bm{QR})(\bm{QR})^\top - \barA(\barA)^\top} + \twonorm{\bm{Q}(\bm{I}_r - \bm{R}\bm{R}^\top)\bm{Q}^\top} \leq 4\twonorm{\bA - \barA} + 4\twonorm{\bm{I}_r - \bm{R}} \leq 4\delta + 8c'\delta \leq 12\delta\leq h$. Therefore,
				\begin{align}
					&\inf_{\hbetak{1}}\sup_{\substack{\{\bbetak{t}\}_{t\in S} \in \mathscr{B}([T], h)}}\tP \bigg(\twonorm{\hbetak{1}-\bbetak{1}} \gtrsim (\zetak{1}h) \wedge \sqrt{\frac{p}{n}}\bigg) \\
					&\geq \inf_{\hbetak{1}}\sup_{\substack{0< c\leq \sigmamax(\bA) \leq \sigmamax(\bA) \leq C \\ \twonorm{\bthetak{1}} \leq C'}}\tP \bigg(\twonorm{\hbetak{1}-\bA\bthetak{1}} \gtrsim (\zetak{1}h) \wedge \sqrt{\frac{p}{n}}\bigg) \\
					&\coloneqq (*),
				\end{align}
				where all $\bAk{t} =$ an arbitrary $\barA \in \mO$, and $\{\bthetak{t}\}_{t=1}^T \subseteq \{\bm{e}_{j}\}_{j=1}^r$ are fixed. 
				
				For any $\bbeta = \bA\bthetak{1}$ and $\widetilde{\bbeta} = \widetilde{\bA}\bthetak{1} \in \mathscr{B}$ with $\bA \neq \widetilde{\bA}$, by Lemma \ref{lem: KL} and equation \eqref{eq: thm 3 eq u2},
				\begin{align}
					&\KL\left(\prod_{t=2}^T \tP_{y|\bx;\barA\bthetak{t}}^{(t)\otimes n} \cdot \tP_{\bx}^{(t) \otimes n}\cdot \tP_{y|\bx;\bA\bthetak{1}}^{(1)\otimes n} \cdot \tP_{\bx}^{(1) \otimes n} \bigg\| \prod_{t=2}^T \tP_{y|\bx;\barA\bthetak{t}}^{(t)\otimes n} \cdot \tP_{\bx}^{(t) \otimes n} \cdot \tP_{y|\bx;\widetilde{\bA}\bthetak{1}}^{(1)\otimes n} \cdot \tP_{\bx}^{(1) \otimes n}\right) \\
					&= \KL\left(\tP_{y|\bx;\bA\bthetak{1}}^{(1)\otimes n} \cdot \tP_{\bx}^{(1) \otimes n} \Big\|  \tP_{y|\bx;\widetilde{\bA}\bthetak{1}}^{(1)\otimes n} \cdot \tP_{\bx}^{(1) \otimes n}\right)\\
					&\lesssim  n\twonorm{(\bA - \widetilde{\bA})\bthetak{1}}^2\\
					&\lesssim n(\zetak{1})^2\delta^2 \\
					&\leq c\log |\mathscr{B}|.
				\end{align}
				Then by Fano's lemma (Lemma \ref{lem: fano}), $(*) \geq 1-\frac{\log 2}{\log |\mathscr{B}|} - c \geq 1/10$.
			\item[$\circ$] If $\sqrt{\frac{r}{n}} > (\zetak{t}h) \wedge \sqrt{\frac{p}{n}}$ for all $t \in [T]$: then we take $S = [T]$ and prove the lower bound
				\begin{equation}\label{eq: r/n lower bound}
					\inf_{\{\hbetak{t}\}_{t=1}^T}\sup_{\substack{\{\bbetak{t}\}_{t\in [T]} \in \mathscr{B}([T], 0)}}\tP \bigg(\twonorm{\hbetak{t_0}-\bbetak{t_0}} \gtrsim \sqrt{\frac{r}{n}}\bigg) \geq \frac{1}{10},
				\end{equation}
				where $t_0 \in \argmin_{t \in [T]}\zetak{t}$. WLOG, assume $t_0 = 1$. Fix all $\bAk{t} = $ some $\bA \in \mO$. Fixing $\bm{\Theta} = \{\bthetak{t}\}_{t =2}^T$ such that $\frac{1}{T}\sum_{t=2}^T \bthetak{t}(\bthetak{t})^\top \succeq \frac{c}{r}\bm{I}_r$ (Hence for any $\bthetak{1}$ with $\twonorm{\bthetak{1}}\leq C \leq \zetak{1}$, we must have $\frac{1}{T}\sum_{t=1}^T \bthetak{t}(\bthetak{t})^\top \succeq \frac{c}{r}\bm{I}_r$).
				We want to show
				\begin{align}
					\inf_{\{\hbetak{t}\}_{t=1}^T} \sup_{\twonorm{\bthetak{1}}\leq C} \tP\left(\twonorm{\hbetak{1} - \bA\bthetak{1}} \geq c\sqrt{\frac{r}{n}}\right) \geq \frac{1}{10}.
				\end{align}
				Denote $\delta = \sqrt{\frac{r}{n}}$. Consider a $c\delta$-packing of $\mB_{\delta}(\bm{0},\mathbb{R}^r, \twonorm{\cdot})$ (denoted as $\mathscr{T}$). By Example 5.8 in \cite{wainwright2019high}, we know that $\log |\mathscr{T}| \gtrsim r$. Note that this also defines a $c\delta$-packing of the space of $\bbetak{1}$ as $\mathscr{M} = \{\bbetak{1} \in \mathbb{R}^p: \bbetak{1} = \bA\btheta, \btheta \in \mathscr{T}\}$, because for any $\bthetak{1} \neq \widetilde{\btheta}^{(1)} \in \mathscr{T}$ we have $\twonorm{\bA(\bthetak{1} - \widetilde{\btheta}^{(1)})} = \twonorm{\bthetak{1} - \widetilde{\btheta}^{(1)}} \geq c\delta$.
				
				Denote the distribution of $\{\yk{t}_i\}_{i=1}^n$ given $\{\bxk{t}_i\}_{i=1}^n$ and $\{\bbetak{t}\}_{t=1}^T$ as $\tP_{y|\bx;\bbetak{t}}^{(t)\otimes n}$ and the distribution of $\{\bxk{t}_i\}_{i=1}^n$ as $\tP_{\bx}^{(t) \otimes n}$. For any $\bbetak{1} \neq \widetilde{\bbeta}^{(1)} \in \mathscr{M}$:
				\begin{align}
					\KL\left(\tP_{y|\bx;\bbetak{1}}^{(1)\otimes n} \cdot \tP_{\bx}^{(1) \otimes n} \bigg\|  \tP_{y|\bx;\bA\widetilde{\bbeta}^{(1)}}^{(1)\otimes n} \cdot \tP_{\bx}^{(1) \otimes n}\right) &\lesssim n\twonorm{\bbetak{1}-\widetilde{\bbeta}^{(1)}}^2\\
					&\lesssim n\delta^2\\
					&\lesssim r \\
					&\leq c\log |\mathscr{M}|,
				\end{align}
				where $c$ is a small constant. Finally, applying Fano's Lemma (Lemma \ref{lem: fano}), we have
				\begin{align}
					\inf_{\{\hbetak{t}\}_{t=1}^T} \sup_{\twonorm{\bthetak{1}}\leq \delta} \tP\left(\twonorm{\hbetak{1} - \bA\bthetak{1}} \geq c\delta\right) \geq 1-\frac{\log 2}{\log |\mathscr{M}|} - c \geq \frac{1}{10}.
				\end{align}
		\end{itemize}
	\item If $h < \frac{1}{\barzeta_{[T]}}\sqrt{\frac{pr}{nT}} + \frac{1}{\barzeta_{[T]}}\sqrt{\frac{r}{n}} + \frac{1}{\barzeta_{[T]}}\frac{\epsilon r}{\sqrt{n}}$: then $\zetak{t}h < \frac{\zetak{t}}{\barzeta_{[T]}}\sqrt{\frac{pr}{nT}} + \frac{\zetak{t}}{\barzeta_{[T]}}\sqrt{\frac{r}{n}} + \frac{\zetak{t}}{\barzeta_{[T]}}\frac{\epsilon r}{\sqrt{n}}$ for all $t \in [T]$. 
		\begin{itemize}
			\item[$\circ$] If $\sqrt{\frac{r}{n}} \geq \frac{\zetak{t_0}}{\barzeta_{[T]}}\sqrt{\frac{pr}{nT}} + \frac{\zetak{t_0}}{\barzeta_{[T]}}\sqrt{\frac{r}{n}} + \frac{\zetak{t_0}}{\barzeta_{[T]}}\frac{\epsilon r}{\sqrt{n}}$ for $t_0 \in \argmin_{t \in [T]}\zetak{t}$: then we take $S = [T]$ and it suffices to prove the lower bound
				\begin{equation}
					\inf_{\{\hbetak{t}\}_{t=1}^T}\sup_{\substack{\{\bbetak{t}\}_{t\in [T]} \in \mathscr{B}([T], 0)}}\tP \bigg(\twonorm{\hbetak{t_0}-\bbetak{t_0}} \gtrsim \sqrt{\frac{r}{n}}\bigg) \geq \frac{1}{10}.
				\end{equation}
				This has already been proved in the previous analysis.
			\item[$\circ$] If $\sqrt{\frac{r}{n}} < \frac{\zetak{t}}{\barzeta_{[T]}}\sqrt{\frac{pr}{nT}} + \frac{\zetak{t}}{\barzeta_{[T]}}\sqrt{\frac{r}{n}} + \frac{\zetak{t}}{\barzeta_{[T]}}\frac{\epsilon r}{\sqrt{n}}$ for all $t \in [T]$:
			\begin{itemize}
				\item[$\star$] If $\sqrt{\frac{r}{n}} \geq \sqrt{\frac{pr}{nT}} + \epsilon\frac{r}{\sqrt{n}}$: then we take $S = [T]$ and it suffices to prove 
					\begin{equation}
						\inf_{\{\hbetak{t}\}_{t=1}^T}\sup_{\substack{\{\bbetak{t}\}_{t\in [T]} \in \mathscr{B}([T], 0)}}\tP \bigg(\twonorm{\hbetak{t_0}-\bbetak{t_0}} \gtrsim \frac{\zetak{t_0}}{\barzeta}\sqrt{\frac{r}{n}}\bigg) \geq \frac{1}{10},
					\end{equation}
					with $t_0 \in \argmin_{t \in [T]}\zetak{t}$, which is automatically true because $\zetak{t_0}/\barzeta \leq 1$ and \eqref{eq: r/n lower bound}.
				\item[$\star$] If $\sqrt{\frac{r}{n}} < \sqrt{\frac{pr}{nT}} + \epsilon\frac{r}{\sqrt{n}}$:
				\begin{itemize}
					\item[$\rhd$] If $\sqrt{\frac{pr}{nT}} \geq \epsilon\frac{r}{\sqrt{n}}$: then we take $S = [T]$ and prove 
						\begin{equation}
							\inf_{\{\hbetak{t}\}_{t=1}^T}\sup_{\substack{\{\bbetak{t}\}_{t\in [T]} \in \mathscr{B}([T], 0)}}\tP \Bigg(\bigcup_{t \in [T]}\Bigg\{\twonorm{\hbetak{t}-\bbetaks{t}} \gtrsim \frac{\zetak{t}}{\barzeta_{[T]}}\sqrt{\frac{pr}{nT}}\Bigg) \geq \frac{1}{10}.
						\end{equation}
						Note that it suffices to prove 
						\begin{equation}
							\inf_{\{\hbetak{t}\}_{t=1}^T}\sup_{\substack{\{\bbetak{t}\}_{t\in [T]} \in \mathscr{B}([T], 0)}}\tP \Bigg(\frac{1}{T}\sum_{t \in [T]}\twonorm{\hbetak{t}-\bbetaks{t}}^2 \gtrsim \frac{pr}{nT}\Bigg) \geq \frac{1}{10},
						\end{equation}
						because if $\frac{1}{T}\sum_{t \in [T]}\twonorm{\hbetak{t}-\bbetaks{t}}^2 \gtrsim \frac{pr}{nT}$, then there must exist $t \in [T]$ such that $\twonorm{\hbetak{t}-\bbetaks{t}}^2 \gtrsim (\frac{\zetak{t}}{\barzeta_{[T]}})^2\frac{pr}{nT}$. 
						
						Now let us prove it. Consider an $\barA \in \mO$ such that the packing number $M(\mB_{\delta}(\barA, \mO, \disttwo), \distf, \alpha\delta) \geq (\frac{C\sqrt{r}}{\alpha})^{r(p-r)}$ in Lemma \ref{lem: ball packing} with $\delta = \frac{1}{\barzeta_{[T]}}\sqrt{\frac{pr}{nT}}$. Let $\mathcal{M} = \{\bA_j\}_{j=1}^{|\mathcal{M}|}$ be the maximum packing of $\mB_{\delta}(\barA, \mO, \disttwo)$ corresponding to the packing number $M(\mB_{\delta}(\barA, \mO, \disttwo), \allowbreak \distf, c\sqrt{r}\delta)$ with a very small constant $c > 0$. This leads to a $c'\barzeta_{[T]}\delta$-packing $\mathscr{M} =\{\{\bbetak{t}\}_{t=1}^T: \bbetak{t} \coloneqq \bA_j\bR_j\bthetak{t}, \bR_j = \argmin_{\bR \in \mOr}\twonorm{\barA-\bA_j\bR}, \text{ for all }t \in [T] \text{ and the same } j \in [|\mathcal{M}|]\}$ of the space $\mathscr{B}([T], 0)$ w.r.t. distance $\sqrt{\frac{1}{T}\sum_{t \in [T]}\twonorm{\bbetak{t} - \widetilde{\bbeta}^{(t)}}^2}$. To verify this, notice that for any $j_1 \neq j_2 \in [|\mathcal{M}|]$, $\bbetak{t} = \bA_{j_1}\bR_{j_1}\bthetak{t}$, $\widetilde{\bbeta}^{(t)} = \bA_{j_2}\bR_{j_2}\bthetak{t}$, with $\bR_{j_1} = \argmin_{\bR \in \mOr}\twonorm{\barA-\bA_{j_1}\bR}$, $\bR_{j_2} = \argmin_{\bR \in \mOr}\twonorm{\barA-\bA_{j_2}\bR}$, by Wedin's $\sin \Theta$-Theorem and Assumption \ref{asmp: theta},
						\begin{align}
							\frac{1}{T}\sum_{t = 1}^T \twonorm{\bbetak{t} - \widetilde{\bbeta}^{(t)}}^2 \geq \frac{1}{T}\fnorm{\bA_{j_1}\bm{\Theta}_{j_1} - \bA_{j_2}\bm{\Theta}_{j_2}}^2 \geq \frac{c}{r}\fnorm{\bA_{j_1}\bA_{j_1}^\top - \bA_{j_2}\bA_{j_2}^\top}^2 \geq (c')^2\barzeta^2\delta^2,
						\end{align}
						where $c'$ is a small constant, $\bm{\Theta}_{j_1} = \{\bR_{j_1}\bthetak{t}\}_{t \in [T]}$, and $\bm{\Theta}_{j_2} = \{\bR_{j_2}\bthetak{t}\}_{t \in [T]}$. On the other hand, notice that for any $j \in [|\mathcal{M}|]$, $\bbetak{t} = \bA_j\bR_j\bthetak{t}$ with $\bR_{j} = \argmin_{\bR \in \mOr}\twonorm{\barA-\bA_{j}\bR}$, by Lemma \ref{lem: distance equivalence}, for all $t \in [T]$, we have
						\begin{align}
							\twonorm{\bbetak{t} - \barA \bthetak{t}} &\leq \twonorm{\bA_j\bR_j -\barA} \twonorm{\bthetak{t}} \\
							&= \zetak{t}\min_{\bR \in \mOr}\twonorm{\bA_j\bR -\barA} \\
							&\leq \sqrt{2}\zetak{t}\twonorm{\bA_j\bA_j^\top - \barA(\barA)^\top} \\
							&\leq \sqrt{2}\zetak{t}\delta,
						\end{align}
						which by triangle inequality leads to
						\begin{equation}\label{eq: thm 3 eq u1}
							\frac{1}{T}\sum_{t \in [T]}\twonorm{\bbetak{t} - \widetilde{\bbeta}^{(t)}}^2 \leq 4C^2\cdot \frac{1}{T}\sum_{t \in [T]}(\zetak{t})^2\cdot \delta^2 \leq 4C^2\barzeta_{[T]}^2\delta^2,
						\end{equation}
						for any $\{\bbetak{t}\}_{t=1}^T$ and $\{\widetilde{\bbeta}^{(t)}\}_{t=1}^T \in \mathscr{M}$.

						By Lemma \ref{lem: A covering}, $|\mathscr{M}| = |\mathcal{M}| \gtrsim C^{r(p-r)} \geq C^{rp/2}$. Denote the distribution of $\{\yk{t}_i\}_{i=1}^n$ given $\{\bxk{t}_i\}_{i=1}^n$ as $\tP_{y|\bx;\bbeta}^{(t)\otimes n}$ and the distribution of $\{\bxk{t}_i\}_{i=1}^n$ as $\tP_{\bx}^{(t) \otimes n}$. Then for any $\{\bbetak{t}\}_{t=1}^T, \{\widetilde{\bbeta}^{(t)}\}_{t=1}^T \in \mathscr{M}$, by Lemma \ref{lem: KL} and equation \eqref{eq: thm 3 eq u1},
						\begin{align}
							\KL\left(\prod_{t=1}^T \tP_{y|\bx;\bbetak{t}}^{(t)\otimes n} \cdot \tP_{\bx}^{(t) \otimes n} \bigg\| \prod_{t=1}^T \tP_{y|\bx;\widetilde{\bbeta}^{(t)}}^{(t)\otimes n} \cdot \tP_{\bx}^{(t) \otimes n}\right) &\lesssim n\sum_{t=1}^T\twonorm{\bbetak{t}-\wtbbeta^{(t)}}^2\\
							&\lesssim nT\barzeta_{[T]}^2\delta^2 \\
							&\leq c\log |\mathscr{M}|,
						\end{align}
						where $c$ is a small constant. Finally, applying Fano's Lemma (Lemma \ref{lem: fano}), we have
						\begin{align}
							\inf_{\{\hbetak{t}\}_{t=1}^T} \sup_{\{\bbetak{t}\}_{t = 1}^T \in \mathscr{B}([T], 0)} \tP\left(\frac{1}{T}\sum_{t \in [T]}\twonorm{\hbetak{t}-\bbetak{t}}^2 \geq c^2\delta^2\right) &\geq 1-\frac{\log 2}{\log |\mathscr{B}|} - c \\
							&\geq \frac{1}{10}.
						\end{align}
					\item[$\rhd$] If $\sqrt{\frac{pr}{nT}} < \epsilon\frac{r}{\sqrt{n}}$: then we consider $t_0 \in \argmin_{t \in [T]}\zetak{t}$, $S \ni t_0$, and prove 
						\begin{equation}
							\inf_{\{\hbetak{t}\}_{t=1}^T}\sup_{S \subseteq \mathcal{S}}\sup_{\substack{\{\bbetak{t}\}_{t\in [T]} \in \mathscr{B}(S, 0) \\ \mathbb{Q}_{S^c}}}\tP \Bigg(\twonorm{\hbetak{t_0}-\bbetak{t_0}} \gtrsim \frac{\zetak{t_0}}{\barzeta_S}\epsilon\frac{r}{\sqrt{n}}\Bigg) \geq \frac{1}{10}.
						\end{equation}
						 Note that $\frac{\zetak{t_0}}{\barzeta_S} \leq 1$ for any $S \subseteq [T]$. Therefore it suffices to prove
						 \begin{equation}
							\inf_{\{\hbetak{t}\}_{t=1}^T}\sup_{S \subseteq \mathcal{S}}\sup_{\substack{\{\bbetak{t}\}_{t\in [T]} \in \mathscr{B}(S, 0) \\ \mathbb{Q}_{S^c}}}\tP \Bigg(\twonorm{\hbetak{t_0}-\bbetak{t_0}} \gtrsim \epsilon\frac{r}{\sqrt{n}}\Bigg) \geq \frac{1}{10}.
						\end{equation}
						 WLOG, assume $t_0 = 1$. Fix all $\bAk{t} = $ some $\bA \in \mO$ and $\{\bthetak{t}\}_{t=1}^T \subseteq \min_{t \in [T]}\zetak{t}\cdot \{\bm{e}_j\}_{j=1}^r \subseteq \mathbb{R}^r$ s.t. $\sigma_r(T^{-1/2}\{\bthetak{t}\}_{t=1}^T) \geq \frac{c}{\sqrt{r}}$ with $\#\{t: \bthetak{t} = \bm{e}_1\} = \lfloor T/r\rfloor$. Without loss of generality, suppose $\bthetak{t} = \min_{t \in [T]}\zetak{t}\cdot\bm{e}_1$ when $t \in  [\lfloor T/r\rfloor]$. Denote $\wtbbeta = \min_{t \in [T]}\zetak{t}\cdot\bA\bm{e}_1$, and $\bbetak{t} = \bA\bthetak{t}$ for $t \geq \lfloor T/r\rfloor + 1$. Consider two data generating mechanisms in Lemma \ref{lem: binomial lower bound}:
						\begin{enumerate}[(I)]
							\item $\{(\bxk{t}_i, \yk{t}_i)\}_{i=1}^n \sim (1-\epsilon')(\tP_{y|\bx;\wtbbeta}^{\otimes n}\cdot \tP_{\bx}^{\otimes n}) + \epsilon' \mathbb{Q}$ independently for $t \in [\lfloor T/r\rfloor ]$, where $\epsilon' = \frac{T\epsilon}{50\lfloor T/r\rfloor}$, and $\{\{\bxk{t}_i\}_{i=1}^n\}_{t = \lfloor T/r\rfloor + 1}^T \sim \mathbb{D}= \prod_{t= \lfloor T/r\rfloor + 1}^T(\tP_{y|\bx;\bbetak{t}}^{\otimes n}\cdot \tP_{\bx}^{\otimes n})$;
							\item With a preserved set $S^c \subseteq [\lfloor T/r\rfloor ]$, generate $\{\bxk{t}\}_{t \in S^c} \sim \mathbb{Q}_{S^c}$ and $\{(\bxk{t}_i, \yk{t}_i)\}_{i=1}^n \sim \tP_{y|\bx;\wtbbeta}^{\otimes n}\cdot \tP_{\bx}^{\otimes n}$ independently for $t \in S  \cap [\lfloor T/r\rfloor ]$, and $\{\{\bxk{t}_i\}_{i=1}^n\}_{t = \lfloor T/r\rfloor + 1}^T \sim \mathbb{D}= \prod_{t= \lfloor T/r\rfloor + 1}^T(\tP_{y|\bx;\bbetak{t}}^{\otimes n}\cdot \tP_{\bx}^{\otimes n})$.
						\end{enumerate}
						Denote the joint distributions of $\{\bxk{t}_i\}_{t \in [\lfloor T/r\rfloor ]}$ in (\Rom{1}) and (\Rom{2}) as $\tP_{(\epsilon, \theta, \mathbb{Q})}$ and $\tP_{(S, \theta, \mathbb{Q}_{S^c})}$, respectively.
					
						Note that by Lemma \ref{lem: KL},
						\begin{align}
							\varpi(\epsilon', \Theta) &\coloneqq \sup\{\|\bbeta_1-\bbeta_2\|_2: \textup{TV}\big(\tP_{y|\bx;\bbeta_1}^{\otimes n}\cdot \tP_{\bx}^{\otimes n}, \tP_{y|\bx;\bbeta_2}^{\otimes n}\cdot \tP_{\bx}^{\otimes n}\big) \leq \epsilon'/(1-\epsilon')\} \\
							&\geq \sup\{\|\bbeta_1-\bbeta_2\|_2: \textup{KL}\big(\tP_{y|\bx;\bbeta_1}^{\otimes n}\cdot \tP_{\bx}^{\otimes n}, \tP_{y|\bx;\bbeta_2}^{\otimes n}\cdot \tP_{\bx}^{\otimes n}\big) \leq 2[\epsilon'/(1-\epsilon')]^2\} \\
							&\geq \sup\{\|\bbeta_1-\bbeta_2\|_2: n\twonorm{\bbeta_1-\bbeta_2}^2 \leq c[\epsilon'/(1-\epsilon')]^2\} \\
							&= c'\frac{\epsilon'}{\sqrt{n}} \\
							&\asymp \frac{r\epsilon}{\sqrt{n}}.
						\end{align}
						Then by Lemma \ref{lem: from chen},
						\begin{equation}
							\inf_{\hbetak{1}} \sup_{\substack{\wtbbeta, \mQ}} (\tP_{(\epsilon, \theta, \mathbb{Q})}\cdot \mathbb{D})\bigg(\twonorm{\hbetak{1}-\bbetaks{1}} \geq \varpi(\epsilon', \Theta)\bigg) \geq \frac{1}{2}.	
						\end{equation}
						Therefore, by Lemma \ref{lem: binomial lower bound}, it follows that
						\begin{align}
							&\inf_{\{\hbetak{t}\}_{t=1}^T}\sup_{S \subseteq \mS}\sup_{\substack{\{\bbetak{t}\}_{t\in S} \in \mathscr{B}(S, 0)  \\ \mQ_{S^c}}}\tP \bigg(\twonorm{\hbetak{1}-\bbetaks{1}} \geq c\frac{\epsilon r}{\sqrt{n}}\bigg) \\
							&\geq \inf_{\hbetak{1}} \sup_{S: |S|\geq T(1-\epsilon)} \sup_{\substack{\{\bbetak{t}\}_{t\in S} \in \mathscr{B}(S, 0)  \\ \mQ_{S^c}}} (\tP_{(S, \theta, \mathbb{Q}_{S^c})}\cdot \mathbb{D})\bigg(\twonorm{\hbetak{1}-\bbetaks{1}} \geq \varpi(\epsilon/50, \Theta)\bigg) \\
							&\geq \frac{1}{10}.
						\end{align}
				\end{itemize}
			\end{itemize}
		\end{itemize}
\end{itemize}

\noindent \underline{\textbf{Part 2:}} Next, we want to show \eqref{eq: lower bound pre 2}.  Consider the case that $S = [T]$ and $h = 0$. Fix all $\bAk{t} = $ some $\bA \in \mO$. Fix $\{\bthetak{t}\}_{t=1}^T \subseteq \{\sqrt{c}\bm{e}_j\}_{j=1}^r$ satisfying $\frac{1}{T}\sum_{t=1}^T \bthetak{t}(\bthetak{t})^\top \succeq \frac{c}{r}\bm{I}_r$. We want to show
\begin{align}
	\inf_{\{\hbetak{t}\}_{t=1}^T} \sup_{\{\bthetak{t}\}_{t = 1}^T \in \Theta([T])} \tP\left(\max_{t\in[T]}\twonorm{\hbetak{t} - \bA\bthetak{t}} \geq c\sqrt{\frac{\log T}{n}}\right) \geq \frac{1}{10},
\end{align}
where 
\begin{equation}
	\Theta([T]) = \bigg\{\{\bthetak{t}\}_{t\in [T]} \subseteq \mathbb{R}^r:  \max_{t\in [T]}\twonorm{\bthetak{t}} \leq C, \frac{1}{T}\sum_{t =1}^T\bthetak{t}(\bthetak{t})^\top \succeq \frac{c}{r}\bm{I}_r\bigg\}.
\end{equation}

Denote $\delta = \sqrt{\frac{\log T}{n}}$. Then $\mathscr{T} = \{\{\widetilde{\btheta}^{(t)}\}_{t=1}^T: \exists t_0 \text{ s.t. } \widetilde{\btheta}^{(t_0)} = \bthetak{t_0}(1+ \delta/\sqrt{c}), \widetilde{\btheta}^{(t)} = \bthetak{t} \text{ for } t \neq t_0\}$ is a $\sqrt{2}\delta$-packing of $\Theta(S)$ w.r.t. distance $\max_{t \in [T]}\twonorm{\bar{\btheta}^{(t)} - \widetilde{\btheta}^{(t)}}$ with $\{\bar{\btheta}^{(t)}\}_{t=1}^T$ and $\{\widetilde{\btheta}^{(t)}\}_{t=1}^T \in \mathscr{T}$. Hence $\mathscr{B} = \{\{\bA\widetilde{\btheta}^{(t)}\}_{t=1}^T: \exists t_0 \text{ s.t. } \widetilde{\btheta}^{(t_0)} = \bthetak{t_0}(1+ \delta/\sqrt{c}), \widetilde{\btheta}^{(t)} = \bthetak{t} \text{ for } t \neq t_0\}$ is a $\sqrt{2}\delta$-packing in $\mathbb{R}^p$ w.r.t. distance $\max_{t \in [T]}\twonorm{\bbetak{t} - \wtbbetak{t}}$ with $\{\bbetak{t}\}_{t=1}^T$ and $\{\wtbbetak{t}\}_{t=1}^T \in \mathscr{T}$. Apparently $|\mathscr{T}| = |\mathscr{B}| = T$. And for any $\{\bbetak{t}\}_{t=1}^T$ and $\{\wtbbetak{t}\}_{t=1}^T \in \mathscr{T}$, they only differ by two components. WLOG, suppose the indices of different components are $t_1$ and $t_2$. Then we have
\begin{align}
	&\KL\left(\prod_{t=1}^T \tP_{y|\bx;\bbetak{t}}^{(t)\otimes n} \cdot \tP_{\bx}^{(t) \otimes n} \bigg\| \prod_{t=1}^T \tP_{y|\bx;\wtbbetak{t}}^{(t)\otimes n} \cdot \tP_{\bx}^{(t) \otimes n}\right) \\
	&=	\KL\left(\tP_{y|\bx;\bA\bthetak{t_1}(1+ \delta/\sqrt{c})}^{(t_1)\otimes n} \cdot \tP_{\bx}^{(t_1) \otimes n}\cdot \tP_{y|\bx;\bA\bthetak{t_2}}^{(t_2)\otimes n} \cdot \tP_{\bx}^{(t_2) \otimes n} \bigg\|  \tP_{y|\bx;\bA\widetilde{\btheta}^{(t_1)}}^{(t_1)\otimes n} \cdot \tP_{\bx}^{(t_1) \otimes n} \cdot \tP_{y|\bx;\bA\bthetak{t_2}(1+ \delta/\sqrt{c})}^{(t_2)\otimes n} \cdot \tP_{\bx}^{(t_2) \otimes n}\right) \\
	&\lesssim n\twonorm{\bA\bthetak{t_1}(1+ \delta/\sqrt{c})-\bA\bthetak{t_1}}^2 + n\twonorm{\bA\bthetak{t_2}-\bA\bthetak{t_2}(1+ \delta/\sqrt{c})}^2\\
	&\lesssim n\delta^2 \\
	&\leq c'\log |\mathscr{T}|,
\end{align}
where $c$ is a small constant. By Fano's Lemma (Lemma \ref{lem: fano}), we have
\begin{align}
	\inf_{\{\hbetak{t}\}_{t=1}^T} \sup_{\{\bthetak{t}\}_{t = 1}^T \in \Theta([T])} \tP\left(\max_{t\in[T]}\twonorm{\hbetak{t} - \bA\bthetak{t}} \geq c\sqrt{\frac{\log T}{n}}\right) \geq 1-\frac{\log 2}{\log |\mathscr{T}|} - c' \geq \frac{1}{10}.
\end{align}

\noindent \underline{\textbf{Part 3:}} Finally, let us discuss how to obtain the final desired lower bound by similar arguments to prove \eqref{eq: lower bound pre 1} and \eqref{eq: lower bound pre 2}. Denote $S'$ as the index set of $t \in [T]$ where $\zetak{t}$ is among the largest $T(1-\frac{c'}{r})$ ones of $\{\zetak{t}\}_{t=1}^T$. 
\begin{itemize}
	\item If $\sqrt{\frac{\log T}{n}} \leq \min_{t \in S'}\Big\{\Big[\frac{\zetak{t}}{\barzeta_{S'}}\sqrt{\frac{pr}{nT}} + \big(\frac{\zetak{t}}{\barzeta_{S'}}\vee 1\big)\sqrt{\frac{r}{n}} + \zetak{t}h + \frac{\zetak{t}}{\barzeta_{S'}}\epsilon\frac{r}{\sqrt{n}}\Big]\wedge \sqrt{\frac{p}{n}}\Big\}$: then by the same arguments we used to prove \eqref{eq: lower bound pre 1}, we can get a lower bound similar to \eqref{eq: lower bound pre 1} by replacing $[T]$ and $\barzeta_{[T]} = \sqrt{\frac{1}{T}\sum_{t \in [T]}(\zetak{t})^2}$ with $S'$ and $\barzeta_{S'} = \sqrt{\frac{1}{|S'|}\sum_{t \in S'}(\zetak{t})^2}$. Then notice that by definition of $S'$, $\barzeta_{S'} \gtrsim \barzeta_{[T]}$, hence we can get the bound
		\begin{align}
			&\inf_{\{\hbetak{t}\}_{t=1}^T}\sup_{S \subseteq \mS}\sup_{\substack{\{\bbetak{t}\}_{t\in S} \in \mathscr{B}(S, h)  \\ \mQ_{S^c}}}\tP \Bigg(\bigcup_{t \in S'}\Bigg\{\twonorm{\hbetak{t}-\bbetak{t}} \gtrsim \bigg[\frac{\zetak{t}}{\barzeta_{S'}}\sqrt{\frac{pr}{nT}} + \bigg(\frac{\zetak{t}}{\barzeta_{S'}} \vee 1\bigg)\sqrt{\frac{r}{n}} + \zetak{t}h\\
			&\quad  + \frac{\zetak{t}}{\barzeta_{S'}}\frac{\epsilon r}{\sqrt{n}}\bigg]\wedge \sqrt{\frac{p}{n}}\Bigg\}\Bigg) \geq \frac{1}{10},
		\end{align} 
		which implies the desired lower bound because $\sqrt{\frac{\log T}{n}} \leq \min_{t \in S'}\Big\{\Big[\frac{\zetak{t}}{\barzeta}\sqrt{\frac{pr}{nT}} + \big(\frac{\zetak{t}}{\barzeta}\vee 1\big)\sqrt{\frac{r}{n}} + \zetak{t}h + \frac{\zetak{t}}{\barzeta}\epsilon\frac{r}{\sqrt{n}}\Big]\wedge \sqrt{\frac{p}{n}}\Big\}$.
	\item If $\sqrt{\frac{\log T}{n}} > \min_{t \in S'}\Big\{\Big[\frac{\zetak{t}}{\barzeta_{S'}}\sqrt{\frac{pr}{nT}} + \big(\frac{\zetak{t}}{\barzeta_{S'}}\vee 1\big)\sqrt{\frac{r}{n}} + \zetak{t}h + \frac{\zetak{t}}{\barzeta_{S'}}\epsilon\frac{r}{\sqrt{n}}\Big]\wedge \sqrt{\frac{p}{n}}\Big\}$: by definition of $S'$, automatically we have $\sqrt{\frac{\log T}{n}} > \min_{t \in [T]\backslash S'}\Big\{\Big[\frac{\zetak{t}}{\barzeta_{S'}}\sqrt{\frac{pr}{nT}} + \big(\frac{\zetak{t}}{\barzeta_{S'}}\vee 1\big)\sqrt{\frac{r}{n}} + \zetak{t}h + \frac{\zetak{t}}{\barzeta_{S'}}\epsilon\frac{r}{\sqrt{n}}\Big]\wedge \sqrt{\frac{p}{n}}\Big\}$. Then by the same argument we used to prove \eqref{eq: lower bound pre 2}, we can get the following bound by replacing $[T]$ in \eqref{eq: lower bound pre 2} with $[T] \backslash S'$:
		\begin{equation}
			\inf_{\{\hbetak{t}\}_{t=1}^T}\sup_{\substack{\{\bbetak{t}\}_{t\in [T]} \in \mathscr{B}([T], 0)}}\tP \Bigg(\max_{t \in [T]\backslash S'}\twonorm{\hbetak{t}-\bbetak{t}} \gtrsim \sqrt{\frac{\log (T-|S'|)}{n}}\Bigg) \geq \frac{1}{10}.
		\end{equation}
		This implies the desired lower bound, because when $T\geq r^{1.01}$ we have $\log (T-|S'|) \gtrsim \log(T/r) \gtrsim \log T^{\frac{0.01}{1.01}} \gtrsim \log T$.
\end{itemize}

\noindent\textbf{A comment:} In Part 1, when we derive the term $\frac{\zetak{t}}{\barzeta}\epsilon\frac{r}{\sqrt{n}}$ in the lower bound, if $\epsilon r > 1$, the lower bound can be strengthened to $\sqrt{\frac{p}{n}}$. This follows by contaminating all but one of the $\lfloor T/r \rfloor$ tasks such that $\bthetak{t} = \min_{t \in [T]}\zetak{t} \cdot \bm{e}_1$. We can then apply the same reasoning used in Part 1 to derive \eqref{eq: additional part 1}, which leads to the lower bound $\sqrt{\frac{p}{n}}$. This result is quite intuitive: the problem effectively reduces to estimating a $p$-dimensional parameter using only $n$ samples, since the coefficients of the other tasks are orthogonal to the current task. This phenomenon justifies the condition $\epsilon r \lesssim 1$ for the lower and upper bounds of both proposed methods.

\subsection{Proof of Theorem \ref{thm: tl}}
If Algorithm \ref{algo: tl} is coupled with Algorithm \ref{algo: mtl}, the proof idea is very similar to the proof of Theorem \ref{thm: mtl}. When the representation learning helps, an argument based on Lemma \ref{lem: a theta a r tl}, Proposition \ref{prop: mtl}, and Lemma \ref{lem: safe net tl}.(\rom{1}) lead to the corresponding term. In the other case, Lemma \ref{lem: safe net tl}.(\rom{2}) guarantees that the single-task rate holds all the time. A combination of these two situations entails the final TL upper bound. The details are as follows.

Denote $\eta = r\sqrt{\frac{p}{nT}} + \sqrt{r}h + \sqrt{r}\sqrt{\frac{r + \log T}{n}} + \frac{|S^c|}{T}\cdot r \cdot \frac{\lambda}{\sqrt{n}}$.

\noindent (\rom{1}) When $\eta \leq C\sqrt{\frac{p}{n_0}}$: note that $\hbarA$ is \textbf{independent} of $\{\bxk{0}_i, \yk{0}_i\}_{i=1}^{n_0}$, which is the key for part (\rom{1}) to be correct by only requiring $n \geq Cr$ (see Remark \ref{rmt: few shot learning}). Hence by Lemma \ref{lem: a theta a r tl} and an argument by first conditioning on $\hbarA$ then taking the expectation, 
\begin{align}
	\twonorm{\hbarA\bthetak{0}_{\hbarA} - \bAks{0}\bthetaks{0}} &\lesssim \twonorm{\hbarA(\hbarA)^\top - \bAks{0}(\bAks{0})^\top} + \sqrt{\frac{r}{n_0}} \\
	&\leq \twonorm{\hbarA(\hbarA)^\top - \barA(\barA)^\top} + \twonorm{\barA(\barA)^\top - \bAks{0}(\bAks{0})^\top}  + \sqrt{\frac{r}{n_0}} \\
	&\lesssim \eta + \sqrt{\frac{r}{n_0}}
\end{align}
\wpr, where we used Proposision \ref{prop: mtl} in the last step. Then by Lemma \ref{lem: safe net tl}.(\rom{1}), $\frac{\gamma}{\sqrt{n_0}} \gtrsim \sqrt{\frac{p}{n_0}} \geq \twonorm{\nabla \fk{t}(\bAks{0}\bthetaks{0})} + C\twonorm{\hbarA\bthetak{0}_{\hbarA} - \bAks{0}\bthetaks{0}}$, which implies that
\begin{equation}
	\twonorm{\hbetak{0} - \bbetaks{0}} = \twonorm{\hbarA\bthetak{0}_{\hbarA} - \bAks{0}\bthetaks{0}} \leq \eta,
\end{equation}
\wpr.

\noindent (\rom{2}) By Lemma \ref{lem: safe net tl}.(\rom{2}), we always have $\twonorm{\hbetak{0}-\bbetaks{0}} \lesssim \sqrt{p/n_0}$ \wpp.

If Algorithm \ref{algo: tl} is coupled with Algorithm \ref{algo: spectral}, we can follow the discussions above and the proof of Theorem \ref{thm: spectral mtl} to get the desired result.

\subsection{Proof of Theorem \ref{thm: tl lower bdd}}
Similar to the proof of Theorem \ref{thm: mtl lower bdd}, we prove the following parts one by one. Combining them together entails the lower bound. Throughout this subsection, we assume the following generative model for tasks in $\{0\}\cup S$:
	\begin{equation}
		y|\bx \sim \tP_{y|\bx, \bbeta} = N(\bx^\top\bbeta, 1), \quad \bx \sim \tP_{\bx},
	\end{equation}
	where $\tP_{\bx}$ is sub-Gaussian with $\bSigma = \tE(\bx\bx^\top)$. Suppose there exist constants $c, C$ such that $0<c \leq \lambda_{\min}(\bSigma) \leq \lambda_{\max}(\bSigma) \leq C < \infty$.
	Then any joint distribution $\tP$ of $(\bx, y)$ can be written as $\tP_{\bx, y} = \tP_{y|\bx; \bbeta}\cdot \tP_{\bx}$.

Denote
\begin{align}
	\mA_0(S, h) &= \Big\{\{\bAk{t}\}_{t\in \{0\}\cup S} \subseteq \mO: \max_{t \in S}\twonorm{\bAk{t}(\bAk{t})^\top - \bAk{0}(\bAk{0})^\top} \leq h\Big\}, \\
	\Theta_0(S) &= \bigg\{\{\bthetak{t}\}_{t\in \{0\}\cup S} \subseteq \mathbb{R}^r:  \max_{t \in \{0\}\cup S}\twonorm{\bthetak{t}} \leq C, \frac{1}{|S|}\sum_{t \in S}\bthetak{t}(\bthetak{t})^\top \succeq \frac{c}{r}\bm{I}_r\bigg\}.
\end{align}

\noindent(\rom{1}) Consider the case $S = [T]$ and $h = 0$. For any $\bA \in \mathbb{R}^{p \times r}$ with $C \geq \sigma_{1}(\bA) \geq \sigma_{2}(\bA) \geq \cdots \geq\sigma_{r}(\bA)  \geq c > 0$, consider its SVD where $\bA = \bm{Q}\bm{\Lambda}\bm{V}$ where $\bm{Q} \in \mO$ and $\bm{V} \in \mathcal{O}^{r \times r}$. Denote $\bm{R} = \bm{\Lambda}\bm{V}$. Then $0 < c \leq \sigma_{\min}(\bm{R}) \leq \sigmamax(\bm{R}) \leq C < \infty$, and $\bA\bthetak{t} = \bm{QR}\bthetak{t} = \bm{Q}\widetilde{\btheta}^{(t)}$. Note that $\bm{Q} \in \mO$ and $\widetilde{\btheta}^{(t)} = \bm{R}\bthetak{t}$ with $\twonorm{\widetilde{\btheta}^{(t)}} \leq \twonorm{\bm{R}}\twonorm{\bthetak{t}} \leq C'$ and $\frac{1}{T}\sum_{t=1}^T \widetilde{\btheta}^{(t)}(\widetilde{\btheta}^{(t)})^\top = \bm{R}\big[\frac{1}{T}\sum_{t=1}^T \btheta^{(t)}(\btheta^{(t)})^\top\big]\bm{R}^\top \succeq \frac{cc'}{r}\bm{I}_r$ if $\frac{1}{T}\sum_{t=1}^T \btheta^{(t)}(\btheta^{(t)})^\top \succeq \frac{c'}{r}\bm{I}_r$. Therefore, fixing $\bthetak{t} \in \{\bm{e}_{j}\}_{j=1}^r$ and $\bthetak{0} = \frac{1}{\sqrt{r}}\bm{1}_r$ s.t. $\{\bthetak{t}\}_{t=0}^T \in \Theta_0([T])$, we know that
\begin{align}
	&\inf_{\hbetak{0}}\sup_{\substack{\{\bbetak{t}\}_{t=1}^T \in \mathscr{B}_0([T], 0)}}\tP \bigg(\twonorm{\hbetak{0}-\bAk{0}\bthetak{0}} \gtrsim \sqrt{\frac{pr}{nT}} \wedge \sqrt{\frac{p}{n_0}}\bigg) \\
	&\geq \inf_{\hbetak{0}}\sup_{\substack{0< c\leq \sigmamax(\bA) \leq \sigmamax(\bA) \leq C \\ \twonorm{\bthetak{0}} \leq C'}}\tP \bigg(\twonorm{\hbetak{0}-\bA\bthetak{0}} \gtrsim \sqrt{\frac{pr}{nT}} \wedge \sqrt{\frac{p}{n_0}}\bigg) \\
	&\coloneqq (*),
\end{align}
where $\bAk{t} = \bA$ for all $t \in \{0\}\cup [T]$. Let $\delta = \sqrt{\frac{pr}{nT}} \wedge \sqrt{\frac{p}{n_0}}$. Consider a $\delta/2$-packing of the ball $\mB_{\delta}(\bm{0}, \mathbb{R}^p, \twonorm{\cdot})$ (denoted as $\mathcal{M}$). By Example 5.8 in \cite{wainwright2019high}, $|\mathcal{M}| \geq 5^p$. Consider 
\begin{equation}
	\bA = \barA + \frac{c'}{\sqrt{r}}\cdot \bu\bm{1}_r^\top, \quad \widetilde{\bA} = \barA + \frac{c'}{\sqrt{r}}\cdot \widetilde{\bu}\bm{1}_r^\top,
\end{equation}
where $\bu \neq \widetilde{\bu} \in \mathcal{M}$, $c' > 0$ is a small constant such that $\sigmamin(\bA), \sigmamin(\widetilde{\bA}) \geq 1-c' \geq c$ and $\sigmamax(\bA), \sigmamax(\widetilde{\bA}) \leq 1+c' \leq C$. Then
\begin{align}
	\twonorm{(\bA - \widetilde{\bA})\bthetak{0}} &= \frac{1}{r}\twonorm{(\bu - \widetilde{\bu})\bm{1}_r^\top\bm{1}_r} = \twonorm{\bu - \widetilde{\bu}} \in \left[\frac{1}{2}\delta, 2\delta\right], \label{eq: thm 5 eq 1}\\
	\twonorm{(\bA - \widetilde{\bA})\bthetak{t}} &=\twonorm{(\bA - \widetilde{\bA})\bm{e}_{j_k}} = \frac{1}{\sqrt{r}}\twonorm{\bu - \widetilde{\bu}} \leq \frac{\delta}{\sqrt{r}}. \label{eq: thm 5 eq 2}
\end{align}
where $\bthetak{t} = \bm{e}_{j_k}$ with $j_t \in [r]$. Therefore $\mathscr{B} = \{\bbeta \in \mathbb{R}^p: \bbeta = \bA\bthetak{0}, \bA = \barA + \frac{1}{\sqrt{r}}\cdot \bu\bm{1}_r^\top, \bu \in \mathcal{M}\}$ becomes a $\delta/2$-packing in $\mathbb{R}^p$ with $|\mathscr{B}| = |\mathcal{M}| \geq 5^p$. Furthermore, for any $\bbeta = \bA\bthetak{0}$ and $\widetilde{\bbeta} = \widetilde{\bA}\bthetak{0} \in \mathscr{B}$ with $\bA \neq \widetilde{\bA}$, by Lemma \ref{lem: KL} and equations \eqref{eq: thm 5 eq 1} and \eqref{eq: thm 5 eq 2},
\begin{align}
	&\KL\left(\prod_{t=1}^T \tP_{y|\bx;\bA\bthetak{t}}^{(t)\otimes n} \cdot \tP_{\bx}^{(t) \otimes n}\cdot \tP_{y|\bx;\bA\bthetak{0}}^{(0)\otimes n_0} \cdot \tP_{\bx}^{(0) \otimes n_0} \bigg\| \prod_{t=1}^T \tP_{y|\bx;\widetilde{\bA}\bthetak{t}}^{(t)\otimes n} \cdot \tP_{\bx}^{(t) \otimes n} \cdot \tP_{y|\bx;\widetilde{\bA}\bthetak{0}}^{(0)\otimes n_0} \cdot \tP_{\bx}^{(0) \otimes n_0}\right) \\
	&\lesssim n\sum_{t=1}^T\twonorm{(\bA - \widetilde{\bA})\bthetak{t}}^2 + n_0\twonorm{(\bA - \widetilde{\bA})\bthetak{0}}^2\\
	&\lesssim nT\cdot \frac{\delta^2}{r} + n_0\delta^2 \\
	&\leq c\log |\mathscr{B}|.
\end{align}
Then by Fano's lemma (Lemma \ref{lem: fano}), $(*) \geq 1-\frac{\log 2}{\log |\mathscr{B}|} - c \geq 1/10$.

\noindent(\rom{2}) Consider the case $S = [T]$. For all $t \in [T]$, fix $\bAk{t} =$ an arbitrary $\barA \in \mO$. Fix $\{\bthetak{t}\}_{t=1}^T \subseteq \{\bm{e}_{j}\}_{j=1}^r$ and $\bthetak{0} = \frac{1}{\sqrt{r}}\bm{1}_r$ s.t. $\{\bthetak{t}\}_{t=0}^T \in \Theta_0([T])$. Set $\delta = \frac{h}{12} \wedge \sqrt{\frac{p}{n_0}}$. Consider a $\delta/2$-packing of the ball $\mB_{\delta}(\bm{0}, \mathbb{R}^p, \twonorm{\cdot})$ (denoted as $\mathcal{M}$). By Example 5.8 in \cite{wainwright2019high}, $|\mathcal{M}| \geq 5^p$. Consider 
\begin{equation}
	\bA = \barA + \frac{c'}{\sqrt{r}}\cdot \bu\bm{1}_r^\top, \quad \widetilde{\bA} = \barA + \frac{c'}{\sqrt{r}}\cdot \widetilde{\bu}\bm{1}_r^\top,
\end{equation}
where $\bu \neq \widetilde{\bu} \in \mathcal{M}$ and $c'$ is a small constant, then
\begin{equation}\label{eq: thm 5 eq 3}
	\twonorm{(\bA - \widetilde{\bA})\bthetak{0}} = \frac{1}{r}\twonorm{(\bu - \widetilde{\bu})\bm{1}_r^\top\bm{1}_r} = \twonorm{\bu - \widetilde{\bu}} \in \left[\frac{c'}{2}\delta, 2c'\delta\right],
\end{equation}
hence $\mathscr{B} = \{\bbeta \in \mathbb{R}^p: \bbeta = \bA\bthetak{0}, \bA = \barA + \frac{c'}{\sqrt{r}}\cdot \bu\bm{1}_r^\top, \bu \in \mathcal{M}\}$ is a $c'\delta/2$-packing in $\mathbb{R}^p$ with $|\mathscr{B}| = |\mathcal{M}| \geq 5^p$. On the other hand, for any $\bA \in \mathbb{R}^{p \times r}$ with $1+2c'\delta \geq \sigma_{1}(\bA) \geq \sigma_{2}(\bA) \geq \cdots \geq\sigma_{r}(\bA)  \geq 1-2c'\delta > 0$ and $\twonorm{\bA - \barA} \leq \delta$, consider its SVD where $\bA = \bm{Q}\bm{\Lambda}\bm{V}$ where $\bm{Q} \in \mO$ and $\bm{V} \in \mathcal{O}^{r \times r}$. Denote $\bm{R} = \bm{\Lambda}\bm{V}$. Then $0 < 1-2c'\delta \leq \sigma_{\min}(\bm{R}) \leq \sigmamax(\bm{R}) \leq 1+2c'\delta < \infty$, and $\bA\bthetak{0} = \bm{QR}\bthetak{0} = \bm{Q}\widetilde{\btheta}^{(0)}$. Note that $\bm{Q} \in \mO$ and $\widetilde{\btheta}^{(0)} = \bm{R}\bthetak{0}$ with $\twonorm{\widetilde{\btheta}^{(0)}} \leq \twonorm{\bm{R}}\twonorm{\bthetak{0}} \leq C[1+2c']$. And $\twonorm{\bm{Q}\bm{Q}^\top - \barA(\barA)^\top} \leq \twonorm{(\bm{QR})(\bm{QR})^\top - \barA(\barA)^\top} + \twonorm{\bm{Q}(\bm{I}_r - \bm{R}\bm{R}^\top)\bm{Q}^\top} \leq 4\twonorm{\bA - \barA} + 4\twonorm{\bm{I}_r - \bm{R}} \leq 4\delta + 8c'\delta \leq 12\delta\leq h$. Therefore,
\begin{align}
	&\inf_{\hbetak{0}}\sup_{\substack{\{\bAk{t}\}_{t\in \{0\}\cup S} \in \mA_0([T], h) \\ \{\bthetak{t}\}_{t \in \{0\}\cup S} \in \Theta_0([T])}}\tP \bigg(\twonorm{\hbetak{0}-\bAk{0}\bthetak{0}} \gtrsim h \wedge \sqrt{\frac{p}{n_0}}\bigg) \\
	&\geq \inf_{\hbetak{0}}\sup_{\substack{0< c\leq \sigmamax(\bA) \leq \sigmamax(\bA) \leq C \\ \twonorm{\bthetak{0}} \leq C'}}\tP \bigg(\twonorm{\hbetak{0}-\bA\bthetak{0}} \gtrsim h \wedge \sqrt{\frac{p}{n_0}}\bigg) \\
	&\coloneqq (*),
\end{align}
where all $\bAk{t} =$ an arbitrary $\barA \in \mO$, and $\{\bthetak{t}\}_{t=1}^T \subseteq \{\bm{e}_{j}\}_{j=1}^r$ are fixed. 

For any $\bbeta = \bA\bthetak{0}$ and $\widetilde{\bbeta} = \widetilde{\bA}\bthetak{0} \in \mathscr{B}$ with $\bA \neq \widetilde{\bA}$, by Lemma \ref{lem: KL} and equation \eqref{eq: thm 5 eq 3},
\begin{align}
	&\KL\left(\prod_{t=1}^T \tP_{y|\bx;\barA\bthetak{t}}^{(t)\otimes n} \cdot \tP_{\bx}^{(t) \otimes n}\cdot \tP_{y|\bx;\bA\bthetak{0}}^{(0)\otimes n_0} \cdot \tP_{\bx}^{(0) \otimes n_0} \bigg\| \prod_{t=1}^T \tP_{y|\bx;\barA\bthetak{t}}^{(t)\otimes n} \cdot \tP_{\bx}^{(t) \otimes n} \cdot \tP_{y|\bx;\widetilde{\bA}\bthetak{0}}^{(0)\otimes n_0} \cdot \tP_{\bx}^{(0) \otimes n_0}\right) \\
	&= \KL\left(\tP_{y|\bx;\bA\bthetak{0}}^{(0)\otimes n_0} \cdot \tP_{\bx}^{(0) \otimes n_0} \Big\|  \tP_{y|\bx;\widetilde{\bA}\bthetak{0}}^{(0)\otimes n_0} \cdot \tP_{\bx}^{(0) \otimes n_0}\right)\\
	&\lesssim  n_0\twonorm{(\bA - \widetilde{\bA})\bthetak{0}}^2\\
	&\lesssim n_0\delta^2 \\
	&\leq c\log |\mathscr{B}|.
\end{align}
Then by Fano's lemma (Lemma \ref{lem: fano}), $(*) \geq 1-\frac{\log 2}{\log |\mathscr{B}|} - c \geq 1/10$.

\noindent(\rom{3}) Consider the case that $S = [T]$ and $h = 0$. Fix all $\bAk{t} = $ some $\bA \in \mO$. Fixing $\{\bthetak{t}\}_{t =1}^T$ such that $\frac{1}{T}\sum_{t=1}^T \bthetak{t}(\bthetak{t})^\top \succeq \frac{c}{r}\bm{I}_r$. We want to show
\begin{align}
	\inf_{\hbetak{0}} \sup_{\twonorm{\bthetak{0}}\leq C} \tP\left(\twonorm{\hbetak{0} - \bA\bthetak{0}} \geq c\sqrt{\frac{r}{n_0}}\right) \geq \frac{1}{10}.
\end{align}
Denote $\delta = \sqrt{\frac{r}{n_0}}$. Consider a $c\delta$-packing of $\mB_{\delta}(\bm{0},\mathbb{R}^r, \twonorm{\cdot})$ (denoted as $\mathscr{T}$). By Example 5.8 in \cite{wainwright2019high}, we know that $\log |\mathscr{T}| \gtrsim r$. Denote the distribution of $\{\yk{0}_i\}_{i=1}^{n_0}$ given $\{\bxk{0}_i\}_{i=1}^{n_0}$ as $\tP_{y|\bx;\bbeta}^{(0)\otimes n_0}$ and the distribution of $\{\bxk{0}_i\}_{i=1}^{n_0}$ as $\tP_{\bx}^{(0) \otimes n_0}$. For any $\bthetak{0} \neq \widetilde{\btheta}^{(0)} \in \mathscr{T}$:
\begin{align}
	&\KL\left(\prod_{t=1}^T \tP_{y|\bx;\bA\bthetak{t}}^{(t)\otimes n} \cdot \tP_{\bx}^{(t) \otimes n}\cdot \tP_{y|\bx;\bA\bthetak{0}}^{(0)\otimes n_0} \cdot \tP_{\bx}^{(0) \otimes n_0} \bigg\| \prod_{t=1}^T \tP_{y|\bx;\bA\bthetak{t}}^{(t)\otimes n} \cdot \tP_{\bx}^{(t) \otimes n} \cdot \tP_{y|\bx;\bA\widetilde{\btheta}^{(0)}}^{(0)\otimes n_0} \cdot \tP_{\bx}^{(0) \otimes n_0}\right) \\
	&\KL\left(\tP_{y|\bx;\bA\bthetak{0}}^{(0)\otimes n_0} \cdot \tP_{\bx}^{(0) \otimes n_0} \bigg\|  \tP_{y|\bx;\bA\widetilde{\btheta}^{(0)}}^{(0)\otimes n_0} \cdot \tP_{\bx}^{(0) \otimes n_0}\right) \\
	&\lesssim n_0\twonorm{\bA\bthetak{0}-\bA\widetilde{\btheta}^{(0)}}^2\\
	&\lesssim r \\
	&\leq c\log |\mathscr{T}|,
\end{align}
where $c$ is a small constant. Finally, applying Fano's Lemma (Lemma \ref{lem: fano}), we have
\begin{align}
	\inf_{\hbetak{0}} \sup_{\twonorm{\bthetak{0}}\leq \delta} \tP\left(\twonorm{\hbetak{0} - \bA\bthetak{0}} \geq c\delta\right) \geq 1-\frac{\log 2}{\log |\mathscr{T}|} - c \geq \frac{1}{10}.
\end{align}

\noindent(\rom{4}) We follow a similar analysis in part (\rom{5}) of the proof of Theorem \ref{thm: mtl lower bdd}. Consider the case $h = 0$. Fix all $\bAk{t} = $ some $\bA \in \mO$ and $\{\bthetak{t}\}_{t=1}^T \subseteq \{\bm{e}_j\}_{j=1}^r \subseteq \mathbb{R}^r$ s.t. $\sigma_r(T^{-1/2}\{\bthetak{t}\}_{t=1}^T) \geq \frac{c}{\sqrt{r}}$ with $\#\{t: \bthetak{t} = \bm{e}_1\} = \lfloor T/r\rfloor$. Without loss of generality, suppose $\bthetak{t} = \bm{e}_1$ when $t \in \{0\} \cup [\lfloor T/r\rfloor]$. Denote $\wtbbeta = \bA\bm{e}_1$, and $\bbetak{t} = \bA\bthetak{t}$ for $t \geq \lfloor T/r\rfloor + 1$. Consider two data generating mechanisms in Lemma \ref{lem: binomial lower bound}:
		\begin{enumerate}[(I)]
			\item $\{(\bxk{t}_i, \yk{t}_i)\}_{i=1}^n \sim (1-\epsilon')(\tP_{y|\bx;\wtbbeta}^{\otimes n}\cdot \tP_{\bx}^{\otimes n}) + \epsilon' \mathbb{Q}$ independently for $t \in [\lfloor T/r\rfloor ]$, where $\epsilon' = \frac{T\epsilon}{50\lfloor T/r\rfloor}$, $\{(\bxk{0}_i, \yk{0}_i)\}_{i=1}^{n_0}$, and $\{\{\bxk{t}_i\}_{i=1}^n\}_{t = \lfloor T/r\rfloor + 1}^T \sim \mathbb{D} = \prod_{t= \lfloor T/r\rfloor + 1}^T(\tP_{y|\bx;\bbetak{t}}^{\otimes n}\cdot \tP_{\bx}^{\otimes n})$;
			\item With a preserved set $S^c \subseteq [\lfloor T/r\rfloor ]$, generate $\{\bxk{t}\}_{t \in S^c} \sim \mathbb{Q}_{S^c}$ and $\{(\bxk{t}_i, \yk{t}_i)\}_{i=1}^n \sim \tP_{y|\bx;\wtbbeta}^{\otimes n}\cdot \tP_{\bx}^{\otimes n}$ independently for $t \in \{0\}\cup (S  \cap [\lfloor T/r\rfloor ])$, and $\{\{\bxk{t}_i\}_{i=1}^n\}_{t = \lfloor T/r\rfloor + 1}^T \sim \mathbb{D}$$\{\{\bxk{t}_i\}_{i=1}^n\}_{t = \lfloor T/r\rfloor + 1}^T \sim \mathbb{D} = \prod_{t= \lfloor T/r\rfloor + 1}^T(\tP_{y|\bx;\bbetak{t}}^{\otimes n}\cdot \tP_{\bx}^{\otimes n})$.
		\end{enumerate}
	Denote the joint distributions of $\{\bxk{t}_i\}_{t \in \{0\}\cup[\lfloor T/r\rfloor ]}$ in (\Rom{1}) and (\Rom{2}) as $\tP_{(\epsilon, \theta, \mathbb{Q})}$ and $\tP_{(S, \theta, \mathbb{Q}_{S^c})}$, respectively.

Note that by Lemma \ref{lem: KL},
\begin{align}
	&\varpi(\epsilon', \Theta) \\
	&\coloneqq \sup\{\|\bbeta_1-\bbeta_2\|_2: \textup{TV}\big(\tP_{y|\bx;\bbeta_1}^{\otimes n}\cdot \tP_{\bx}^{\otimes n}, \tP_{y|\bx;\bbeta_2}^{\otimes n}\cdot \tP_{\bx}^{\otimes n}\big) \leq \epsilon'/(1-\epsilon'), \\
	&\quad\quad\quad\quad   \textup{TV}\big(\tP_{y|\bx;\bbeta_1}^{\otimes n_0}\cdot \tP_{\bx}^{\otimes n}, \tP_{y|\bx;\bbeta_2}^{\otimes n_0}\cdot \tP_{\bx}^{\otimes n_0}\big) \leq 1/20\} \\
	&\geq \sup\{\|\bbeta_1-\bbeta_2\|_2: \textup{KL}\big(\tP_{y|\bx;\bbeta_1}^{\otimes n}\cdot \tP_{\bx}^{\otimes n}, \tP_{y|\bx;\bbeta_2}^{\otimes n}\cdot \tP_{\bx}^{\otimes n}\big) \leq 2[\epsilon'/(1-\epsilon')]^2, \\
	&\quad\quad\quad\quad \textup{KL}\big(\tP_{y|\bx;\bbeta_1}^{\otimes n_0}\cdot \tP_{\bx}^{\otimes n}, \tP_{y|\bx;\bbeta_2}^{\otimes n_0}\cdot \tP_{\bx}^{\otimes n_0}\big) \leq 1/200\} \\
	&\geq \sup\{\|\bbeta_1-\bbeta_2\|_2: n\twonorm{\bbeta_1-\bbeta_2}^2 \leq c[\epsilon'/(1-\epsilon')]^2, n_0\twonorm{\bbeta_1-\bbeta_2}^2 \leq c\} \\
	&= c'\frac{\epsilon'}{\sqrt{n}}\wedge \frac{1}{\sqrt{n_0}} \\
	&\asymp \frac{r\epsilon}{\sqrt{n}}\wedge \frac{1}{\sqrt{n_0}}.
\end{align}
Then by Lemma \ref{lem: from chen second},
\begin{equation}
	\inf_{\hbetak{0}} \sup_{\substack{\wtbbeta, \mQ}} (\tP_{(\epsilon, \theta, \mathbb{Q})}\cdot \mathbb{D})\bigg(\twonorm{\hbetak{0}-\bbetaks{0}} \geq \varpi(\epsilon', \Theta)\bigg) \geq \frac{1}{2}.	
\end{equation}
Therefore, by Lemma \ref{lem: transfer binomial lower bound}, it follows that
\begin{align}
	&\inf_{\hbetak{0}}\sup_{S \subseteq \mS}\sup_{\substack{\{\bbetak{t}\}_{t\in \{0\}\cup S} \in \mathscr{B}(S, 0)  \\ \mQ_{S^c}}}\tP \bigg(\twonorm{\hbetak{0}-\bbetaks{0}} \geq c\frac{\epsilon r}{\sqrt{n}}\wedge \frac{1}{\sqrt{n_0}}\bigg) \\
	&\geq \inf_{\hbetak{0}} \sup_{S: |S|\geq T(1-\epsilon)} \sup_{\substack{\{\bbetak{t}\}_{t\in \{0\}\cup S} \in \mathscr{B}(S, 0)  \\ \mQ_{S^c}}} (\tP_{(S, \theta, \mathbb{Q}_{S^c})}\cdot \mathbb{D})\bigg(\twonorm{\hbetak{0}-\bbetaks{0}} \geq \varpi(\epsilon/50, \Theta)\bigg) \\
	&\geq \frac{1}{10}.
\end{align}

\subsection{Proofs of Lemmas and Propositions}

Denote $\hSigmak{t} = \sum_{i=1}^n\frac{1}{n}\bxk{t}_i(\bxk{t}_i)^\top$.

\subsubsection{Proof of Lemma \ref{lem: a theta a}}
By optimality of $\bthetak{t}_{\bA}$ and $\bthetak{t}_{\bAks{t}}$,
\begin{equation}\label{eq: lemma 1 def}
	\bA^\top\nabla \fk{t}(\bA \bthetak{t}_{\bA}) = (\bAks{t})^\top\nabla \fk{t}(\bAks{t} \bthetak{t}_{\bAks{t}}) = \bm{0}_r,
\end{equation}
which entails 
\begin{align}
	\bA^\top &\left[\nabla \fk{t}(\bAks{t} \bthetaks{t}) + \hSigmak{t}(\bA \bthetak{t}_{\bA} - \bAks{t} \bthetaks{t})\right] = \bm{0}, \\
	\bthetak{t}_{\bA} &= (\bA^\top \hSigmak{t} \bA)^{-1}\bA^\top \hSigmak{t}\bAks{t}\bthetaks{t} - (\bA^\top \hSigmak{t} \bA)^{-1}\bA^\top\nabla \fk{t}(\bAks{t} \bthetaks{t}).
\end{align}
Therefore,
\begin{align}
	\twonorm{\bA\bthetak{t}_{\bA}-\bAks{t}\bthetaks{t}} &\leq \underbrace{\twonorm{[\bm{I}-\bA(\bA^\top \hSigmak{t} \bA)^{-1}\bA^\top \hSigmak{t}]\bAks{t} \bthetaks{t}}}_{[1]} \\
	&\quad + \underbrace{\twonorm{\bA(\bA^\top \hSigmak{t} \bA)^{-1}\bA^\top \nabla \fk{t}(\bAks{t} \bthetaks{t})}}_{[2]}.
\end{align}
Note that 
\begin{align}
	[1] &\lesssim \twonorm{[\bm{I}-\bA(\bA^\top \hSigmak{t} \bA)^{-1}\bA^\top \hSigmak{t}]\bAks{t}}\twonorm{\bthetaks{t}} \\
	&\lesssim  \twonorm{(\hSigmak{t})^{-1/2}}\cdot \twonorm{(\hSigmak{t})^{1/2}\bAks{t} - (\hSigmak{t})^{1/2}\bA(\bA^\top \hSigmak{t} \bA)^{-1}\bA^\top(\hSigmak{t})^{1/2}(\hSigmak{t})^{1/2}\bAks{t}} \twonorm{\bthetaks{t}} \quad \footnotemark \\
	&\lesssim \twonorm{[\bm{I}-\bm{B}(\bm{B}^\top\bm{B})^{-1}\bm{B}^\top]\bm{B}^{(t)*}}\twonorm{\bthetaks{t}} \\
	&\lesssim \twonorm{P_{\bm{B}^{\perp}}\bm{B}^{(t)*}}\twonorm{\bthetaks{t}}, \label{eq: lemma 1 eq 1}
\end{align}\footnotetext{This step holds because of Lemma \ref{lem: cov hat}, which shows that $\lambdamin(\hSigmak{t}) > 0$ (so that $\hSigmak{t}$ is invertible) w.p. at least $1-e^{-C(p+\log T)}$.}
w.p.  at least $1-e^{-Cp}$, where $\bm{B} = (\hSigmak{t})^{1/2}\bA$, $\bm{B}^{(t)*} = (\hSigmak{t})^{1/2}\bAks{t}$, $\bA^{\perp} \in \{\tilde{\bA} \in \mathcal{O}^{p\times (p-r)}\}: \tilde{\bA}\tilde{\bA}^\top + \bA\bA^\top = \bm{I}_p\}$, $\bm{B}^{\perp} = (\hSigmak{t})^{-1/2}\bA^{\perp}$, $P_{\bm{B}^{\perp}} = \bm{B}^{\perp}((\bm{B}^{\perp})^\top\bm{B}^{\perp})^{-1}(\bm{B}^{\perp})^\top = \bm{I}-P_{\bm{B}}$. Further, notice that
\begin{align}
	\twonorm{P_{\bm{B}^{\perp}}\bm{B}^{(t)*}} &= \twonorm{(\hSigmak{t})^{-1/2}\bA^{\perp}[(\bA^{\perp})^\top (\hSigmak{t})^{-1} \bA^{\perp}]^{-1}(\bA^{\perp})^\top\bAks{t}} \\
	&\leq \twonorm{(\hSigmak{t})^{-1/2}}\cdot \lambdamin^{-1}((\bA^{\perp})^\top (\hSigmak{t})^{-1} \bA^{\perp})\cdot \twonorm{(\bA^{\perp})^\top\bAks{t}} \\
	&\lesssim \twonorm{(\bA^{\perp})^\top\bAks{t}} \\
	&\lesssim  \twonorm{\bA\bA^\top - \bAks{t}(\bAks{t})^\top}, \label{eq: lemma 1 eq 2}
\end{align}
w.p. at least $1-e^{-C(p+\log T)}$. And
\begin{align}
	[2] &\leq \twonorm{\bA(\bA^\top \hSigmak{t} \bA)^{-1}\bA^\top \nabla \fk{t}(\bAks{t} \bthetak{t}_{\bAks{t}})} \\
	&\quad + \twonorm{\bA(\bA^\top \hSigmak{t} \bA)^{-1}\bA^\top [\nabla \fk{t}(\bAks{t} \bthetaks{t})-\nabla \fk{t}(\bAks{t} \bthetak{t}_{\bAks{t}})]} \\
	&\lesssim \twonorm{\bA^\top(\bAks{t})^{\perp}}\cdot \twonorm{\nabla \fk{t}(\bAks{t} \bthetak{t}_{\bAks{t}})} + \twonorm{\bthetaks{t} - \bthetak{t}_{\bAks{t}}} \label{eq: lemma 1 eq 4}\\
	&\lesssim \twonorm{\bA\bA^\top - \bAks{t}(\bAks{t})^\top}(\twonorm{\nabla \fk{t}(\bAks{t} \bthetaks{t})} + C\twonorm{\bthetaks{t} - \bthetak{t}_{\bAks{t}}}) + \twonorm{\bthetaks{t} - \bthetak{t}_{\bAks{t}}} \\
	&\lesssim \twonorm{\bA\bA^\top - \bAks{t}(\bAks{t})^\top}\sqrt{\frac{p+\log T}{n}} + \sqrt{\frac{r+\log T}{n}}, \label{eq: lemma 1 eq 3}
\end{align}
w.p. at least $1-e^{-C(r+\log T)}$, where we used the fact that $\twonorm{\bthetaks{t} - \bthetak{t}_{\bAks{t}}} \lesssim \sqrt{\frac{r+\log T}{n}}$ for all $t \in S$ w.p. at least $1-e^{-C(r+\log T)}$ (which is the standard rate of $r$-dimensional linear regression). Inequality \eqref{eq: lemma 1 eq 4} holds because there exists $\bm{a} \in \mathbb{R}^r$ such that $\nabla f^{(t)}(\bAks{t}\bthetaks{t}) = (\bAks{t})^\perp \bm{a}$ (due to \eqref{eq: lemma 1 def}) and $\twonorm{\bm{a}} = \twonorm{\nabla f^{(t)}(\bAks{t}\bthetaks{t})} \lesssim \sqrt{\frac{p+\log T}{n}}$ w.p. at least $1-e^{C(p+\log T)}$. Putting \eqref{eq: lemma 1 eq 1}, \eqref{eq: lemma 1 eq 2}, and \eqref{eq: lemma 1 eq 3} together, we complete the proof.

\subsubsection{Proof of Lemma \ref{lem: theta bdded}}
By Lemma \ref{lem: a theta a}, w.p. at least $1-e^{-C(r+\log T)}$,
\begin{align}
	\twonorm{\bA\bthetak{t}_{\bA}} -\twonorm{\bAks{t}\bthetaks{t}} &\leq \twonorm{\bA\bthetak{t}_{\bA}-\bAks{t}\bthetaks{t}}\cdot \zetak{t} \\
	&\lesssim \twonorm{\bA\bA^\top - \bAks{t}(\bAks{t})^\top} \cdot \zetak{t}  + \sqrt{\frac{r+\log T}{n}} \\
	&\leq C'\zetak{t}.
\end{align}
Note that $\twonorm{\bA\bthetak{t}_{\bA}} = \twonorm{\bthetak{t}_{\bA}}$ and $\twonorm{\bAks{t}\bthetaks{t}}= \twonorm{\bthetaks{t}} \leq \zetak{t}$.

\subsubsection{Proof of Lemma \ref{lem: a theta a extention}}
The proof is almost the same as the proof of Lemma \ref{lem: a theta a}, thus omitted.

\subsubsection{Proof of Lemma \ref{lem: gradient 2 norm}}
Denote $\bm{G} = \big\{\nabla \fk{t}(\bAks{t}\bthetaks{t})\big\}_{t \in S}$ and $\Delta^{(t)} = \barA\bthetak{t}_{\barA} - \bAks{t}\bthetaks{t}$. By Lemma \ref{lem: a theta a}, w.p. at least $1-e^{-C(r+\log T)}$, for all $t \in S$,
\begin{equation}
	\twonorm{\Delta^{(t)}} \lesssim h\zetak{t} + \sqrt{\frac{r+\log T}{n}},
\end{equation}
leading to
\begin{align}
	\twonorm{\{\bDelta^{(t)}\}_{t \in S}} &= \sup_{\twonorm{\bu}=\twonorm{\bm{v}}=1}\norma{\sum_{t \in S}u_t (\bDelta^{(t)})^\top\bm{v}}  \\
	&\leq \sup_{\twonorm{\bu}=\twonorm{\bm{v}}=1} \twonorm{\bu}\cdot \sqrt{\sum_{t \in S}((\bDelta^{(t)})^\top\bm{v})^2} \\
	&\lesssim \sqrt{T}\left(h\barzeta + \sqrt{\frac{r+\log T}{n}}\right). \label{eq: lemma 5 eq 1}
\end{align}
By Lemma D.3 in \cite{duan2023adaptive} or Theorem 5.39 in \cite{vershynin2010introduction}, w.p. at least $1-e^{-C(p+\log T)}$, 
\begin{equation}
	\twonorm{\bm{G}} \lesssim \sqrt{\frac{p+T}{n}}. \label{eq: lemma 5 eq 2}
\end{equation}
Combining \eqref{eq: lemma 5 eq 1}, \eqref{eq: lemma 5 eq 2}, and the fact that $\twonorm{\widebar{\bm{G}}} \leq \twonorm{\bm{G}} + \twonorm{\{\bDelta^{(t)}\}_{t \in S}}$ implies the desired result.

\subsubsection{Proof of Lemma \ref{lem: matrix 2 norm}}
Note that
\begin{align}
	\twonorm{\widebar{\bm{G}}-\bm{G}} &= \sup_{\twonorm{\bu}=\twonorm{\bm{v}}=1}\norma{\sum_{j =1}^d u_j (\widebar{\bm{g}}_j - \bm{g}_j)^\top\bm{v}}  \\
	&\leq \sup_{\twonorm{\bu}=\twonorm{\bm{v}}=1} \twonorm{\bu}\cdot \sqrt{\sum_{j =1}^d[(\widebar{\bm{g}}_j - \bm{g}_j)^\top\bm{v}]^2} \\
	&\leq \sqrt{d}\Delta,
\end{align}
which completes the proof.

\subsubsection{Proof of Lemma \ref{lem: the same A}}
By convexity of $\fk{t}$,
\begin{align}
	&\fk{t}(\bAk{t}\bthetak{t}_{\bAk{t}}) - \fk{t}(\hbarA\bthetak{t}_{\hbarA}) + \frac{\lambda}{\sqrt{n}}\twonorm{\bAk{t}(\bAk{t})^\top - \hbarA(\hbarA)^\top} \\
	&\geq \frac{\lambda}{\sqrt{n}}\twonorm{\bAk{t}(\bAk{t})^\top - \hbarA(\hbarA)^\top} + \nabla \fk{t}(\hbarA\bthetak{t}_{\hbarA})^\top(\bAk{t}\bthetak{t}_{\bAk{t}}-\hbarA\bthetak{t}_{\hbarA}) \\
	&\geq \frac{\lambda}{\sqrt{n}}\twonorm{\bAk{t}(\bAk{t})^\top - \hbarA(\hbarA)^\top} - \twonorm{\nabla \fk{t}(\hbarA\bthetak{t}_{\hbarA})^\top\bAk{t}\bthetak{t}_{\bAk{t}}} \label{eq: lemma 9 eq 1}\\
	&\geq \frac{\lambda}{\sqrt{n}}\twonorm{\bAk{t}(\bAk{t})^\top - \hbarA(\hbarA)^\top} - \twonorm{\nabla \fk{t}(\hbarA\bthetak{t}_{\hbarA})}\cdot \twonorm{((\hbarA)^{\perp})^\top\bAk{t}}\cdot \twonorm{\bthetak{t}_{\bAk{t}}} \label{eq: lemma 9 eq 2}\\
	&\geq \bigg(\frac{\lambda}{\sqrt{n}} - C \twonorm{\nabla \fk{t}(\hbarA\bthetak{t}_{\hbarA})}\cdot \zetak{t}\bigg)\cdot \twonorm{\bAk{t}(\bAk{t})^\top - \hbarA(\hbarA)^\top} \label{eq: lemma 9 eq 3}\\
	&> 0,
\end{align}
when $\bAk{t}(\bAk{t})^\top \neq \hbarA(\hbarA)^\top$, where \eqref{eq: lemma 9 eq 1} and \eqref{eq: lemma 9 eq 2} are due to the definition of $\bthetak{t}_{\hbarA}$ (which leads to $\nabla \fk{t}(\hbarA\bthetak{t}_{\hbarA})^\top\hbarA = \bm{0}$ and $\nabla \fk{t}(\hbarA\bthetak{t}_{\hbarA}) = (\hbarA)^{\perp}\bu$ with $\twonorm{\bu} = \twonorm{\fk{t}(\hbarA\bthetak{t}_{\hbarA})}$), and \eqref{eq: lemma 9 eq 3} is due to Lemma \ref{lem: distance equivalence}. Note that the above strict inequality leads to a contradiction with the fact that $(\bAk{t}, \bthetak{t})$ is a minimizer, therefore we must have $\bAk{t}(\bAk{t})^\top \neq \hbarA(\hbarA)^\top$ for all $t \in S$.

\subsubsection{Proof of Lemma \ref{lem: small fraction diversity}}
It is straightforward to see that
\begin{align}
	\frac{1}{|S'|}\sum_{t\in S'} \bthetaks{t}(\bthetaks{t})^\top &= \frac{|S|}{|S'|}\left(\frac{1}{|S|}\sum_{t\in S} \bthetaks{t}(\bthetaks{t})^\top - \frac{1}{|S|}\sum_{t\in S \backslash S'} \bthetaks{t}(\bthetaks{t})^\top\right) \\
	&\succeq \frac{|S|}{|S'|}\left(\frac{c}{r}\barzeta^2 - \alpha\max_{t \in S}(\zetak{t})^2\right)\bm{I}_r \\
	&\succeq \frac{c}{2r}\barzeta^2,
\end{align}
because $|S|/|S'| \geq 1$ and $\alpha \leq c/r\cdot \barzeta^2/[2\max_{t \in S}(\zetak{t})^2]$ with a very small $c'' > 0$.

\subsubsection{Proof of Lemma \ref{lem: step 1 control}}
By definition, we have
\begin{equation}
	0 \leq \fk{t}(\bAks{t}\bthetaks{t}) - \fk{t}(\hAk{t}\hthetak{t}) + \frac{\lambda}{\sqrt{n}}\twonorm{\bAks{t}(\bAks{t})^\top - \widehat{\barA} (\widehat{\barA})^\top} - \frac{\lambda}{\sqrt{n}}\twonorm{\hAk{t}(\hAk{t})^\top - \widehat{\barA} (\widehat{\barA})^\top},
\end{equation}
which implies
\begin{align}
	\frac{\lambda}{\sqrt{n}}\twonorm{\bAks{t}(\bAks{t})^\top - \hAk{t}(\hAk{t})^\top} &\geq \fk{t}(\hAk{t}\hthetak{t}) - \fk{t}(\bAks{t}\bthetaks{t})\\
	&\geq \frac{1}{2}(\hAk{t}\hthetak{t}-\bAks{t}\bthetaks{t})^\top\hSigmak{t}(\hAk{t}\hthetak{t}-\bAks{t}\bthetaks{t}) \\
	&\quad - \twonorm{\nabla \fk{t}(\bAks{t}\bthetaks{t})}\cdot \twonorm{\hAk{t}\hthetak{t}-\bAks{t}\bthetaks{t}} \\
	&\geq c\twonorm{\hAk{t}\hthetak{t}-\bAks{t}\bthetaks{t}}^2 - C\sqrt{\frac{p +\log T}{n}}\twonorm{\hAk{t}\hthetak{t}-\bAks{t}\bthetaks{t}}, \label{eq: initial bound r}
\end{align}
w.p. at least $1-e^{-C'(p+\log T)}$. 

Next, we will argue how to upper bound $\twonorm{\bAks{t}(\bAks{t})^\top - \hAk{t}(\hAk{t})^\top}$ by $\twonorm{\hAk{t}\hthetak{t}-\bAks{t}\bthetaks{t}}$ so that solving the inequality can lead to the desired conclusion. Note that upper bounding $\twonorm{\bAks{t}(\bAks{t})^\top - \hAk{t}(\hAk{t})^\top}$ by $\twonorm{\hAk{t}\hthetak{t}-\bAks{t}\bthetaks{t}}$ is not always applicable. Fortunately, for the analysis purpose, we can pick any $\bAks{t}$ whose column space is the same and the corresponding $\bthetaks{t}$ without changing their product $\bbetaks{t}$. The same argument holds for $\hAk{t}$ and $\hthetak{t}$ as well. So WLOG, we denote $\hAk{t} = (\widehat{\bm{a}}_1, \hA_{-1})$, $\bAks{t} = (\bm{a}_1^*, \bm{A}_{-1}^*)$, and consider $\htheta = (\widehat{\theta}_1, \bm{0}_{r-1}^\top)^\top$, $\bthetaks{t} = (\theta_1^*, \bm{0}_{r-1}^\top)^\top$, where $\widehat{\theta}_1\widehat{\bm{a}}_1 = \hbetak{t}$ and $\theta_1^*\bm{a}_1^* = \bbetaks{t}$, no matter which $\hA_{-1}$ and $\bm{A}_{-1}^*$ we pick to encode $\textup{Col}(\hAk{t})$ and $\textup{Col}(\bAks{t})$. 

Consider $\bA^*_{-1} = (\bm{I} - \bm{P}_{\bm{a}_1^*})\hA_{-1}\bm{R}$, where $\bm{R} = (\hA_{-1}^\top(\bm{I} - \bm{P}_{\bm{a}_1^*})\hA_{-1})^{-1/2}$ serving as a normalization matrix which makes $(\bA^*_{-1})^\top \bA^*_{-1} = \bm{I}_r$. Here $\bm{P}_{\bm{a}_1^*} = \bm{a}_1^*(\bm{a}_1^*)^\top/\twonorm{\bm{a}_1^*}^2$ is the projection matrix corresponding to $\bm{a}_1^*$.

Putting everything together, note that
\begin{equation}\label{eq: A bound hat ori}
	\twonorm{\hAk{t}(\hAk{t})^\top - \bAks{t}(\bAks{t})^\top} \leq \twonorm{\bm{a}_1^*(\bm{a}_1^*)^\top - \widehat{\bm{a}}_1(\widehat{\bm{a}}_1)^\top} + \twonorm{\hA_{-1}(\hA_{-1})^\top - \bA_{-1}^*(\bA_{-1}^*)^\top},
\end{equation}
where
\begin{align}
  &\twonorm{\hA_{-1}(\hA_{-1})^\top - \bA_{-1}^*(\bA_{-1}^*)^\top} \\
  &\leq \twonorm{(\bm{I} - \bm{P}_{\widehat{\bm{a}}_1})\hA_{-1}(\hA_{-1})^\top(\bm{I} - \bm{P}_{\widehat{\bm{a}}_1}) - (\bm{I} - \bm{P}_{\bm{a}_1^*})\hA_{-1}\bm{R}\bm{R}^\top(\hA_{-1})^\top(\bm{I} - \bm{P}_{\bm{a}_1^*})} \\
  &\leq \twonorm{\bm{P}_{\bm{a}^*_1}(\bm{I} - \bm{P}_{\widehat{\bm{a}}_1})} + \twonorm{(\bm{I} - \bm{P}_{\widehat{\bm{a}}_1})\bm{P}_{\bm{a}^*_1}}\ + \twonorm{(\bm{I} - \bm{P}_{\bm{a}_1^*})\hA_{-1}(\bm{I}-\bm{R}\bm{R}^\top)(\hA_{-1})^\top(\bm{I} - \bm{P}_{\bm{a}_1^*})} \\
  &\lesssim \twonorm{\bm{a}_1^*(\bm{a}_1^*)^\top - \widehat{\bm{a}}_1(\widehat{\bm{a}}_1)^\top} + \twonorm{\bm{I}-\bm{R}\bm{R}^\top}. 
\end{align}
Recall that $\bm{R}\bm{R}^\top = [\hA_{-1}^\top(\bm{I} - \bm{P}_{\bm{a}_1^*})\hA_{-1}]^{-1} = (\bm{I} - \hA_{-1}^\top\bm{P}_{\bm{a}_1^*}\hA_{-1})^{-1}$. Therefore,
\begin{equation}
	\twonorm{\bm{I}-\bm{R}\bm{R}^\top} \lesssim \twonorm{(\bm{I} - \hA_{-1}^\top\bm{P}_{\bm{a}_1^*}\hA_{-1})^{-1}} \cdot \twonorm{\bm{I} - \hA_{-1}^\top\bm{P}_{\bm{a}_1^*}\hA_{-1} - \bm{I}} \lesssim \twonorm{\bm{a}_1^*(\bm{a}_1^*)^\top - \widehat{\bm{a}}_1(\widehat{\bm{a}}_1)^\top}.
\end{equation}
Plugging these bounds back to \eqref{eq: A bound hat ori}, we have
\begin{equation}
	\twonorm{\hAk{t}(\hAk{t})^\top - \bAks{t}(\bAks{t})^\top} \lesssim \twonorm{\bm{a}_1^*(\bm{a}_1^*)^\top - \widehat{\bm{a}}_1(\widehat{\bm{a}}_1)^\top}.
\end{equation}
Then by Lemma \ref{lem: no step 2 when r = 1}, \wppp, for all $t \in S$,
\begin{equation}
	\zetak{t}\twonorm{\bm{a}_1^*(\bm{a}_1^*)^\top - \widehat{\bm{a}}_1(\widehat{\bm{a}}_1)^\top} \lesssim \twonorm{\widehat{\theta}_1\widehat{\bm{a}}_1 - \theta_1^*\bm{a}_1^*} + \sqrt{\frac{p+\log T}{n}} = \twonorm{\hAk{t}\hthetak{t} - \bAks{t}\bthetaks{t}} + \sqrt{\frac{p+\log T}{n}}.
\end{equation}
Plugging it back to \eqref{eq: initial bound r}, we have
\begin{align}
	\twonorm{\hAk{t}\hthetak{t}-\bAks{t}\bthetaks{t}} \lesssim \sqrt{\frac{p+\log T}{n}} + \frac{\lambda}{\sqrt{n}\zetak{t}} + \Big(\frac{p+\log T}{n}\Big)^{1/4}\bigg(\frac{\lambda}{\zetak{t}\sqrt{n}}\bigg)^{1/2}
\end{align}
for all $t \in S$, \wppp.

\subsubsection{Proof of Lemma \ref{lem: no step 2 when r = 1}}
The proof idea is very similar to the proof of Lemma \ref{lem: a theta a}. By the optimality condition, $\bA^\top\nabla \fk{t}(\bA\bthetak{t}_{\bA}) = 0$, implying that
\begin{align}
	\bA^\top&\left[\nabla \fk{t}(\bAks{t}\bthetaks{t}) + \hSigmak{t}(\bA\bthetak{t}_{\bA} - \bAks{t}\bthetaks{t})\right] = \bm{0},\\
	\bthetak{t}_{\bA} &= (\bA^\top\hSigmak{t}\bA)^{-1}\bA^\top\hSigmak{t}\bAks{t}\bthetaks{t} - (\bA^\top\hSigmak{t}\bA)^{-1}\bA^\top \nabla \fk{t}(\bAks{t}\bthetaks{t}).
\end{align}
Note that $\bthetaks{t} \in \mathbb{R}$ and $\bAks{t} \in \mathbb{R}^p$ when $r = 1$. Hence, \wppp, we have
\begin{align}
	&\twonorm{\bA \bthetak{A}_{\bA} - \bAk{t}\bthetaks{t}} \\
	&\geq \twonorm{[\bm{I} - \bA(\bA^\top\hSigmak{t}\bA)^{-1}\bA^\top\hSigmak{t}]\bAks{t}\bthetaks{t}} - \twonorm{\bA(\bA^\top\hSigmak{t}\bA)^{-1}\bA^\top\nabla \fk{t}(\bAks{t}\bthetaks{t})} \\
	&\geq \twonorm{[\bm{I} - \bA(\bA^\top\hSigmak{t}\bA)^{-1}\bA^\top\hSigmak{t}]\bAks{t}}\zetak{t} -\twonorm{\bA}\cdot \twonorm{\bA(\bA^\top\hSigmak{t}\bA)^{-1}} \cdot \twonorm{\bA}\cdot \twonorm{\nabla \fk{t}(\bAks{t}\bthetaks{t})} \\
	&\geq \zetak{t}\lambdamin((\hSigmak{t})^{-1/2}) \cdot \twonorm{(\hSigmak{t})^{1/2}\bAks{t} - (\hSigmak{t})^{1/2}\bA(\bA^\top\hSigmak{t}\bA)^{-1}\bA^\top\hSigmak{t}\bAks{t}} - C\sqrt{\frac{p+\log T}{n}} \\
	&\geq C'\zetak{t}\twonorm{[\bm{I}-\bm{B}(\bm{B}^\top\bm{B})^{-1})\bm{B}^\top]\bm{B}^{(t)*}} - C\sqrt{\frac{p+\log T}{n}},  \label{eq: lemma 12 eq 1}
\end{align}
where $\bm{B} = (\hSigmak{t})^{1/2}\bA$, $\bm{B}^{(t)*} = (\hSigmak{t})^{1/2}\bAks{t}$, $\bA^{\perp} \in \{\tilde{\bA} \in \mathcal{O}^{p\times (p-1)}\}: \tilde{\bA}\tilde{\bA}^\top + \bA\bA^\top = \bm{I}_p\}$, $\bm{B}^{\perp} = (\hSigmak{t})^{-1/2}\bA^{\perp}$, $P_{\bm{B}^{\perp}} = \bm{B}^{\perp}((\bm{B}^{\perp})^\top\bm{B}^{\perp})^{-1}(\bm{B}^{\perp})^\top = \bm{I}-P_{\bm{B}}$. Further, notice that
\begin{align}
	\twonorm{P_{\bm{B}^{\perp}}\bm{B}^{(t)*}} &= \twonorm{(\hSigmak{t})^{-1/2}\bA^{\perp}[(\bA^{\perp})^\top (\hSigmak{t})^{-1} \bA^{\perp}]^{-1}(\bA^{\perp})^\top\bAks{t}} \\
	&\geq \lambdamin^{1/2}([(\bA^{\perp})^\top (\hSigmak{t})^{-1} \bA^{\perp}]^{-1})\cdot \twonorm{(\bA^{\perp})^\top\bAks{t}} \\
	&= \twonorm{(\bA^{\perp})^\top (\hSigmak{t})^{-1} \bA^{\perp}}^{-1/2}\cdot \twonorm{(\bA^{\perp})^\top\bAks{t}} \\
	&\gtrsim \twonorm{\bA\bA^\top - \bAks{t}(\bAks{t})^\top},  \label{eq: lemma 12 eq 2}
\end{align}
w.p. at least $1-e^{-C(p+\log T)}$. Combining \eqref{eq: lemma 12 eq 1} and \eqref{eq: lemma 12 eq 2} completes the proof.

\subsubsection{Proof of Lemma \ref{lem: safe net mtl}}
First, by Lemma \ref{lem: cov hat}, $0 < c \leq \min_{t \in S}\lambdamin(\hSigmak{t}) \leq \max_{t \in S}\lambdamax(\hSigmak{t}) \leq C < \infty$ \wppp. Then (\rom{1})  is the direct consequence of Lemmas E.3 in \cite{duan2023adaptive}. (\rom{2}) and (\rom{3}) come from Lemma E.2 in \cite{duan2023adaptive}.

\subsubsection{Proof of Lemma \ref{lem: a theta a r tl}}
We modify a few steps in the proof of Lemma \ref{lem: a theta a}. The key point here is that $\bA \in \mO$ is fixed.

By optimality of $\bthetak{t}_{\bA}$ and $\bthetak{t}_{\bAks{t}}$,
\begin{equation}
	\bA^\top\nabla \fk{t}(\bA \bthetak{t}_{\bA}) = (\bAks{t})^\top\nabla \fk{t}(\bAks{t} \bthetak{t}_{\bAks{t}}) = \bm{0}_r,
\end{equation}
which entails 
\begin{align}
	\bA^\top &\left[\nabla \fk{t}(\bAks{t} \bthetaks{t}) + \hSigmak{t}(\bA \bthetak{t}_{\bA} - \bAks{t} \bthetaks{t})\right] = \bm{0}, \\
	\bthetak{t}_{\bA} &= (\bA^\top \hSigmak{t} \bA)^{-1}\bA^\top \hSigmak{t}\bAks{t}\bthetaks{t} - (\bA^\top \hSigmak{t} \bA)^{-1}\bA^\top\nabla \fk{t}(\bAks{t} \bthetaks{t}).
\end{align}
Therefore,
\begin{align}
	\twonorm{\bA\bthetak{t}_{\bA}-\bAks{t}\bthetaks{t}} &\leq \underbrace{\twonorm{[\bm{I}-\bA(\bA^\top \hSigmak{t} \bA)^{-1}\bA^\top \hSigmak{t}]\bAks{t} \bthetaks{t}}}_{[1]} \\
	&\quad + \underbrace{\twonorm{\bA(\bA^\top \hSigmak{t} \bA)^{-1}\bA^\top \nabla \fk{t}(\bAks{t} \bthetaks{t})}}_{[2]}.
\end{align}
Note that 
\begin{align}
	[1] &\lesssim \twonorm{\bAks{t}-\bA(\bA^\top \hSigmak{t} \bA)^{-1}\bA^\top \hSigmak{t}\bAks{t}} \\
	&= \twonorm{\bAks{t}-\bA(\bA^\top \hSigmak{t} \bA)^{-1}\bA^\top \hSigmak{t}(\bA\bA^\top + \bA^{\perp}(\bA^{\perp})^\top)\bAks{t}} \\
	&\leq \twonorm{(\bm{I} - \bA\bA^\top)\bAks{t}} + \twonorm{\bA(\bA^\top \hSigmak{t} \bA)^{-1}\bA^\top \hSigmak{t}\bA^{\perp}(\bA^{\perp})^\top\bAks{t}} \\
	&\lesssim \twonorm{(\bA^{\perp})^\top\bAks{t}} + \twonorm{(\bA^\top \hSigmak{t} \bA)^{-1}} \cdot \twonorm{\bA^\top \hSigmak{t}\bA^{\perp}} \cdot\twonorm{(\bA^{\perp})^\top\bAks{t}}
\end{align}
By Lemma \ref{lem: cov hat r} and the condition $n_0 \geq Cr$,
\begin{align}
	 \twonorm{(\bA^\top \hSigmak{t} \bA)^{-1}} &= \lambda_{\min}^{-1}(\bA^\top \hSigmak{t} \bA) \\
	 &\leq [\lambda_{\min}(\bA^\top \bSigmak{t} \bA) - \twonorm{\bA^\top \hSigmak{t} \bA - \bA^\top \bSigmak{t} \bA}]^{-1} \\
	 &\leq \big[\lambda_{\min}(\bA^\top \bSigmak{t} \bA) - C\sqrt{r/n_0}\big]^{-1} \\
	 &\lesssim C'',
\end{align}
\wprp. And similarly
\begin{equation}
	\twonorm{\bA^\top \hSigmak{t}\bA^{\perp}} \lesssim \twonorm{\bA^\top \bSigmak{t}\bA^{\perp}} + \twonorm{\bA^\top (\hSigmak{t}-\bSigmak{t})\bA^{\perp}} \lesssim C,
\end{equation}
\wprp. 
Therefore by Lemma \ref{lem: distance equivalence}, \wpr,
\begin{equation}\label{eq: lemma 19 eq 1}
	[1] \lesssim \twonorm{\bA\bA^\top - \bAks{t}(\bAks{t})^\top}.
\end{equation}

And w.p. at least $1-e^{-C(r+\log T)}$,
\begin{equation}\label{eq: lemma 19 eq 2}	
	[2] \lesssim \twonorm{(\bA^\top \hSigmak{t} \bA)^{-1}} \cdot \twonorm{\bA^\top \nabla \fk{t}(\bAks{t} \bthetaks{t})}  \lesssim \sqrt{\frac{r}{n_0}},
\end{equation}
Denote $\{\bbeta_j\}_{j=1}^N$ as a $1/2$-cover (whose components are inside the set to be covered) of $\mathscr{B} = \{\bbeta \in \mathbb{R}^p: \bbeta = \bA\bu, \twonorm{\bu} \leq 1\}$ (which is isomorphic to the unit ball in $\mathbb{R}^r$), with $N \leq 5^r$ (by Example 5.8 in \cite{wainwright2019high}). The second inequality in \eqref{eq: lemma 19 eq 2} holds because
\begin{align}
	\twonorm{\bA^\top \nabla \fk{t}(\bAks{t} \bthetaks{t})} &= \sup_{\twonorm{\bu} \leq 1} (\bA\bu)^\top\left[\frac{2}{n_0}(\bXk{0})^\top\bepsilonk{0}\right] \\
	&= \sup_{\bbeta \in \mathscr{B}} \bbeta^\top\left[\frac{2}{n_0}(\bXk{0})^\top\bepsilonk{0}\right] \\
	&\leq \sup_{\bbeta \in \mathscr{B}} (\bbeta-\bbeta^\top_{j_0})^\top\left[\frac{2}{n_0}(\bXk{0})^\top\bepsilonk{0}\right] + \max_{j\in[N]} \bbeta^\top_j\left[\frac{2}{n_0}(\bXk{0})^\top\bepsilonk{0}\right] \\
	&\leq \sup_{\bbeta = \bA\bu\in \mathscr{B}} (\bu - \bu_{j_0})^\top\bA\nabla \fk{t}(\bAks{t} \bthetaks{t}) + \max_{j\in[N]} \bbeta^\top_j\left[\frac{2}{n_0}(\bXk{0})^\top\bepsilonk{0}\right] \\
	&\leq \frac{1}{2}\twonorm{\bA\nabla \fk{t}(\bAks{t} \bthetaks{t})} + \max_{j\in[N]} \bbeta^\top_j\left[\frac{2}{n_0}(\bXk{0})^\top\bepsilonk{0}\right],
\end{align}
leading to
\begin{equation}
	\twonorm{\bA^\top \nabla \fk{t}(\bAks{t} \bthetaks{t})} \leq 2\max_{j\in[N]} \bbeta^\top_j\left[\frac{2}{n_0}(\bXk{0})^\top\bepsilonk{0}\right],
\end{equation}
where $\bbeta_{j_0} = \bA\bu_{j_0}$ with some $\twonorm{\bu_{j_0}} \leq 1$ is the one which is closest to $\bbeta = \bA\bu$ in the cover (depending on $\bbeta$). Hence, $\twonorm{\bu_{j_0} - \bu} = \twonorm{\bbeta_{j_0} - \bbeta} \leq 1/2$. Showing $\max_{j\in[N]} \bbeta^\top_j\big[\frac{2}{n_0}(\bXk{0})^\top\bepsilonk{0}\big] \lesssim \sqrt{r/n_0}$ \wpr\, is standard.

Finally, combining \eqref{eq: lemma 19 eq 1} and \eqref{eq: lemma 19 eq 2}, we complete the proof.

\subsubsection{Proof of Lemma \ref{lem: a theta a tl}}
The proof is exactly the same as the proof of Lemma \ref{lem: a theta a}, so omitted.

\subsubsection{Proof of Lemma \ref{lem: safe net tl}}
The proof is the same as the proof of Lemma \ref{lem: safe net mtl}, so we omit it here. 

\subsubsection{Proof of Proposition \ref{prop: mtl}}
Define the random event $\mathcal{E}_1$ as the event that Lemmas \ref{lem: a theta a}, \ref{lem: theta bdded}, \ref{lem: a theta a extention}, \ref{lem: gradient 2 norm}, and \ref{lem: the same A} hold, hence $\tP(\mathcal{E}_1) \geq 1-Ce^{-(r+\log T)}$ with some constant $C > 0$. Define another event
\begin{equation}
	\mathcal{E}_2 = \bigg\{\max_{t\in [T]}\twonorm{\hSigmak{t}-\bSigmak{t}} \lesssim \sqrt{\frac{p+\log T}{n}}\bigg\}.
\end{equation}
By Lemma \ref{lem: cov hat}, we have $\tP(\mathcal{E}_2) \geq 1-Ce^{-(p+\log T)}$ with some constant $C > 0$. It suffices to prove the upper bounds in Proposition \ref{prop: mtl} conditioned on $\mathcal{E}_1 \cap \mathcal{E}_2$, therefore we condition on $\mathcal{E}_1 \cap \mathcal{E}_2$ in the remaining proof without stating it explicitly and all the arguments are deterministic.

Define
\begin{align}
	G(\bA) &= \sum_{t = 1}^T \min_{\bAk{t} \in \mO, \bthetak{t} \in \mathbb{R}^r}\left\{\frac{1}{T}\fk{t}(\bAk{t}\bthetak{t}) + \frac{\sqrt{n}}{nT}\lambda \twonorm{\bAk{t}(\bAk{t})^\top - \bA\bA^\top}\right\}, \\
	G_S(\bA) &= \sum_{t \in S} \min_{\bAk{t} \in \mO, \bthetak{t} \in \mathbb{R}^r}\left\{\frac{1}{T}\fk{t}(\bAk{t}\bthetak{t}) + \frac{\sqrt{n}}{nT}\lambda \twonorm{\bAk{t}(\bAk{t})^\top - \bA\bA^\top}\right\}, \\
	G_{S^c}(\bA) &= \sum_{t \in S^c} \min_{\bAk{t} \in \mO, \bthetak{t} \in \mathbb{R}^r}\left\{\frac{1}{T}\fk{t}(\bAk{t}\bthetak{t}) + \frac{\sqrt{n}}{nT}\lambda \twonorm{\bAk{t}(\bAk{t})^\top - \bA\bA^\top}\right\}, \\
	\hbarA &\in \argmin_{\bA \in \mO}G(\bA).
\end{align}
Our first step is to prove the same bound for $\twonorm{\hbarA\hbarA^\top - \barA\barA^\top}$. Let's prove it by contradiction. Denote the desired rate as
\begin{equation}
	\eta = \barzeta^{-1}r\sqrt{\frac{p}{nT}} + \barzeta^{-1}\sqrt{r}\sqrt{\frac{r+\log T}{n}} + \sqrt{r}h + \frac{\lambda r}{\sqrt{n}\barzeta^2}\epsilon.
\end{equation}
Suppose $\twonorm{\hbarA\hbarA^\top - \barA\barA^\top} > C\eta$ with some constant $C > 0$. 

Consider any minimizer $(\bAk{t}, \bthetak{t}) = \argmin_{\bA \in \mO, \btheta \in \mathbb{R}^r}\{\frac{1}{T}\fk{t}(\bA\btheta) + \frac{\sqrt{n}}{nT}\lambda \twonorm{\bA\bA^\top - \hbarA(\hbarA)^\top}\}$. Define two index sets
\begin{align}
	\mA_1 &= \left\{t \in S: \twonorm{\bAk{t}(\bAk{t})^\top - \hbarA(\hbarA)^\top} \geq c\twonorm{\hbarA(\hbarA)^\top - \barA \barA^\top}\cdot \frac{1}{\sqrt{r}}\right\}, \\
	\mA_2 &= \left\{t \in S: \twonorm{\bAk{t}(\bAk{t})^\top - \hbarA(\hbarA)^\top} < c\twonorm{\hbarA(\hbarA)^\top - \barA \barA^\top}\cdot \frac{1}{\sqrt{r}}\right\},
\end{align}
where $c > 0$ is a small constant. Then we have
\begin{align}
	&\sum_{t \in \mA_1}\frac{1}{T}[\fk{t}(\bAk{t}\bthetak{t})-\fk{t}(\barA\bthetak{t}_{\barA})] +  \sum_{t \in \mA_1}\frac{\sqrt{n}}{nT}\twonorm{\bAk{t}(\bAk{t})^\top - \hbarA(\hbarA)^\top} \\
	&\geq -\frac{1}{T}\norma{\left\<\{\nabla \fk{t}(\barA\bthetak{t}_{\barA})\}_{t \in \mA_1}, \{\bAk{t}\bthetak{t} - \barA\bthetak{t}_{\barA}\}_{t \in \mA_1}\right\>} \\
	&\quad + \frac{1}{2}\frac{1}{T}\fnorm{\{\bAk{t}\bthetak{t} - \barA\bthetak{t}_{\barA}\}_{t \in \mA_1}}^2 + \sum_{t \in \mA_1}\frac{1}{\sqrt{n}T}\lambda \twonorm{\bAk{t}(\bAk{t})^\top - \hbarA(\hbarA)^\top} \\
	&\geq -\frac{1}{T}\norma{\left\<\{\nabla \fk{t}(\bAks{t}\bthetaks{t})\}_{t \in \mA_1}, \{\bAk{t}\bthetak{t} - \hbarA\bthetak{t}_{\hbarA} + \hbarA\bthetak{t}_{\hbarA}- \barA\bthetak{t}_{\barA}\}_{t \in \mA_1}\right\>}\\
	&\quad - \frac{1}{T}\fnorm{\{\nabla \fk{t}(\barA\bthetak{t}_{\barA}) - \nabla \fk{t}(\bAks{t}\bthetaks{t})\}} \fnorm{\{\bAk{t}\bthetak{t} - \barA\bthetak{t}_{\barA}\}_{t \in \mA_1}} \\
	&\quad + \frac{1}{2}\frac{1}{T}\fnorm{\{\bAk{t}\bthetak{t} - \barA\bthetak{t}_{\barA}\}_{t \in \mA_1}}^2 + \sum_{t \in \mA_1}\frac{1}{\sqrt{n}T}\lambda \twonorm{\bAk{t}(\bAk{t})^\top - \hbarA(\hbarA)^\top} \\
	&\geq -\frac{1}{T}\sum_{t \in \mA_1}\twonorm{\nabla \fk{t}(\bAks{t}\bthetaks{t})}\cdot (\twonorm{\bAk{t}\bthetak{t} - \hbarA\bthetak{t}_{\hbarA}} + \twonorm{\hbarA\bthetak{t}_{\hbarA}- \barA\bthetak{t}_{\barA}})\\
	&\quad -\frac{1}{T}C\left[\sqrt{T}\sqrt{\frac{r+\log T}{n}} + h\sqrt{\sum_{t \in \mA_1}(\zetak{t})^2}\right]\cdot \fnorm{\{\bAk{t}\bthetak{t} - \barA\bthetak{t}_{\barA}\}_{t \in \mA_1}} \\
	&\quad + \frac{1}{2}\frac{1}{T}\fnorm{\{\bAk{t}\bthetak{t} - \barA\bthetak{t}_{\barA}\}_{t \in \mA_1}}^2 + \sum_{t \in \mA_1}\frac{1}{\sqrt{n}T}\lambda \twonorm{\bAk{t}(\bAk{t})^\top - \hbarA(\hbarA)^\top} \label{eq: prop 1 eq 2}\\
	&\geq -\frac{C}{T}\sum_{t \in \mA_1}\sqrt{\frac{p+\log T}{n}}\cdot \zetak{t}(\twonorm{\bAk{t}(\bAk{t})^\top - \hbarA(\hbarA)^\top} + \twonorm{\hbarA(\hbarA)^\top- \barA\barA^\top})\\
	&\quad -\frac{1}{T}C\left[\sqrt{T}\sqrt{\frac{r+\log T}{n}} + h\sqrt{\sum_{t \in \mA_1}(\zetak{t})^2}\right]\cdot \fnorm{\{\bAk{t}\bthetak{t} - \barA\bthetak{t}_{\barA}\}_{t \in \mA_1}} \\
	&\quad + \frac{1}{2}\frac{1}{T}\fnorm{\{\bAk{t}\bthetak{t} - \barA\bthetak{t}_{\barA}\}_{t \in \mA_1}}^2 + \sum_{t \in \mA_1}\frac{1}{\sqrt{n}T}\lambda \twonorm{\bAk{t}(\bAk{t})^\top - \hbarA(\hbarA)^\top} \label{eq: prop 1 eq idk} \\
	&\geq -C\left(h^2\cdot \frac{1}{T}\sum_{t \in \mA_1}(\zetak{t})^2 + \frac{r+\log T}{n}\right) + \frac{|\mA_1|}{2\sqrt{n}T}\lambda \cdot \frac{1}{\sqrt{r}}\twonorm{\hbarA(\hbarA)^\top - \barA \barA^\top}, \label{eq: prop 1 eq 1}
\end{align}
where \eqref{eq: prop 1 eq 2} holds because $\twonorm{\{\nabla \fk{t}(\bAks{t}\bthetaks{t})\}_{t \in \mA_1}} \leq \sqrt{\frac{p}{n}} + h\sqrt{\sum_{t \in \mA_1}(\zetak{t})^2} + \sqrt{T}\sqrt{\frac{r+\log T}{n}}$ (Lemma \ref{lem: gradient 2 norm}) and $\fnorm{\{\nabla \fk{t}(\barA\bthetak{t}_{\barA}) - \nabla \fk{t}(\bAks{t}\bthetaks{t})\}} = \\ \sqrt{\sum_{t \in \mA_1}\twonorm{\hSigmak{t}(\barA\bthetak{t}_{\barA}-\bAks{t}\bthetaks{t})}^2} \lesssim \sqrt{\sum_{t \in \mA_1} \twonorm{\barA\bthetak{t}_{\barA}-\bAks{t}\bthetaks{t}}^2} \lesssim \sqrt{h^2\sum_{t \in \mA_1}(\zetak{t})^2 + T\cdot \frac{r+\log T}{n}}$ (Lemma \ref{lem: a theta a}). Inequality \eqref{eq: prop 1 eq idk} holds because
\begin{align}
	\twonorm{\bAk{t}\bthetak{t}} &\lesssim \twonorm{\bAk{t}(\bAk{t})^\top-\hbarA(\hbarA)^\top}(\twonorm{\bthetak{t}_{\hbarA}} + \twonorm{\nabla f^{(t)}(\hbarA\bthetak{t}_{\hbarA})}) \\
	&\lesssim \twonorm{\bAk{t}(\bAk{t})^\top-\hbarA(\hbarA)^\top}\Big(\zetak{t}+\twonorm{\hbarA\bthetak{t}_{\hbarA} - \bAks{t}\bthetaks{t}} + \twonorm{\nabla f^{(t)}(\bAks{t}\bthetaks{t})}\Big) \\
	&\lesssim \twonorm{\bAk{t}(\bAk{t})^\top-\hbarA(\hbarA)^\top}\zetak{t},
\end{align} 
where we used Lemmas \ref{lem: a theta a}, \ref{lem: theta bdded}, and \ref{lem: a theta a extention}. Inequality \eqref{eq: prop 1 eq 1} holds because
\begin{align}
	&\frac{1}{\sqrt{n}T}\lambda \sum_{t \in \mA_1}\twonorm{\bAk{t}(\bAk{t})^\top - \hbarA(\hbarA)^\top} \\
	&\quad -\frac{1}{T}\sum_{t \in \mA_1}\sqrt{\frac{p+\log T}{n}}\cdot \zetak{t} (\twonorm{\bAk{t}(\bAk{t})^\top - \hbarA(\hbarA)^\top} + \twonorm{\hbarA(\hbarA)^\top- \barA\barA^\top}) \\
	&\geq \frac{1}{\sqrt{n}T}\lambda \sum_{t \in \mA_1}\twonorm{\bAk{t}(\bAk{t})^\top - \hbarA(\hbarA)^\top} - \frac{1}{T}\sum_{t \in \mA_1}C\sqrt{\frac{r(p+\log T)}{n}}\cdot\zetak{t}\twonorm{\bAk{t}(\bAk{t})^\top - \hbarA(\hbarA)^\top} \\
	&\geq \frac{1}{2\sqrt{n}T}\lambda \sum_{t \in \mA_1}\twonorm{\bAk{t}(\bAk{t})^\top - \hbarA(\hbarA)^\top} \\
	&\geq \frac{|\mA_1|}{2\sqrt{n}T}\lambda \cdot \frac{1}{\sqrt{r}}\twonorm{\hbarA(\hbarA)^\top - \barA \barA^\top},
\end{align}
when $\frac{\lambda}{\sqrt{n}} \geq 2C\max_{t \in S}\zetak{t}\cdot \sqrt{\frac{r(p+\log T)}{n}}$.

For any $t \in \mA_2$, since $\twonorm{\bAk{t}\bthetak{t}_{\bAk{t}} - \hbarA\bthetak{t}_{\hbarA}} \leq \twonorm{\bAk{t}(\bAk{t})^\top - \hbarA(\hbarA)^\top}(\twonorm{\bthetak{t}_{\hbarA}}+\twonorm{\nabla \fk{t}(\hbarA \bthetak{t}_{\hbarA})}) \leq c\zetak{t}\twonorm{\hbarA(\hbarA)^\top - \barA \barA^\top} \frac{1}{\sqrt{r}}$, by triangle inequality, we have
\begin{equation}
	\twonorm{\bAk{t}\bthetak{t}_{\bAk{t}} - \barA\bthetak{t}_{\barA}} \geq \twonorm{\hbarA\bthetak{t}_{\hbarA} - \barA\bthetak{t}_{\barA}} - \frac{c}{\sqrt{r}}\zetak{t}\twonorm{\hbarA(\hbarA)^\top - \barA \barA^\top}.
\end{equation}
With similar arguments to obtain \eqref{eq: prop 1 eq 1}, here we have
\begin{align}
	&\sum_{t \in \mA_2}\frac{1}{T}[\fk{t}(\bAk{t}\bthetak{t})-\fk{t}(\barA\bthetak{t}_{\barA})] +  \sum_{t \in \mA_2}\frac{\sqrt{n}}{nT}\twonorm{\bAk{t}(\bAk{t})^\top - \hbarA(\hbarA)^\top} \\
	&\geq \frac{1}{2T}\fnorm{\{\bAk{t}\bthetak{t} - \barA\bthetak{t}_{\barA}\}_{t \in \mA_2}}^2   - \frac{1}{T}\sqrt{\frac{p+T}{n}}\cdot \sqrt{2r}\fnorma{\{\hbarA\bthetak{t}_{\hbarA}- \barA\bthetak{t}_{\barA}\}_{t \in \mA_2}}\\
	&\quad - \frac{C'}{T}\left(\sqrt{\frac{pr}{n}} + \sqrt{T}\sqrt{\frac{r+\log T}{n}} + h\sqrt{\sum_{t \in \mA_2}(\zetak{t})^2}\right)\cdot \fnorm{\{\bAk{t}\bthetak{t} - \barA\bthetak{t}_{\barA}\}_{t \in \mA_2}}\\
	&\quad +  \frac{1}{T}\sum_{t \in \mA_2}\frac{\lambda}{\sqrt{n}}\twonorm{\bAk{t}(\bAk{t})^\top - \hbarA(\hbarA)^\top} -\frac{C}{T}\sum_{t \in \mA_2}\sqrt{\frac{p+\log T}{n}}\cdot \zetak{t}\twonorm{\bAk{t}(\bAk{t})^\top - \hbarA(\hbarA)^\top} \label{eq: prop 1 eq new}\\
	&\geq -C\left(\frac{pr}{nT} + h^2\cdot \frac{1}{T}\sum_{t \in \mA_2}(\zetak{t})^2 + \frac{r+\log T}{n}\right) + \frac{C}{T}\fnorm{\{\bAk{t}\bthetak{t} - \barA\bthetak{t}_{\barA}\}_{t \in \mA_2}}^2 \\
	&\quad -\frac{1}{T}\sqrt{\frac{p+T}{n}}\cdot \sqrt{2r}\fnorma{\{\hbarA\bthetak{t}_{\hbarA}- \barA\bthetak{t}_{\barA}\}_{t \in \mA_2}}\\
	&= -C\left(\frac{pr}{nT} + h^2\cdot \frac{1}{T}\sum_{t \in \mA_2}(\zetak{t})^2 + \frac{r+\log T}{n}\right) + \frac{C}{T}\sum_{t\in\mA_2}\twonorm{\bAk{t}\bthetak{t} - \barA\bthetak{t}_{\barA}}^2  \\
	&\quad - \frac{1}{T}\sqrt{\frac{p+T}{n}}\cdot \sqrt{2r}\fnorma{\{\hbarA\bthetak{t}_{\hbarA}- \barA\bthetak{t}_{\barA}\}_{t \in \mA_2}} \label{eq: prop 1 eq imp} \\
	&\geq -C\left(\frac{pr}{nT} + h^2\cdot \frac{1}{T}\sum_{t \in \mA_2}(\zetak{t})^2 + \frac{r+\log T}{n}\right) + \frac{C'}{T}\sum_{t\in\mA_2}\twonorm{\hbarA\bthetak{t}_{\hbarA} - \barA\bthetak{t}_{\barA}}^2 \\
	&\quad -\frac{1}{T}\sqrt{\frac{p+T}{n}}\cdot \sqrt{2r}\fnorma{\{\hbarA\bthetak{t}_{\hbarA}- \barA\bthetak{t}_{\barA}\}_{t \in \mA_2}} -\frac{c}{Tr}\sum_{t \in \mA_2}(\zetak{t})^2\cdot \twonorm{\hbarA(\hbarA)^\top - \barA \barA^\top}^2 \\
	&\geq -C\left(\frac{pr}{nT} + h^2\cdot \frac{1}{T}\sum_{t \in \mA_2}(\zetak{t})^2 + \frac{r+\log T}{n}\right) + \frac{C'}{2T}\sum_{t\in\mA_2}\twonorm{\hbarA\bthetak{t}_{\hbarA} - \barA\bthetak{t}_{\barA}}^2 \\
	&\quad + \frac{C'}{2T}\fnorma{\{\hbarA\bthetak{t}_{\hbarA}- \barA\bthetak{t}_{\barA}\}_{t \in \mA_2}}^2 -\frac{1}{T}\sqrt{\frac{p+T}{n}}\cdot \sqrt{2r}\fnorma{\{\hbarA\bthetak{t}_{\hbarA}- \barA\bthetak{t}_{\barA}\}_{t \in \mA_2}} \\
	&\quad -\frac{c}{Tr}\sum_{t \in \mA_2}(\zetak{t})^2\cdot \twonorm{\hbarA(\hbarA)^\top - \barA \barA^\top}^2.\\
	&\geq -C\left(\frac{pr}{nT} + h^2\cdot \frac{1}{T}\sum_{t \in \mA_2}(\zetak{t})^2 + \frac{r+\log T}{n}\right) + \frac{C'}{2T}\sum_{t\in\mA_2}\twonorm{\hbarA\bthetak{t}_{\hbarA} - \barA\bthetak{t}_{\barA}}^2 \\
	&\quad -\frac{c}{Tr}\sum_{t \in \mA_2}(\zetak{t})^2\cdot \twonorm{\hbarA(\hbarA)^\top - \barA \barA^\top}^2.  \label{eq: prop 1 eq 3}
\end{align}
Compared to the derivation of \eqref{eq: prop 1 eq 1}, here we used a different way to bound the intermediate term $\frac{1}{T}|\<\{\nabla \fk{t}(\bAks{t}\bthetaks{t})\}_{t \in \mA_2}, \{\bAk{t}\bthetak{t} - \hbarA\bthetak{t}_{\hbarA} + \hbarA\bthetak{t}_{\hbarA}- \barA\bthetak{t}_{\barA}\}_{t \in \mA_2}\>|$ to obtain \eqref{eq: prop 1 eq new}. More specifically,
\begin{align}
	&\frac{1}{T}\Big|\Big\<\{\nabla \fk{t}(\bAks{t}\bthetaks{t})\}_{t \in \mA_2}, \{\bAk{t}\bthetak{t} - \hbarA\bthetak{t}_{\hbarA} + \hbarA\bthetak{t}_{\hbarA}- \barA\bthetak{t}_{\barA}\}_{t \in \mA_2}\Big\>\Big| \\
	&\leq \frac{1}{T}\sum_{t \in \mA_2}\twonorm{\nabla \fk{t}(\bAks{t}\bthetaks{t})}\twonorm{\bAk{t}\bthetak{t} - \hbarA\bthetak{t}_{\hbarA}} + \frac{1}{T}\twonorm{\{\nabla \fk{t}(\bAks{t}\bthetaks{t})\}_{t \in \mA_2}} \big\|\{\hbarA\bthetak{t}_{\hbarA}- \barA\bthetak{t}_{\barA}\}_{t \in \mA_2}\big\|_* \\
	&\leq \frac{1}{T}\sum_{t \in \mA_2}\sqrt{\frac{p+\log T}{n}}\zetak{t}\twonorm{\bAk{t}(\bAk{t})^\top - \hbarA(\hbarA)^\top}   + \frac{1}{T}\sqrt{\frac{p+T}{n}}\cdot \sqrt{2r}\fnorma{\{\hbarA\bthetak{t}_{\hbarA}- \barA\bthetak{t}_{\barA}\}_{t \in \mA_2}},
\end{align}
where we used the fact that $|\<\bm{A}, \bm{B}\>| \leq \twonorm{\bm{A}}\|\bm{B}\|_* \leq \twonorm{\bm{A}}\fnorm{\bm{B}}\cdot \textup{rank}(\bm{B})$ for any matrix $\bm{A}$ and $\bm{B}$ of the same dimension and $\textup{rank}\big(\{\hbarA\bthetak{t}_{\hbarA}- \barA\bthetak{t}_{\barA}\}_{t \in \mA_2}\big) \leq 2r$.

Finally, notice that by triangle inequality,
\begin{align} 
	G_{S^c}(\hbarA) - G_{S^c}(\barA) &\geq \sum_{t \in S^c} \min_{\bAk{t} \in \mO, \bthetak{t} \in \mathbb{R}^r}\left\{\frac{\sqrt{n}}{nT}\lambda (\twonorm{\bAk{t}(\bAk{t})^\top - \hbarA(\hbarA)^\top} - \twonorm{\bAk{t}(\bAk{t})^\top - \barA(\barA)^\top})\right\} \\
	&\geq -\frac{\sqrt{n}}{nT}\lambda |S^c|\cdot \twonorm{\hbarA(\hbarA)^\top - \barA(\barA)^\top}. \label{eq: prop 1 eq 7}
\end{align}

\noindent \underline{\textbf{Case 1:}} $|\mA_2| \geq (1-c/r\cdot \barzeta^2/\max_{t\in S}(\zetak{t})^2)|S|$ with some small constant $c > 0$:

Then by \eqref{eq: prop 1 eq 1}, \eqref{eq: prop 1 eq 3}, and \eqref{eq: prop 1 eq 7},
\begin{align}
	G(\hbarA) - G(\barA) &= G_S(\hbarA) - G_S(\barA) + G_{S^c}(\hbarA) - G_{S^c}(\barA) \\
	&\geq -C\left(\frac{pr}{nT} + h^2 \barzeta^2 + \frac{r+\log T}{n}\right) + \frac{C'}{T}\sum_{t\in\mA_2}\twonorm{\hbarA\bthetak{t}_{\hbarA} - \barA\bthetak{t}_{\barA}}^2 \\
	&\quad -\frac{c}{Tr}\sum_{t\in\mA_2}(\zetak{t})^2\cdot \twonorm{\hbarA(\hbarA)^\top - \barA \barA^\top}^2 \\
	&\quad -\frac{\sqrt{n}}{nT}\lambda |S^c|\cdot \twonorm{\hbarA(\hbarA)^\top - \barA\barA^\top}. \label{eq: prop 1 eq 8}
\end{align}
By Assumption \ref{asmp: theta}, Wedin's $\sin \Theta$-Theorem, and applying Lemma \ref{lem: small fraction diversity} on $\{\bthetak{t}_{\barA}\}_{t \in \mathcal{A}_2}$,
\begin{equation}\label{eq: prop 1 eq x}
	\twonorm{\hbarA(\hbarA)^\top - \barA(\barA)^\top} \leq \fnorm{\hbarA(\hbarA)^\top - \barA(\barA)^\top} \lesssim \sqrt{\frac{r}{|S|}}\barzeta^{-1} \cdot \fnorm{\{\hbarA\bthetak{t}_{\hbarA} - \barA\bthetak{t}_{\barA}\}_{t \in \mA_2}}.
\end{equation}
Here we used the fact that $\frac{1}{|S|}\sum_{t \in S}\bthetak{t}_{\barA}(\bthetak{t}_{\barA})^\top \succeq \frac{c}{r}\barzeta^2 \bm{I}_p$. To see this, notice that
\begin{align}
	\sigma_r\left(\frac{1}{\sqrt{|S|}}\{\barA\bthetak{t}_{\barA}\}_{t \in S}\right) &= \sigma_r\left(\frac{1}{\sqrt{|S|}}\{\bbetaks{t}\}_{t \in S}\right) - \frac{1}{\sqrt{|S|}}\twonorm{\{\barA\bthetak{t}_{\barA}\}_{t \in S}-\{\bbetaks{t}\}_{t \in S}} \\
	&\geq \frac{c}{\sqrt{r}}\barzeta - \left(\barzeta h+\sqrt{\frac{r+\log T}{n}}\right) \\
	&\geq \frac{c'}{\sqrt{r}}\barzeta,
\end{align}
where by Lemmas \ref{lem: a theta a} and \ref{lem: matrix 2 norm},
\begin{equation}
	\frac{1}{\sqrt{|S|}}\twonorm{\{\barA\bthetak{t}_{\barA}\}_{t \in S}-\{\bbetaks{t}\}_{t \in S}} \leq \sqrt{\sum_{t \in S} \twonorm{\barA\bthetak{t}_{\barA} - \bAks{t}\bthetaks{t}}^2} \lesssim h\barzeta + \sqrt{\frac{r+\log T}{n}},
\end{equation}
and the last step comes from conditions (\rom{1}) and (\rom{2}) by noticing that
\begin{equation}
	\sqrt{\frac{r+\log T}{n}} \lesssim \frac{\barzeta}{\sqrt{r}}\cdot \sqrt{\frac{p+\log T}{n}}\cdot \frac{1}{\zetak{t}} \lesssim \frac{\barzeta}{\sqrt{r}}.
\end{equation}
This leads to
\begin{equation}
	\lambda_r\left(\frac{1}{|S|}\sum_{t \in S}\bthetak{t}_{\barA}(\bthetak{t}_{\barA})^\top\right) = \lambda_r\left(\frac{1}{|S|}\sum_{t \in S}\bthetak{t}_{\barA}(\barA)^\top\barA(\bthetak{t}_{\barA})^\top\right) \geq \frac{c'}{r}\barzeta^2.
\end{equation}
Plugging \eqref{eq: prop 1 eq x} into \eqref{eq: prop 1 eq 8}, we have
\begin{align}
	G(\hbarA) - G(\barA) &\geq -C\left(h^2\barzeta^2 + \frac{r+\log T}{n}\right) + \frac{C'}{T}\cdot \frac{|S|}{r}\barzeta^2 \twonorm{\hbarA(\hbarA)^\top - \barA\barA^\top}^2 \\
	&\quad -\frac{c|S|}{Tr}\barzeta^2\twonorm{\hbarA(\hbarA)^\top - \barA \barA^\top}^2 - C\sqrt{\frac{r(p+T)}{nT}} \cdot 
	\twonorm{\hbarA(\hbarA)^\top - \barA \barA^\top}\sqrt{\frac{1}{T}\sum_{t \in \mA_2}(\zetak{t})^2} \\
	&\quad -\frac{\sqrt{n}}{nT}\lambda |S^c|\cdot \twonorm{\hbarA(\hbarA)^\top - \barA\barA^\top} \\
	&\geq  \frac{C}{T}\cdot \frac{|S|}{r}\barzeta^2 \twonorm{\hbarA(\hbarA)^\top - \barA(\barA)^\top}^2 -\frac{\sqrt{n}}{nT}\lambda |S^c|\cdot \twonorm{\hbarA(\hbarA)^\top - \barA(\barA)^\top} \\
	&\quad -C\left(\frac{pr}{nT} + h^2\barzeta^2 + \frac{r+\log T}{n}\right) - C\sqrt{\frac{r(p+T)}{nT}}\barzeta \cdot  \\
	&> 0,
\end{align}
because $\twonorm{\hbarA(\hbarA)^\top - \barA \barA^\top} \geq C\eta$ with some constant $C > 0$.

\noindent\underline{\textbf{Case 2:}} $|\mA_1| \geq c/r\cdot \barzeta^2/\max_{t\in S}(\zetak{t})^2\cdot |S|$ with some small constant $c > 0$: 

By inequality \eqref{eq: prop 1 eq imp} and triangle inequality, we have
\begin{align}
	&\sum_{t \in \mA_2}\frac{1}{T}[\fk{t}(\bAk{t}\bthetak{t})-\fk{t}(\barA\bthetak{t}_{\barA})] +  \sum_{t \in \mA_2}\frac{\sqrt{n}}{nT}\twonorm{\bAk{t}(\bAk{t})^\top - \hbarA(\hbarA)^\top} \\
	&= -C\left(\frac{pr}{nT} + h^2\cdot \frac{1}{T}\sum_{t \in \mA_2}(\zetak{t})^2 + \frac{r+\log T}{n}\right) + \frac{C}{T}\sum_{t\in\mA_2}\twonorm{\bAk{t}\bthetak{t} - \barA\bthetak{t}_{\barA}}^2  \\
	&\quad - \frac{1}{T}\sqrt{\frac{p+T}{n}}\cdot \sqrt{2r}\fnorma{\{\hbarA\bthetak{t}_{\hbarA}- \barA\bthetak{t}_{\barA}\}_{t \in \mA_2}}\\
	&\geq -C\left(\frac{pr}{nT} + h^2\cdot \frac{1}{T}\sum_{t \in \mA_2}(\zetak{t})^2 + \frac{r+\log T}{n}\right) + \frac{C}{T}\sum_{t\in\mA_2}\twonorm{\bAk{t}\bthetak{t} - \barA\bthetak{t}_{\barA}}^2  \\
	&\quad - \frac{1}{T}\sqrt{\frac{p+T}{n}}\cdot \sqrt{2r}\fnorma{\{\bAk{t}\bthetak{t}- \barA\bthetak{t}_{\barA}\}_{t \in \mA_2}} - \frac{1}{T}\sqrt{\frac{p+T}{n}}\cdot \sqrt{2r}\fnorma{\{\bAk{t}\bthetak{t} - \hbarA\bthetak{t}_{\hbarA}\}_{t \in \mA_2}} \\
	&\geq -C\left(\frac{pr}{nT} + h^2\cdot \frac{1}{T}\sum_{t \in \mA_2}(\zetak{t})^2 + \frac{r+\log T}{n}\right) + \frac{C}{T}\fnorma{\{\bAk{t}\bthetak{t}- \barA\bthetak{t}_{\barA}\}_{t \in \mA_2}}^2  \\
	&\quad - \frac{1}{T}\sqrt{\frac{p+T}{n}}\cdot \sqrt{2r}\fnorma{\{\bAk{t}\bthetak{t}- \barA\bthetak{t}_{\barA}\}_{t \in \mA_2}} \\
	&\quad - \sqrt{\frac{p+T}{nT}}\cdot \sqrt{2r} \cdot \sqrt{\frac{1}{T}\sum_{t \in \mA_2}\twonorm{\bAk{t}(\bAk{t})^\top - \hoA(\hoA)^\top}^2 (\zetak{t})^2} \\
	&\geq -C\left(\frac{pr}{nT} + h^2\cdot \frac{1}{T}\sum_{t \in \mA_2}(\zetak{t})^2 + \frac{r+\log T}{n}\right) + \frac{C}{T}\fnorma{\{\bAk{t}\bthetak{t}- \barA\bthetak{t}_{\barA}\}_{t \in \mA_2}}^2  \\
	&\quad - \frac{1}{T}\sqrt{\frac{p+T}{n}}\cdot \sqrt{2r}\fnorma{\{\bAk{t}\bthetak{t}- \barA\bthetak{t}_{\barA}\}_{t \in \mA_2}} - \sqrt{\frac{p+T}{nT}}\cdot\barzeta  \twonorm{\hoA(\hoA)^\top - \oA\oA^\top} \\
	&\geq -C\left(\frac{pr}{nT} + h^2\barzeta^2 + \frac{r+\log T}{n}\right)- \sqrt{\frac{p+T}{nT}}\cdot\barzeta  \twonorm{\hoA(\hoA)^\top - \oA\oA^\top} \label{eq: prop 1 eq 9}
\end{align}
By the condition of the proposition, we must have
\begin{equation}\label{eq: prop 1 eq regime}
	\frac{\min_{t \in S}\zetak{t}}{\barzeta}\cdot r \lesssim \sqrt{p+\log T}, \quad \frac{\min_{t \in S}\zetak{t}}{\barzeta}\cdot \frac{r}{\sqrt{T}} \lesssim 1.
\end{equation}

By \eqref{eq: prop 1 eq 1}, \eqref{eq: prop 1 eq 7}, and \eqref{eq: prop 1 eq 9}, we have
\begin{align}
	&G(\hbarA) - G(\barA) \\
	&\geq \frac{C\lambda}{\sqrt{n}T}\left(\frac{|\mA_1|}{\sqrt{r}} - |S^c|\right)\cdot \twonorm{\hbarA(\hbarA)^\top - \barA \barA^\top}- \sqrt{\frac{p+T}{nT}}\cdot\barzeta  \twonorm{\hoA(\hoA)^\top - \oA\oA^\top}  \\
	&\quad -C\left(\frac{pr}{nT} + h^2\cdot \barzeta^2 + \frac{r+\log T}{n}\right)\\
	&\geq  \frac{C\lambda}{\sqrt{n}T}\left(\frac{|S|}{r^{3/2}}\cdot \frac{\barzeta^2}{\max_{t \in S}(\zetak{t})^2} - |S^c|\right)\cdot \twonorm{\hbarA(\hbarA)^\top - \barA \barA^\top}- \sqrt{\frac{p+T}{nT}}\cdot\barzeta  \twonorm{\hoA(\hoA)^\top - \oA\oA^\top} \\
	&\quad -C\left(\frac{pr}{nT} + h^2\cdot \barzeta^2 + \frac{r+\log T}{n}\right)\\
	&\geq  \frac{C\lambda}{\sqrt{n}}r^{-3/2}\cdot \frac{\barzeta^2}{\max_{t \in S}(\zetak{t})^2} \cdot \twonorm{\hbarA(\hbarA)^\top - \barA \barA^\top} - \sqrt{\frac{p+T}{nT}}\cdot\barzeta  \twonorm{\hoA(\hoA)^\top - \oA\oA^\top} \\
	&\quad -C\left(\frac{pr}{nT} + h^2\cdot \barzeta^2 + \frac{r+\log T}{n}\right)\\
	&\geq \frac{C\lambda}{2\sqrt{n}}r^{-3/2}\cdot \frac{\barzeta^2}{\max_{t \in S}(\zetak{t})^2} \cdot \twonorm{\hbarA(\hbarA)^\top - \barA \barA^\top} -C\left(\frac{pr}{nT} + h^2\cdot \barzeta^2 + \frac{r+\log T}{n}\right)
\end{align}
due to the assumption that $|S^c|/T \leq cr^{-3/2}\barzeta/\max_{t \in S}\zetak{t}$ with a small constant $c > 0$ and
\begin{align}
	\lambda &\asymp \sqrt{r(p+\log T)}\cdot \frac{\max_{t \in S}(\zetak{t})^2}{\min_{t \in S}\zetak{t}} \\
	&\gtrsim \bigg[\sqrt{pr}\frac{\barzeta}{\min_{t \in S}\zetak{t}} + \sqrt{r(p+\log T)}\cdot \frac{\barzeta}{\min_{t \in S}\zetak{t}}\bigg] \cdot \frac{\max_{t \in S}(\zetak{t})^2}{\barzeta},
\end{align}
where we used \eqref{eq: prop 1 eq regime} in the second inequality. Furthermore, by $\min_{t \in S}\zetak{t} \gtrsim \sqrt{\frac{p+\log T}{n}}$, we have
\begin{align}
	&G(\hbarA) - G(\barA) \\
	&\geq C''\sqrt{\frac{p+\log T}{n}}r^{-1}\cdot \frac{\barzeta^2}{\min_{t \in S}\zetak{t}} \cdot \twonorm{\hbarA(\hbarA)^\top - \barA \barA^\top} -C\left(\frac{pr}{nT} + h^2\cdot \barzeta^2 + \frac{r+\log T}{n}\right) \\
	&\geq C''r^{-1}\cdot \barzeta^2 \cdot \twonorm{\hbarA(\hbarA)^\top - \barA \barA^\top} -C\left(\frac{pr}{nT} + h^2\cdot \barzeta^2 + \frac{r+\log T}{n}\right).
\end{align}
Therefore, if $\twonorm{\hoA(\hoA)^\top - \oA\oA^\top} \geq C\eta$, we must have
\begin{equation}
	G(\hbarA) - G(\barA) > 0,
\end{equation}
which implies that $\hbarA$ is not a minimizer of $G$ and contradicts with the optimality of $\hoA$. Hence we must have $\twonorm{\hoA(\hoA)^\top - \oA\oA^\top} \leq C\eta \asymp \barzeta^{-1}r\sqrt{\frac{p}{nT}} + \barzeta^{-1}\sqrt{r}\sqrt{\frac{r+\log T}{n}} + \sqrt{r}h + \frac{\lambda r}{\sqrt{n}\barzeta^2}\epsilon$ with some constant $C > 0$. This completes our proof.


\section{Proofs for GLMs}

\subsection{Lemmas}

\begin{lemma}\label{lem: glm two norm gradient}
	Suppose Assumptions \ref{asmp: x} and \ref{asmp: glm} hold, and $n \geq C(p+\log T)$ with a sufficiently large $C$. Then when $\delta \leq C$ with some constant $C > 0$,
	\begin{equation}
		\twonorm{\nabla \fk{t}(\bbetaks{t})} \lesssim \delta,
	\end{equation}
	w.p. at least $1-\exp\{-Cn\delta^2 + C'p\}$. By taking $\delta \asymp \sqrt{\frac{p+\log T}{n}}$, this implies that:
	\begin{enumerate}[(i)]
		\item $\max_{t\in S}\twonorm{\nabla \fk{t}(\bbetaks{t})} \lesssim \sqrt{\frac{p+\log T}{n}}$, \wpp;
		\item $\twonorm{\nabla \fk{0}(\bbetaks{0})} \lesssim \sqrt{\frac{p}{n_0}}$, \wpp.
	\end{enumerate}

\end{lemma}

\begin{lemma}\label{lem: glm gradient 2 norm}
	Suppose Assumptions \ref{asmp: x} and \ref{asmp: glm} hold, and $n \geq C(p+\log T)$ with a sufficiently large $C$. Then when $t \leq C'$ with some constant $C' > 0$, for any $S \subseteq [T]$,
	\begin{equation}
		\twonorm{\{\nabla \fk{t}(\bbetaks{t})\}_{t \in S}} \lesssim \sqrt{\frac{p+|S|+\delta}{n}},
	\end{equation}
	w.p. at least $1-e^{-C'\delta}$.
\end{lemma}

\begin{lemma}[A revised version of Proposition 1 in \cite{loh2015regularized}]\label{lem: glm rsc}
	Suppose Assumptions \ref{asmp: x} and \ref{asmp: glm} hold, and $n \geq C(p+\log T)$ with a sufficiently large $C$. Then there exist constants $C_1, C_2$, such that \wppp, for all $\bbeta \in \mathbb{R}^p$, $t \in [T]$, we have
	\begin{equation}\label{eq: lem 23 eq 1}
		\fk{t}(\bbeta) - \fk{t}(\bbetaks{t}) - \nabla \fk{t}(\bbetaks{t})^\top(\bbeta-\bbetaks{t}) \geq \begin{cases}
			C_1\twonorm{\bbeta-\bbetaks{t}}^2, \text{ when } \twonorm{\bbeta-\bbetaks{t}} \leq 1;\\
			C_2\twonorm{\bbeta-\bbetaks{t}}, \text{ when } \twonorm{\bbeta-\bbetaks{t}} > 1.
		\end{cases}
	\end{equation}
\end{lemma}

\begin{lemma}\label{lem: glm rsc cor}
	Suppose Assumptions \ref{asmp: x} and \ref{asmp: glm} hold, and $n \geq C(p+\log T)$ with a sufficiently large $C$. Then there exist constants $C_1, C_2, C_3, C_4$, such that for all $\bbeta, \wtbbeta \in \mathbb{R}^p$, $t \in [T]$, \wpp,
	\begin{equation}
		\fk{t}(\bbeta) - \fk{t}(\wtbbeta) - \nabla \fk{t}(\bbetaks{t})^\top(\bbeta-\wtbbeta) \geq \begin{cases}
			C_1\twonorm{\bbeta - \wtbbeta}^2 - C_2\twonorm{\wtbbeta - \bbetaks{t}}^2, &\text{if }\twonorm{\bbeta - \bbetaks{t}} \leq 1;\\
			C_3\twonorm{\bbeta - \wtbbeta} - C_4\twonorm{\wtbbeta - \bbetaks{t}}^2, &\text{if }\twonorm{\bbeta - \bbetaks{t}} > 1.
			\end{cases}
	\end{equation}
\end{lemma}

\begin{lemma}\label{lem: glm bdded theta 1}
	Suppose Assumptions \ref{asmp: x} and \ref{asmp: glm} hold, and $n \geq C(p+\log T)$ with a sufficiently large $C$. Then \wppp, for all $t \in S$, for all $\bA \in \mO$ with $\twonorm{\bA\bA^\top - \bAks{t}(\bAks{t})^\top} \leq c(\zetak{t})^{-1}$, where $c > 0$ is a small constant, we have 
	\begin{enumerate}[(i)]
		\item $\twonorm{\bthetak{t}_{\bA}} \leq C''\zetak{t}$;
		\item $\twonorm{\bA\bthetak{t}_{\bA} - \bAks{t}\bthetaks{t}} \leq c'$ with a small constant $c' > 0$.
	\end{enumerate}
\end{lemma}

\begin{lemma}\label{lem: glm bdded theta 2}
	Suppose Assumptions \ref{asmp: x} and \ref{asmp: glm} hold, and $n \geq C(p+\log T)$ with a sufficiently large $C$. Assume there exists $\wtbA  \in \mO$ such that $\twonorm{\wtbA\bthetak{t}_{\wtbA} - \bAks{t}\bthetaks{t}} \leq c$ with a small constant $c$. Then \wppp, for all $t \in S$, for all $\bA \in \mO$ with $\twonorm{\bA\bA^\top - \wtbA(\wtbA)^\top}\leq c'(\zetak{t})^{-1}$, where $c' > 0$ is a small constant, we have 
	\begin{enumerate}[(i)]
		\item $\twonorm{\bthetak{t}_{\bA}} \leq C''\zetak{t}$;
		\item $\twonorm{\bA\bthetak{t}_{\bA} - \bAks{t}\bthetaks{t}} \leq c''$ with a small constant $c'' > 0$.
	\end{enumerate}
\end{lemma}

\begin{lemma}\label{lem: glm a theta}
	Suppose Assumptions \ref{asmp: x} and \ref{asmp: glm} hold, and $n \geq C(p+\log T)$ with a sufficiently large $C$. Then \wppp, for all $t \in S$, for all $\bA, \wtbA \in \mO$ with $\twonorm{\bA\bthetak{t}_{\bA} - \bAks{t}\bthetaks{t}}, \twonorm{\wtbA\bthetak{t}_{\wtbA}  - \bAks{t}\bthetaks{t}}\leq c'$, where $c' > 0$ is a small constant, we have
	\begin{equation}
		\twonorm{\bA\bthetak{t}_{\bA} - \wtbA\bthetak{t}_{\wtbA}} \lesssim \twonorm{\bA\bA^\top - \wtbA(\wtbA)^\top}(\twonorm{\bthetak{t}_{\wtbA}}+\twonorm{\nabla \fk{t}(\wtbA\bthetak{t}_{\wtbA})}).
	\end{equation}
\end{lemma}

\begin{lemma}\label{lem: glm a theta star}
	Suppose Assumptions \ref{asmp: x} and \ref{asmp: glm} hold, and $n \geq C(p+\log T)$ with a sufficiently large $C$. Then \wppp, for all $t \in S$, for all $\bA, \wtbA \in \mO$ with $\twonorm{\bA\bthetak{t}_{\bA} - \bAks{t}\bthetaks{t}} \leq c'$, where $c' > 0$ is a small constant, we have
	\begin{equation}
		\twonorm{\bA\bthetak{t}_{\bA} - \bAks{t}\bthetaks{t}} \lesssim \twonorm{\bA\bA^\top - \bAks{t}(\bAks{t})^\top}\zetak{t} + \sqrt{\frac{r+\log T}{n}}.
	\end{equation}
\end{lemma}

\begin{lemma}\label{lem: glm a theta not far}
	Suppose Assumptions \ref{asmp: x} and \ref{asmp: glm} hold, and $n \geq C[(p+\log T) \vee \lambda^2]$ with a sufficiently large $C$. Given $\wtbA \in \mO$, denote $(\bAk{t}, \bthetak{t}) = \argmin_{\bA \in \mO, \btheta \in \mathbb{R}^r}\{\fk{t}(\bA\btheta) + \frac{\lambda}{\sqrt{n}}\twonorm{\bA\bA^\top - \wtbA(\wtbA)^\top}\}$. Then \wpp, for any $\wtbA \in \mO$, we have $\twonorm{\bAk{t}\bthetak{t} - \bbetaks{t}} \leq c$, where $c > 0$ is a small constant.
\end{lemma}

\subsection{Proof of Theorem \ref{thm: glm mtl}}
For the penalized ERM, the proof logic is almost the same as the logic underlying the proofs of Proposition \ref{prop: mtl} and Theorem \ref{thm: mtl}. We only point out the difference and skip the details here. We need to bound $\twonorm{\hbarA(\hbarA)^\top - \barA(\barA)^\top}$ first. Then we know that $\twonorm{\hbarA(\hbarA)^\top - \barA(\barA)^\top}$ and $\twonorm{\hbarA\bthetak{t}_{\hbarA} - \bAks{t}\bthetaks{t}}$ are small w.h.p. Then a direct application of Lemma \ref{lem: glm bdded theta 2} implies that $\twonorm{\hbarA\bthetaks{t}_{\hbarA} - \bAks{t}\bthetaks{t}}$ is small w.h.p. Finally, applying Lemma \ref{lem: glm a theta}, we have $\twonorm{\hbarA\bthetaks{t}_{\hbarA} - \barA\bthetak{t}_{\barA}} \lesssim \twonorm{\hbarA(\hbarA)^\top - \barA(\barA)^\top}(\zetak{t} +\twonorm{\nabla \fk{t}(\barA\bthetak{t}_{\barA})}) \lesssim \twonorm{\hbarA(\hbarA)^\top - \barA(\barA)^\top}\zetak{t}$ w.h.p. which gives us the ideal bound for $\twonorm{\hbarA\bthetaks{t}_{\hbarA} - \barA\bthetak{t}_{\barA}}$ for all $t \in S$. 

\noindent(\rom{1}) In the proof of Proposition \ref{prop: mtl}, for $t \in \mA_1$, we know $\twonorm{\bAk{t}\bthetak{t} - \bAks{t}\bthetaks{t}} \leq $ a small constant $c$ w.h.p. by Lemma \ref{lem: glm a theta not far}. Then by Lemmas \ref{lem: glm gradient 2 norm},  \ref{lem: glm rsc cor}, and \ref{lem: glm a theta star}, \wprp,
\begin{align}
	&\frac{1}{T}\sum_{t \in \mA_1}\left[\fk{t}(\bAk{t}\bthetak{t}) - \fk{t}(\barA\bthetak{t}_{\barA})\right] \\
	&\geq -\frac{1}{T}\norma{\left\<\{\nabla \fk{t}(\bAks{t}\bthetaks{t})\}_{t \in \mA_1}, \{\bAk{t}\bthetak{t} - \hbarA\bthetak{t}_{\hbarA} + \hbarA\bthetak{t}_{\hbarA}- \barA\bthetak{t}_{\barA}\}_{t \in \mA_1}\right\>}\\
	&\quad + \frac{C_1}{T}\fnorm{\{\bAk{t}\bthetak{t} - \barA\bthetak{t}_{\barA}\}_{t \in \mA_1}}^2 -  \frac{C_2}{T}\fnorm{\{\bAks{t}\bthetaks{t} - \barA\bthetak{t}_{\barA}\}_{t \in \mA_1}}^2\\
	&\geq -\frac{C}{T}\sum_{t \in \mA_1}\sqrt{\frac{p+\log T}{n}}\cdot \zetak{t}(\twonorm{\bAk{t}(\bAk{t})^\top - \hbarA(\hbarA)^\top} + \twonorm{\hbarA(\hbarA)^\top- \barA\barA^\top})\\
	&\quad + \frac{C_1}{T}\fnorm{\{\bAk{t}\bthetak{t} - \barA\bthetak{t}_{\barA}\}_{t \in \mA_1}}^2 -C\left(h^2\cdot \frac{1}{T}\sum_{t \in \mA_1}(\zetak{t})^2 + \frac{r+\log T}{n}\right) \\
	&\geq -C\left(h^2\cdot \frac{1}{T}\sum_{t \in \mA_1}(\zetak{t})^2 + \frac{r+\log T}{n}\right)-\frac{C'}{T}\sum_{t \in \mA_1}\sqrt{\frac{p+\log T}{n}}\cdot \zetak{t}\sqrt{r}\twonorm{\bAk{t}(\bAk{t})^\top - \hbarA(\hbarA)^\top}.
\end{align}
Therefore, the calculations in \eqref{eq: prop 1 eq 1} are still correct under GLMs.

\noindent(\rom{2}) In the proof of Proposition \ref{prop: mtl}, for $t \in \mA_2$, we used the fact that \wpr,
\begin{equation}\label{eq: thm glm mtl eq 1}
	\twonorm{\bAk{t}\bthetak{t}_{\bAk{t}} - \hbarA\bthetak{t}_{\hbarA}} \lesssim \twonorm{\bAk{t}(\bAk{t})^\top - \hbarA(\hbarA)^\top}(\twonorm{\bthetak{t}_{\hbarA}}+\twonorm{\nabla \fk{t}(\hbarA\bthetak{t}_{\hbarA})}).
\end{equation}
Here, this still holds because
\begin{itemize}
	\item $\twonorm{\bAk{t}\bthetak{t}_{\bAk{t}} - \bAks{t}\bthetaks{t}} \leq $ a small constant $c$ by Lemma \ref{lem: glm a theta not far};
	\item For $t \in \mA_2$, we know that $\twonorm{\bAk{t}(\bAk{t})^\top - \hbarA(\hbarA)^\top} \leq \frac{c}{\sqrt{r}} \leq c'(\zetak{t})^{-1}$ with a small constant $c' > 0$. Then by Lemma \ref{lem: glm bdded theta 2}, $\twonorm{\bthetak{t}_{\hbarA}} \leq C\zetak{t}$ and $\twonorm{\hbarA\bthetak{t}_{\hbarA} - \bAks{t}\bthetaks{t}} \leq Cc'$ with a small constant $c' > 0$.
\end{itemize}
Then \eqref{eq: thm glm mtl eq 1} holds by Lemma \ref{lem: glm a theta}.

Finally, Lemma \ref{lem: safe net mtl} still holds for the case of GLMs. Note that in Lemma \ref{lem: safe net mtl}, (\rom{1}) only requires the maximum eigenvalue of the Hessian to be upper bounded (see Lemma E.3 in \cite{duan2023adaptive}), which is true by Assumptions \ref{asmp: x} and \ref{asmp: glm}. By Lemma E.2 in \cite{duan2023adaptive}, (\rom{2}) and (\rom{3}) require the minimum eigenvalue of Hessian to be lower bounded when $\twonorm{\bbetak{t} - \bbetaks{t}} \leq C$ with some constant $C > 0$ w.h.p. and $\gamma/\sqrt{n_0} \leq C$, both of which are true. 

For the spectral method, the proof is almost the same as the proof of Theorem \ref{thm: spectral mtl}, hence we do not repeat it here.

\subsection{Proof of Theorem \ref{thm: glm tl}}
We only outline the proof when Algorithm \ref{algo: tl} is coupled with Algorithm \ref{algo: mtl}. If Algorithm \ref{algo: tl} is coupled with Algorithm \ref{algo: spectral}, we can follow a similar argument with the proof of Theorem \ref{thm: spectral mtl}.

Denote $\eta = r\sqrt{\frac{p}{nT}} + \sqrt{r}h + \sqrt{r}\sqrt{\frac{r + \log T}{n}} + \frac{|S^c|}{T}\cdot r \cdot \frac{\lambda}{\sqrt{n}}$.

\noindent(\rom{1}) When $\eta \leq C\sqrt{\frac{p}{n_0}}$: since $\twonorm{\hbarA(\hbarA)^\top - \bAks{0}(\bAks{0})^\top} \leq $ a small constant $c$, by Lemma \ref{lem: glm a theta not far}, \wpp, $\twonorm{\hbarA\bthetak{t}_{\hbarA} - \bAks{0}\bthetaks{0}} \leq $ a small constant $c'$. Then by Lemma \ref{lem: glm a theta star} and Theorem \ref{thm: glm mtl}, \wpr, $\twonorm{\hbarA\bthetak{t}_{\hbarA} - \bAks{0}\bthetaks{0}} \lesssim \twonorm{\hbarA(\hbarA)^\top - \bAks{0}(\bAks{0})^\top} \lesssim \eta$. As we commented in the proof of Theorem \ref{thm: glm mtl}, Lemma \ref{lem: safe net tl}.(\rom{1}) still holds for GLMs because the maximum eigenvalue of the Hessian is upper bounded w.h.p. 

\noindent(\rom{2}) When $\eta > C\sqrt{\frac{p}{n_0}}$: the result comes from Lemma \ref{lem: safe net tl}.(\rom{2}), which still holds when $\gamma/\sqrt{n_0} \lesssim $ some constant $C$ because of Lemma E.2 in \cite{duan2023adaptive}.

\subsection{Proofs of Lemmas}

\subsubsection{Proof of Lemma \ref{lem: glm two norm gradient}}
First, note that $\nabla \fk{t}(\bbetaks{t}) = \frac{2}{n}\sum_{i=1}^n \bxk{t}_i[\yk{t}_i-\psi'((\bxk{t}_i)^\top\bbetaks{t})]$. Consider a $1/2$-cover of the unit ball in $\mathbb{R}^p$ w.r.t Euclidean norm $\{\bu_j\}_{j=1}^N$ with $N \leq 5^p$ (Example 5.8 in \cite{wainwright2019high}). By a standard argument (see the proof of Theorem 6.5 in \cite{wainwright2019high}), we have
\begin{equation}
	\twonorm{\nabla \fk{t}(\bbetaks{t})} \leq 2\max_{j\in [N]}\left\{\frac{2}{n}\sum_{i=1}^n \bu_j^\top\bxk{t}_i[\yk{t}_i-\psi'((\bxk{t}_i)^\top\bbetaks{t})]\right\}.
\end{equation}
By union bound,
\begin{equation}
	\tE \exp\{\lambda\twonorm{\nabla \fk{t}(\bbetaks{t})}\} \leq N \max_{j\in [N]}\tE \exp\left\{\frac{4\lambda}{n}\sum_{i=1}^n \bu_j^\top\bxk{t}_i[\yk{t}_i-\psi'((\bxk{t}_i)^\top\bbetaks{t})]\right\} = (*).
\end{equation}
Note that by the GLM density function, for any $a \in \mathbb{R}$, 
\begin{align}
	\tE \exp\left\{a[\yk{t}_i-\psi'((\bxk{t}_i)^\top\bbetaks{t})]\right\} &\leq \int \rho(y)\exp\left\{[a+(\bxk{t}_i)^\top\bbetaks{t}]\yk{t}_i- \psi(a+(\bxk{t}_i)^\top\bbetaks{t})\right\}\dint \mu (y) \\
	&\quad \cdot \exp\left\{\psi(a+(\bxk{t}_i)^\top\bbetaks{t}) - \psi((\bxk{t}_i)^\top\bbetaks{t}) -a\psi'((\bxk{t}_i)^\top\bbetaks{t})\right\} \\
	&\leq \exp\left\{\frac{1}{2}a^2\psi''\big(\delta[a+(\bxk{t}_i)^\top\bbetaks{t}] + (1-\delta)(\bxk{t}_i)^\top\bbetaks{t}\big)\right\} \\
	&\leq \exp\left\{Ca^2\right\}, \label{eq: lemma grad eq 1}
\end{align}
where $\delta \in [0, 1]$. Therefore, for any positive $\lambda \leq Cn$ with some $C > 0$,
\begin{equation}
	(*) \leq N\max_{j\in [N]}\tE \exp\left\{C'\frac{\lambda^2}{n^2}\sum_{i=1}^n (\bu_j^\top\bxk{t}_i)^2\right\}  \leq \exp\left\{\frac{C''\lambda^2}{n} + C'''p\right\},
\end{equation}
where the second inequality comes from the property of sub-exponential variables (see property 3 in Proposition 2.7.1 of \cite{vershynin2018high}). Finally, for any $\delta > 0$, by Chernoff's bound,
\begin{equation}
	\tP(\twonorm{\nabla \fk{t}(\bbetaks{t})} > \delta) \leq e^{-\lambda t}\tE \exp\{\lambda \twonorm{\nabla \fk{t}(\bbetaks{t})}\}  \leq \exp\left\{\frac{C''\lambda^2}{n} + C'''p - \lambda \delta\right\}.
\end{equation}
Let $\lambda = C''n\delta/2$ ($\delta <$ some constant $C$), leading to 
\begin{equation}
	\tP(\twonorm{\nabla \fk{t}(\bbetaks{t})} > \delta) \leq \exp\left\{-Cn\delta^2 + C'p\right\},
\end{equation}
for all $t \in S$. 

\subsubsection{Proof of Lemma \ref{lem: glm gradient 2 norm}}
Note that $\nabla \fk{t}(\bbetaks{t}) = \frac{2}{n}\sum_{i=1}^n \bxk{t}_i[\yk{t}_i-\psi'((\bxk{t}_i)^\top\bbetaks{t})] \coloneqq \bm{g}^{(t)} \in \mathbb{R}^p$. Denote $\{\bm{g}^{(t)}\}_{t \in S}$ as $\bm{G} \in \mathbb{R}^{p \times |S|}$. Consider two $1/4$-covers of the unit ball in $\mathbb{R}^p$ and $\mathbb{R}^{|S|}$ w.r.t Euclidean norm, as $\{\bu^{(j_1)}\}_{j_1=1}^{N_1}$ and $\{\bv^{(j_2)}\}_{j_2=1}^{N_2}$ respectively, with $N_1 \leq 9^p$ and $N_2 \leq 9^{|S|}$ (Example 5.8 in \cite{wainwright2019high}). By a standard argument (see the proof of Theorem 6.5 in \cite{wainwright2019high})
, we have
\begin{equation}
	\twonorm{\bm{G}} \lesssim \max_{\substack{j_1 \in [N_1] \\ j_2 \in [N_2]}}\norm{(\bu^{(j_1)})^\top\bm{G}\bv^{(j_2)}} = \max_{\substack{j_1 \in [N_1] \\ j_2 \in [N_2]}}\norma{\sum_{t \in S}(\bu^{(j_1)})^\top\bm{g}^{(t)}v^{j_2}_k}.
\end{equation}
It follows that for any $\lambda \in \mathbb{R}$,
\begin{align}
	\tE e^{\lambda \twonorm{\bm{G}}} &\lesssim \sum_{j_1=1}^{N_1}\sum_{j_2=1}^{N_2}\prod_{t=1}^{|S|}\tE\exp\{\lambda\norm{(\bu^{(j_1)})^\top\bm{g}^{(t)}v^{j_2}_k}\} \\
	&\leq \sum_{j_1=1}^{N_1}\sum_{j_2=1}^{N_2}\prod_{t=1}^{|S|}\big[\tE\exp\{\lambda(\bu^{(j_1)})^\top\bm{g}^{(t)}v^{j_2}_k\} + \tE\exp\{-\lambda(\bu^{(j_1)})^\top\bm{g}^{(t)}v^{j_2}_k\} \big].
\end{align}
Similar to the calculations in \eqref{eq: lemma grad eq 1}, we can obtain that
\begin{equation}
	\tE\exp\{\lambda(\bu^{(j_1)})^\top\bm{g}^{(t)}v^{(j_2)}_k\}, \tE\exp\{-\lambda(\bu^{(j_1)})^\top\bm{g}^{(t)}v^{(j_2)}_k\} \lesssim \exp\left\{\frac{C}{n}\lambda^2\cdot |v^{(j_2)}_k|^2\right\},
\end{equation}
hence
\begin{equation}
		\tE e^{\lambda \twonorm{\bm{G}}} \lesssim \sum_{j_1=1}^{N_1}\sum_{j_2=1}^{N_2}2^{|S|}\cdot \exp\left\{\frac{C}{n}\lambda^2\right\} \lesssim \exp\left\{\frac{C}{n}\lambda^2 + C'p + C'|S|\right\}.
\end{equation}
Then similar to the proof of Lemma \ref{lem: glm two norm gradient}, the proof can be finished by Chernoff's bound.

\subsubsection{Proof of Lemma \ref{lem: glm rsc}}
The proof of part (\rom{1}) follows the idea in the proof of Theorem 9.36 in \cite{wainwright2019high}, and the proof of part (\rom{2}) is the same as the proof of equation (39b) in Proposition 1 of \cite{loh2015regularized}.

(\rom{1}) If $\twonorm{\bbeta-\bbetaks{t}} \leq 1$: Denote $\bDelta = \bbeta-\bbetaks{t}$, $\varphi_{\tau}(u) = u^2\mathds{1}(\norm{u} \leq \tau)$ with any $\tau > 0$. By Taylor expansion, with some $\delta \in [0, 1]$ and any constant $T > 0$,
\begin{align}
	\text{LHS of } \eqref{eq: lem 23 eq 1} &= \frac{1}{n}\sum_{i=1}^n\psi''((\bbetaks{t})^\top\bxk{t}_i + \delta\bDelta^\top\bxk{t}_i)\cdot (\bDelta^\top\bxk{t}_i)^2 \\
	&\geq \frac{1}{n}\sum_{i=1}^n\psi''((\bbetaks{t})^\top\bxk{t}_i + \delta\bDelta^\top\bxk{t}_i)\varphi_{\tau}(\bDelta^\top\bxk{t}_i)\mathds{1}(\norm{(\bbetaks{t})^\top\bxk{t}_i} \leq T)\\
	&\geq \min_{\norm{u} \leq \tau + T}\psi''(u) \cdot \underbrace{\frac{1}{n}\sum_{i=1}^n\varphi_{\tau}(\bDelta^\top\bxk{t}_i)\mathds{1}(\norm{(\bbetaks{t})^\top\bxk{t}_i} \leq T)}_{(*)}
\end{align}
It suffices to show that by fixing some $\tau, T > 0$, $(*) \geq C\twonorm{\bDelta}^2$ w.p. at least $1-e^{-C'n}$. In fact, we only need to consider the case $\twonorm{\bDelta} = 1$. If the results hold when $\twonorm{\bDelta} = 1$, then when $\twonorm{\bDelta} \in (0,1)$, since $\varphi_{\tau}$ is non-decreasing on $\tau$, plugging in $\bDelta/\twonorm{\bDelta}$ implies $(*) \geq C\twonorm{\bDelta}^2$. So in the following analysis, we consider $\twonorm{\bDelta} = 1$.

Define
\begin{equation}
	Z_n = \sup_{\twonorm{\bDelta} = 1} \left\{\frac{1}{n}\sum_{i=1}^n\varphi_{\tau}(\bDelta^\top\bxk{t}_i)\mathds{1}(\norm{(\bbetaks{t})^\top\bxk{t}_i} \leq T) - \tE [\varphi_{\tau}(\bDelta^\top\bxk{t}_i)\mathds{1}(\norm{(\bbetaks{t})^\top\bxk{t}_i} \leq T)]\right\}.
\end{equation}
By bounded difference inequality (Corollary 2.21 in \cite{wainwright2019high}) or functional Hoeffding inequality (Theorem 3.26 in \cite{wainwright2019high}),
\begin{equation}\label{eq: lemma 23 eq 1}
	\tP(Z_n \geq \tE Z_n + \delta) \leq \exp\{-Cn\delta^2/\tau^4\}.
\end{equation}
Denote $\{\epsilon_i\}_{i=1}^n$ as independent Rademacher variables. By symmetrization,
\begin{align}
	\tE Z_n &\lesssim \tE_{\bxk{t}, \epsilon}\left[\sup_{\twonorm{\bDelta} = 1}\norma{\frac{1}{n}\sum_{i=1}^n\varphi_{\tau}(\bDelta^\top\bxk{t}_i)\mathds{1}(\norm{(\bbetaks{t})^\top\bxk{t}_i} \leq T)\cdot \epsilon_i}\right] \\
	&\lesssim \tau \tE_{\bxk{t}, \epsilon}\left[\sup_{\twonorm{\bDelta} = 1}\norma{\frac{1}{n}\sum_{i=1}^n (\bDelta^\top\bxk{t}_i)\epsilon_i}\right] \label{eq: lemma 23 eq 4}\\
	&\lesssim \tau \cdot \frac{1}{n}\tE_{\bxk{t}, \epsilon} \sqrt{\sup_{\twonorm{\bDelta} = 1} \sum_{i=1}^n (\bDelta^\top\bxk{t}_i)^2\epsilon_i^2} \\
	&\lesssim \tau \cdot \frac{1}{n} \sqrt{n\tE\twonorm{\bxk{t}_i}^2}\label{eq: lemma 23 eq 5}  \\
	&\lesssim \tau \sqrt{\frac{p}{n}}, \label{eq: lemma 23 eq 2}
\end{align}
where \eqref{eq: lemma 23 eq 4} and \eqref{eq: lemma 23 eq 5} are due to Rademacher contraction inequality (equation (5.61) in \cite{wainwright2019high}, because the function $\varphi_{\tau}(\cdot)\mathds{1}(\norm{(\bbetaks{t})^\top\bxk{t}_i} \leq T)$ is $2\tau$-Lipschitz) and Jensen's inequality. On the other hand,
\begin{align}
	&\sup_{\twonorm{\bDelta} = 1} \tE [\varphi_{\tau}(\bDelta^\top\bxk{t}_i)\mathds{1}(\norm{(\bbetaks{t})^\top\bxk{t}_i}] \leq T)] \\
	&\geq \sup_{\twonorm{\bDelta} = 1}\left\{\tE [(\bDelta^\top\bxk{t}_i)^2\cdot \mathds{1}(\norm{(\bbetaks{t})^\top\bxk{t}_i}] \leq T)] - \tE [(\bDelta^\top\bxk{t}_i)^2\cdot \mathds{1}(\norm{\bDelta^\top\bxk{t}_i} > \tau)\cdot \mathds{1}(\norm{(\bbetaks{t})^\top\bxk{t}_i} \leq T)] \right\} \\
	&\geq \sup_{\twonorm{\bDelta} = 1}\tE (\bDelta^\top\bxk{t}_i)^2 - \tP(\norm{(\bbetaks{t})^\top\bxk{t}_i} > T) - \sup_{\twonorm{\bDelta} = 1}\sqrt{\tE (\bDelta^\top\bxk{t}_i)^4}\cdot \sqrt{\tP(\norm{\bDelta^\top\bxk{t}_i} > \tau)} \\
	&\geq C - \exp\{-C'T^2/(\zetak{t})^2\} - e^{-C''\tau^2} \\
	&\geq C/2, \label{eq: lemma 23 eq 3}
\end{align}
when $T$ and $\tau$ are large.

Putting \eqref{eq: lemma 23 eq 1}, \eqref{eq: lemma 23 eq 2}, and \eqref{eq: lemma 23 eq 3} together, and setting $\delta = $ a small constant $c$, w.p. at least $1-e^{-C''n}$, we have 
\begin{align}
	(*) &\geq \sup_{\twonorm{\bDelta} = 1} \tE [\varphi_{\tau}(\bDelta^\top\bxk{t}_i)\mathds{1}(\norm{(\bbetaks{t})^\top\bxk{t}_i}] \leq T)] - Z_n  \\
	&\geq C/2 - C'\sqrt{\frac{p}{n}} - \delta \\
	&\geq C/4,
\end{align}
when $n \gtrsim p+\log T$.

(\rom{2}) If $\twonorm{\bbeta-\bbetaks{t}} > 1$: Denote $\bDelta = \bbeta-\bbetaks{t}$, $\delta = \frac{1}{\twonorm{\bDelta}}$. By convexity of $\fk{t}$,
\begin{equation}
	\fk{t}(\delta\bbeta + (1-\delta)\bbetaks{t}) \leq \delta\fk{t}(\bbeta) + (1-\delta)\fk{t}(\bbetaks{t}),
\end{equation}
which combining with part (\rom{1}) implies that w.p. at least $1-e^{-C''n}$,
\begin{align}
	\fk{t}(\bbeta) - \fk{t}(\bbetaks{t}) - \nabla \fk{t}(\bbetaks{t})^\top(\bbeta - \bbetaks{t}) &\geq \frac{\fk{t}(\bbetaks{t}+\delta\bDelta) - \fk{t}(\bbetaks{t}) - \delta\nabla \fk{t}(\bbetaks{t})^\top\bDelta}{\delta} \\
	&\geq \delta C\twonorm{\bDelta}^2 \\
	&= C\twonorm{\bDelta},
\end{align}
because $\twonorm{\delta\bDelta} = 1$. 

\subsubsection{Proof of Lemma \ref{lem: glm rsc cor}}
It is easy to see that
\begin{align}
	\fk{t}(\bbeta) - \fk{t}(\wtbbeta) - \nabla \fk{t}(\bbetaks{t})^\top(\bbeta - \wtbbeta) &= \underbrace{\fk{t}(\bbeta) - \fk{t}(\bbetaks{t}) - \nabla \fk{t}(\bbetaks{t})^\top(\bbeta - \bbetaks{t})}_{[1]} \\
	&\quad + \underbrace{\fk{t}(\bbetaks{t}) - \fk{t}(\wtbbeta) + \nabla \fk{t}(\bbetaks{t})^\top(\wtbbeta - \bbetaks{t})}_{[2]}.
\end{align}
By Lemma \ref{lem: glm rsc}, w.p. at least $1-e^{-C''n}$,
\begin{equation}
	[1] \geq C\twonorm{\bbeta - \bbetaks{t}}^2.
\end{equation} 
And with some $\delta, \delta'\in [0, 1]$, by Assumption \ref{asmp: glm}, \wpp,
\begin{align}
	[2] &\leq \nabla \fk{t}(\delta\bbetaks{t} + (1-\delta)\wtbbeta)^\top(\bbetaks{t}-\wtbbeta) + \nabla \fk{t}(\bbetaks{t})^\top(\wtbbeta - \bbetaks{t}) \\
	&= (\wtbbeta - \bbetaks{t})^\top\nabla^2 \fk{t}(\delta'[\delta\bbetaks{t} + (1-\delta)\wtbbeta] + (1-\delta')\bbetaks{t})^\top(\wtbbeta - \bbetaks{t}) \\
	&\lesssim \frac{1}{n}\sum_{i=1}^n [(\wtbbeta - \bbetaks{t})^\top\bxk{t}_i]^2 \\
	&\lesssim \twonorm{\wtbbeta - \bbetaks{t}}^2,
\end{align}
where the last inequality comes from the fact that $\twonorm{\hSigmak{t}} \leq C$ \wppp (by Lemma \ref{lem: cov hat}). Putting all pieces together leads to the desired conclusion.

\subsubsection{Proof of Lemma \ref{lem: glm bdded theta 1}}
Define $\widetilde{\btheta}^{(t)}_{\bA} = \bA^\top\bAks{t}\bthetaks{t}$. It is easy to see that
\begin{equation}
	\twonorm{\bA\widetilde{\btheta}^{(t)}_{\bA} - \bAks{t}\bthetaks{t}} \leq \twonorm{(\bA\bA^\top - \bm{I})\bAks{t}\bthetaks{t}} \lesssim \twonorm{(\bA^{\perp})^\top\bAks{t}}\zetak{t} \leq c.
\end{equation}
If $\twonorm{\bA\bthetak{t}_{\bA}-\bAks{t}\bthetaks{t}} > 1$, then by Lemma \ref{lem: glm rsc cor}, \wpp,
\begin{align}
	\fk{t}(\bA\bthetak{t}_{\bA}) - \fk{t}(\bA\widetilde{\btheta}^{(t)}_{\bA}) &\geq -\twonorm{\nabla \fk{t}(\bAks{t}\bthetaks{t})}\cdot \twonorm{\bA\bthetak{t}_{\bA}-\bA\widetilde{\btheta}^{(t)}_{\bA}} + C\twonorm{\bA\bthetak{t}_{\bA}-\bAks{t}\bthetaks{t}} \\
	&\quad - C'\twonorm{\bA\widetilde{\btheta}^{(t)}_{\bA} - \bAks{t}\bthetaks{t}}^2 \\
	&\geq (C-\twonorm{\nabla \fk{t}(\bAks{t}\bthetaks{t})})\cdot \twonorm{\bA\bthetak{t}_{\bA}-\bA\widetilde{\btheta}^{(t)}_{\bA}}\\
	&\quad - C \twonorm{\bA\widetilde{\btheta}^{(t)}_{\bA} - \bAks{t}\bthetaks{t}}-C'\twonorm{\bA\widetilde{\btheta}^{(t)}_{\bA} - \bAks{t}\bthetaks{t}}^2 \\
	&\geq \frac{1}{2}C\twonorm{\bA\bthetak{t}_{\bA}-\bA\widetilde{\btheta}^{(t)}_{\bA}} - Cc - C'c^2 \\
	&\geq \frac{1}{2}C\twonorm{\bA\bthetak{t}_{\bA}-\bAks{t}\bthetaks{t}} - \frac{1}{2}Cc - Cc - C'c^2 \\
	&>0,
\end{align}
which is contradicted with the definition of $\bthetak{t}_{\bA}$. Therefore, we must have $\twonorm{\bA\bthetak{t}_{\bA}-\bAks{t}\bthetaks{t}} \leq 1$. By Lemma \ref{lem: glm rsc cor}, \wpp,
\begin{align}
	\fk{t}(\bA\bthetak{t}_{\bA}) - \fk{t}(\bA\widetilde{\btheta}^{(t)}_{\bA}) &\geq -\twonorm{\nabla \fk{t}(\bAks{t}\bthetaks{t})}\cdot \twonorm{\bA\bthetak{t}_{\bA}-\bA\widetilde{\btheta}^{(t)}_{\bA}} + C\twonorm{\bA\bthetak{t}_{\bA}-\bA\widetilde{\btheta}^{(t)}_{\bA}}^2 \\
	&\quad - C'\twonorm{\bA\widetilde{\btheta}^{(t)}_{\bA} - \bAks{t}\bthetaks{t}}^2 \\
	&\geq -C''\sqrt{\frac{p+\log T}{n}} + C\twonorm{\bA\bthetak{t}_{\bA}-\bA\widetilde{\btheta}^{(t)}_{\bA}}^2 - C'c^2.
\end{align}
We know that $\fk{t}(\bA\bthetak{t}_{\bA}) - \fk{t}(\bA\widetilde{\btheta}^{(t)}_{\bA}) \geq 0$ by optimality, so we must have
\begin{equation}
	-C''\sqrt{\frac{p+\log T}{n}} + C\twonorm{\bA\bthetak{t}_{\bA}-\bA\widetilde{\btheta}^{(t)}_{\bA}}^2 - C'c^2 \leq 0,
\end{equation}
i.e., $\twonorm{\bA\bthetak{t}_{\bA}-\bA\widetilde{\btheta}^{(t)}_{\bA}} \leq \sqrt{\frac{C'}{C}}c + (C'')^{1/2}(\frac{p+\log T}{n})^{1/4} \leq $ a small constant $c'$, \wpp. This implies that $\twonorm{\bthetak{t}_{\bA}} = \twonorm{\bA\bthetak{t}_{\bA}} \leq \twonorm{\bA\bthetak{t}_{\bA}-\bA\widetilde{\btheta}^{(t)}_{\bA}} + \twonorm{\bA\widetilde{\btheta}^{(t)}_{\bA} - \bAks{t}\bthetaks{t}} + \twonorm{\bAks{t}\bthetaks{t}} \leq C\zetak{t}$, \wpp, which completes the proof.

\subsubsection{Proof of Lemma \ref{lem: glm bdded theta 2}}
Define $\widetilde{\btheta}^{(t)}_{\bA} = \bA^\top\wtbA\bthetak{t}_{\wtbA}$. It is easy to see that
\begin{equation}
	\twonorm{\bA\widetilde{\btheta}^{(t)}_{\bA} - \wtbA\bthetak{t}_{\wtbA}} \leq \twonorm{(\bA\bA^\top - \bm{I})\wtbA\bthetak{t}_{\wtbA}} \lesssim \twonorm{(\bA^{\perp})^\top\wtbA}\zetak{t} \leq c.
\end{equation}
If $\twonorm{\bA\bthetak{t}_{\bA}-\bAks{t}\bthetaks{t}} > 1$, then by Lemma \ref{lem: glm rsc cor}, \wpp,
\begin{align}
	\fk{t}(\bA\bthetak{t}_{\bA}) - \fk{t}(\bA\widetilde{\btheta}^{(t)}_{\bA}) &\geq -\twonorm{\nabla \fk{t}(\bAks{t}\bthetaks{t})}\cdot \twonorm{\bA\bthetak{t}_{\bA}-\bA\widetilde{\btheta}^{(t)}_{\bA}} + C\twonorm{\bA\bthetak{t}_{\bA}-\bAks{t}\bthetaks{t}} \\
	&\quad - C'\twonorm{\bA\widetilde{\btheta}^{(t)}_{\bA} - \bAks{t}\bthetaks{t}}^2 \\
	&\geq (C-\twonorm{\nabla \fk{t}(\bAks{t}\bthetaks{t})})\cdot \twonorm{\bA\bthetak{t}_{\bA}-\bA\widetilde{\btheta}^{(t)}_{\bA}}\\
	&\quad - C \twonorm{\bA\widetilde{\btheta}^{(t)}_{\bA} - \bAks{t}\bthetaks{t}}-C'\twonorm{\bA\widetilde{\btheta}^{(t)}_{\bA} - \bAks{t}\bthetaks{t}}^2 \\
	&\geq \frac{1}{2}C\twonorm{\bA\bthetak{t}_{\bA}-\bA\widetilde{\btheta}^{(t)}_{\bA}} - C\twonorm{\bA\widetilde{\btheta}^{(t)}_{\bA} - \wtbA\bthetak{t}_{\wtbA}}- C \twonorm{\wtbA\bthetak{t}_{\wtbA} - \bAks{t}\bthetaks{t}}\\
	&\quad -C'\twonorm{\bA\widetilde{\btheta}^{(t)}_{\bA} - \wtbA\bthetak{t}_{\wtbA}}^2 -C'\twonorm{\wtbA\bthetak{t}_{\wtbA} - \bAks{t}\bthetaks{t}}^2 \\
	&\geq \frac{1}{2}C\twonorm{\bA\bthetak{t}_{\bA}-\bA\widetilde{\btheta}^{(t)}_{\bA}} - Cc - C'c^2 \\
	&\geq \frac{1}{2}C\twonorm{\bA\bthetak{t}_{\bA}-\bAks{t}\bthetaks{t}} - \frac{1}{2}Cc - Cc - C'c^2 \\
	&>0,
\end{align}
which is contradicted with the definition of $\bthetak{t}_{\bA}$. Therefore, we must have $\twonorm{\bA\bthetak{t}_{\bA}-\bAks{t}\bthetaks{t}} \leq 1$. By Lemma \ref{lem: glm rsc cor}, \wpp,
\begin{align}
	\fk{t}(\bA\bthetak{t}_{\bA}) - \fk{t}(\bA\widetilde{\btheta}^{(t)}_{\bA}) &\geq -\twonorm{\nabla \fk{t}(\bAks{t}\bthetaks{t})}\cdot \twonorm{\bA\bthetak{t}_{\bA}-\bA\widetilde{\btheta}^{(t)}_{\bA}} + C\twonorm{\bA\bthetak{t}_{\bA}-\bA\widetilde{\btheta}^{(t)}_{\bA}}^2 \\
	&\quad - C'\twonorm{\bA\widetilde{\btheta}^{(t)}_{\bA} - \bAks{t}\bthetaks{t}}^2 \\
	&\geq -\twonorm{\nabla \fk{t}(\bAks{t}\bthetaks{t})}\cdot \twonorm{\bA\bthetak{t}_{\bA}-\bA\widetilde{\btheta}^{(t)}_{\bA}} + C\twonorm{\bA\bthetak{t}_{\bA}-\bA\widetilde{\btheta}^{(t)}_{\bA}}^2 \\
	&\quad - \frac{1}{2}C\twonorm{\bA\widetilde{\btheta}^{(t)}_{\bA} - \wtbA\bthetak{t}_{\wtbA}}^2 - C''\twonorm{\wtbA\bthetak{t}_{\wtbA} - \bAks{t}\bthetaks{t}}^2 \\
	&\geq -C''\sqrt{\frac{p+\log T}{n}} + \frac{1}{2}C\twonorm{\bA\bthetak{t}_{\bA}-\bA\widetilde{\btheta}^{(t)}_{\bA}}^2 - 4Cc^2.
\end{align}
We know that $\fk{t}(\bA\bthetak{t}_{\bA}) - \fk{t}(\bA\widetilde{\btheta}^{(t)}_{\bA}) \geq 0$ by optimality, so we must have
\begin{equation}
	-C''\sqrt{\frac{p+\log T}{n}} + C\twonorm{\bA\bthetak{t}_{\bA}-\bA\widetilde{\btheta}^{(t)}_{\bA}}^2 - C'c^2 \leq 0,
\end{equation}
i.e., $\twonorm{\bA\bthetak{t}_{\bA}-\bA\widetilde{\btheta}^{(t)}_{\bA}} \leq \sqrt{\frac{C'}{C}}c + (C'')^{1/2}(\frac{p+\log T}{n})^{1/4} \leq $ a small constant $c'$, \wpp. This implies that $\twonorm{\bthetak{t}_{\bA}} = \twonorm{\bA\bthetak{t}_{\bA}} \leq \twonorm{\bA\bthetak{t}_{\bA}-\bA\widetilde{\btheta}^{(t)}_{\bA}} + \twonorm{\bA\widetilde{\btheta}^{(t)}_{\bA} - \bAks{t}\bthetaks{t}} + \twonorm{\bAks{t}\bthetaks{t}} \leq C\zetak{t}$, \wpp, which completes the proof.

\subsubsection{Proof of Lemma \ref{lem: glm a theta}}
Similar to the proof of Lemma \ref{lem: a theta a extention}, by the optimality of $\bthetak{t}_{\bA}$,
\begin{equation}
	\bA^\top(\nabla \fk{t}(\wtbA\bthetak{t}_{\wtbA}) + \widetilde{\bSigma}^{(t)}(\bA\bthetak{t}_{\bA} - \wtbA\bthetak{t}_{\wtbA})) = \bm{0},
\end{equation}
where $\widetilde{\bSigma}^{(t)} = \frac{1}{n}\sum_{i=1}^n \bxk{t}_i(\bxk{t}_i)^\top \psi''\big((\bxk{t}_i)^\top\wtbA\bthetak{t}_{\wtbA} + \delta(\bxk{t}_i)^\top(\bA\bthetak{t}_{\bA} - \wtbA\bthetak{t}_{\wtbA})\big)$, $\delta \in [0, 1]$. With similar arguments used in the proof of Lemma \ref{lem: glm rsc}, it can be shown that $\lambdamin(\widetilde{\bSigma}^{(t)}) > 0$ \wpp. The remaining steps are the same as the steps in the proof of Lemma \ref{lem: a theta a extention}.

\subsubsection{Proof of Lemma \ref{lem: glm a theta star}}
The proof is almost the same as the proof of Lemma \ref{lem: glm a theta}, so we omit it here.

\subsubsection{Proof of Lemma \ref{lem: glm a theta not far}}
If $\twonorm{\bAk{t}\bthetak{t} - \bAks{t}\bthetaks{t}} > 1$, by Lemmas \ref{lem: glm two norm gradient} and \ref{lem: glm rsc}, \wppp, for any $\wtbA \in \mO$,
\begin{align}
	&\fk{t}(\bAk{t}\bthetak{t}) + \frac{\lambda}{\sqrt{n}}\twonorm{\bAk{t}(\bAk{t})^\top - \wtbA(\wtbA)^\top} - \fk{t}(\bAks{t}\bthetaks{t}) - \frac{\lambda}{\sqrt{n}}\twonorm{\bAks{t}(\bAks{t})^\top - \wtbA(\wtbA)^\top} \\
	&\geq \nabla \fk{t}(\bAks{t}\bthetaks{t})^\top(\bAk{t}\bthetak{t} - \bAks{t}\bthetaks{t}) - 2\frac{\lambda}{\sqrt{n}} + C\twonorm{\bAk{t}\bthetak{t} - \bAks{t}\bthetaks{t}} \\
	&\geq \left(C - C'\sqrt{\frac{p+\log T}{n}}\right)\twonorm{\bAk{t}\bthetak{t} - \bAks{t}\bthetaks{t}} - 2\frac{\lambda}{\sqrt{n}}\\
	&\geq \frac{1}{2}C - \frac{1}{4}C \\
	&>0, 
\end{align}
which contradicts to the definitions of $\bAk{t}$ and $\bthetak{t}$. Hence $\twonorm{\bAk{t}\bthetak{t} - \bAks{t}\bthetaks{t}} \leq 1$, \wppp. Then by Lemmas \ref{lem: glm two norm gradient} and \ref{lem: glm rsc} again, \wppp, for any $\wtbA \in \mO$,
\begin{align}
	&\fk{t}(\bAk{t}\bthetak{t}) + \frac{\lambda}{\sqrt{n}}\twonorm{\bAk{t}(\bAk{t})^\top - \wtbA(\wtbA)^\top} - \fk{t}(\bAks{t}\bthetaks{t}) - \frac{\lambda}{\sqrt{n}}\twonorm{\bAks{t}(\bAks{t})^\top - \wtbA(\wtbA)^\top} \\
	&\geq \nabla \fk{t}(\bAks{t}\bthetaks{t})^\top(\bAk{t}\bthetak{t} - \bAks{t}\bthetaks{t}) - 2\frac{\lambda}{\sqrt{n}} + C\twonorm{\bAk{t}\bthetak{t} - \bAks{t}\bthetaks{t}}^2 \\
	&\geq C\twonorm{\bAk{t}\bthetak{t} - \bAks{t}\bthetaks{t}}^2 - C'\sqrt{\frac{p+\log T}{n}}\twonorm{\bAk{t}\bthetak{t} - \bAks{t}\bthetaks{t}}- 2\frac{\lambda}{\sqrt{n}}.
\end{align}
By the optimality of $\bAk{t}\bthetak{t}$, we must have the RHS of above $\leq 0$ \wppp, which entails that $\twonorm{\bAk{t}\bthetak{t} - \bAks{t}\bthetaks{t}} \lesssim \sqrt{\frac{p+\log T}{n}} + \lambda^{1/2}n^{-1/4} \leq $ a small constant $c$, which completes the proof.

\section{Proofs for Non-linear Regression Models}

\subsection{Lemmas}

\begin{lemma}\label{lem: non-linear rsc}
	Suppose Assumptions \ref{asmp: x}, \ref{asmp: non-linear}, and \ref{asmp: non-linear n} hold. Then \wpp, for any $\bbeta$, $\wtbbeta \in \mathbb{R}^p$, for any $t \in S$,
	\begin{equation}
		\fk{t}(\bbeta) - \fk{t}(\wtbbeta) - \nabla \fk{t}(\bbetaks{t})^\top(\bbeta - \wtbbeta) \geq C'\twonorm{\bbeta - \wtbbeta}^2  -C''\twonorm{\wtbbeta - \bbetaks{t}}^2.
	\end{equation}
\end{lemma}

\begin{lemma}\label{lem: non-linear two norm gradient}
	Suppose Assumptions \ref{asmp: x}, \ref{asmp: non-linear}, and \ref{asmp: non-linear n} hold. Then when $\delta \leq C$ with some constant $C > 0$,
	\begin{equation}
		\twonorm{\nabla \fk{t}(\bbetaks{t})} \lesssim \delta,
	\end{equation}
	w.p. at least $1-\exp\{-Cn\delta^2 + C'p\}$. By taking $t \asymp \sqrt{\frac{p+\log T}{n}}$, this implies that:
	\begin{enumerate}[(i)]
		\item $\max_{t\in S}\twonorm{\nabla \fk{t}(\bbetaks{t})} \lesssim \sqrt{\frac{p+\log T}{n}}$, \wpp;
		\item $\twonorm{\nabla \fk{0}(\bbetaks{0})} \lesssim \sqrt{\frac{p}{n_0}}$, \wpp.
	\end{enumerate}

\end{lemma}

\begin{lemma}\label{lem: non-linear gradient 2 norm}
	Suppose Assumptions \ref{asmp: x}, \ref{asmp: non-linear}, and \ref{asmp: non-linear n} hold. Then when $t \leq C$ with some constant $C > 0$, for any $S \subseteq [T]$,
	\begin{equation}
		\twonorm{\{\nabla \fk{t}(\bbetaks{t})\}_{t \in S}} \lesssim \sqrt{\frac{p+|S|+\delta}{n}},
	\end{equation}
	w.p. at least $1-e^{-C'\delta}$.
\end{lemma}

\begin{lemma}\label{lem: non-linear a theta}
	Suppose Assumptions \ref{asmp: x}, \ref{asmp: non-linear}, and \ref{asmp: non-linear n} hold. Then \wppp, for all $t \in S$, for all $\bA, \wtbA \in \mO$ with $\twonorm{\bA\bthetak{t}_{\bA} - \bAks{t}\bthetaks{t}}, \twonorm{\wtbA\bthetak{t}_{\wtbA}  - \bAks{t}\bthetaks{t}}\leq c'$, where $c' > 0$ is a small constant, we have
	\begin{equation}
		\twonorm{\bA\bthetak{t}_{\bA} - \wtbA\bthetak{t}_{\wtbA}} \lesssim \twonorm{\bA\bA^\top - \wtbA(\wtbA)^\top}(\twonorm{\bthetak{t}_{\wtbA}}+\twonorm{\nabla \fk{t}(\wtbA\bthetak{t}_{\wtbA})}).
	\end{equation}
\end{lemma}

\begin{lemma}\label{lem: non-linear a theta star}
	Suppose Assumptions \ref{asmp: x}, \ref{asmp: non-linear}, and \ref{asmp: non-linear n} hold. Then \wppp, for all $t \in S$, for all $\bA, \wtbA \in \mO$ with $\twonorm{\bA\bthetak{t}_{\bA} - \bAks{t}\bthetaks{t}} \leq c'$, where $c' > 0$ is a small constant, we have
	\begin{equation}
		\twonorm{\bA\bthetak{t}_{\bA} - \bAks{t}\bthetaks{t}} \lesssim \twonorm{\bA\bA^\top - \bAks{t}(\bAks{t})^\top}\zetak{t} + \sqrt{\frac{r+\log T}{n}}.
	\end{equation}
\end{lemma}

\begin{lemma}\label{lem: non-linear bdded theta 1}
	Suppose Assumptions \ref{asmp: x}, \ref{asmp: non-linear}, and \ref{asmp: non-linear n} hold. Then \wppp, for all $t \in S$, for all $\bA \in \mO$ with $\twonorm{\bA\bA^\top - \bAks{t}(\bAks{t})^\top} \leq c(\zetak{t})^{-1}$, where $c > 0$ is a small constant, we have 
	\begin{enumerate}[(i)]
		\item $\twonorm{\bthetak{t}_{\bA}} \leq C''\zetak{t}$;
		\item $\twonorm{\bA\bthetak{t}_{\bA} - \bAks{t}\bthetaks{t}} \leq c'$ with a small constant $c' > 0$.
	\end{enumerate}
\end{lemma}

\begin{lemma}\label{lem: non-linear bdded theta 2}
	Suppose Assumptions \ref{asmp: x}, \ref{asmp: non-linear}, and \ref{asmp: non-linear n} hold. Assume there exists $\wtbA  \in \mO$ such that $\twonorm{\wtbA\bthetak{t}_{\wtbA} - \bAks{t}\bthetaks{t}} \leq c$, where $c > 0$ is a small constant. Then \wppp, for all $t \in S$, for all $\bA \in \mO$ with $\twonorm{\bA\bA^\top - \wtbA(\wtbA)^\top}\leq c'(\zetak{t})^{-1}$, where $c' > 0$ is a small constant, we have 
	\begin{enumerate}[(i)]
		\item $\twonorm{\bthetak{t}_{\bA}} \leq C''\zetak{t}$;
		\item $\twonorm{\bA\bthetak{t}_{\bA} - \bAks{t}\bthetaks{t}} \leq c''$ with a small constant $c'' > 0$.
	\end{enumerate}
\end{lemma}

\begin{lemma}\label{lem: non-linear a theta not far}
	Suppose Assumptions \ref{asmp: x}, \ref{asmp: non-linear}, and \ref{asmp: non-linear n} hold. Let $n \geq C\lambda^2$ with a sufficiently large $C$. Given $\wtbA \in \mO$, denote $(\bAk{t}, \bthetak{t}) = \argmin_{\bA \in \mO, \btheta \in \mathbb{R}^r}\{\fk{t}(\bA\btheta) + \frac{\lambda}{\sqrt{n}}\twonorm{\bA\bA^\top - \wtbA(\wtbA)^\top}\}$. Then \wpp, for any $\wtbA \in \mO$, we have $\twonorm{\bAk{t}\bthetak{t} - \bbetaks{t}} \leq c$ with a small constant $c > 0$.
\end{lemma}

\subsection{Proof of Theorem \ref{thm: non-linear mtl}}
For the penalized ERM, the proof logic is almost the same as the logic underlying the proofs of Proposition \ref{prop: mtl} and Theorem \ref{thm: mtl}. Similar to the proof of Theorem \ref{thm: glm mtl}, we only point out the differences and skip the details. We need to bound $\twonorm{\hbarA(\hbarA)^\top - \barA(\barA)^\top}$ first. Then we know that $\twonorm{\hbarA(\hbarA)^\top - \barA(\barA)^\top}$ and $\twonorm{\hbarA\bthetak{t}_{\hbarA} - \bAks{t}\bthetaks{t}}$ are small w.h.p. Then a direct application of Lemma \ref{lem: non-linear bdded theta 2} implies that $\twonorm{\hbarA\bthetaks{t}_{\hbarA} - \bAks{t}\bthetaks{t}}$ is small w.h.p. Finally, applying Lemma \ref{lem: non-linear a theta}, we have $\twonorm{\hbarA\bthetaks{t}_{\hbarA} - \barA\bthetak{t}_{\barA}} \lesssim \twonorm{\hbarA(\hbarA)^\top - \barA(\barA)^\top}(\twonorm{\bthetak{t}_{\barA}} +\twonorm{\nabla \fk{t}(\barA\bthetak{t}_{\barA})}) \lesssim \twonorm{\hbarA(\hbarA)^\top - \barA(\barA)^\top}\zetak{t}$ w.h.p. which gives us the ideal bound for $\twonorm{\hbarA\bthetaks{t}_{\hbarA} - \barA\bthetak{t}_{\barA}}$ for all $t \in S$. 

\noindent (\rom{1}) In the proof of Proposition \ref{prop: mtl}, for $t \in \mA_1$, we know $\twonorm{\bAk{t}\bthetak{t} - \bAks{t}\bthetaks{t}} \leq c$ with a small constant $c > 0$ w.h.p. by Lemma \ref{lem: non-linear a theta not far}. Then by Lemmas \ref{lem: non-linear rsc}, \ref{lem: non-linear gradient 2 norm},  and \ref{lem: non-linear a theta star}, \wprp,
\begin{align}
	&\frac{1}{T}\sum_{t \in \mA_1}\left[\fk{t}(\bAk{t}\bthetak{t}) - \fk{t}(\barA\bthetak{t}_{\barA})\right] \\
	&\geq -\frac{1}{T}\norma{\left\<\{\nabla \fk{t}(\bAks{t}\bthetaks{t})\}_{t \in \mA_1}, \{\bAk{t}\bthetak{t} - \hbarA\bthetak{t}_{\hbarA} + \hbarA\bthetak{t}_{\hbarA}- \barA\bthetak{t}_{\barA}\}_{t \in \mA_1}\right\>}\\
	&\quad + \frac{C_1}{T}\fnorm{\{\bAk{t}\bthetak{t} - \barA\bthetak{t}_{\barA}\}_{t \in \mA_1}}^2 -  \frac{C_2}{T}\fnorm{\{\bAks{t}\bthetaks{t} - \barA\bthetak{t}_{\barA}\}_{t \in \mA_1}}^2\\
	&\geq -\frac{C}{T}\sum_{t \in \mA_1}\sqrt{\frac{p+\log T}{n}}\cdot \zetak{t}(\twonorm{\bAk{t}(\bAk{t})^\top - \hbarA(\hbarA)^\top} + \twonorm{\hbarA(\hbarA)^\top- \barA\barA^\top})\\
	&\quad + \frac{C_1}{T}\fnorm{\{\bAk{t}\bthetak{t} - \barA\bthetak{t}_{\barA}\}_{t \in \mA_1}}^2 -C\left(h^2\cdot \frac{1}{T}\sum_{t \in \mA_1}(\zetak{t})^2 + \frac{r+\log T}{n}\right) \\
	&\geq -C\left(h^2\cdot \frac{1}{T}\sum_{t \in \mA_1}(\zetak{t})^2 + \frac{r+\log T}{n}\right)-\frac{C'}{T}\sum_{t \in \mA_1}\sqrt{\frac{p+\log T}{n}}\cdot \zetak{t}\sqrt{r}\twonorm{\bAk{t}(\bAk{t})^\top - \hbarA(\hbarA)^\top}.
\end{align}
Therefore the calculations in \eqref{eq: prop 1 eq 1} are still correct under non-linear regression model \eqref{eq: non-linear model}.

\noindent (\rom{2}) In the proof of Proposition \ref{prop: mtl}, for $t \in \mA_2$, we used the fact that \wpr,
\begin{equation}\label{eq: thm non-linear mtl eq 1}
	\twonorm{\bAk{t}\bthetak{t}_{\bAk{t}} - \hbarA\bthetak{t}_{\hbarA}} \lesssim \twonorm{\bAk{t}(\bAk{t})^\top - \hbarA(\hbarA)^\top}(\twonorm{\bthetak{t}_{\hbarA}} +\twonorm{\nabla \fk{t}(\hbarA\bthetak{t}_{\hbarA})}).
\end{equation}
Here this still holds because
\begin{itemize}
	\item $\twonorm{\bAk{t}\bthetak{t}_{\bAk{t}} - \bAks{t}\bthetaks{t}} \leq c$ with a small constant $c > 0$, by Lemma \ref{lem: non-linear a theta not far};
	\item For $t \in \mA_2$, we know that $\twonorm{\bAk{t}(\bAk{t})^\top - \hbarA(\hbarA)^\top} \leq \frac{c}{\sqrt{r}} \leq c'(\zetak{t})^{-1}$ with a small constant $c' > 0$. Then by Lemma \ref{lem: non-linear bdded theta 2}, $\twonorm{\bthetak{t}_{\hbarA}} \leq C\zetak{t}$ and $\twonorm{\hbarA\bthetak{t}_{\hbarA} - \bAks{t}\bthetaks{t}} \leq Cc'$ with a small constant $c' > 0$.
\end{itemize}
Then \eqref{eq: thm non-linear mtl eq 1} holds by Lemma \ref{lem: non-linear a theta}.

Finally, Lemma \ref{lem: safe net mtl} still holds for non-linear regression model \eqref{eq: non-linear model}. Note that in Lemma \ref{lem: safe net mtl}, (\rom{1}) only requires the maximum eigenvalue of Hessian to be upper bounded when the evaluation point is close to the true $\bbetaks{t}$ (see Lemma E.3 in \cite{duan2023adaptive}), which is true by Assumptions \ref{asmp: x} and \ref{asmp: non-linear} (see our analysis in the proof of Lemma \ref{lem: non-linear a theta}). By Lemma E.2 in \cite{duan2023adaptive}, (\rom{2}) and (\rom{3}) require the minimum eigenvalue of Hessian to be lower bounded when $\twonorm{\bbetak{t} - \bbetaks{t}} \leq $ some constant $C$ w.h.p. and $\gamma/\sqrt{n_0} \leq C$, both of which are true. 

For the spectral method, the proof is almost the same as the proof of Theorem \ref{thm: spectral mtl}, hence we do not repeat it here.. 

\subsection{Proof of Theorem \ref{thm: non-linear tl}}
We only outline the proof when Algorithm \ref{algo: tl} is coupled with Algorithm \ref{algo: mtl}. If Algorithm \ref{algo: tl} is coupled with Algorithm \ref{algo: spectral}, we can follow a similar argument with the proof of Theorem \ref{thm: spectral mtl}.

Denote $\eta = r\sqrt{\frac{p}{nT}} + \sqrt{r}h + \sqrt{r}\sqrt{\frac{r + \log T}{n}} + \frac{|S^c|}{T}\cdot r \cdot \frac{\lambda}{\sqrt{n}}$.

\noindent (\rom{1}) When $\eta \leq C\sqrt{\frac{p}{n_0}}$: since $\twonorm{\hbarA(\hbarA)^\top - \bAks{0}(\bAks{0})^\top} \leq $ a small constant $c$, by Lemma \ref{lem: non-linear a theta not far}, \wpp, $\twonorm{\hbarA\bthetak{t}_{\hbarA} - \bAks{0}\bthetaks{0}} \leq $ a small constant $c'$. Then by Lemma \ref{lem: non-linear a theta star} and Theorem \ref{thm: non-linear mtl}, \wpr, $\twonorm{\hbarA\bthetak{t}_{\hbarA} - \bAks{0}\bthetaks{0}} \lesssim \twonorm{\hbarA(\hbarA)^\top - \bAks{0}(\bAks{0})^\top} \lesssim \eta$. As we commented in the proof of Theorem \ref{thm: non-linear mtl}, Lemma \ref{lem: safe net tl}.(\rom{1}) still holds for non-linear regression model \eqref{eq: non-linear model} because the maximum eigenvalue of Hessian is upper bounded w.h.p.  when the evaluation point is close to the true $\bbetaks{t}$

\noindent (\rom{2}) When $\eta > C\sqrt{\frac{p}{n_0}}$: the result comes from Lemma \ref{lem: safe net tl}.(\rom{2}), which still holds when $\gamma/\sqrt{n_0} \lesssim $ some constant $C$ because of Lemma E.2 in \cite{duan2023adaptive}.

\subsection{Proofs of Lemmas}

\subsubsection{Proof of Lemma \ref{lem: non-linear rsc}}
First,
\begin{align}
	\fk{t}(\bbeta) - \fk{t}(\wtbbeta) - \nabla \fk{t}(\wtbbeta)^\top(\bbeta - \wtbbeta) &\geq \underbrace{\fk{t}(\bbeta) - \fk{t}(\bbetaks{t}) - \nabla \fk{t}(\bbetaks{t})^\top(\bbeta - \bbetaks{t})}_{[1]}\\
	&\quad + \underbrace{\fk{t}(\bbetaks{t}) - \fk{t}(\wtbbeta) - \nabla \fk{t}(\bbetaks{t})^\top(\bbetaks{t}-\wtbbeta)}_{[2]}.
\end{align}
By Taylor expansion, 
\begin{align}
	[1] &= \left\<\int_0^1 \nabla \fk{t}(\delta\bbeta + (1-\delta)\bbetaks{t})\dint t - \nabla \fk{t}(\bbetaks{t}), \bbeta - \bbetaks{t}\right\> \\
	&=\frac{2}{n}\sum_{i=1}^n\int_0^1 [g(\delta(\bxk{t}_i)^\top\bbeta + (1-\delta)(\bxk{t}_i)^\top\bbetaks{t}) - g((\bxk{t}_i)^\top\bbetaks{t})]g'((\bxk{t}_i)^\top\delta\bbeta + (1-\delta)\bbetaks{t})\dint \delta\\
	&\quad \cdot [(\bxk{t}_i)^\top(\bbeta - \bbetaks{t})] - \frac{2}{n}\sum_{i=1}^n\epsilonk{t}_i\cdot \int_0^1 [g'(\delta(\bxk{t}_i)^\top\bbeta + (1-\delta)(\bxk{t}_i)^\top\bbetaks{t}) - g'((\bxk{t}_i)^\top\bbetaks{t})]\dint \delta \\
	&\quad \cdot [(\bxk{t}_i)^\top(\bbeta - \bbetaks{t})]\\
	&= \frac{2}{n}\sum_{i=1}^n \int_0^1 \underbrace{g'((\bxk{t}_i)^\top\bbeta_i)}_{\geq C}\cdot \underbrace{g'(\delta(\bxk{t}_i)^\top\bbeta + (1-\delta)(\bxk{t}_i)^\top\bbetaks{t})}_{\geq C} \delta\dint \delta\cdot [(\bxk{t}_i)^\top(\bbeta - \bbetaks{t})]^2 \\
	&\quad - \frac{2}{n}\sum_{i=1}^n \epsilonk{t}_i\cdot \int_0^1\int_0^1 g''(w(\delta(\bxk{t}_i)^\top\bbeta + (1-\delta)(\bxk{t}_i)^\top\bbetaks{t}) + (1-w)(\bxk{t}_i)^\top\bbetaks{t})\delta\dint \delta\dint w \\
	&\quad \cdot [(\bxk{t}_i)^\top(\bbeta - \bbetaks{t})]^2\\
	&\geq C(\bbeta - \bbetaks{t})^\top\hSigmak{t}(\bbeta - \bbetaks{t}) - \frac{2}{n}\sum_{i=1}^n \epsilonk{t}_ib_i^{(t)}[(\bxk{t}_i)^\top(\bbeta - \bbetaks{t})]^2,
\end{align}
where $b_i^{(t)} = \int_0^1\int_0^1 g''(w(\delta(\bxk{t}_i)^\top\bbeta + (1-\delta)(\bxk{t}_i)^\top\bbetaks{t}) + (1-w)(\bxk{t}_i)^\top\bbetaks{t})\delta\dint \delta\dint w$ and $|b_i^{(t)}| \leq C'$. Therefore, by H\"older's inequality,
\begin{align}
	\norma{\frac{2}{n}\sum_{i=1}^n \epsilonk{t}_ib_i^{(t)}[(\bxk{t}_i)^\top(\bbeta - \bbetaks{t})]^2} &\leq \frac{2C'}{n}\sum_{i=1}^n |\epsilonk{t}_i|[(\bxk{t}_i)^\top(\bbeta - \bbetaks{t})]^2 \\
	&\leq \frac{2C'}{n}\sup_{\twonorm{\bu} \leq 1}\twonorma{\sum_{i=1}^n |\epsilonk{t}_i|\cdot [(\bxk{t}_i)^\top\bu]^2}\cdot \twonorm{\bbeta - \bbetaks{t}}^2.
\end{align}
Suppose $\{\bu_j\}_{j=1}^N$ is a $1/64$-packing of the unit ball in $\mathbb{R}^p$, with $N \lesssim 129^p$. By standard arguments,
\begin{equation}
	\frac{1}{n}\sup_{\twonorm{\bu} \leq 1}\twonorma{\sum_{i=1}^n |\epsilonk{t}_i|\cdot [(\bxk{t}_i)^\top\bu]^2} \lesssim \frac{1}{n}\max_{j \in [N]}\left\{\sum_{i=1}^n |\epsilonk{t}_i|\cdot [(\bxk{t}_i)^\top\bu_j]^2\right\}.	
\end{equation}
It is easy to see that for any $\delta > 0$,
\begin{align}
	\tP\left(|\epsilonk{t}_i|\cdot [(\bxk{t}_i)^\top\bu_j]^2 - \tE\{|\epsilonk{t}_i|\cdot [(\bxk{t}_i)^\top\bu_j]^2\} > \delta\right) &\leq \tP\left(|\epsilonk{t}_i|\cdot [(\bxk{t}_i)^\top\bu_j]^2 > \delta\right) \\
	&\leq \tP(|\epsilonk{t}_i| > \delta^{1/3}) + \tP(|(\bxk{t}_i)^\top\bu_j|>\delta^{1/3}) \\
	&\lesssim \exp\{-C\delta^{2/3}\},
\end{align}
and $\tE\left\{\{|\epsilonk{t}_i| [(\bxk{t}_i)^\top\bu_j]^2 - \tE[|\epsilonk{t}_i| [(\bxk{t}_i)^\top\bu_j]^2]\}^2 \cdot \mathds{1}\big(|\epsilonk{t}_i| [(\bxk{t}_i)^\top\bu_j]^2 - \tE[|\epsilonk{t}_i| [(\bxk{t}_i)^\top\bu_j]^2] <0\big)\right\} \leq \sqrt{\tE \big\{|\epsilonk{t}_i|^2 [(\bxk{t}_i)^\top\bu_j]^4\big\}} \leq C < \infty$. By Corollary 2 in \cite{bakhshizadeh2020sharp},
\begin{equation}
	\tP\left(|\epsilonk{t}_i|\cdot [(\bxk{t}_i)^\top\bu_j]^2 - \tE\{|\epsilonk{t}_i|\cdot [(\bxk{t}_i)^\top\bu_j]^2\} > \delta\right) \leq \exp\{-Cn\delta\} + \exp\{-C(n\delta)^{2/3}\},
\end{equation}
which entails that \wppp,
\begin{align}
	\frac{1}{n}\max_{j \in [N]}\left\{\sum_{i=1}^n |\epsilonk{t}_i|\cdot [(\bxk{t}_i)^\top\bu_j]^2\right\} &\lesssim \max_{j \in [N]}\tE\{|\epsilonk{t}_i|\cdot [(\bxk{t}_i)^\top\bu_j]^2\} + \sqrt{\frac{p+\log T}{n}}\vee \frac{(p+\log T)^{3/2}}{n} \\
	&\leq C''.
\end{align}
Hence by Lemma \ref{lem: cov hat} and the assumption on $g''$, \wppp,
\begin{equation}\label{eq: lem non-linear rsc eq 1}
	[1] \geq C\twonorm{\bbeta - \bbetaks{t}}^2 \geq \frac{C}{2}\twonorm{\bbeta - \wtbbeta}^2 - C\twonorm{\wtbbeta - \bbetaks{t}}^2.
\end{equation}
On the other hand, similarly, we can show that \wppp,
\begin{equation}\label{eq: lem non-linear rsc eq 2}
	[2] \lesssim \twonorm{\wtbbeta - \bbetaks{t}}^2.
\end{equation}
Combining \eqref{eq: lem non-linear rsc eq 1} and \eqref{eq: lem non-linear rsc eq 2} completes the proof.

\subsubsection{Proof of Lemma \ref{lem: non-linear two norm gradient}}
Note that $\nabla \fk{t}(\bbetaks{t}) = -\frac{2}{n}\sum_{i = 1}^n \epsilonk{t}_i g'((\bxk{t}_i)^\top\bbetaks{t})\bxk{t}_i$, where $\epsilonk{t}_i g'((\bxk{t}_i)^\top\bbetaks{t})\bxk{t}_i$ is sub-exponential in the sense that $\tE \exp\{\lambda\epsilonk{t}_i g'((\bxk{t}_i)^\top\bbetaks{t})(\bxk{t}_i)^\top \bu \} \leq \exp\{C^2\lambda^2\twonorm{\bu}^2\}$ for any $\bu \in \mathbb{R}^p$ and $\lambda \in \mathbb{R}$ with $|\lambda|\twonorm{\bu} \leq C'$. Then the proof is very similar to the proof of Lemma \ref{lem: glm two norm gradient}, so we omit the details.

\subsubsection{Proof of Lemma \ref{lem: non-linear gradient 2 norm}}
The proof is very similar to the proof of Lemma \ref{lem: glm gradient 2 norm}, so omitted.

\subsubsection{Proof of Lemma \ref{lem: non-linear a theta}}
Denote $\bbeta_{\delta} = \delta\bbeta + (1-\delta)\wtbbeta$ with some $t \in [0, 1]$. It suffices to show that w.h.p, for all $t  \in [0, 1]$ and all $t \in S$, the minimum and maximum eigenvalues of $\nabla^2 \fk{t}(\bbeta_{\delta})$ are bounded away from zero and infinity.

Note that
\begin{align}
	\nabla^2 \fk{t}(\bbeta_{\delta}) &=  \frac{1}{n}\sum_{i=1}^n [g'((\bxk{t}_i)^\top \bbeta_{\delta})]^2\cdot \bxk{t}_i(\bxk{t}_i)^\top \\
	&\quad + \frac{1}{n}\sum_{i=1}^n [g((\bxk{t}_i)^\top \bbeta_{\delta}) - \yk{t}_i]\cdot g''((\bxk{t}_i)^\top\bbeta_{\delta})\cdot \bxk{t}_i(\bxk{t}_i)^\top \\
	&= \underbrace{\frac{1}{n}\sum_{i=1}^n [g'((\bxk{t}_i)^\top \bbeta_{\delta})]^2\cdot \bxk{t}_i(\bxk{t}_i)^\top}_{[1]} \\
	&\quad + \underbrace{\frac{1}{n}\sum_{i=1}^n [g((\bxk{t}_i)^\top \bbeta_{\delta}) - g((\bxk{t}_i)^\top \bbetaks{t})]\cdot g''((\bxk{t}_i)^\top\bbeta_{\delta})\cdot \bxk{t}_i(\bxk{t}_i)^\top}_{[2]} \\
	&\quad - \underbrace{\frac{1}{n}\sum_{i=1}^n \epsilonk{t}_i\cdot g''((\bxk{t}_i)^\top\bbeta_{\delta})\cdot \bxk{t}_i(\bxk{t}_i)^\top}_{[3]}.
\end{align}
It is straightforward to see that \wppp,
\begin{equation}
	0 <C \leq \min_{t \in S}\lambdamin([1]) \leq \max_{t \in S}\lambdamax([1]) \leq C' < \infty,
\end{equation}
by the condition on $g'$ and Lemma \ref{lem: cov hat}. And similar to the analysis in the proof of Lemma \ref{lem: non-linear rsc}, \wppp,
\begin{align}
	\lambdamax([3])  \leq c,
\end{align}
where $c$ is small. Similarly, it can be shown that
\begin{align}
	&[2] = \frac{1}{n}\sum_{i=1}^n \int_0^1 g'(w(\bxk{t}_i)^\top\bbeta_{\delta} + (1-w)(\bxk{t}_i)^\top\bbetaks{t})\dint w \cdot g''((\bxk{t}_i)^\top\bbeta_{\delta})\cdot \bxk{t}_i(\bxk{t}_i)^\top\cdot (\bxk{t}_i)^\top(\bbeta_{\delta} - \bbetaks{t}),\\
	&\lambdamax([2]) \leq C\twonorm{\bbeta_{\delta} - \bbetaks{t}} \lesssim \twonorm{\bbeta - \bbetaks{t}} +  \twonorm{\wtbbeta - \bbetaks{t}} \leq c,
\end{align}
\wppp, where $c$ is small. Putting all the pieces together, we have
\begin{equation}
	0 <C \leq \min_{t \in S}\lambdamin(\nabla^2 \fk{t}(\bbeta_{\delta})) \leq \max_{t \in S}\lambdamax(\nabla^2 \fk{t}(\bbeta_{\delta})) \leq C' < \infty,
\end{equation}
\wpp. The remaining steps are the same as in the proof of Lemma \ref{lem: a theta a extention}.

\subsubsection{Proof of Lemma \ref{lem: non-linear a theta star}}
The proof is similar to the proofs of Lemmas \ref{lem: non-linear a theta} and \ref{lem: a theta a}, so omitted.

\subsubsection{Proof of Lemma \ref{lem: non-linear bdded theta 1}}
Define $\widetilde{\btheta}^{(t)}_{\bA} = \bA^\top\bAks{t}\bthetaks{t}$. It is easy to see that
\begin{equation}
	\twonorm{\bA\widetilde{\btheta}^{(t)}_{\bA} - \bAks{t}\bthetaks{t}} \leq \twonorm{(\bA\bA^\top - \bm{I})\bAks{t}\bthetaks{t}} \lesssim \twonorm{(\bA^{\perp})^\top\bAks{t}}\zetak{t} \leq c.
\end{equation}
Then by Lemma \ref{lem: non-linear rsc}, \wpp,
\begin{equation}
	\fk{t}(\bA\bthetak{t}_{\bA}) - \fk{t}(\bA\widetilde{\btheta}^{(t)}_{\bA}) \geq C\twonorm{\bA\bthetak{t}_{\bA} - \bA\widetilde{\btheta}^{(t)}_{\bA}}^2 - C'\twonorm{\bA\widetilde{\btheta}^{(t)}_{\bA} - \bbetaks{t}}^2 - C'\twonorm{\nabla \fk{t}(\bbetaks{t})}\cdot \twonorm{\bA\bthetak{t}_{\bA} - \bA\widetilde{\btheta}^{(t)}_{\bA}},
\end{equation}
where the RHS must be non-positive, otherwise it is contradicted by the definition of $\bthetak{t}_{\bA}$. Therefore, $\twonorm{\bA\bthetak{t}_{\bA} - \bA\widetilde{\btheta}^{(t)}_{\bA}} \lesssim \twonorm{\bA\widetilde{\btheta}^{(t)}_{\bA} - \bbetaks{t}} + \twonorm{\nabla \fk{t}(\bbetaks{t})}^{1/2} \leq c$, where $c > 0$ is a small constant, \wpp. This also implies that $\twonorm{\bthetak{t}_{\bA}} \leq C\zetak{t}$ \wppp.

\subsubsection{Proof of Lemma \ref{lem: non-linear bdded theta 2}}
Define $\widetilde{\btheta}^{(t)}_{\bA} = \bA^\top\widetilde{\bA}\bthetak{t}_{\widetilde{\bA}}$. It is easy to see that
\begin{equation}
	\twonorm{\bA\widetilde{\btheta}^{(t)}_{\bA} - \widetilde{\bA}\bthetak{t}_{\widetilde{\bA}}} \leq \twonorm{(\bA\bA^\top - \bm{I})\widetilde{\bA}\bthetak{t}_{\widetilde{\bA}}} \lesssim \twonorm{(\bA^{\perp})^\top\widetilde{\bA}}\zetak{t} \leq c.
\end{equation}
Then by Lemma \ref{lem: non-linear rsc}, \wpp,
\begin{equation}
	\fk{t}(\bA\bthetak{t}_{\bA}) - \fk{t}(\bA\widetilde{\btheta}^{(t)}_{\bA}) \geq C\twonorm{\bA\bthetak{t}_{\bA} - \bA\widetilde{\btheta}^{(t)}_{\bA}}^2 - C'\twonorm{\bA\widetilde{\btheta}^{(t)}_{\bA} - \bbetaks{t}}^2 - C'\twonorm{\nabla \fk{t}(\bbetaks{t})}\cdot \twonorm{\bA\bthetak{t}_{\bA} - \bA\widetilde{\btheta}^{(t)}_{\bA}},
\end{equation}
where the RHS must be non-positive, otherwise it is contradicted by the definition of $\bthetak{t}_{\bA}$. Therefore, $\twonorm{\bA\bthetak{t}_{\bA} - \bA\widetilde{\btheta}^{(t)}_{\bA}} \lesssim \twonorm{\bA\widetilde{\btheta}^{(t)}_{\bA} - \bbetaks{t}} + \twonorm{\nabla \fk{t}(\bbetaks{t})}^{1/2} \leq $ a small constant $c$ \wpp. This also implies that $\twonorm{\bA\bthetak{t}_{\bA} - \bbetaks{t}} \leq \twonorm{\bA\bthetak{t}_{\bA} - \bA\widetilde{\btheta}^{(t)}_{\bA}} + \twonorm{\bA\widetilde{\btheta}^{(t)}_{\bA} - \widetilde{\bA}\bthetak{t}_{\widetilde{\bA}}} +  \twonorm{\widetilde{\bA}\bthetak{t}_{\widetilde{\bA}} - \bbetaks{t}} \leq $ a small constant $c$ and $\twonorm{\bthetak{t}_{\bA}} \leq C$ \wppp, which completes the proof.

\subsubsection{Proof of Lemma \ref{lem: non-linear a theta not far}}
By Lemmas \ref{lem: non-linear rsc} and \ref{lem: non-linear gradient 2 norm}, \wpp, for any $\wtbA \in \mO$,
\begin{align}
	&\fk{t}(\bAk{t}\bthetak{t}) + \frac{\lambda}{\sqrt{n}}\twonorm{\bAk{t}(\bAk{t})^\top - \wtbA(\wtbA)^\top} - \fk{t}(\bAks{t}\bthetaks{t}) - \frac{\lambda}{\sqrt{n}}\twonorm{\bAks{t}(\bAks{t})^\top - \wtbA(\wtbA)^\top} \\
	&\geq \nabla \fk{t}(\bAks{t}\bthetaks{t})^\top(\bAk{t}\bthetak{t} - \bAks{t}\bthetaks{t}) - 2\frac{\lambda}{\sqrt{n}} + C\twonorm{\bAk{t}\bthetak{t} - \bAks{t}\bthetaks{t}}^2 \\
	&\geq C\twonorm{\bAk{t}\bthetak{t} - \bAks{t}\bthetaks{t}}^2 - C'\sqrt{\frac{p+\log T}{n}}\twonorm{\bAk{t}\bthetak{t} - \bAks{t}\bthetaks{t}}- 2\frac{\lambda}{\sqrt{n}}.
\end{align}
By definition of $\bAk{t}\bthetak{t}$, the RHS must be non-positive, which gives us
\begin{equation}
	\twonorm{\bAk{t}\bthetak{t} - \bAks{t}\bthetaks{t}} \lesssim \sqrt{\frac{p+\log T}{n}} + \frac{\lambda^{1/2}}{n^{1/4}} \leq c,
\end{equation}
where $c > 0$ is a small constant, \wppp.

\section{Proofs for Estimation of Intrinsic Dimension $r$}


\subsection{Proof of Theorem \ref{thm: adaptation unknown r}}
We already mentioned the motivation of the thresholding strategy briefly in Section \ref{sec: unknown r}. Intuitively, there is a gap between $\sigma_r(\bm{B}_S^*)$ and $\sigma_{r+1}(\bm{B}_S^*)$ when $h$ and $|S^c|$ are small. But in practice, we do not have access to the singular values of $\bm{B}_S^*$. We have to estimate $\bm{B}_S^*$ first, then use the singular value of this estimate. As long as the singular values of the estimate are not far from the singular values of $\bm{B}_S^*$, thresholding can work well. We give a rigorous proof in the following. Most parts of the proof follow the arguments in the proof of Theorem \ref{thm: spectral mtl}.
\vskip0.2cm

\noindent Consider the decomposition
\begin{equation}
	\bbetaks{t} = \bP_{\oA}\bbetaks{t} + \bP_{\oA^\perp}\bbetaks{t} = \oA(\oA)^\top \bbetaks{t} + \oA^\perp (\oA^\perp)^\top \bbetaks{t} \coloneqq  \oA\bthetaks{t}_{\oA} + \bdeltaks{t}, \quad \forall t \in S.
\end{equation}
Denote $\widebar{\bm{\Theta}}_S = \{\bthetaks{t}_{\barA}\}_{t \in S}$, $\widebar{\bB}_S = \{\oA\bthetaks{t}_{\barA}\}_{t \in S}$, $\widebar{\bB} = \barA \widebar{\bm{\Theta}} = (\widebar{\bB}_S \quad \bm{0}_{p \times |S^c|})$, $\bB^*_S = \{\bbetaks{t}\}_{t \in S}$, $\widetilde{\bB}_S = \{\widetilde{\bbeta}^{(t)}\}_{t \in S}$, $\widehat{\bB}_S = \{\prod_R(\widetilde{\bbeta}^{(t)})\}_{t \in S}$, $\widehat{\bB} = \{\prod_R(\widetilde{\bbeta}^{(t)})\}_{t=1}^T$, and $\bm{D}_S^* = \{\bdeltaks{t}\}_{t \in S}$.  By a similar argument as in the proof of Theorem \ref{thm: spectral mtl},
\begin{align}
	\twonorm{\widebar{\bB}_S/\sqrt{T} - \bB^*_S/\sqrt{T}} &= \frac{1}{\sqrt{T}}\twonorm{\bm{D}^*_S} \\
	&\lesssim \barzeta h \cdot \bigg[\frac{\sigmamax(\bm{D}^*_S)}{\sqrt{r}\sigmamin(\bm{D}^*_S)} \wedge 1\bigg]  \\
	&\lesssim \frac{\barzeta}{\sqrt{r}} \bigg[\frac{\sigmamax(\bm{D}^*_S)}{\sigmamin(\bm{D}^*_S)} \wedge \sqrt{r}\bigg]\cdot h \cdot \frac{\min_{t \in S}\zetak{t}}{\sqrt{(p+\log T)/n}}\\
	&\lesssim \frac{\barzeta}{\sqrt{r}}\\
	&\leq \widetilde{C}\sqrt{\frac{p+\log T}{n}},
\end{align}
when $\min_{t \in S}\zetak{t}\cdot  h\Big[\frac{\sigmamax(\bm{D}^*_S)}{\sigmamin(\bm{D}^*_S)} \wedge \sqrt{r}\Big] \lesssim \sqrt{\frac{p+\log T}{n}}$. Because $n \gtrsim p+\log T$, we have
\begin{equation}\label{eq: proof thm adaption unknown r eq 1}
	\twonorm{\widebar{\bB}_S/\sqrt{T} - \bB^*_S/\sqrt{T}} \leq \widetilde{C}\sqrt{\frac{p+\log T}{n}},
\end{equation}
\wppp. Since $R = \texttt{quantile}(\{\widetilde{\bbeta}^{(t)}\}_{t=1}^T, 1-\bar{\epsilon})$, we have $\prod_R(\widetilde{\bbeta}^{(t)}) = \widetilde{\bbeta}^{(t)}$ for $\bar{\epsilon}$-proportion of $S$ among all $t \in S$. In addition, by the fact that $\max_{t \in S}\twonorm{\widetilde{\bbeta}^{(t)}-\bbetaks{t}} \lesssim \sqrt{\frac{p+\log T}{n}}$, \wppp, the assumption $\min_{t \in S}\zetak{t} \gtrsim \sqrt{\frac{p+\log T}{n}}$, and $\bar{\epsilon} \geq \epsilon$, we have
\begin{equation}\label{eq: R upper bdd}
	R \leq \max_{t \in S}\twonorm{\widetilde{\bbeta}^{(t)}} \lesssim  \max_{t \in S}\zetak{t},
\end{equation}
\wppp. Therefore,
\begin{equation}
	\twonorm{\widehat{\bB}_S/\sqrt{T} - \widetilde{\bB}_S/\sqrt{T}} \leq \fnorm{\widehat{\bB}_S/\sqrt{T} - \widetilde{\bB}_S/\sqrt{T}} = \sqrt{\frac{1}{T}\sum_{t \in S}\twonorm{\Pi_R(\widetilde{\bbeta}^{(t)})-\widetilde{\bbeta}^{(t)}}^2} \leq C\sqrt{\bar{\epsilon}}R.
\end{equation}
By Lemma 5.39 in \cite{vershynin2010introduction}, we have
\begin{equation}
	\twonorm{\widetilde{\bB}_S/\sqrt{T} - \bm{B}^*_S/\sqrt{T}} \leq C''\sqrt{\frac{p+T}{nT}} \leq C\sqrt{\frac{p+\log T}{n}},
\end{equation}
\wppp. By the equation \eqref{eq: proof thm adaption unknown r eq 1}, and the single-task rate, \wppp,
\begin{align}
	\twonorm{\widehat{\bB}_S/\sqrt{T} - \bB^*_S/\sqrt{T}} &\leq \twonorm{\widehat{\bB}_S/\sqrt{T} - \widetilde{\bB}_S/\sqrt{T}} + \twonorm{\widetilde{\bB}_S/\sqrt{T} - \bB^*_S/\sqrt{T}}\\
	&\leq C\sqrt{\bar{\epsilon}}R + C\sqrt{\frac{p+\log T}{n}}.
\end{align}
Moreover, by the projection, $\twonorm{\widehat{\bB}_{S^c}/\sqrt{T}} \leq \fnorm{\widehat{\bB}_{S^c}/\sqrt{T}}\leq \sqrt{\frac{|S^c|}{T}}\cdot R \leq \sqrt{\bar{\epsilon}}R$, \wppp. Hence \wppp,
\begin{align}
	\twonorm{\widehat{\bB}/\sqrt{T} - \widebar{\bB}/\sqrt{T}} &\leq \twonorm{\widehat{\bB}_S/\sqrt{T} - \widebar{\bB}_S/\sqrt{T}} + \twonorm{\widehat{\bB}_{S^c}/\sqrt{T}} \\
	&\leq \twonorm{\widehat{\bB}_S/\sqrt{T} - \bB^*_S/\sqrt{T}} + \twonorm{\bB^*_S/\sqrt{T} - \widebar{\bB}_S/\sqrt{T}} + \twonorm{\widehat{\bB}_{S^c}/\sqrt{T}} \\
	&\leq (\widetilde{C}+C)\sqrt{\frac{p+\log T}{n}} + (C+1)\sqrt{\bar{\epsilon}}R.
\end{align}
Then by Weyl's inequality, for all $r' \leq r$,
\begin{align}
	\sigma_{r'}(\widehat{\bB}/\sqrt{T}) &\geq \sigma_{r'}(\widebar{\bB}/\sqrt{T}) - (\widetilde{C}+C)\sqrt{\frac{p+\log T}{n}} - (C+1)\sqrt{\bar{\epsilon}}R \\
	&\geq \sigma_{r'}(\bB^*_S/\sqrt{T}) - (2\widetilde{C}+C)\sqrt{\frac{p+\log T}{n}} - (C+1)\sqrt{\bar{\epsilon}}R\\
	&\geq \frac{c}{\sqrt{r}}\barzeta - (2\widetilde{C}+C)\sqrt{\frac{p+\log T}{n}} - (C+1)\sqrt{\bar{\epsilon}}R\\ 
	&\geq 2(2\widetilde{C}+C)\sqrt{\frac{p+\log T}{n}} + 2(C+1)\sqrt{\bar{\epsilon}}R,
\end{align}
\wppp, when $n \geq \frac{36(2\widetilde{C}+C)^2}{c^2\barzeta^2}r(p+\log T)$ and $\frac{c}{2\sqrt{r}}\barzeta \geq 3(C+1)\sqrt{\bar{\epsilon}}R$. The second condition holds because of $\bar{\epsilon} \leq \frac{c'}{r}\Big(\frac{\barzeta}{\max_{t \in S}\zetak{t}}\Big)^2$ and \eqref{eq: R upper bdd}.

On the other hand, by Weyl's inequality, for all $r' > r$, \wppp,
\begin{equation}
	\sigma_{r'}(\widehat{\bB}/\sqrt{T}) \leq \sigma_{r'}(\widebar{\bB}/\sqrt{T}) + (\widetilde{C}+C)\sqrt{\frac{p+\log T}{n}} + (C+1)\sqrt{\bar{\epsilon}}R = (2\widetilde{C}+C)\sqrt{\frac{p+\log T}{n}} + (C+1)\sqrt{\bar{\epsilon}}R,
\end{equation}
because $\sigma_{r'}(\widebar{\bB}/\sqrt{T}) = 0$ when $r' > r$. Finally, by setting $T_1 \in (2\widetilde{C}+C, 4\widetilde{C}+2C]$ and $T_2 \in (C+1, 2C+2]$, we complete the proof.

\end{document}